\documentclass{article} %
\usepackage{preprint,times}
\usepackage[table]{xcolor}

\usepackage{amsmath,amsfonts,bm}

\def\eqref#1{equation~\ref{#1}}

\def\1{\bm{1}}

\def\ra{{\textnormal{a}}}
\def\rb{{\textnormal{b}}}
\def\rc{{\textnormal{c}}}
\def\rd{{\textnormal{d}}}

\def\ry{{\textnormal{y}}}

\def\va{{\bm{a}}}
\def\vb{{\bm{b}}}

\def\vs{{\bm{s}}}

\DeclareMathAlphabet{\mathsfit}{\encodingdefault}{\sfdefault}{m}{sl}
\SetMathAlphabet{\mathsfit}{bold}{\encodingdefault}{\sfdefault}{bx}{n}

\def\tS{{\tens{S}}}

\usepackage{hyperref}
\usepackage{url}
\usepackage{graphicx}
\usepackage{booktabs}
\usepackage{multirow}
\usepackage{array}
\usepackage{float}
\usepackage{wrapfig}
\usepackage{xspace}
\usepackage{tikz}
\usepackage{placeins}
\usetikzlibrary{calc,positioning}
\definecolor{tblue}{RGB}{0,190,225}
\definecolor{tgreen}{RGB}{255,0,190}
\definecolor{tred}{RGB}{255,140,0}

\newcommand{\telescopeemoji}{%
  \raisebox{-0.18em}{%
    \includegraphics[height=1.05em]{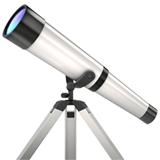}%
  }%
}

\title{
T3lescope\,\telescopeemoji:
Arbitrary-Resolution High-Fidelity Generative Surface Reconstruction from Images
}

\author{Atsuhiro Noguchi$^{1}$\quad Tianhan Xu$^{1}$\quad Yiming Liang$^{1}$\quad Yuta Kikuchi$^{1}$ \\
{\bf Masahiro Ishiyama$^{1}$\quad Shintaro Takagi$^{2}$\quad Hitoshi Murai$^{2}$\quad Eiichi Matsumoto$^{1}$} \\
{\normalfont $^{1}$Preferred Networks, Inc.\quad $^{2}$The University of Tokyo}
}

\newcommand{\method}{T3lescope\xspace}

\begin{document}

\maketitle
\lhead{Preprint}
\newcommand{\relrect}[6]{%
  \coordinate (rA) at ($(#1.north west)!#3!(#1.north east)$);
  \coordinate (rB) at ($(#1.north west)!{#3+#5}!(#1.north east)$);
  \coordinate (rC) at ($(#1.north west)!#4!(#1.south west)$);
  \coordinate (rD) at ($(#1.north west)!{#4+#6}!(#1.south west)$);
  \coordinate (bTL) at (rA |- rC); \coordinate (bTR) at (rB |- rC);
  \coordinate (bBL) at (rA |- rD); \coordinate (bBR) at (rB |- rD);
  \draw[#2, line width=\bw] (bTL) rectangle (bBR);}
\newlength{\arrU}
\newlength{\arrw}
\newcommand{\blockarrowbare}[1]{%
  \begin{scope}[x=\arrU, y=\arrU, scale=#1]
    \filldraw[fill=white, draw=black, line width=0.7pt]
      (-0.040,-0.011) -- (-0.040,0.011) -- (0.020,0.011) -- (0.020,0.024)
      -- (0.048,0) -- (0.020,-0.024) -- (0.020,-0.011) -- cycle;
  \end{scope}}
\newcommand{\blockarrow}[2]{%
  \blockarrowbare{#1}
  \begin{scope}[x=\arrU, y=\arrU, scale=#1]
    \node[inner sep=0] at (-0.010,0) {\resizebox{#1\arrw}{!}{#2}};
  \end{scope}}

\newcommand{\gaparrow}[2]{%
  \coordinate (gm) at ($(#1.east)!0.5!(#2.west)$);
  \begin{scope}[shift={(gm)}]
    \blockarrow{1}{Upscale}
    \node[inner sep=0, anchor=south] at (-0.0012\arrU,0.0086\arrU)
      {\includegraphics[height=0.0496\arrU]{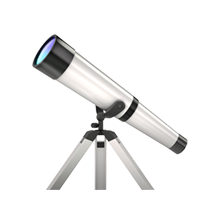}};
  \end{scope}}

\newcommand{\bwh}{0.275pt}
\newcommand{\framein}[2]{%
  \draw[#2, line width=\bw]
    ($(#1.north west)+(\bwh,-\bwh)$) rectangle ($(#1.south east)+(-\bwh,\bwh)$);}

\newcommand{\linkbox}[3]{%
  \relrect{#1}{#3}{0.3}{0.1}{0.5}{0.5}
  \draw[#3, line width=\bw, line join=round]
    ($(#2.north west)+(\bwh,-\bwh)$) -- (bTL) -- (bBL) -- ($(#2.south west)+(\bwh,\bwh)$);}

\begin{figure}[H]
  \centering
  \setlength{\tabcolsep}{0pt}
  \newcommand{\tsc}{0.193\linewidth}
  \setlength{\arrU}{\linewidth}
  \setlength{\arrw}{0.050\arrU}
  \newcommand{\bw}{0.55pt}
  \newcommand{\tsin}{0.20\linewidth}
  \newcommand{\tsar}{0.13\linewidth}
  \newcommand{\tshero}{0.645\linewidth}

  \newsavebox{\stripbox}
  \sbox{\stripbox}{%
  \begin{tikzpicture}
    \node[inner sep=0] (p1) {\includegraphics[width=\tsc]{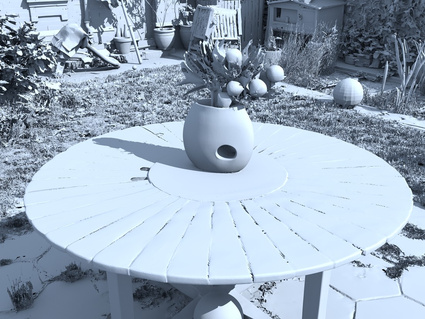}};
    \node[inner sep=0, right=2pt of p1] (p2) {\includegraphics[width=\tsc]{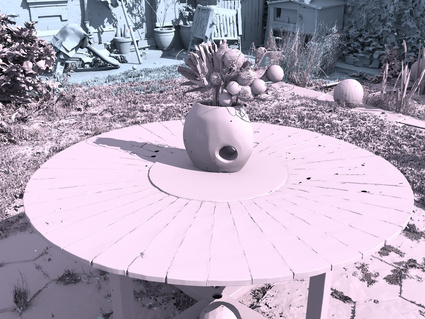}};
    \node[inner sep=0, right=0.4pt of p2] (p3) {\includegraphics[width=\tsc]{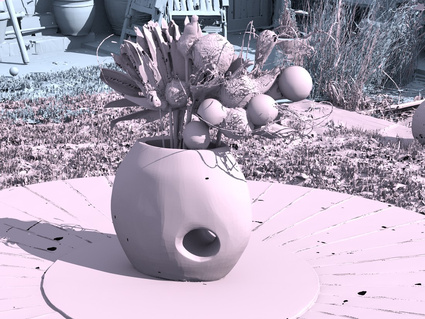}};
    \node[inner sep=0, right=2pt of p3] (p4) {\includegraphics[width=\tsc]{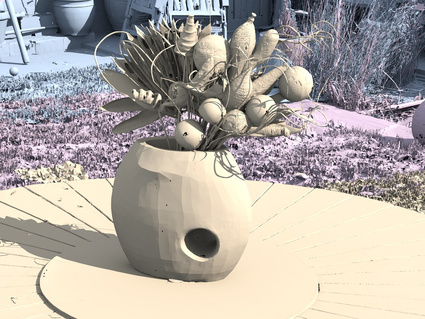}};
    \node[inner sep=0, right=0.4pt of p4] (p5) {\includegraphics[width=\tsc]{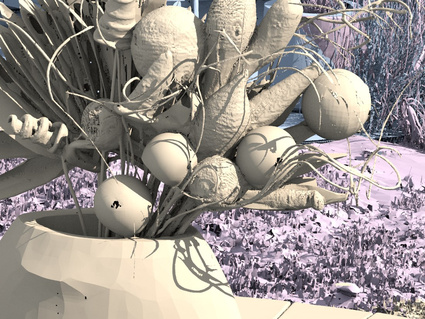}};
    \framein{p3}{tgreen}
    \framein{p5}{tred}
    \linkbox{p2}{p3}{tgreen}
    \linkbox{p4}{p5}{tred}
    \gaparrow{p1}{p2}
    \gaparrow{p3}{p4}
    \node[below=1pt of p1, inner sep=0, font=\footnotesize] {Cascade level 0};
    \node[inner sep=0, font=\footnotesize, anchor=north, yshift=-1pt] at ($(p2.south)!0.5!(p3.south)$) {Cascade level 1};
    \node[inner sep=0, font=\footnotesize, anchor=north, yshift=-1pt] at ($(p4.south)!0.5!(p5.south)$) {Cascade level 2};
  \end{tikzpicture}}
  \usebox{\stripbox}\\[3pt]

  \makebox[\wd\stripbox][s]{%
    \begin{minipage}[c]{\tsin}
      \centering
      \includegraphics[width=\linewidth]{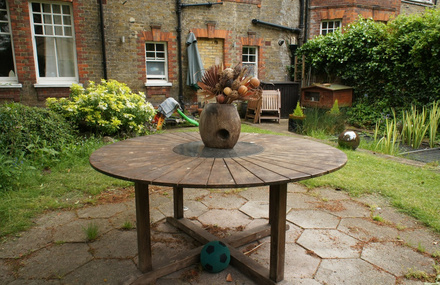} \\[1pt]
      \includegraphics[width=\linewidth]{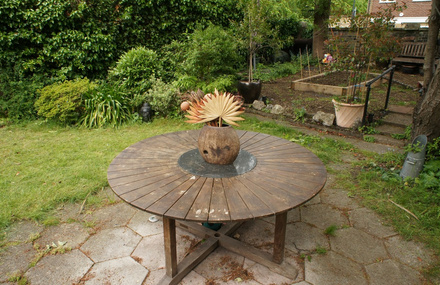} \\[1pt]
      \makebox[\linewidth][c]{\footnotesize\begin{tabular}{@{}c@{}}185 views\\[-2pt] $5187\times3361$ resolution\end{tabular}}
    \end{minipage}\hfill
    \begin{minipage}[c]{\tsar}
      \centering
      \raisebox{9pt}{\begin{tikzpicture}
        \blockarrowbare{1}
        \node[inner sep=0, anchor=south, font=\footnotesize, align=center]
          at (0.004\arrU,0.027\arrU) {Generative\\[-1pt]reconstruction};
      \end{tikzpicture}}
    \end{minipage}\hfill
    \begin{minipage}[c]{\tshero}
      \centering
      \begin{tikzpicture}
        \node[inner sep=0] (hero) {\includegraphics[width=\linewidth]{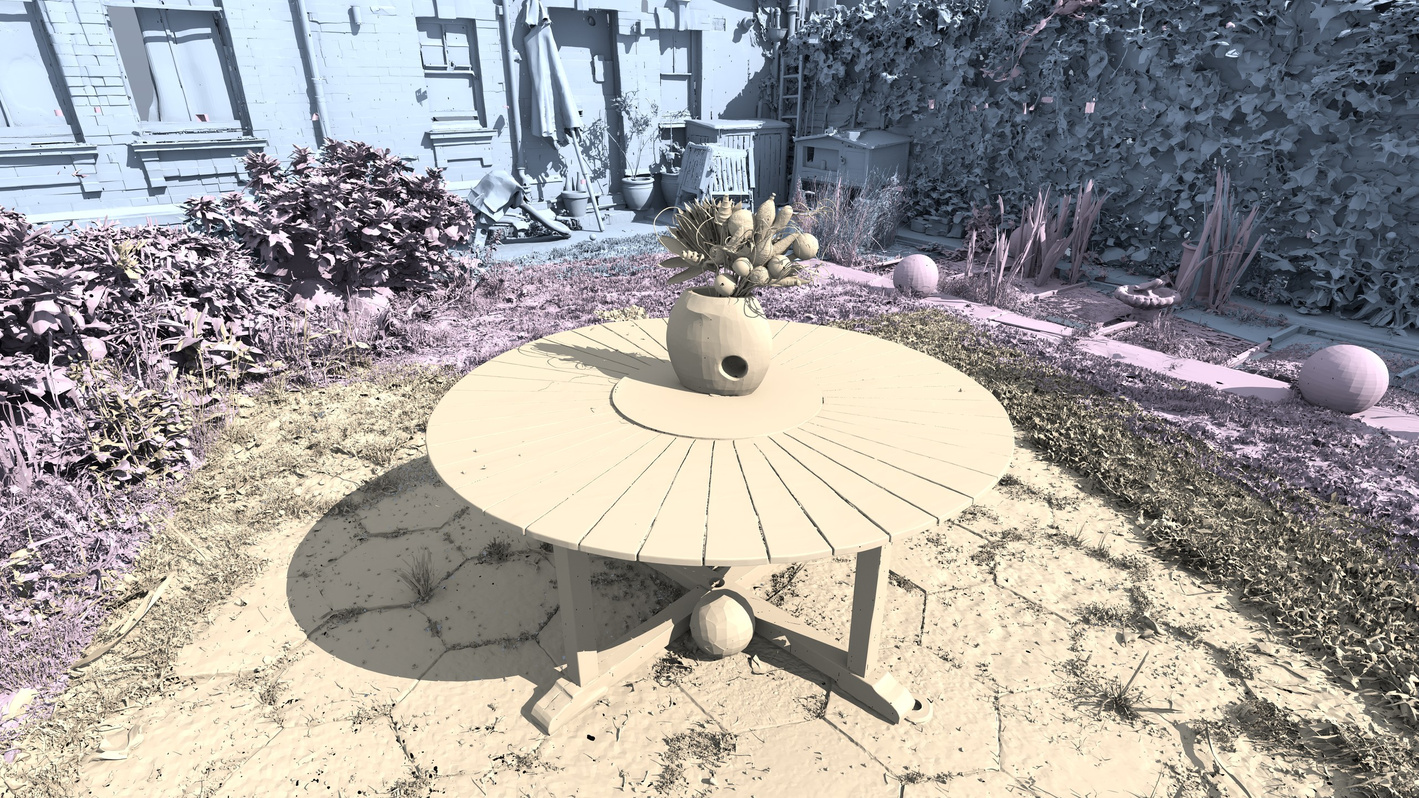}};
        \node[anchor=north, inner sep=0, font=\footnotesize] at ([yshift=-2pt]hero.south) {Reconstructed mesh (three cascade levels)};
      \end{tikzpicture}
    \end{minipage}}

  \caption{\textbf{High-fidelity mesh reconstruction without per-scene optimization.}
  Our method takes multi-view images as input and directly outputs a high-resolution mesh
  with mesh flow models. We adopt a coarse-to-fine cascade: the entire scene is generated
  at a coarse resolution first, and regions of interest are then upscaled. The cascade
  layout and the mesh resolution are chosen at inference time. Colors mark the cascade
  level each surface was generated at.}
  \label{fig:teaser}
\end{figure}

\begin{abstract}
We reconstruct high-fidelity 3D scene meshes from posed multi-view images without per-scene optimization, across scales ranging from single objects to large outdoor scenes. Per-scene optimization methods lack the learned 3D prior needed when observations are sparse or surfaces are glossy or transparent. Existing generative methods leverage such priors to complete geometry in sparsely observed regions, but typically operate at a fixed resolution over a limited spatial extent, trading spatial coverage against detail. Reconstructing a large scene therefore often requires partitioning it into independently processed overlapping local regions, making it difficult to maintain global geometric consistency.
To address these issues, we propose \method, which applies a single fixed-resolution generator across scene scales in an inference-time coarse-to-fine cascade. A coarse level establishes the scene layout, and finer levels perturb and denoise geometry inherited from the coarser level within progressively finer spatial cells to recover surface detail. The model is trained on individual cells at multiple scales and shares its weights across all levels, so no hierarchy is fixed during training, and the number of levels, cell scales, and cell locations are determined at inference time.
On indoor, outdoor, and city-scale scenes, \method outperforms feed-forward and generative baselines, matches or surpasses per-scene optimization, and recovers fine structures as well as glossy and transparent surfaces. These results show that our method generalizes across diverse scenes, view counts, and image resolutions.
Project page: {\color[RGB]{0,102,204}\url{https://pfnet-research.github.io/t3lescope/}}
\end{abstract}

\section{Introduction}
\label{sec:intro}

Reconstructing high-fidelity 3D meshes from posed multi-view images is important for a broad range of real-world applications, including robotic perception, simulation, digital twins, and immersive environments. These applications require both fine surface detail and globally consistent scene geometry, across targets ranging from single objects to large outdoor scenes. Existing approaches broadly fall into three categories: per-scene optimization, feed-forward prediction, and generative reconstruction.
Optimization-based methods fit a scene-specific 3D representation to the input images and can recover accurate surfaces from dense observations~\citep{wang2021neus,huang20242dgs,guedon2025milo}. However, they require optimization for every scene and, lacking a learned 3D prior, fail under sparse views, occlusion, or glossy, reflective, and transparent surfaces.

Feed-forward methods address the cost of per-scene optimization by learning priors across many scenes and directly predicting geometry such as depth or point maps~\citep{wang2024dust3r,lin2025da3,keetha2026mapanything}. However, their outputs are typically image-aligned and must be fused across views to form a complete mesh, where cross-view inconsistencies can introduce noise, holes, or duplicate surfaces.
Generative reconstruction addresses both issues by conditioning a learned 3D generator on the input views. Like feed-forward methods, it avoids per-scene optimization, while its learned prior directly predicts a complete 3D representation, including weakly constrained surfaces~\citep{xiang2025trellis,xiang2025trellis2}. Its main limitation is scale: current 3D generators typically operate on a fixed voxel grid at a single scene scale, trading spatial coverage against geometric detail.
GenRecon~\citep{schmid2026genrecon}, the closest prior work, extends such a fixed-resolution generator to large scenes using overlapping local cells with MultiDiffusion~\citep{bartal2023multidiffusion}. While this increases spatial coverage, distant regions interact only indirectly through overlaps, making global geometry difficult to maintain. Coarse-to-fine 3D generators offer another solution, but existing approaches fix their hierarchy, spatial extent, and output resolution during training~\citep{ren2024xcube,ren2024scube}. These limitations motivate a different question: can a single fixed-resolution generator be composed across scene scales at inference time, from global structure to progressively finer detail?

To this end, we propose \method, an inference-time coarse-to-fine cascade that reuses the same generator across multiple scene scales. A coarse level reconstructs the full scene and establishes its global structure, while each finer level refines inherited geometry by halving the cell size, perturbing it with noise, and denoising it to recover finer surface detail. The same weights are shared across all levels and trained on individual cells at multiple scene scales, without paired coarse-to-fine supervision. Thus, cell size, layout, and cascade depth determine spatial coverage and output resolution at inference time.
This flexibility, however, makes image conditioning nontrivial: cells vary in position and physical scale across scenes and cascade levels, while each level requires image features at a different resolution. To address this, we precompute reusable multi-resolution image features and condition each voxel using its projected location and projected size. This makes conditioning independent of absolute scene position and scale, allowing cells to be placed at arbitrary positions and scales while providing the detail required at each cascade level.

We evaluate \method on two public benchmarks, ScanNet++ and Tanks and Temples, and on our own city-block dataset, under both sparse- and dense-view settings. It outperforms feed-forward and generative baselines and is competitive with per-scene optimization. It recovers fine structures as well as glossy and transparent surfaces. Results also show that our method generalizes across diverse scenes, view counts, and image resolutions (Figures~\ref{fig:teaser} and~\ref{fig:main-scannetpp}). Ablations validate both parent initialization and image conditioning.
Our contributions are:
\begin{itemize}
\setlength{\topsep}{1pt}\setlength{\itemsep}{1pt}\setlength{\parsep}{0pt}
\item We propose an inference-time coarse-to-fine cascade that reuses one fixed-resolution 3D generator across scene scales, with coverage and resolution chosen at inference time.
\item We develop a cell-independent multi-scale conditioning scheme that reuses image features across cells, levels, and sampling steps while allowing flexible cell placement and scale.
\item We show that \method outperforms feed-forward and generative baselines and is competitive with per-scene optimization on indoor, outdoor, and city-scale scenes.
\end{itemize}

\section{Related Work}
\label{sec:related}

\textbf{Mesh reconstruction.}
Mesh reconstruction from posed images has traditionally relied on multi-view stereo,
which matches pixels across images and fuses the recovered
geometry~\citep{schoenberger2016mvs}. Neural implicit and Gaussian-based methods instead
optimize a 3D representation for each
scene~\citep{wang2021neus,yu2022monosdf,huang20242dgs,chen2024pgsr,guedon2025milo,gomez2026gaussianwrapping},
with recent methods scaling to city-scale scenes~\citep{liu2025citygaussianv2}. These approaches recover
accurate geometry from dense views and can be regularized with learned monocular or
geometric cues. However, they require per-scene optimization and tend to degrade under sparse observations or with challenging glossy and transparent surfaces. Feed-forward methods instead learn scene priors that generalize to unseen scenes and predict
image-aligned depth or point maps in a single forward
pass~\citep{wang2024dust3r,lin2025da3,keetha2026mapanything,guo2025murre}. While they avoid per-scene optimization, their predictions must be fused across views, making the resulting mesh sensitive to cross-view inconsistencies. In contrast, our method
directly generates the mesh with a 3D generator, without either per-scene optimization or multi-view fusion.

\textbf{Generative 3D reconstruction.}
Recent 3D generative models directly produce high-resolution meshes through sparse
structures and structured latents~\citep{xiang2025trellis2}. Pixel-aligned image conditioning improves geometric fidelity via explicit 2D--3D correspondence~\citep{li2026pixal3d}. GenRecon extends an object-level 3D generator to scene-scale reconstruction by tiling the scene with local overlapping cells and jointly sampling them with
MultiDiffusion~\citep{schmid2026genrecon,bartal2023multidiffusion}. This extends
generation beyond the model's native spatial extent, but each denoiser operates on a local 
cell, and distant regions interact only indirectly through neighboring overlaps. Instead, our cascade propagates global structure from the coarse reconstruction to each finer cell.

\textbf{Hierarchical generation.}
Coarse-to-fine 3D generators reach high resolution with hierarchies whose levels and
scales are fixed during training, often with a separate model at each
level~\citep{ren2024xcube,ren2024scube,meng2025lt3sd}. Methods for larger
scenes scale spatial extent by progressively extending the generated region~\citep{wu2024blockfusion,lu2025infinicube,joshi2026worldflow3d,li2025worldgrow,meng2026seen2scene}.
In 2D, cascaded diffusion models similarly train a separate model for each
resolution~\citep{ho2022cascaded}, while training-free methods reuse the same model to progressively refine upsampled samples~\citep{meng2022sdedit,du2024demofusion}. We adapt this idea to 3D. At each finer level, we reduce the cell size while keeping the generator resolution fixed. This allows the same model to be reused across scales, with the hierarchy configured at inference time.

\section{Method}
\label{sec:method}

We reconstruct a mesh $\mathcal M$ from posed multi-view images with a
3D generative model, without
per-scene optimization. Since the generator has a fixed resolution, we
reconstruct the scene coarse-to-fine. The coarsest level captures the scene, and finer
levels refine it (Section~\ref{sec:method:cascade}). At each level, we partition the scene
into overlapping cells (Section~\ref{sec:method:singlelevel}) and apply the generator with corresponding image features (Section~\ref{sec:method:attention}). We train a single model across scales (Section~\ref{sec:method:training}) and configure the hierarchy and spatial scale at inference time.
Figure~\ref{fig:pipeline} shows the pipeline.

\subsection{Single-Level Scene Generation}
\label{sec:method:singlelevel}

We adopt TRELLIS.2~\citep{xiang2025trellis2}, a sparse-voxel mesh generator, as our
generative model. It first generates the sparse structure (SS) of surface voxels and then generates a mesh over that structure. We refer to these two
steps as the SS and mesh stages. The input is a set of posed images
$\mathcal I=\{(I_i,K_i,T_i)\}_{i=1}^{N}$, where $I_i$ is an RGB image, $K_i$ its intrinsics, and
$T_i$ its camera pose. Both stages use 3D VAEs to encode their representations and Diffusion Transformers~\citep{peebles2023scalable} for flow matching. The generator operates on a fixed $R^3$ sparse voxel grid and is
conditioned on DINOv3~\citep{simeoni2025dinov3} image features together with camera intrinsics and poses.

For scene generation, we follow GenRecon~\citep{schmid2026genrecon}
and partition the scene into overlapping cells that share a scene-wide latent updated by MultiDiffusion~\citep{bartal2023multidiffusion} in both the SS and mesh stages.
At each flow step, we denoise each cell independently, then blend the overlapping predictions to update the shared scene latent. The world-space edge length $s$ of each cell, the
corresponding voxel spacing $\Delta=s/R$, and the cell placement are all set at inference time.
Because the generator operates in canonical cell coordinates and receives neither absolute world position nor metric cell size, the same model can be applied to cells at arbitrary locations and scales.

However, GenRecon's MultiDiffusion-based inference extends generation beyond a single $R^3$ cell, while each denoiser still operates with only local cell context. Distant regions therefore interact only indirectly through neighboring overlaps, motivating our coarse-to-fine cascade, described in the next section, which propagates global context across scales.

\definecolor{tsel}{RGB}{225,45,45}
\tikzset{
  pl/.style={font=\footnotesize, align=center, inner sep=1pt},
  pt/.style={font=\small\bfseries, anchor=west, inner sep=0},
  pn/.style={font=\scriptsize, align=center, text=black!65, inner sep=1pt},
  box/.style={draw=black!18, fill=black!2.5, rounded corners=4pt, line width=0.5pt},
  flow/.style={draw=black!55, fill=white, rounded corners=3pt, line width=0.6pt,
               font=\footnotesize, align=center, minimum height=8.5mm, minimum width=15mm},
  ar/.style={-{latex}, line width=0.8pt, black!60},
  bigar/.style={-{latex}, line width=1.6pt, black!45},
  att/.style={-{latex}, line width=0.7pt, black!55, dashed},
}
\def\panW{3.75}\def\panH{2.5}
\def\sceneobjs{0.15/0.22/3.60/0.30, 0.22/0.30/1.10/2.28,
               1.55/0.92/3.30/1.06, 1.82/0.30/2.00/0.92, 2.85/0.30/3.03/0.92,
               1.95/0.46/2.90/0.54}
\newcommand{\pipehit}[4]{%
  \gdef\hitflag{0}%
  \foreach \oa/\ob/\oc/\od in \sceneobjs{%
    \pgfmathsetmacro{\meet}{((#1<\oc)&&(#3>\oa)&&(#2<\od)&&(#4>\ob))?1:0}%
    \pgfmathsetmacro{\insd}{((#1>\oa)&&(#3<\oc)&&(#2>\ob)&&(#4<\od))?1:0}%
    \ifdim\meet pt>0.5pt \ifdim\insd pt<0.5pt \gdef\hitflag{1}\fi\fi}}
\newcommand{\pipelevel}[4]{%
  \pgfmathsetmacro{\vs}{\panW/(#1*#2)}%
  \pgfmathtruncatemacro{\NX}{#1*#2}\pgfmathtruncatemacro{\NY}{\panH/\vs}%
  \foreach \i in {0,...,\numexpr\NX-1\relax}{\foreach \j in {0,...,\numexpr\NY-1\relax}{%
    \pgfmathsetmacro{\xa}{\i*\vs}\pgfmathsetmacro{\xb}{(\i+1)*\vs}%
    \pgfmathsetmacro{\ya}{\j*\vs}\pgfmathsetmacro{\yb}{(\j+1)*\vs}%
    \pipehit{\xa}{\ya}{\xb}{\yb}%
    \ifnum\hitflag=1 \fill[#3!48] (\xa,\ya) rectangle (\xb,\yb);
      \draw[#3!70, line width=0.12pt] (\xa,\ya) rectangle (\xb,\yb);\fi}}%
  \pipecells{#1}{#4}}
\newcommand{\pipecells}[2]{%
  \pgfmathsetmacro{\S}{\panW/#1}\pgfmathtruncatemacro{\M}{\panH/\S}%
  \foreach \i in {0,...,\numexpr#1-1\relax}{\foreach \j in {0,...,\numexpr\M-1\relax}{%
    \pgfmathsetmacro{\xa}{\i*\S}\pgfmathsetmacro{\xb}{(\i+1)*\S}%
    \pgfmathsetmacro{\ya}{\j*\S}\pgfmathsetmacro{\yb}{(\j+1)*\S}%
    \pipehit{\xa}{\ya}{\xb}{\yb}%
    \ifnum\hitflag=1 \draw[#2!75!black, line width=0.55pt] (\xa,\ya) rectangle (\xb,\yb);\fi}}%
}
\newcommand{\pipejrect}[5]{%
  \pgfmathsetmacro{\qxa}{#1+#5*(rnd-0.5)}\pgfmathsetmacro{\qya}{#2+#5*(rnd-0.5)}%
  \pgfmathsetmacro{\qxb}{#3+#5*(rnd-0.5)}\pgfmathsetmacro{\qyb}{#2+#5*(rnd-0.5)}%
  \pgfmathsetmacro{\qxc}{#3+#5*(rnd-0.5)}\pgfmathsetmacro{\qyc}{#4+#5*(rnd-0.5)}%
  \pgfmathsetmacro{\qxd}{#1+#5*(rnd-0.5)}\pgfmathsetmacro{\qyd}{#4+#5*(rnd-0.5)}%
  \draw (\qxa,\qya) -- (\qxb,\qyb) -- (\qxc,\qyc) -- (\qxd,\qyd) -- cycle;}
\newcommand{\pipeest}[1]{%
  \pgfmathsetseed{7}%
  \begin{scope}[black!80, line width=0.6pt, line join=round]
    \draw (0,0.26) \foreach \i in {1,...,6}{-- ({\i*\panW/6},{0.26+#1*(rnd-0.5)})};
    \pipejrect{0.22}{0.30}{1.10}{2.28}{#1}
    \pipejrect{1.55}{0.92}{3.30}{1.06}{#1}
    \pipejrect{1.82}{0.30}{2.00}{0.92}{#1}
    \pipejrect{2.85}{0.30}{3.03}{0.92}{#1}
    \pipejrect{1.95}{0.46}{2.90}{0.54}{#1}
  \end{scope}}
\newcommand{\pipemesh}{%
  \begin{scope}[black!80, line width=0.7pt, line join=round]
    \draw (0,0.26) -- (\panW,0.26);
    \foreach \oa/\ob/\oc/\od in \sceneobjs{\draw[fill=black!6] (\oa,\ob) rectangle (\oc,\od);}
  \end{scope}}
\newcommand{\pipepoints}{%
  \pgfmathsetseed{9}%
  \foreach \oa/\ob/\oc/\od in \sceneobjs{%
    \pgfmathsetmacro{\ww}{\oc-\oa}\pgfmathsetmacro{\hh}{\od-\ob}%
    \pgfmathtruncatemacro{\NP}{2*(\ww+\hh)*5+1}%
    \foreach \t in {1,...,\NP}{%
      \pgfmathsetmacro{\dd}{rnd*2*(\ww+\hh)}%
      \pgfmathsetmacro{\px}{(\dd<\ww)?(\oa+\dd):((\dd<\ww+\hh)?\oc:((\dd<2*\ww+\hh)?(\oc-(\dd-\ww-\hh)):\oa))}%
      \pgfmathsetmacro{\py}{(\dd<\ww)?\ob:((\dd<\ww+\hh)?(\ob+\dd-\ww):((\dd<2*\ww+\hh)?\od:(\od-(\dd-2*\ww-\hh))))}%
      \fill[black!65] (\px+0.14*rnd-0.07,\py+0.14*rnd-0.07) circle (0.035);}}%
}
\newcommand{\pipepyr}[6]{%
  \pgfmathsetmacro{\ph}{#3/1.544}%
  \node[anchor=north west, inner sep=0] at (#1,#2)
    {\includegraphics[width=#3cm]{imgs/teaser/input#6.jpg}};
  \begin{scope}
    \clip (#1,#2) rectangle ++(#3,-\ph);
    \foreach \i in {1,...,6}{%
      \draw[white, line width=0.4pt] (#1+\i*#4,#2) -- ++(0,-\ph);
      \draw[white, line width=0.4pt] (#1,#2-\i*#4) -- ++(#3,0);}
  \end{scope}
  \draw[#5, line width=0.6pt] (#1,#2) rectangle ++(#3,-\ph);}
\def\tS{0.34}
\newcommand{\pipeview}[6]{%
  \pipepyr{#1}{#2}{1.30}{\tS}{black!40}{#3}
  \pipepyr{#1+1.45}{#2}{0.87}{\tS}{black!40}{#3}
  \pipepyr{#1+2.47}{#2}{0.52}{\tS}{black!40}{#3}
  \pgfmathsetmacro{\lx}{#1+(#4==0?0:(#4==1?1.45:2.47))}%
  \draw[tsel, line width=0.9pt] (\lx+#5*\tS,#2-#6*\tS) rectangle ++(\tS,-\tS);}
\newcommand{\pipetoken}[7]{%
  \pgfmathsetmacro{\tw}{#4*#7/\tS}%
  \begin{scope}
    \clip (#1,#2) rectangle ++(#7,-#7);
    \node[anchor=north west, inner sep=0] at (#1-#5*#7,#2+#6*#7)
      {\includegraphics[width=\tw cm]{imgs/teaser/input#3.jpg}};
  \end{scope}
  \draw[tsel, line width=0.8pt] (#1,#2) rectangle ++(#7,-#7);}
\def\cellshape{0.24/0.352/1.60/0.48, 0.352/0.48/1.60/1.60}
\newcommand{\pipecell}[6]{%
  \pgfmathsetmacro{\gs}{#4/#3}%
  \begin{scope}[shift={(#1-#4/2,#2-#4/2)}]
    \fill[white] (0,0) rectangle (#4,#4);
    \foreach \i in {0,...,\numexpr#3-1\relax}{\foreach \j in {0,...,\numexpr#3-1\relax}{%
      \pgfmathsetmacro{\ua}{\i/#3}\pgfmathsetmacro{\ub}{(\i+1)/#3}%
      \pgfmathsetmacro{\va}{\j/#3}\pgfmathsetmacro{\vb}{(\j+1)/#3}%
      \gdef\hitflag{0}%
      \foreach \ra/\rb/\rc/\rd in \cellshape{%
        \pgfmathsetmacro{\meet}{((\ua<\rc)&&(\ub>\ra)&&(\va<\rd)&&(\vb>\rb))?1:0}%
        \pgfmathsetmacro{\insd}{((\ua>\ra)&&(\ub<\rc)&&(\va>\rb)&&(\vb<\rd))?1:0}%
        \ifdim\meet pt>0.5pt \ifdim\insd pt<0.5pt \gdef\hitflag{1}\fi\fi}%
      \ifnum\hitflag=1
        \pgfmathtruncatemacro{\ii}{max(0,min(#3-1,\i+round(#6*(2*rnd-1))))}%
        \pgfmathtruncatemacro{\jj}{max(0,min(#3-1,\j+round(#6*(2*rnd-1))))}%
        \fill[#5] (\ii*\gs,\jj*\gs) rectangle ++(\gs,\gs);\fi}}%
    \draw[black!15, line width=0.15pt] (0,0) grid[step=\gs] (#4,#4);
    \draw[black!55, line width=0.5pt] (0,0) rectangle (#4,#4);
  \end{scope}}

\begin{figure}[t]
  \centering
  \resizebox{\linewidth}{!}{%
  \begin{tikzpicture}[x=1cm,y=1cm]
    \pgfmathsetseed{11}
    \def\pS{0.66}
    \pgfmathsetmacro{\pw}{\pS*\panW}\pgfmathsetmacro{\ph}{\pS*\panH}
    \def\ry{4.25}                                  %
    \pgfmathsetmacro{\rm}{\ry+\ph/2}               %
    \def\xa{3.85}\def\xb{6.925}\def\xc{10.00}\def\xo{13.075}   %

    \node[pt] at (0,6.25) {(a)\;Inference-time coarse-to-fine cascade};
    \foreach \k/\d in {3/0, 2/0.12, 1/0.24}{
      \node[anchor=north west, inner sep=0, draw=white, line width=0.6pt]
        at (\d,\rm+0.58-\d) {\includegraphics[width=1.05cm]{imgs/teaser/input\k.jpg}};}
    \node[pl, anchor=north] at (0.62,\rm-0.40) {posed\\[-1pt]images};
    \draw[ar] (1.36,\rm) -- (1.66,\rm);
    \begin{scope}[shift={(1.72,\rm-0.375)}, scale=0.30]
      \fill[white] (0,0) rectangle (\panW,\panH);
      \pipepoints
      \draw[black!25, line width=0.4pt] (0,0) rectangle (\panW,\panH);
    \end{scope}
    \node[pl, anchor=north] at (2.28,\rm-0.42) {DA3 points\\[-1pt]{\scriptsize view selection}};
    \draw[ar] (2.93,\rm) -- (3.80,\rm);
    \node[pn, anchor=south, font=\tiny] at (3.36,\rm+0.03) {cell\\[-2pt]placement};
    \foreach \l/\bx/\nc/\col/\pcol/\s/\jit in {0/\xa/3/tblue/tblue/{s_0}/0,
        1/\xb/6/tgreen/tblue/{s_0/2}/0.10, 2/\xc/12/tred/tgreen/{s_0/4}/0.045}{
      \begin{scope}[shift={(\bx,\ry)}, scale=\pS]
        \fill[white] (0,0) rectangle (\panW,\panH);
        \ifnum\l=0 \pipecells{\nc}{\col}\else \pipelevel{\nc}{4}{\pcol}{\col}\pipeest{\jit}\fi
        \draw[black!25, line width=0.4pt] (0,0) rectangle (\panW,\panH);
      \end{scope}
      \pgfmathtruncatemacro{\lp}{\l-1}
      \node[pl, anchor=north] at (\bx+\pw/2,\ry-0.08)
        {\ifnum\l>0 \textbf{\color{\pcol!80!black}L\lp{} mesh} + \fi
         \textbf{\color{\col!80!black}L\l{} cells}\\[-1pt]{\scriptsize cell size $\s$}};}
    \foreach \x in {6.39,9.465}{\draw[bigar] (\x,\rm) -- ++(0.40,0);}
    \draw[ar] (12.54,\rm) -- (12.96,\rm);
    \begin{scope}[shift={(\xo,\ry)}, scale=\pS]
      \fill[white] (0,0) rectangle (\panW,\panH);
      \pipemesh
      \draw[black!25, line width=0.4pt] (0,0) rectangle (\panW,\panH);
    \end{scope}
    \node[pl, anchor=north] at (\xo+\pw/2,\ry-0.08) {\textbf{\color{tred!80!black}L2 mesh}};
    \pgfmathsetmacro{\cs}{\pS*\panW/6}
    \draw[tsel, line width=1.0pt] (\xb,\ry) rectangle ++(\cs,\cs);

    \draw[box] (0,0) rectangle (3.60,3.35);
    \node[pt] at (0.15,3.10) {(b)\;Multi-scale tile bank};
    \pipeview{0.27}{2.85}{1}{0}{1}{1}
    \pipeview{0.27}{1.93}{2}{1}{1}{0}
    \pipeview{0.27}{1.01}{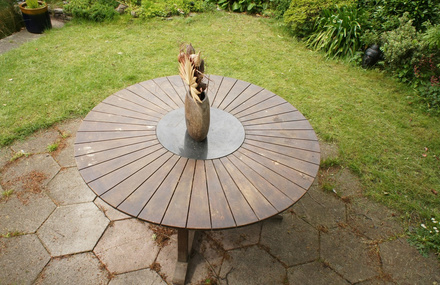}{2}{0}{0}

    \draw[box] (3.85,0) rectangle (15.55,3.35);
    \draw[ar] (2.28,\rm-1.03) -- (2.28,3.52) -- (4.30,3.52) -- (4.30,1.62);
    \node[pt, anchor=center] at (9.70,3.10) {(c)\;Cell generation at one level};
    \draw[line width=0.8pt, black, -{latex}] (3.60,1.55) -- (5.02,1.55);
    \node[pn, anchor=north] at (4.40,1.50) {select};
    \foreach \k/\w/\c/\r/\yy in {1/1.30/1/1/2.18, 2/0.87/1/0/1.73, 3/0.52/0/0/1.28}{\pipetoken{5.12}{\yy}{\k}{\w}{\c}{\r}{0.38}}
    \node[pn, anchor=north] at (5.31,0.84) {tile\\features};
    \def\gy{1.60}
    \pipecell{6.60}{\gy}{10}{1.05}{black!45}{1.2}
    \draw[tsel, line width=1.0pt] (6.075,\gy-0.525) rectangle ++(1.05,1.05);
    \draw[tsel, line width=0.7pt, densely dotted, -{latex}]
      (\xb,\ry+0.5*\cs) -- (6.60,\ry+0.5*\cs) -- (6.60,\gy+0.56);
    \node[pn, anchor=north west, align=left] at (6.30,\gy-0.58) {L0: pure noise\\else: noised parent};
    \node[flow] (ss) at (8.60,\gy) {SS\\[-2pt]generator};
    \pipecell{10.60}{\gy}{10}{1.05}{black!45}{0}
    \node[pn, anchor=south] at (10.60,\gy+0.55) {sparse structure};
    \node[flow] (sh) at (12.65,\gy) {mesh\\[-2pt]generator};
    \pipecell{14.70}{\gy}{20}{1.05}{black!30}{0}
    \begin{scope}[shift={(14.70-0.525,\gy-0.525)}, black!90, line width=0.8pt]
      \clip (0,0) rectangle (1.05,1.05);
      \foreach \ra/\rb/\rc/\rd in \cellshape{\draw (\ra*1.05,\rb*1.05) rectangle (\rc*1.05,\rd*1.05);}
    \end{scope}
    \draw[black!55, line width=0.5pt] (14.175,\gy-0.525) rectangle ++(1.05,1.05);
    \node[pn, anchor=south, align=center] at (14.70,\gy+0.55) {mesh on\\$2\times$ finer SS};
    \draw[ar] (7.15,\gy) -- (ss.west);  \draw[ar] (ss.east) -- (10.05,\gy);
    \draw[ar] (11.15,\gy) -- (sh.west); \draw[ar] (sh.east) -- (14.15,\gy);
    \fill[tred, opacity=0.18] (14.175,\gy-0.525) rectangle ++(0.525,0.525);
    \draw[tred!85!black, line width=1.0pt] (14.175,\gy-0.525) rectangle ++(0.525,0.525);
    \node[anchor=west, inner sep=0] at (14.53,\gy-0.85)
      {\includegraphics[height=0.50cm]{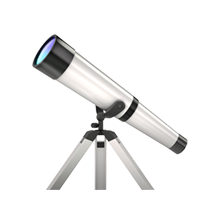}};
    \draw[ar, tred!85!black] (14.4375,\gy-0.525) -- (14.4375,0.20) -- (6.25,0.20) -- (6.25,\gy-0.56);
\node[pn, anchor=south, text=tred!85!black] at (10.60,0.22) {to the child cell at the next level};
    \draw[att] (5.31,2.20) -- (5.31,2.55) -- (12.65,2.55) -- (sh.north);
    \draw[att] (8.60,2.55) -- (ss.north);
    \node[pn, anchor=south] at (10.60,2.57) {keys and values for voxel-queried attention};
  \end{tikzpicture}}
  \caption{\textbf{Method overview.} L$n$ denotes level $n$. \textbf{(a)}~DA3 points place the L0
  cells and select their views. Each finer level halves the cell size and refines the mesh of the
  level above. \textbf{(b)}~Each image is encoded once as a tile pyramid, and from each view a cell
  takes the tiles at the pyramid level that matches its voxel spacing. \textbf{(c)}~All cells at all
  levels share the SS and mesh generators, which attend to these tiles. At finer levels, the SS
  stage starts from the noised parent (SDEdit).}
  \label{fig:pipeline}
\end{figure}

\subsection{Inference-Time Coarse-to-Fine Cascade}
\label{sec:method:cascade}

To overcome the limited context of single-level generation, we introduce a coarse-to-fine cascade. The coarsest level covers a large spatial extent and establishes the global structure, while each subsequent level refines the inherited geometry to recover finer surface detail. Each level halves the cell size, with
cell size at level $l$ given by $s_l=s_0/2^{l}$ for $l=0,\dots,L-1$
(Figure~\ref{fig:pipeline}(a)).

At every level, we derive the child sparse structure from the parent mesh decoder $D_{\mathrm{shape}}$. The decoder upsamples the parent shape latent through subdivision stages, each doubling the spatial resolution. We use the output at resolution $2R$, where a child cell with half the parent edge length corresponds to an $R^3$ block of voxels, and
take this block as the child's sparse structure. Because the parent geometry is estimated at a coarser resolution, subdivision alone leaves the surface coarse and misaligned with the finer child grid. This provides a noisy initialization, which we refine with the SS flow model at the child resolution (Figure~\ref{fig:pipeline}(c)).

To refine this noisy initialization, we first encode the inherited sparse structure into the SS latent, $\bar x_0^{(l+1)}
  =A_{\mathrm{SS}}\!\left(E_{\mathrm{SS}}\!\left(\hat O^{(l)}_{\mathrm{sub}}\right)\right),
$
where $\hat O^{(l)}_{\mathrm{sub}}$ is the $R^3$ block of the parent's subdivision occupancy
that covers the child cell,
$E_{\mathrm{SS}}$ is the SS encoder, and $A_{\mathrm{SS}}$ is the latent normalization used for flow matching.
We then perturb $\bar x_0^{(l+1)}$ to an intermediate time $t_s$ on the forward path,
$x_{t_s}=(1-t_s)\,\bar x_0^{(l+1)}+\sigma_{t_s}\epsilon, \quad \epsilon\sim\mathcal N(0,I),$ with $\sigma_t=\sigma_{\min}+(1-\sigma_{\min})t$, and integrate the SS flow back to $t=0$
with $K$ Euler steps over $t_s=\tau_K>\dots>\tau_0=0$,
\begin{equation}
  x_{\tau_{k-1}}=x_{\tau_k}-(\tau_k-\tau_{k-1})\,v_\theta\!\left(x_{\tau_k},\tau_k;\mathcal I\right),
  \quad k=K,\dots,1,
  \qquad \hat x_0^{(l+1)}=x_{\tau_0}.
  \label{eq:cascade-flow}
\end{equation}
This applies SDEdit~\citep{meng2022sdedit} between cascade levels using the deterministic flow sampler. Since the child is conditioned on finer-scale image features, the refinement recovers new surface detail while preserving global structure inherited from the parent.
The parameter $t_s$ controls the trade-off between preserving the parent geometry and incorporating finer image details available at the child level.
As $t_s\rightarrow0$, the child remains close to the parent. As
$t_s\rightarrow1$, it relies mainly on local high-resolution image features and loses the global context that the parent provides. The global context helps resolve ambiguities that cannot be determined from a single cell alone. To balance local detail and global context, we set $t_s=0.8$ empirically for all benchmarks, without
tuning per scene, dataset, or view count.

The parent also determines the child-cell layout by retaining only cells that intersect its predicted geometry, restricting generation to relevant regions.
To initialize the cells of the coarsest level and per-cell view selection, we use the point cloud from Depth Anything 3 (DA3)~\citep{lin2025da3}. We use the depth only for these two choices, and never fuse it into the output geometry (Appendix~\ref{app:da3-selection}).

\subsection{Multi-Scale Training of a Shared Model}
\label{sec:method:training}

Since the parent prediction is used only to initialize the next level at inference time via
Equation~\ref{eq:cascade-flow}, training does not require paired coarse and fine examples. We
train one SS flow and one mesh flow on individual cells, using standard conditional flow
matching~\citep{lipman2023flowmatching,liu2023rectifiedflow}. We map cells of several edge lengths $s$ to the same canonical coordinates and $R^3$ grid and condition the model on the images that observe each cell.
Cells at different scales therefore differ only in voxel spacing $\Delta=s/R$.
Neither $s$ nor $\Delta$ is given to the model, so the same weights apply across scales.
 Unlike XCube and SCube~\citep{ren2024xcube,ren2024scube}, our hierarchy is not fixed during training, so its depth, cell layout, and physical scales can be chosen at inference time without retraining. Appendix~\ref{app:training} provides more details about training.
\subsection{Cell-Independent Multi-Scale Image Conditioning}
\label{sec:method:attention}

In the MultiDiffusion formulation above, every cell is evaluated at every flow step of the SS and mesh stages, making feature reuse important for efficient inference. Pixel-aligned
conditioning~\citep{li2026pixal3d,schmid2026genrecon} projects each voxel into the input views,
bilinearly samples one feature per view, and aggregates them before the 3D generator. However, the DINOv3 token grid is coarse, so the sampled features can lose the information needed for detailed surface reconstruction. Pixal3D~\citep{li2026pixal3d} addresses this by upsampling features to
image resolution, but the upsampled maps become costly to cache as image resolution
and view count increase, and must then be recomputed at every flow step. Aggregating before
the generator also keeps the view weights from depending on the current voxel state. We
address these with voxel-queried image attention over compact cached
features and a cell-independent tile bank that provides each cell with scale-matched image features without per-cell encoding.

\textbf{Voxel-queried image attention.}
For a voxel $p$, we gather the DINO tokens within a $w\times w$ window around its projection in each view,
and let the current voxel feature $h_p$ attend to all of them with a single softmax over
tokens and views. Instead of bilinear sampling, we apply relative positional encoding to the continuous offset from the
voxel projection to the token center to preserve sub-token position
and local image context.
As the query is the current voxel feature, the attention weights adapt to the evolving 3D state across transformer blocks and flow steps. We also encode the projected voxel size and ray direction~\citep{li2025prope} (Appendix~\ref{app:relative-attention}).
For finer detail, the mesh stage also samples compact multi-stride ConvNeXt features~\citep{liu2022convnext}
at the same projected location. The DINO features remain
on their native token grid, while the ConvNeXt features are lightweight,
allowing both to be cached for all views and reused across cells and flow steps.

\textbf{Cell-independent tile bank.}
The base generator~\citep{xiang2025trellis2} is pretrained on $512\times512$ images
that frame the target region at a particular image scale. To match this pretraining setup, each
view should provide a crop that covers the cell projection, with pixel spacing matched to
the cell's voxel spacing. GenRecon~\citep{schmid2026genrecon} instead resizes each full image to a single
resolution. In wide scenes, where an image covers a large area, this can leave a fine
cell with only a few pixels.
Instead of center-cropping the images around each cell projection, which would require separate image
encoding for every cell--view pair and dominate the cost, we build an isotropic
pyramid for each input image and divide each level into a fixed grid of $512\times512$ tiles
(Figure~\ref{fig:pipeline}(b)). Each tile is defined by its source image, pyramid level, and
crop origin, independently of the 3D cells. For each view, a cell
selects the level that matches its voxel spacing and the tiles covering its projection, so fine
cells use the original image while coarser cells use downsampled levels. We treat each tile as a virtual posed view by incorporating its scale and crop origin into the
intrinsics, which keeps the input format of pretraining. At each level, we encode only the
selected tile union and reuse it across cells, the SS and mesh stages,
and all flow steps. The tile bank also benefits training, where the same tiles are reused across cell scales and 3D layouts.
Appendices~\ref{app:conditioning} and~\ref{app:ablations} give details and ablations.

\section{Experiments}
\label{sec:experiments}
\begin{figure}[t]
  \centering
  \setlength{\tabcolsep}{0.5pt}
  \renewcommand{\arraystretch}{1.0}
  \newlength{\qcA}\setlength{\qcA}{\dimexpr(\textwidth-5pt)/5\relax}
  \newlength{\qcR}\setlength{\qcR}{\dimexpr(\textwidth-2.5pt)/3\relax}
  \newlength{\qcH}\setlength{\qcH}{\dimexpr(0.5625\qcR-2pt)/2\relax}
  \newcommand{\qcLab}[1]{\makebox[\qcA][c]{\scriptsize #1}}
  \begin{tabular}{@{}ccccc@{}}
    \includegraphics[width=\qcA]{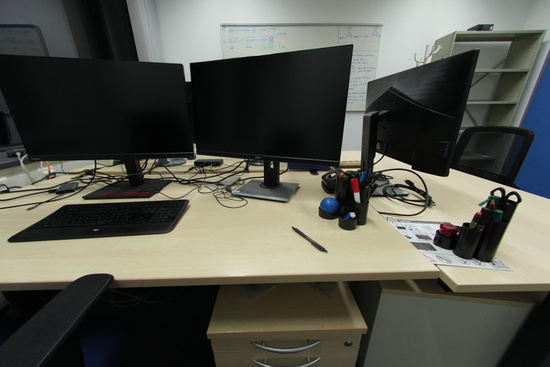} & \includegraphics[width=\qcA]{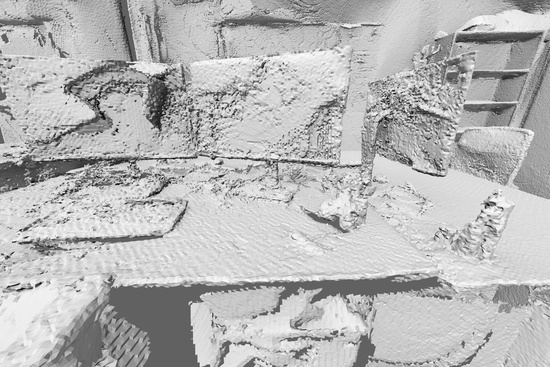} & \includegraphics[width=\qcA]{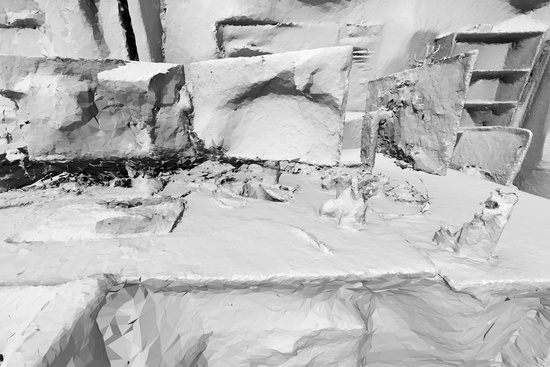} & \includegraphics[width=\qcA]{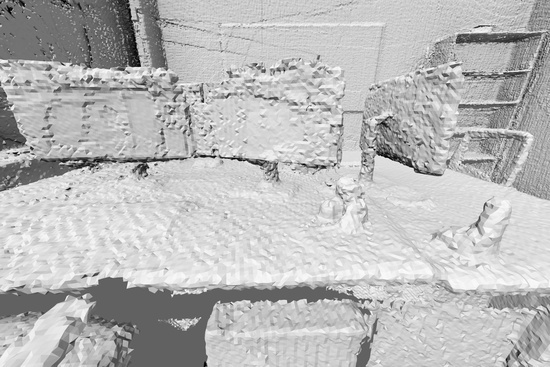} & \includegraphics[width=\qcA]{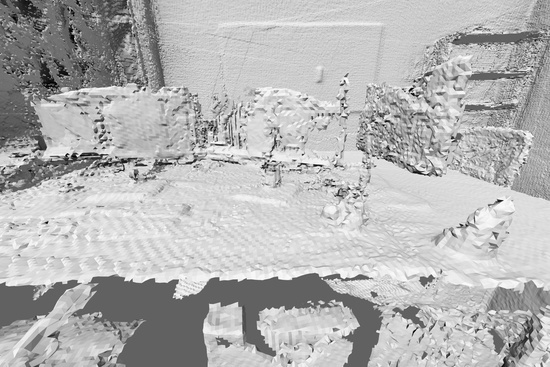} \\[-5pt]
    \qcLab{Image} & \qcLab{2DGS} & \qcLab{MILo} & \qcLab{DA3} & \qcLab{MapAnything} \\[0pt]
    \includegraphics[width=\qcA]{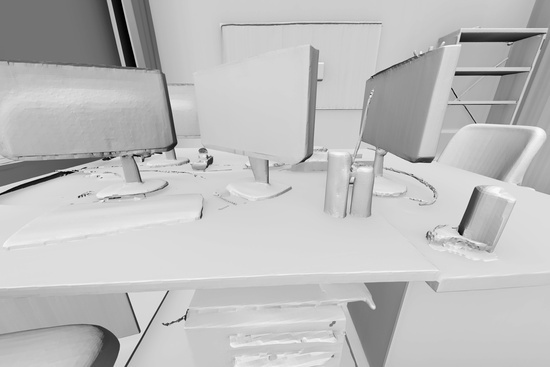} & \includegraphics[width=\qcA]{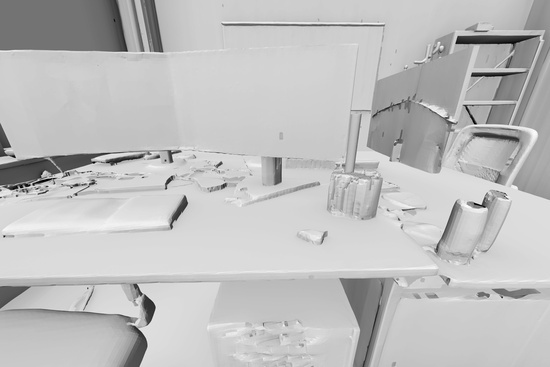} & \includegraphics[width=\qcA]{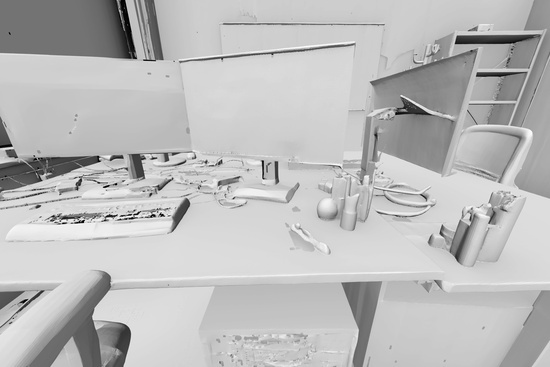} & \includegraphics[width=\qcA]{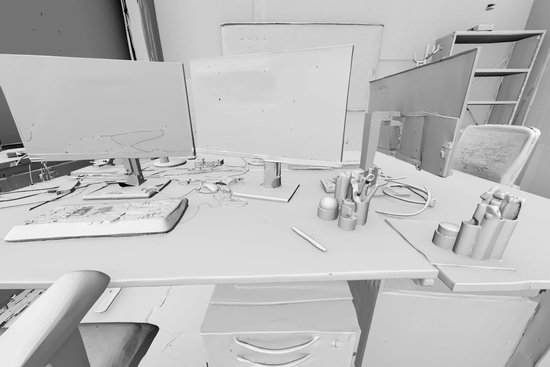} & \includegraphics[width=\qcA]{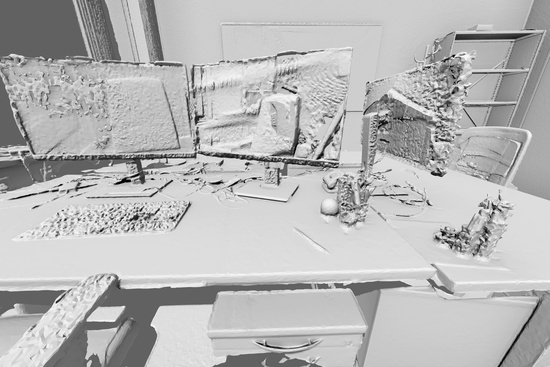} \\[-5pt]
    \qcLab{GenRecon} & \qcLab{Ours (L0)} & \qcLab{Ours (L1)} & \qcLab{Ours (L2)} & \qcLab{Ground truth} \\[-3pt]
    \multicolumn{5}{c}{\scriptsize (a) ScanNet++ \texttt{7b6477cb95}, 32 input views at $1752\times1168$ resolution} \\[1pt]
    \includegraphics[width=\qcA]{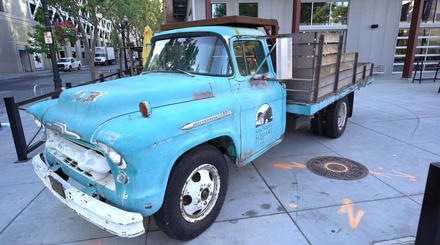} & \includegraphics[width=\qcA]{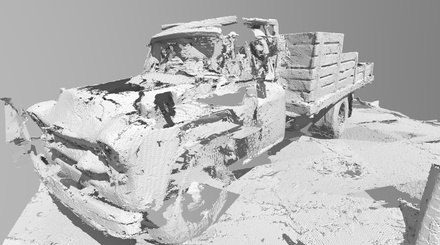} & \includegraphics[width=\qcA]{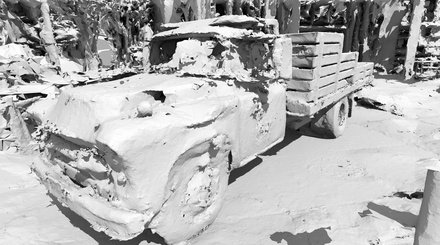} & \includegraphics[width=\qcA]{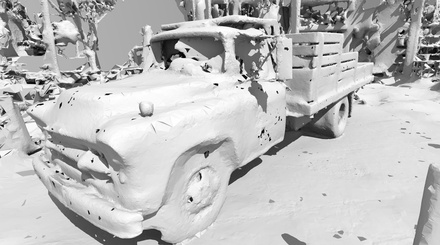} & \includegraphics[width=\qcA]{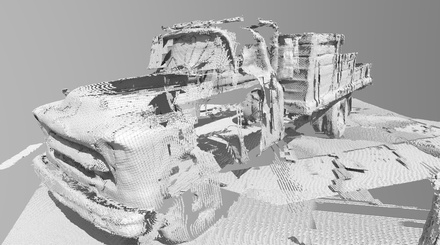} \\[-5pt]
    \qcLab{Image} & \qcLab{2DGS} & \qcLab{MILo} & \qcLab{GaussianWrapping} & \qcLab{DA3} \\[0pt]
    \includegraphics[width=\qcA]{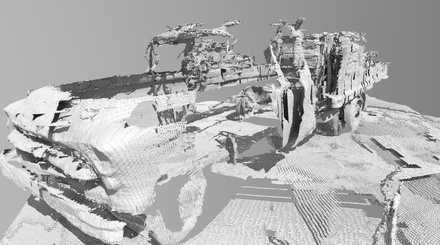} & \includegraphics[width=\qcA]{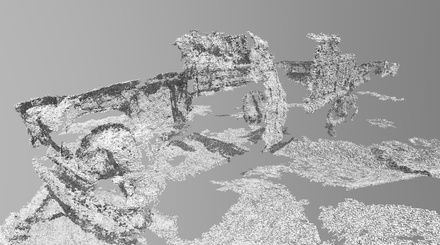} & \includegraphics[width=\qcA]{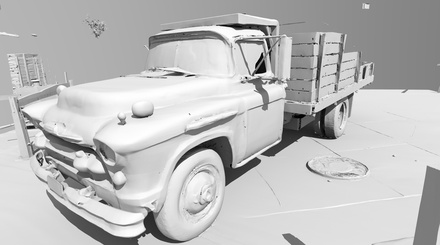} & \includegraphics[width=\qcA]{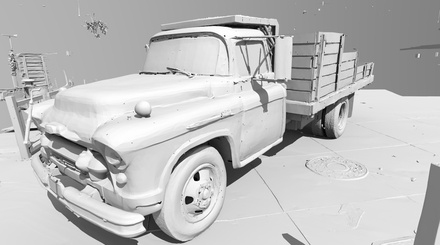} & \includegraphics[width=\qcA]{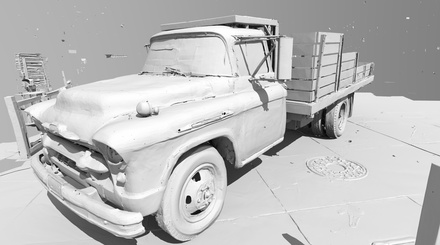} \\[-5pt]
    \qcLab{MapAnything} & \qcLab{Murre} & \qcLab{Ours (L0)} & \qcLab{Ours (L1)} & \qcLab{Ours (L2)} \\[-3pt]
    \multicolumn{5}{c}{\scriptsize (b) Tanks and Temples \texttt{truck}, 16 input views at $1958\times1091$ resolution} \\
  \end{tabular}\\[1pt]
  \begin{tabular}{@{}cccc@{}}
    \begin{tabular}[b]{@{}c@{}}\includegraphics[width=\qcR]{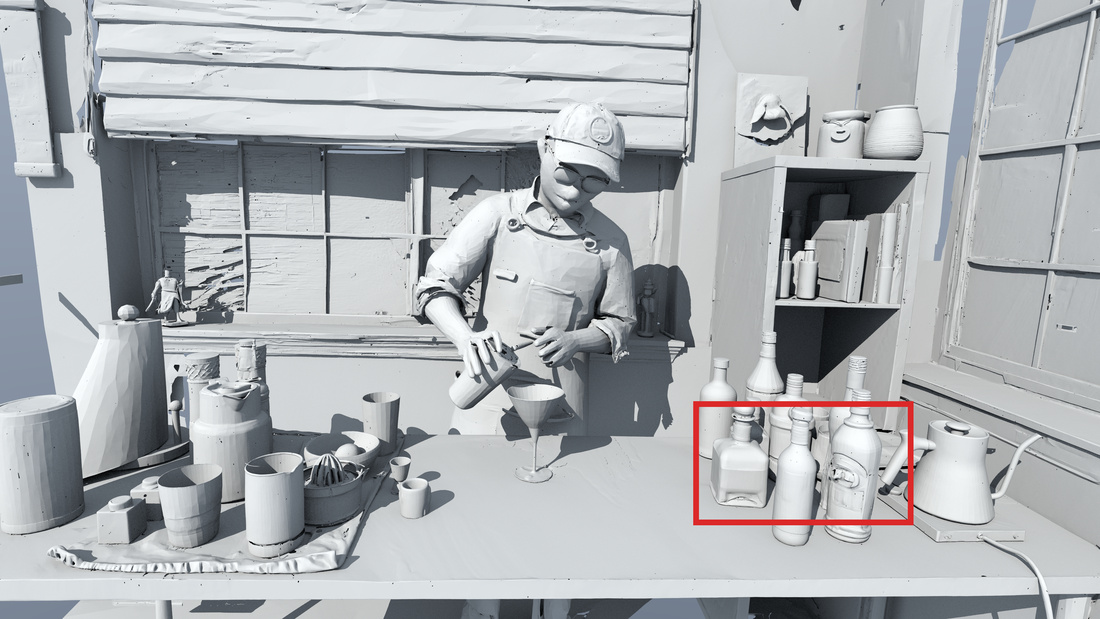}\end{tabular} &
    \begin{tabular}[b]{@{}c@{}}
      \includegraphics[height=\qcH]{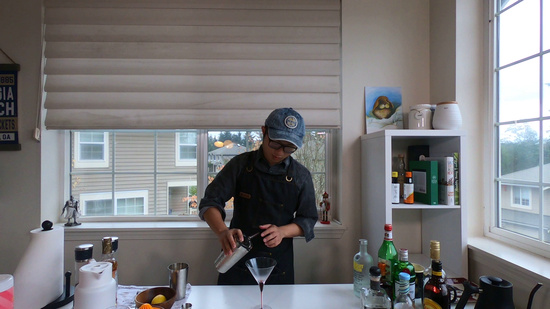} \\[1pt]
      \includegraphics[height=\qcH]{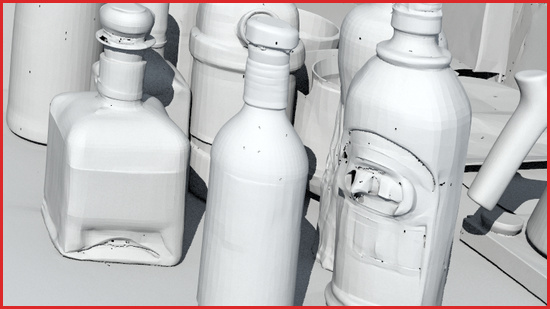}
    \end{tabular} &
    \begin{tabular}[b]{@{}c@{}}\includegraphics[width=\qcR]{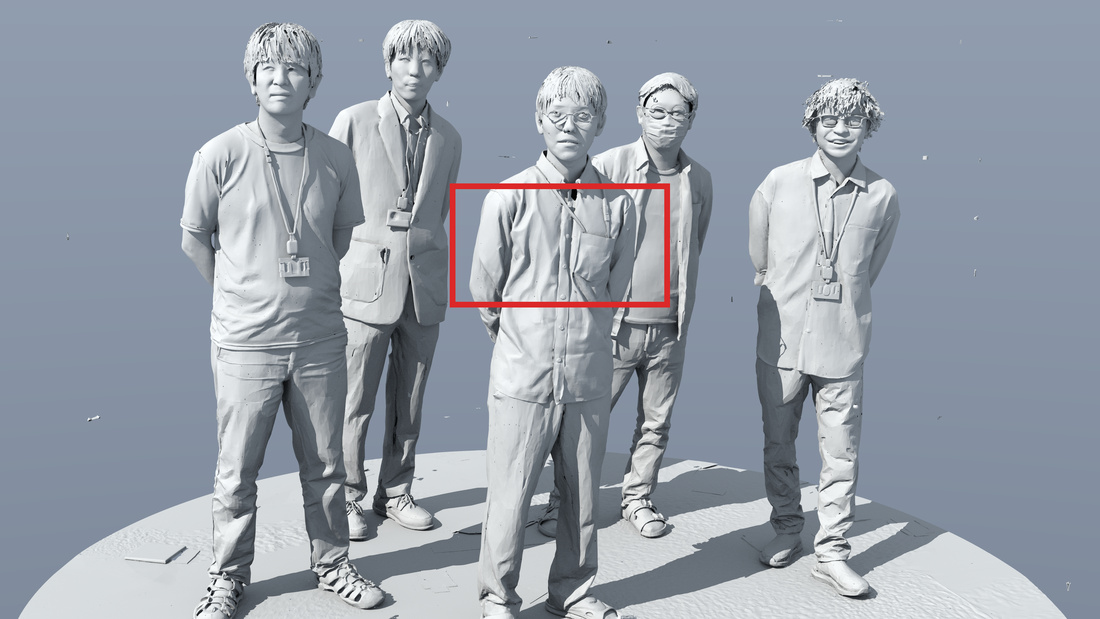}\end{tabular} &
    \begin{tabular}[b]{@{}c@{}}
      \includegraphics[height=\qcH]{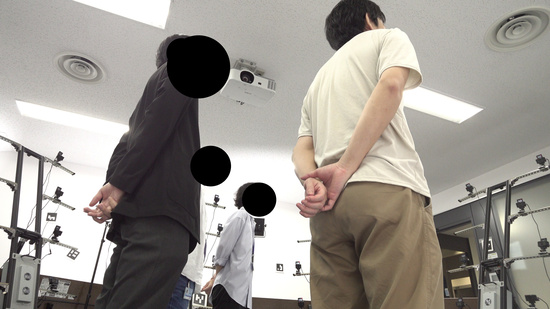} \\[1pt]
      \includegraphics[height=\qcH]{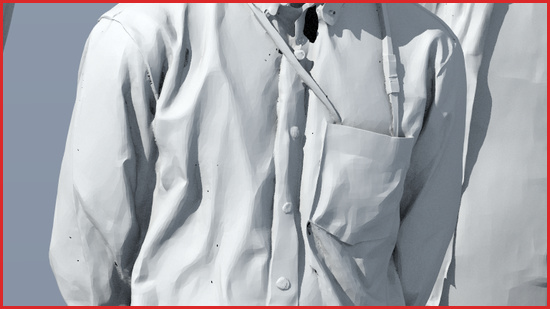}
    \end{tabular} \\[-5pt]
    \scriptsize \begin{tabular}[t]{@{}c@{}}\texttt{coffee\_martini} (N3DV)\\[-2pt] 18 views, $2704\times2028$ resolution\end{tabular} & \scriptsize input / close-up & \scriptsize \begin{tabular}[t]{@{}c@{}}in-house person capture\\[-2pt] 56 views, $3840\times2160$ resolution\end{tabular} & \scriptsize input / close-up \\[-3pt]
    \multicolumn{4}{c}{\scriptsize (c) Real-world captures (ours)} \\
  \end{tabular}
  \caption{\textbf{Qualitative results.} (a) ScanNet++ \texttt{7b6477cb95} and
  (b) Tanks and Temples \texttt{truck}.
  (c) Real-world captures (ours): mesh, one input view, and a close-up of the boxed
  region.}
  \label{fig:main-scannetpp}
\end{figure}

\subsection{Experimental Setup}
\label{sec:exp:setup}

\textbf{Models and training data.}
We train two models (Appendix~\ref{app:training}): a general scene reconstruction model and a city-block
reconstruction model. We initialize both from a pretrained
TRELLIS.2~\citep{xiang2025trellis2} model, first fine-tune it for multi-view
conditioning on TexVerse~\citep{zhang2025texverse}, and then fine-tune the two models
independently on their respective datasets. The general model is trained exclusively on synthetic data: the single-room
datasets SAGE-10K~\citep{sage10k} and 3D-FRONT~\citep{fu2021front3d}, together with our
procedurally generated multi-room houses and grass scenes. The
city-block model is fine-tuned on our dataset of photographs captured in Japanese cities
and their paired building information models (BIMs). The exterior geometry of the BIMs defines the
training target and contains no people or vehicles (Figure~\ref{fig:bim-qualitative}; Appendix~\ref{app:city-dataset}).

\textbf{Inference.}
In all experiments, each cell has an SS grid with $R=64$,
and the mesh decoder produces meshes at $512^3$ resolution. We apply margin-based chunk decoding
to obtain a complete scene-wide mesh (Appendix~\ref{app:halo-stitching}).
We place the cells at a stride of $s/2$ at every level. We use 12 flow-matching steps in the SS and
mesh stages, with classifier-free guidance~\citep{ho2022classifierfree} scales of 7.5 and 4.0, respectively.
At finer levels, the SS stage starts from the parent at $t_s=0.8$ and runs $K=10$ steps.
Each cell uses at most 64 conditioning image tiles. We use three cascade levels, numbered from level 0 (coarsest) to level 2 (finest), and four on the city blocks.

\textbf{Datasets.}
We evaluate on ScanNet++~\citep{yeshwanth2023scannetpp}, six scenes from Tanks and Temples~\citep{knapitsch2017tanks}, and our city-block dataset.
For ScanNet++, we use the 25 validation scenes with the largest ground-truth bounding-box volumes, with 32 or 256 input views selected by farthest-point sampling.
For Tanks and Temples, we use 16 or 256 input views of the 2K images. Appendix~\ref{app:tnt-results} provides results with 32 views and 4K inputs. We use all views for scenes with fewer than 256.
For the city blocks, we use three held-out blocks with 256 input views.
We also show \texttt{garden} from Mip-NeRF~360~\citep{barron2022mipnerf360} (Figure~\ref{fig:teaser}), and \texttt{coffee\_martini} from N3DV~\citep{li2022n3dv} and an in-house capture of five people (Figure~\ref{fig:main-scannetpp}(c)).

\textbf{Baselines.}
Our baselines include per-scene optimization methods (2DGS~\citep{huang20242dgs}, MILo~\citep{guedon2025milo}, GaussianWrapping~\citep{gomez2026gaussianwrapping}, and CityGaussianV2~\citep{liu2025citygaussianv2}), feed-forward methods (Depth Anything~3 (DA3)~\citep{lin2025da3}, MapAnything~\citep{keetha2026mapanything}, and Murre~\citep{guo2025murre}), and the generative method GenRecon~\citep{schmid2026genrecon}.
We evaluate all but CityGaussianV2 on ScanNet++, also drop GenRecon on Tanks and Temples, and use 2DGS, CityGaussianV2, DA3, and Murre on the city blocks.
All but MILo, GaussianWrapping, and GenRecon are meshed by a shared TSDF fusion (Appendix~\ref{app:baselines}).

\begin{table}[t]
  \centering
  \caption{\textbf{Quantitative comparison on ScanNet++.} Mean over the 25 largest scenes. 2D metrics compare depth and normals rendered at held-out views, and 3D metrics compare the mesh surfaces visible from those views. L0 to L2 are our cascade levels. The top three results are highlighted.}
  \label{tab:main-scannetpp}
  \scriptsize
  \setlength{\tabcolsep}{3pt}
  \definecolor{rank1}{HTML}{FFB3B3}
  \definecolor{rank2}{HTML}{FFD9B3}
  \definecolor{rank3}{HTML}{FFFFB3}
  \definecolor{emphrow}{HTML}{EDEDED}
  \ifdefined\paperautocellwidth\else\newlength{\paperautocellwidth}\fi
  \ifdefined\paperautoprobe\else\newlength{\paperautoprobe}\fi
  \settowidth{\paperautocellwidth}{\textbf{0.0000}}
  \settowidth{\paperautoprobe}{32 Views}
  \ifdim\dimexpr\paperautoprobe-12\tabcolsep\relax>7\paperautocellwidth  \setlength{\paperautocellwidth}{\dimexpr(\paperautoprobe-12\tabcolsep)/7\relax}\fi
  \settowidth{\paperautoprobe}{256 Views}
  \ifdim\dimexpr\paperautoprobe-12\tabcolsep\relax>7\paperautocellwidth  \setlength{\paperautocellwidth}{\dimexpr(\paperautoprobe-12\tabcolsep)/7\relax}\fi
  \settowidth{\paperautoprobe}{2D}
  \ifdim\dimexpr\paperautoprobe-6\tabcolsep\relax>4\paperautocellwidth  \setlength{\paperautocellwidth}{\dimexpr(\paperautoprobe-6\tabcolsep)/4\relax}\fi
  \settowidth{\paperautoprobe}{3D}
  \ifdim\dimexpr\paperautoprobe-4\tabcolsep\relax>3\paperautocellwidth  \setlength{\paperautocellwidth}{\dimexpr(\paperautoprobe-4\tabcolsep)/3\relax}\fi
  \settowidth{\paperautoprobe}{MAE$\downarrow$}
  \ifdim\dimexpr\paperautoprobe-0\tabcolsep\relax>1\paperautocellwidth  \setlength{\paperautocellwidth}{\dimexpr(\paperautoprobe-0\tabcolsep)/1\relax}\fi
  \settowidth{\paperautoprobe}{AbsRel$\downarrow$}
  \ifdim\dimexpr\paperautoprobe-0\tabcolsep\relax>1\paperautocellwidth  \setlength{\paperautocellwidth}{\dimexpr(\paperautoprobe-0\tabcolsep)/1\relax}\fi
  \settowidth{\paperautoprobe}{Ang$\downarrow$}
  \ifdim\dimexpr\paperautoprobe-0\tabcolsep\relax>1\paperautocellwidth  \setlength{\paperautocellwidth}{\dimexpr(\paperautoprobe-0\tabcolsep)/1\relax}\fi
  \settowidth{\paperautoprobe}{Comp$\uparrow$}
  \ifdim\dimexpr\paperautoprobe-0\tabcolsep\relax>1\paperautocellwidth  \setlength{\paperautocellwidth}{\dimexpr(\paperautoprobe-0\tabcolsep)/1\relax}\fi
  \settowidth{\paperautoprobe}{CD$\downarrow$}
  \ifdim\dimexpr\paperautoprobe-0\tabcolsep\relax>1\paperautocellwidth  \setlength{\paperautocellwidth}{\dimexpr(\paperautoprobe-0\tabcolsep)/1\relax}\fi
  \settowidth{\paperautoprobe}{F@2$\uparrow$}
  \ifdim\dimexpr\paperautoprobe-0\tabcolsep\relax>1\paperautocellwidth  \setlength{\paperautocellwidth}{\dimexpr(\paperautoprobe-0\tabcolsep)/1\relax}\fi
  \settowidth{\paperautoprobe}{NC$\uparrow$}
  \ifdim\dimexpr\paperautoprobe-0\tabcolsep\relax>1\paperautocellwidth  \setlength{\paperautocellwidth}{\dimexpr(\paperautoprobe-0\tabcolsep)/1\relax}\fi
  \providecommand{\paperautospan}[2]{\makebox[\dimexpr#1\paperautocellwidth+#1\tabcolsep+#1\tabcolsep-\tabcolsep-\tabcolsep\relax][c]{#2}}
  \providecommand{\paperautocell}[1]{\paperautospan{1}{#1}}
  \resizebox{\linewidth}{!}{%
  \begin{tabular}{lcccccccccccccc}
    \toprule
     & \multicolumn{7}{c}{\paperautospan{7}{32 Views}} & \multicolumn{7}{c}{\paperautospan{7}{256 Views}} \\
    \cmidrule(lr){2-8}\cmidrule(lr){9-15}
     & \multicolumn{4}{c}{\paperautospan{4}{2D}} & \multicolumn{3}{c}{\paperautospan{3}{3D}} & \multicolumn{4}{c}{\paperautospan{4}{2D}} & \multicolumn{3}{c}{\paperautospan{3}{3D}} \\
    \cmidrule(lr){2-5}\cmidrule(lr){6-8}\cmidrule(lr){9-12}\cmidrule(lr){13-15}
    Method & \multicolumn{1}{c}{\paperautospan{1}{MAE$\downarrow$}} & \multicolumn{1}{c}{\paperautospan{1}{AbsRel$\downarrow$}} & \multicolumn{1}{c}{\paperautospan{1}{Ang$\downarrow$}} & \multicolumn{1}{c}{\paperautospan{1}{Comp$\uparrow$}} & \multicolumn{1}{c}{\paperautospan{1}{CD$\downarrow$}} & \multicolumn{1}{c}{\paperautospan{1}{F@2$\uparrow$}} & \multicolumn{1}{c}{\paperautospan{1}{NC$\uparrow$}} & \multicolumn{1}{c}{\paperautospan{1}{MAE$\downarrow$}} & \multicolumn{1}{c}{\paperautospan{1}{AbsRel$\downarrow$}} & \multicolumn{1}{c}{\paperautospan{1}{Ang$\downarrow$}} & \multicolumn{1}{c}{\paperautospan{1}{Comp$\uparrow$}} & \multicolumn{1}{c}{\paperautospan{1}{CD$\downarrow$}} & \multicolumn{1}{c}{\paperautospan{1}{F@2$\uparrow$}} & \multicolumn{1}{c}{\paperautospan{1}{NC$\uparrow$}} \\
    \midrule
    \multicolumn{15}{l}{\textbf{\textit{w/ per-scene optimization}}} \\
    2DGS & \paperautocell{0.1426} & \paperautocell{0.1141} & \paperautocell{38.376} & \paperautocell{0.9058} & \paperautocell{0.0723} & \paperautocell{0.4837} & \paperautocell{0.7181} & \paperautocell{0.0929} & \paperautocell{0.0846} & \paperautocell{32.208} & \paperautocell{0.9747} & \paperautocell{0.0513} & \paperautocell{0.5345} & \paperautocell{0.8006} \\
    MILo & \paperautocell{0.2523} & \paperautocell{0.1780} & \paperautocell{42.156} & \paperautocell{0.9848} & \paperautocell{0.1127} & \paperautocell{0.4241} & \paperautocell{0.6910} & \paperautocell{0.1406} & \paperautocell{0.1226} & \paperautocell{33.612} & \cellcolor{rank1}\paperautocell{0.9998} & \paperautocell{0.0817} & \paperautocell{0.5508} & \paperautocell{0.7985} \\
    GaussianWrapping & \paperautocell{0.3467} & \paperautocell{0.2762} & \paperautocell{35.974} & \paperautocell{0.9648} & \paperautocell{0.2830} & \paperautocell{0.4413} & \paperautocell{0.7505} & \paperautocell{0.3197} & \paperautocell{0.3071} & \paperautocell{30.118} & \paperautocell{0.9761} & \paperautocell{0.2136} & \paperautocell{0.6075} & \paperautocell{0.8113} \\
    \midrule
    \multicolumn{15}{l}{\textbf{\textit{w/o per-scene optimization}}} \\
    DA3 & \paperautocell{0.0811} & \paperautocell{0.0800} & \paperautocell{28.000} & \paperautocell{0.9682} & \cellcolor{rank2}\paperautocell{0.0212} & \cellcolor{rank1}\paperautocell{0.7765} & \paperautocell{0.8658} & \paperautocell{0.0818} & \paperautocell{0.0792} & \paperautocell{20.789} & \cellcolor{rank3}\paperautocell{0.9958} & \cellcolor{rank1}\paperautocell{0.0193} & \cellcolor{rank1}\paperautocell{0.7882} & \cellcolor{rank3}\paperautocell{0.9128} \\
    MapAnything & \paperautocell{0.1019} & \paperautocell{0.1213} & \paperautocell{30.090} & \paperautocell{0.9275} & \paperautocell{0.0249} & \paperautocell{0.6716} & \paperautocell{0.8458} & \paperautocell{0.0907} & \paperautocell{0.0969} & \paperautocell{23.982} & \paperautocell{0.9905} & \paperautocell{0.0255} & \paperautocell{0.5761} & \paperautocell{0.8916} \\
    Murre & \paperautocell{0.1990} & \paperautocell{0.2041} & \paperautocell{40.515} & \paperautocell{0.9431} & \paperautocell{0.0638} & \paperautocell{0.3867} & \paperautocell{0.7180} & \paperautocell{0.2759} & \paperautocell{0.2792} & \paperautocell{40.780} & \paperautocell{0.9803} & \paperautocell{0.0916} & \paperautocell{0.3471} & \paperautocell{0.6952} \\
    GenRecon & \paperautocell{0.1175} & \paperautocell{0.1228} & \paperautocell{20.928} & \paperautocell{0.9764} & \paperautocell{0.0604} & \paperautocell{0.2846} & \paperautocell{0.8652} & \paperautocell{0.1049} & \paperautocell{0.1099} & \paperautocell{20.030} & \paperautocell{0.9782} & \paperautocell{0.0611} & \paperautocell{0.2834} & \paperautocell{0.8723} \\
    \cellcolor{emphrow}\textbf{Ours (L0)} & \cellcolor{rank3}\paperautocell{0.0714} & \cellcolor{rank3}\paperautocell{0.0746} & \cellcolor{rank3}\paperautocell{19.207} & \cellcolor{rank1}\paperautocell{0.9984} & \paperautocell{0.0235} & \paperautocell{0.6336} & \cellcolor{rank3}\paperautocell{0.9090} & \cellcolor{rank3}\paperautocell{0.0683} & \cellcolor{rank3}\paperautocell{0.0677} & \cellcolor{rank3}\paperautocell{18.952} & \cellcolor{rank2}\paperautocell{0.9990} & \paperautocell{0.0233} & \paperautocell{0.6251} & \paperautocell{0.9125} \\
    \cellcolor{emphrow}\textbf{Ours (L1)} & \cellcolor{rank2}\paperautocell{0.0569} & \cellcolor{rank2}\paperautocell{0.0575} & \cellcolor{rank2}\paperautocell{17.098} & \cellcolor{rank2}\paperautocell{0.9983} & \cellcolor{rank1}\paperautocell{0.0199} & \cellcolor{rank3}\paperautocell{0.6959} & \cellcolor{rank2}\paperautocell{0.9232} & \cellcolor{rank2}\paperautocell{0.0571} & \cellcolor{rank2}\paperautocell{0.0535} & \cellcolor{rank2}\paperautocell{16.684} & \paperautocell{0.9892} & \cellcolor{rank3}\paperautocell{0.0217} & \cellcolor{rank3}\paperautocell{0.6874} & \cellcolor{rank2}\paperautocell{0.9224} \\
    \cellcolor{emphrow}\textbf{Ours (L2)} & \cellcolor{rank1}\paperautocell{0.0565} & \cellcolor{rank1}\paperautocell{0.0536} & \cellcolor{rank1}\paperautocell{16.330} & \cellcolor{rank3}\paperautocell{0.9962} & \cellcolor{rank2}\paperautocell{0.0212} & \cellcolor{rank2}\paperautocell{0.7155} & \cellcolor{rank1}\paperautocell{0.9247} & \cellcolor{rank1}\paperautocell{0.0557} & \cellcolor{rank1}\paperautocell{0.0499} & \cellcolor{rank1}\paperautocell{15.624} & \paperautocell{0.9902} & \cellcolor{rank2}\paperautocell{0.0202} & \cellcolor{rank2}\paperautocell{0.7176} & \cellcolor{rank1}\paperautocell{0.9283} \\
    \bottomrule
  \end{tabular}%
  }
\end{table}

\textbf{Metrics.}
We report Chamfer distance (CD) in meters, F-score, and normal consistency (NC). F@$\tau$ denotes the F-score at a threshold of $\tau$~cm. We report F@2 on ScanNet++ and F@10 on the city blocks. For Tanks and Temples, we use the official per-scene threshold $\tau$, which ranges from 3 to 25~mm. Because scene scale varies widely across its scenes, we also divide CD by $\tau$ before averaging and report CD$/\tau$.
ScanNet++ and Tanks and Temples are evaluated using the virtual-scan protocol of GaussianWrapping~\citep{gomez2026gaussianwrapping}, which scores only surfaces seen from the cameras of the capture. For ScanNet++, we also report the held-out-view 2D metrics used by GenRecon~\citep{schmid2026genrecon}, including depth MAE and AbsRel, normal angular error, and completeness. For the city blocks, the BIM meshes model surfaces unseen from the street, such as roofs, so we evaluate whole blocks, cropping only the prediction to the ground-truth footprint. See Appendix~\ref{app:evaluation} for details.

\subsection{Comparison with Prior Reconstruction Methods}
\label{sec:exp:comparison}

\textbf{ScanNet++.}
Table~\ref{tab:main-scannetpp} reports results on the 25 largest ScanNet++ scenes with
32 and 256 input views. Our finest level outperforms GenRecon, the closest generative
baseline, on all seven metrics in both settings. DA3 obtains a higher F@2 in both
settings and a slightly lower Chamfer distance and higher completeness with 256 views,
but its depth and normal errors are larger than ours. In
Figure~\ref{fig:main-scannetpp}(a), DA3 produces bumpy surfaces and misses small objects
on the desk, while our method recovers flat surfaces, the keyboard, and pens.
Our accuracy changes little between 32 and 256 views, whereas all optimization-based
methods degrade with 32 views (e.g., F@2 of 2DGS drops from 0.53 to 0.48). Without
per-scene optimization, our method outperforms all optimization-based baselines on all three 3D metrics.
Appendix~\ref{app:scannetpp-results} gives further results on all 50 validation scenes.

\textbf{Tanks and Temples.}
Table~\ref{tab:main}(a) reports the mean over three small and three large scenes. Optimization-based methods achieve higher F-scores with dense observations. Still, our method achieves the lowest or a competitive Chamfer distance in all settings and the highest F-score on the large scenes with 16 views. Since our method does
not fit the input images pixel by pixel, its surfaces can be slightly
offset from the ground truth, which the tight thresholds of 3 to 25\,mm penalize. Our meshes are nevertheless more complete and detailed
(Figures~\ref{fig:main-scannetpp}(b) and~\ref{fig:baseline-comparison-tnt}). None of the baselines reconstructs the
transparent, reflective windows of \texttt{truck} as closed surfaces, while our
generative prior completes them. Appendix~\ref{app:tnt-results}
gives per-scene results, results with 4K input images, and further qualitative comparisons.

\begin{table*}[t]
  \centering
  \caption{\textbf{Quantitative comparison on Tanks and Temples and the city-block dataset.} (a) Mean over the three small (Caterpillar, Ignatius, Truck) and three large (Barn, Courthouse, Meetingroom) scenes with 2K input images. F@$\tau$ and CD$/\tau$ use the per-scene threshold $\tau$. (b) Mean over three blocks, evaluated against the BIM ground truth. L$x$ denotes level $x$, with its cell size in parentheses. The top three results are highlighted.}
  \label{tab:main}
  \begin{minipage}[b]{0.64\textwidth}
    \centering
  \scriptsize
  \setlength{\tabcolsep}{3pt}
  \definecolor{rank1}{HTML}{FFB3B3}
  \definecolor{rank2}{HTML}{FFD9B3}
  \definecolor{rank3}{HTML}{FFFFB3}
  \definecolor{emphrow}{HTML}{EDEDED}
  \ifdefined\paperautocellwidth\else\newlength{\paperautocellwidth}\fi
  \ifdefined\paperautoprobe\else\newlength{\paperautoprobe}\fi
  \settowidth{\paperautocellwidth}{\textbf{000.00}}
  \settowidth{\paperautoprobe}{Small, 16 Views}
  \ifdim\dimexpr\paperautoprobe-2\tabcolsep\relax>2\paperautocellwidth  \setlength{\paperautocellwidth}{\dimexpr(\paperautoprobe-2\tabcolsep)/2\relax}\fi
  \settowidth{\paperautoprobe}{Small, 256 Views}
  \ifdim\dimexpr\paperautoprobe-2\tabcolsep\relax>2\paperautocellwidth  \setlength{\paperautocellwidth}{\dimexpr(\paperautoprobe-2\tabcolsep)/2\relax}\fi
  \settowidth{\paperautoprobe}{Large, 16 Views}
  \ifdim\dimexpr\paperautoprobe-2\tabcolsep\relax>2\paperautocellwidth  \setlength{\paperautocellwidth}{\dimexpr(\paperautoprobe-2\tabcolsep)/2\relax}\fi
  \settowidth{\paperautoprobe}{Large, 256 Views}
  \ifdim\dimexpr\paperautoprobe-2\tabcolsep\relax>2\paperautocellwidth  \setlength{\paperautocellwidth}{\dimexpr(\paperautoprobe-2\tabcolsep)/2\relax}\fi
  \settowidth{\paperautoprobe}{F@$\tau$$\uparrow$}
  \ifdim\dimexpr\paperautoprobe-0\tabcolsep\relax>1\paperautocellwidth  \setlength{\paperautocellwidth}{\dimexpr(\paperautoprobe-0\tabcolsep)/1\relax}\fi
  \settowidth{\paperautoprobe}{CD$/\tau$$\downarrow$}
  \ifdim\dimexpr\paperautoprobe-0\tabcolsep\relax>1\paperautocellwidth  \setlength{\paperautocellwidth}{\dimexpr(\paperautoprobe-0\tabcolsep)/1\relax}\fi
  \providecommand{\paperautospan}[2]{\makebox[\dimexpr#1\paperautocellwidth+#1\tabcolsep+#1\tabcolsep-\tabcolsep-\tabcolsep\relax][c]{#2}}
  \providecommand{\paperautocell}[1]{\paperautospan{1}{#1}}
  \resizebox{\linewidth}{!}{%
  \begin{tabular}{lcccccccc}
    \toprule
     & \multicolumn{2}{c}{\paperautospan{2}{Small, 16 Views}} & \multicolumn{2}{c}{\paperautospan{2}{Small, 256 Views}} & \multicolumn{2}{c}{\paperautospan{2}{Large, 16 Views}} & \multicolumn{2}{c}{\paperautospan{2}{Large, 256 Views}} \\
    \cmidrule(lr){2-3}\cmidrule(lr){4-5}\cmidrule(lr){6-7}\cmidrule(lr){8-9}
    Method & \multicolumn{1}{c}{\paperautospan{1}{F@$\tau$$\uparrow$}} & \multicolumn{1}{c}{\paperautospan{1}{CD$/\tau$$\downarrow$}} & \multicolumn{1}{c}{\paperautospan{1}{F@$\tau$$\uparrow$}} & \multicolumn{1}{c}{\paperautospan{1}{CD$/\tau$$\downarrow$}} & \multicolumn{1}{c}{\paperautospan{1}{F@$\tau$$\uparrow$}} & \multicolumn{1}{c}{\paperautospan{1}{CD$/\tau$$\downarrow$}} & \multicolumn{1}{c}{\paperautospan{1}{F@$\tau$$\uparrow$}} & \multicolumn{1}{c}{\paperautospan{1}{CD$/\tau$$\downarrow$}} \\
    \midrule
    \multicolumn{9}{l}{\textbf{\textit{w/ per-scene optimization}}} \\
    2DGS & \cellcolor{rank1}\paperautocell{0.355} & \paperautocell{9.42} & \cellcolor{rank3}\paperautocell{0.448} & \paperautocell{9.64} & \paperautocell{0.101} & \paperautocell{44.95} & \paperautocell{0.233} & \paperautocell{28.08} \\
    MILo & \cellcolor{rank2}\paperautocell{0.346} & \paperautocell{14.79} & \cellcolor{rank2}\paperautocell{0.527} & \paperautocell{11.63} & \paperautocell{0.053} & \paperautocell{54.04} & \cellcolor{rank2}\paperautocell{0.268} & \paperautocell{51.29} \\
    GaussianWrapping & \cellcolor{rank3}\paperautocell{0.345} & \paperautocell{8.80} & \cellcolor{rank1}\paperautocell{0.650} & \cellcolor{rank1}\paperautocell{5.54} & \paperautocell{0.094} & \paperautocell{37.61} & \cellcolor{rank1}\paperautocell{0.370} & \paperautocell{42.51} \\
    \midrule
    \multicolumn{9}{l}{\textbf{\textit{w/o per-scene optimization}}} \\
    DA3 & \paperautocell{0.148} & \paperautocell{12.48} & \paperautocell{0.199} & \paperautocell{9.25} & \paperautocell{0.114} & \paperautocell{35.34} & \paperautocell{0.180} & \paperautocell{24.26} \\
    MapAnything & \paperautocell{0.071} & \paperautocell{19.23} & \paperautocell{0.077} & \paperautocell{27.90} & \paperautocell{0.014} & \paperautocell{194.69} & \paperautocell{0.039} & \paperautocell{86.44} \\
    Murre & \paperautocell{0.051} & \paperautocell{47.77} & \paperautocell{0.058} & \paperautocell{31.85} & \paperautocell{0.051} & \paperautocell{42.13} & \paperautocell{0.060} & \paperautocell{49.72} \\
    \cellcolor{emphrow}\textbf{Ours (L0)} & \paperautocell{0.219} & \cellcolor{rank3}\paperautocell{8.19} & \paperautocell{0.249} & \paperautocell{8.04} & \cellcolor{rank3}\paperautocell{0.175} & \cellcolor{rank3}\paperautocell{15.71} & \paperautocell{0.232} & \cellcolor{rank3}\paperautocell{7.08} \\
    \cellcolor{emphrow}\textbf{Ours (L1)} & \paperautocell{0.278} & \cellcolor{rank2}\paperautocell{7.57} & \paperautocell{0.313} & \cellcolor{rank3}\paperautocell{7.29} & \cellcolor{rank1}\paperautocell{0.195} & \cellcolor{rank2}\paperautocell{12.45} & \paperautocell{0.240} & \cellcolor{rank1}\paperautocell{6.90} \\
    \cellcolor{emphrow}\textbf{Ours (L2)} & \paperautocell{0.299} & \cellcolor{rank1}\paperautocell{7.28} & \paperautocell{0.354} & \cellcolor{rank2}\paperautocell{6.95} & \cellcolor{rank2}\paperautocell{0.189} & \cellcolor{rank1}\paperautocell{11.11} & \cellcolor{rank3}\paperautocell{0.248} & \cellcolor{rank2}\paperautocell{7.00} \\
    \bottomrule
  \end{tabular}%
  }
    \\[2pt] {\footnotesize (a) Tanks and Temples}
  \end{minipage}%
  \hfill
  \begin{minipage}[b]{0.32\textwidth}
    \centering
  \scriptsize
  \setlength{\tabcolsep}{3pt}
  \definecolor{rank1}{HTML}{FFB3B3}
  \definecolor{rank2}{HTML}{FFD9B3}
  \definecolor{rank3}{HTML}{FFFFB3}
  \definecolor{emphrow}{HTML}{EDEDED}
  \ifdefined\paperautocellwidth\else\newlength{\paperautocellwidth}\fi
  \ifdefined\paperautoprobe\else\newlength{\paperautoprobe}\fi
  \settowidth{\paperautocellwidth}{\textbf{0.000}}
  \settowidth{\paperautoprobe}{CD $\downarrow$}
  \ifdim\dimexpr\paperautoprobe-0\tabcolsep\relax>1\paperautocellwidth  \setlength{\paperautocellwidth}{\dimexpr(\paperautoprobe-0\tabcolsep)/1\relax}\fi
  \settowidth{\paperautoprobe}{F@10 $\uparrow$}
  \ifdim\dimexpr\paperautoprobe-0\tabcolsep\relax>1\paperautocellwidth  \setlength{\paperautocellwidth}{\dimexpr(\paperautoprobe-0\tabcolsep)/1\relax}\fi
  \settowidth{\paperautoprobe}{NC $\uparrow$}
  \ifdim\dimexpr\paperautoprobe-0\tabcolsep\relax>1\paperautocellwidth  \setlength{\paperautocellwidth}{\dimexpr(\paperautoprobe-0\tabcolsep)/1\relax}\fi
  \providecommand{\paperautospan}[2]{\makebox[\dimexpr#1\paperautocellwidth+#1\tabcolsep+#1\tabcolsep-\tabcolsep-\tabcolsep\relax][c]{#2}}
  \providecommand{\paperautocell}[1]{\paperautospan{1}{#1}}
  \resizebox{\linewidth}{!}{%
  \begin{tabular}{lccc}
    \toprule
    Method & \multicolumn{1}{c}{\paperautospan{1}{CD $\downarrow$}} & \multicolumn{1}{c}{\paperautospan{1}{F@10 $\uparrow$}} & \multicolumn{1}{c}{\paperautospan{1}{NC $\uparrow$}} \\
    \midrule
    \multicolumn{4}{l}{\textbf{\textit{w/ per-scene optimization}}} \\
    2DGS & \paperautocell{4.858} & \paperautocell{0.076} & \paperautocell{0.174} \\
    CityGaussianV2 & \paperautocell{5.155} & \paperautocell{0.112} & \paperautocell{0.228} \\
    \midrule
    \multicolumn{4}{l}{\textbf{\textit{w/o per-scene optimization}}} \\
    DA3 & \paperautocell{5.184} & \paperautocell{0.015} & \paperautocell{0.048} \\
    Murre & \paperautocell{9.998} & \paperautocell{0.001} & \paperautocell{0.010} \\
    \cellcolor{emphrow}\textbf{Ours (L0, 80\,m)} & \paperautocell{0.872} & \paperautocell{0.245} & \paperautocell{0.269} \\
    \cellcolor{emphrow}\textbf{Ours (L1, 40\,m)} & \cellcolor{rank3}\paperautocell{0.714} & \cellcolor{rank3}\paperautocell{0.296} & \cellcolor{rank3}\paperautocell{0.315} \\
    \cellcolor{emphrow}\textbf{Ours (L2, 20\,m)} & \cellcolor{rank2}\paperautocell{0.662} & \cellcolor{rank2}\paperautocell{0.351} & \cellcolor{rank2}\paperautocell{0.350} \\
    \cellcolor{emphrow}\textbf{Ours (L3, 10\,m)} & \cellcolor{rank1}\paperautocell{0.645} & \cellcolor{rank1}\paperautocell{0.367} & \cellcolor{rank1}\paperautocell{0.361} \\
    \bottomrule
  \end{tabular}%
  }
    \\[2pt] {\footnotesize (b) City blocks, 256 input views}
  \end{minipage}%
\end{table*}

\begin{figure*}[t]
  \centering
  \begin{minipage}[b]{0.162\textwidth}
    \centering
  \setlength{\tabcolsep}{0.5pt}
  \renewcommand{\arraystretch}{1.0}
  \begin{tabular}[b]{c}
    \includegraphics[width=\dimexpr(\linewidth-1pt)/1\relax]{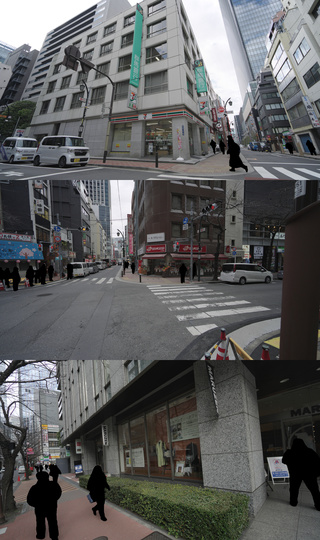} \\[1.5pt]
    \multicolumn{1}{c}{\footnotesize (a) Input views}
  \end{tabular}
  \end{minipage}%
  \hfill
  \begin{minipage}[b]{0.826\textwidth}
    \centering
  \setlength{\tabcolsep}{0.5pt}
  \renewcommand{\arraystretch}{1.0}
  \begin{tabular}[b]{cc@{\hspace{3pt}}cc}
    \includegraphics[width=\dimexpr(\linewidth-7pt)/4\relax]{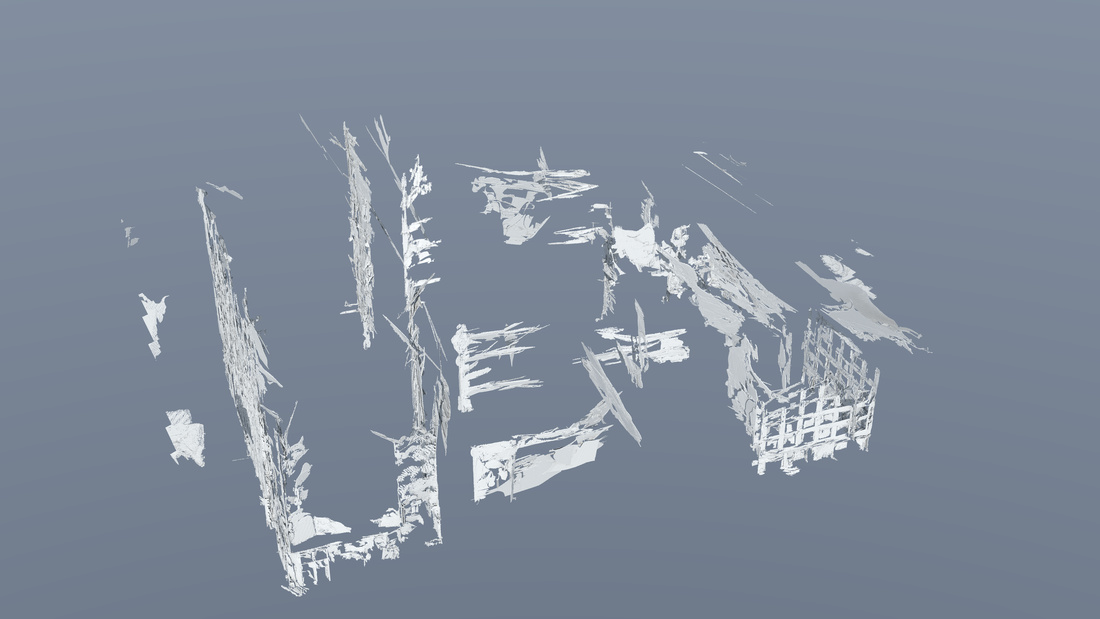} & \includegraphics[width=\dimexpr(\linewidth-7pt)/4\relax]{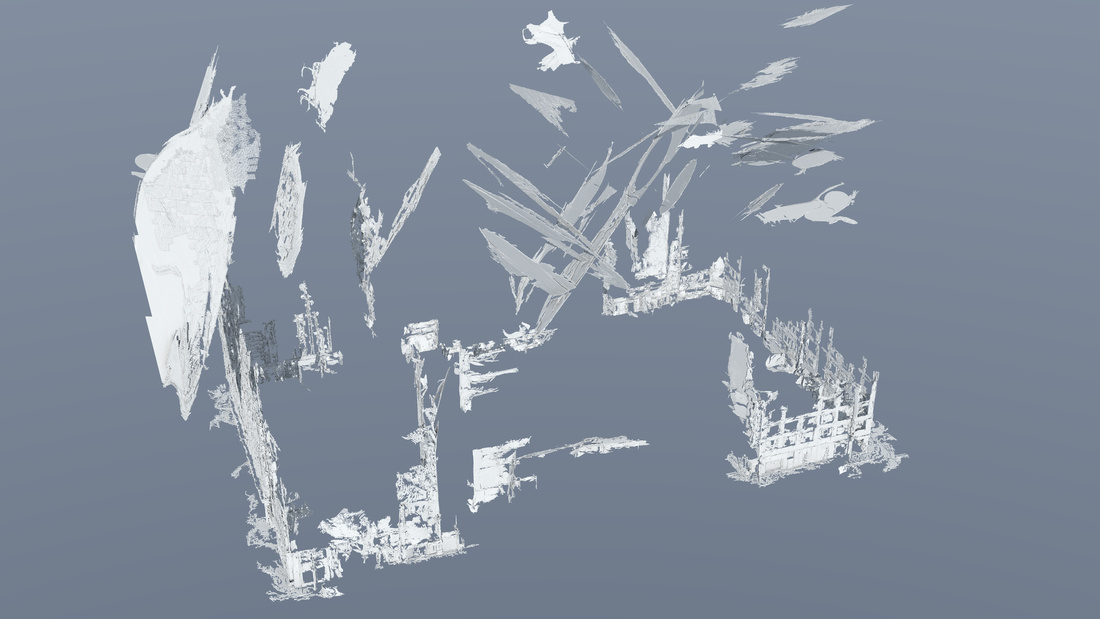} & \includegraphics[width=\dimexpr(\linewidth-7pt)/4\relax]{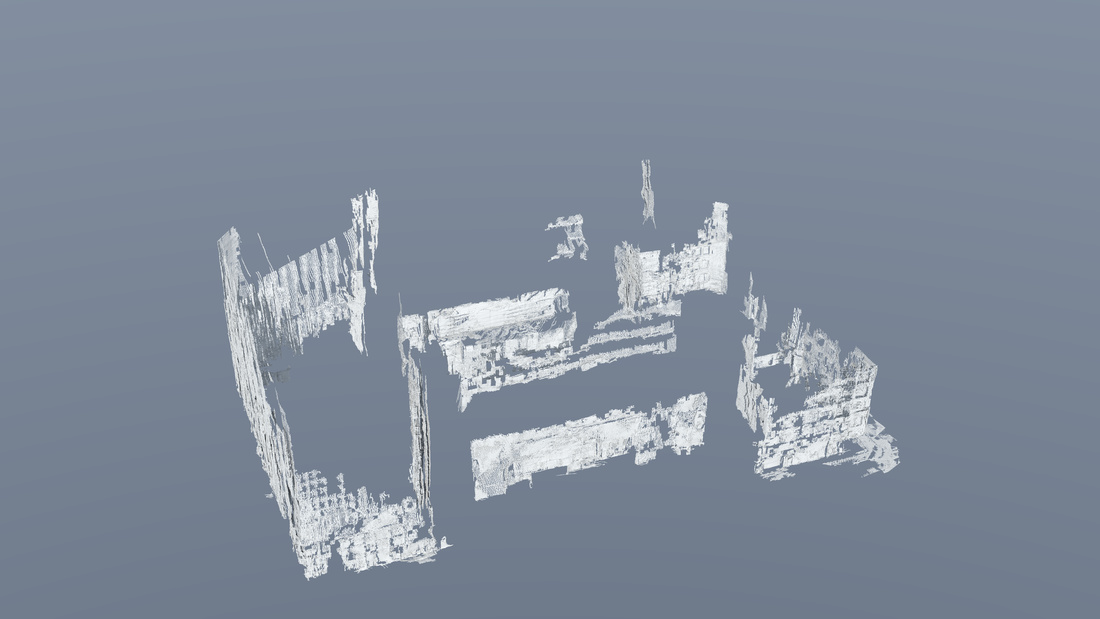} & \includegraphics[width=\dimexpr(\linewidth-7pt)/4\relax]{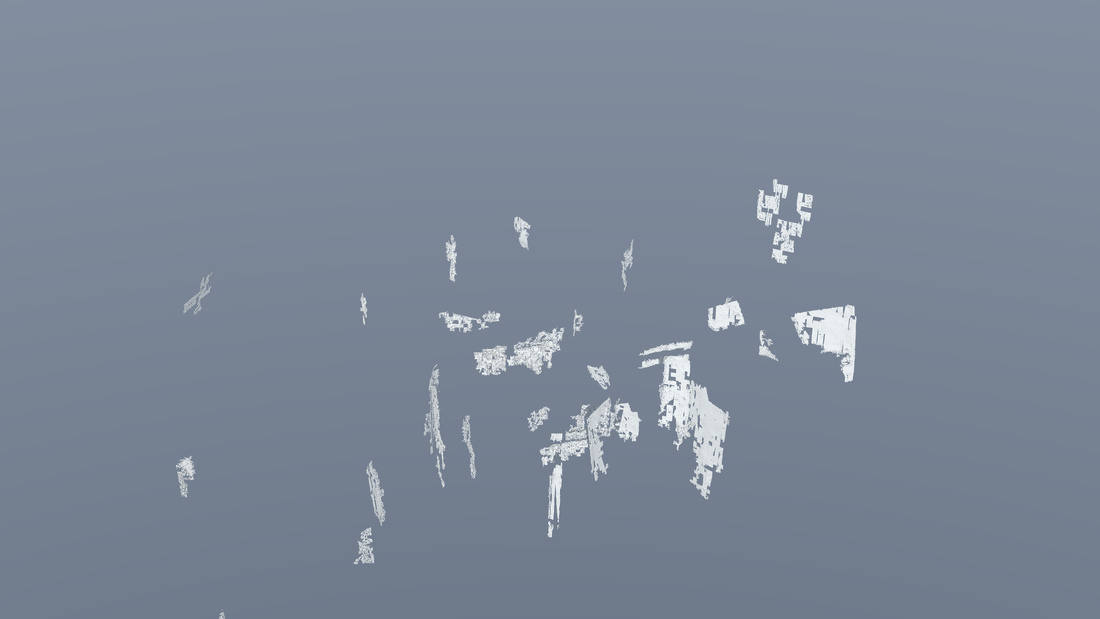} \\[-4.5pt]
    \makebox[\dimexpr(\linewidth-7pt)/4\relax][c]{\footnotesize 2DGS} & \makebox[\dimexpr(\linewidth-7pt)/4\relax][c]{\footnotesize CityGaussianV2} & \makebox[\dimexpr(\linewidth-7pt)/4\relax][c]{\footnotesize DA3} & \makebox[\dimexpr(\linewidth-7pt)/4\relax][c]{\footnotesize Murre} \\[3pt]
    \includegraphics[width=\dimexpr(\linewidth-7pt)/4\relax]{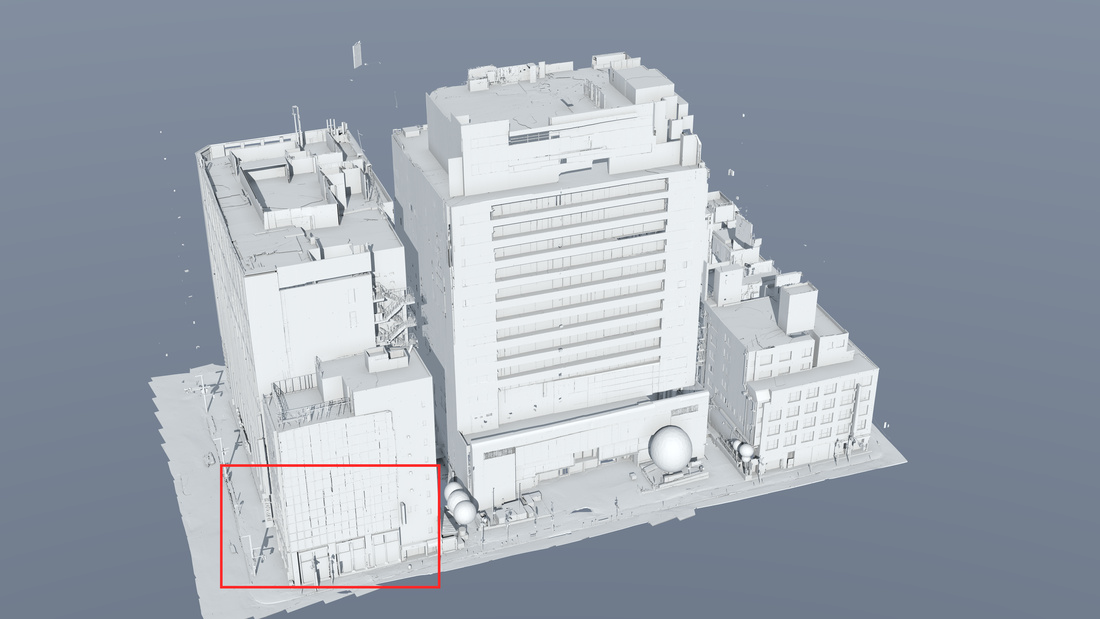} & \includegraphics[width=\dimexpr(\linewidth-7pt)/4\relax]{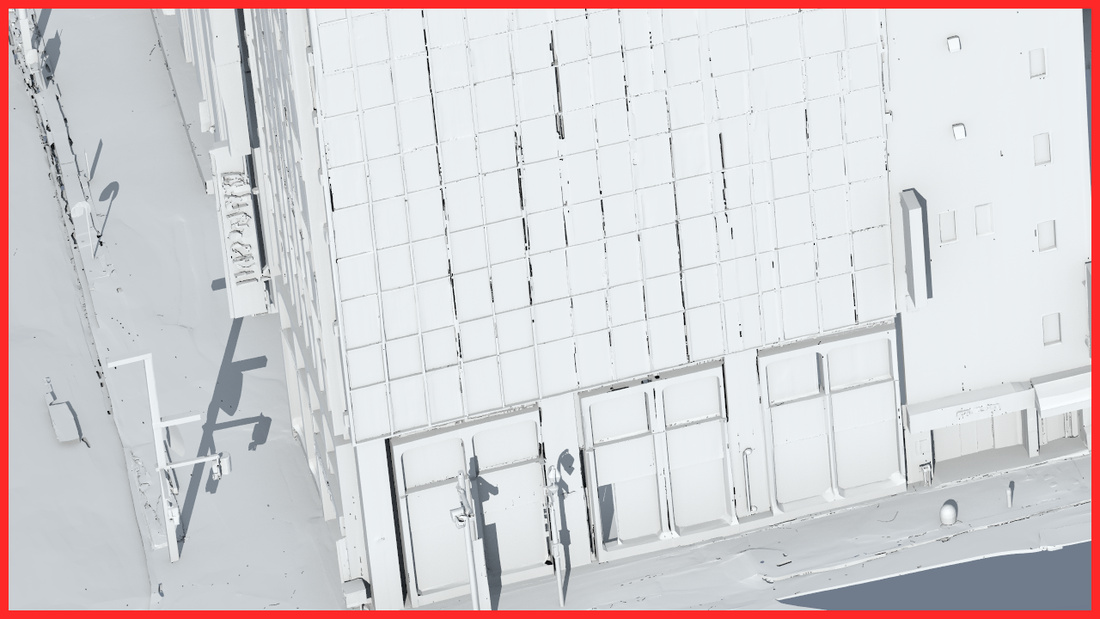} & \includegraphics[width=\dimexpr(\linewidth-7pt)/4\relax]{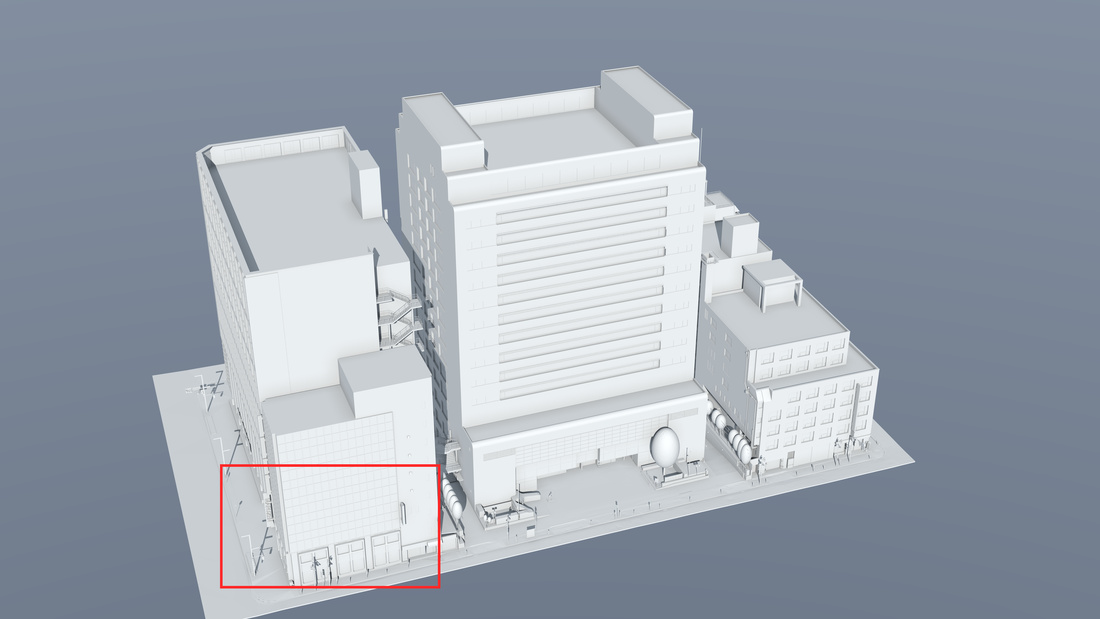} & \includegraphics[width=\dimexpr(\linewidth-7pt)/4\relax]{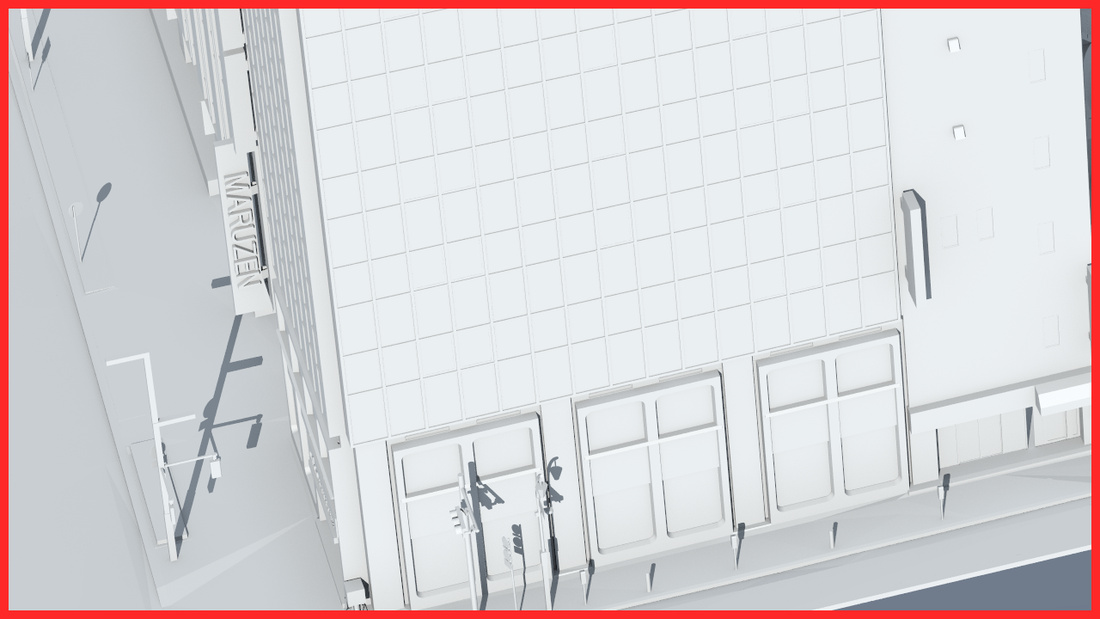} \\[-4.5pt]
    \multicolumn{2}{c}{\makebox[\dimexpr2\dimexpr\dimexpr(\linewidth-7pt)/4\relax\relax+1pt\relax][c]{\footnotesize Ours}} & \multicolumn{2}{c}{\makebox[\dimexpr2\dimexpr\dimexpr(\linewidth-7pt)/4\relax\relax+1pt\relax][c]{\footnotesize Ground truth}} \\[1.5pt]
    \multicolumn{4}{c}{\footnotesize (b) Reconstruction results}
  \end{tabular}
  \end{minipage}%
  \caption{\textbf{Qualitative comparison on the city-block dataset.} (a) Three of the 256 input views. (b) Each method's reconstruction, with zoomed crops of ours and the ground truth to compare detail.}
  \label{fig:bim-qualitative}
\end{figure*}

\textbf{City blocks.}
Table~\ref{tab:main}(b) reports results on the three held-out blocks with 256 input
views. These scenes are challenging because the street-level photographs contain depths
from a few meters to over a hundred meters. All baselines fail in this setting, with Chamfer distances of about 5\,m or
more, whereas our method reaches 0.65\,m and outperforms them on all three metrics.
Each finer cascade level improves all three metrics, and
Figure~\ref{fig:bim-qualitative} shows the reconstructed blocks with close-ups.

\textbf{Real-world captures.}
Figure~\ref{fig:teaser} shows \texttt{garden}, where finer levels around the table
recover the plants in the vase, the gaps between the table planks, the metal plate at
its center, and a glossy sphere in the background.
Appendix~\ref{app:inthewild-results} compares this scene with GaussianWrapping. In the
other two captures (Figure~\ref{fig:main-scannetpp}(c)), we recover transparent glass
bottles, a badge lanyard, and shirt wrinkles.

\subsection{Ablations and Analysis}
\label{sec:exp:ablation}

\begin{figure*}[t]
  \centering
  \begin{minipage}[b]{0.4\textwidth}
    \centering
  \footnotesize
  \setlength{\tabcolsep}{3pt}
  \definecolor{rank1}{HTML}{FFB3B3}
  \definecolor{rank2}{HTML}{FFD9B3}
  \definecolor{rank3}{HTML}{FFFFB3}
  \begin{tabular}{lcccc}
    \toprule
     & \multicolumn{2}{c}{2D} & \multicolumn{2}{c}{3D} \\
    \cmidrule(lr){2-3}\cmidrule(lr){4-5}
    $t_s$ & MAE $\downarrow$ & Ang $\downarrow$ & CD $\downarrow$ & F@2 $\uparrow$ \\
    \midrule
    0.8 & \textbf{0.0565} & \textbf{16.33} & \textbf{0.0212} & \textbf{0.7155} \\
    0.9 & 0.1450 & 19.16 & 0.0621 & 0.6779 \\
    1.0 & 0.2170 & 22.15 & 0.1051 & 0.6306 \\
    \bottomrule
  \end{tabular}
    \\[0pt] {\footnotesize (a) ScanNet++, 32 input views}
  \end{minipage}%
  \hfill
  \begin{minipage}[b]{0.56\textwidth}
    \centering
  \setlength{\tabcolsep}{0.5pt}
  \renewcommand{\arraystretch}{1.0}
  \begin{tabular}{cc}
    \includegraphics[width=\dimexpr(\linewidth-2pt)/2\relax]{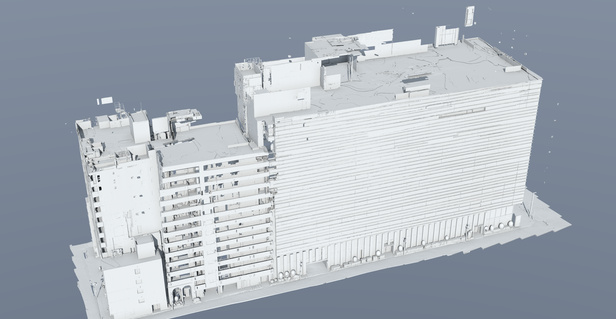} & \includegraphics[width=\dimexpr(\linewidth-2pt)/2\relax]{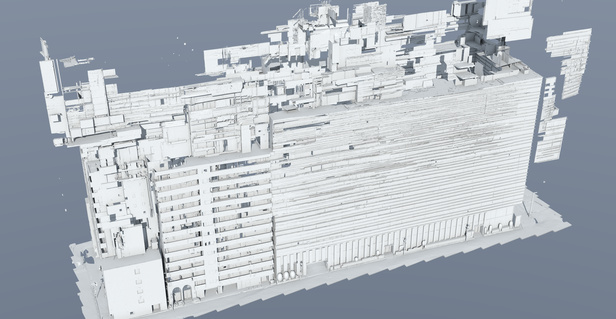} \\[-4.5pt]
    \makebox[\dimexpr(\linewidth-2pt)/2\relax][c]{\scriptsize $t_s = 0.8$} & \makebox[\dimexpr(\linewidth-2pt)/2\relax][c]{\scriptsize $t_s = 1.0$} \\
  \end{tabular}
    \\[0pt] {\footnotesize (b) City block, 256 input views}
  \end{minipage}%
  \caption{\textbf{Effect of parent initialization.} Every finer cascade level starts from a latent noised to $t_s$, where $t_s = 1$ generates from pure noise. (a) Mean over the 25 largest ScanNet++ scenes, with the best results in bold. (b) One block of the city-block dataset.}
  \label{fig:ablation-parent}
\end{figure*}

\textbf{Parent initialization.}
Each finer level starts its flow sampling from the parent's prediction noised to $t_s$
instead of pure noise (Section~\ref{sec:method:cascade}). We compare our default
$t_s=0.8$ with $t_s=0.9$ and $t_s=1$ at every finer level
(Figure~\ref{fig:ablation-parent}(a)). At $t_s=1$, the finer levels start from pure
noise and inherit only the cell placement. On the large ScanNet++
scenes with 32 views, lowering $t_s$ improves every metric and reduces the Chamfer
distance by 80\%. Results with 256 views are in Appendix~\ref{app:cascade-ablation}.
Figure~\ref{fig:ablation-parent}(b) shows a city block. Overlapping cells keep
neighboring surfaces continuous, but information from distant cells barely reaches a
cell, and the roof is not visible from the street. Without the parent, the roof breaks
into fragments above the building. With the
parent, all cells start from the same coarse building, and the roof stays in place.

\begin{figure*}[t]
  \centering
  \begin{minipage}[b]{0.5\textwidth}
    \centering
  \footnotesize
  \setlength{\tabcolsep}{3pt}
  \definecolor{rank1}{HTML}{FFB3B3}
  \definecolor{rank2}{HTML}{FFD9B3}
  \definecolor{rank3}{HTML}{FFFFB3}
  \begin{tabular}{lcccc}
    \toprule
     & \multicolumn{2}{c}{2D} & \multicolumn{2}{c}{3D} \\
    \cmidrule(lr){2-3}\cmidrule(lr){4-5}
    Conditioning & MAE $\downarrow$ & Ang $\downarrow$ & CD $\downarrow$ & F@2 $\uparrow$ \\
    \midrule
    Baseline & 0.0610 & 16.44 & 0.0228 & 0.6424 \\
    Voxel attn. & 0.0609 & 16.34 & 0.0235 & 0.6495 \\
    + ConvNeXt & \textbf{0.0557} & \textbf{15.62} & \textbf{0.0202} & \textbf{0.7176} \\
    \bottomrule
  \end{tabular}
    \\[0pt] {\footnotesize (a) ScanNet++, 256 input views}
  \end{minipage}%
  \hfill
  \begin{minipage}[b]{0.46\textwidth}
    \centering
  \setlength{\tabcolsep}{0.5pt}
  \renewcommand{\arraystretch}{1.0}
  \begin{tabular}{ccc}
    \includegraphics[width=\dimexpr(\linewidth-3pt)/3\relax]{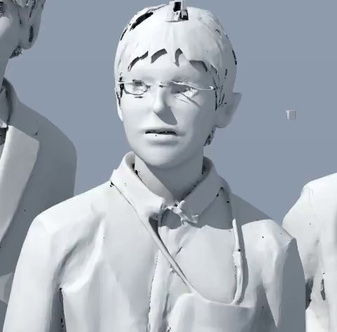} & \includegraphics[width=\dimexpr(\linewidth-3pt)/3\relax]{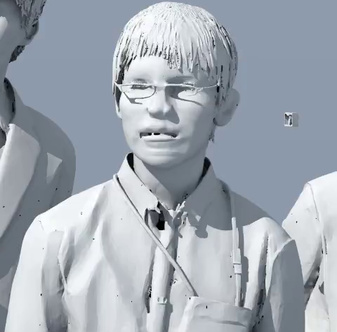} & \includegraphics[width=\dimexpr(\linewidth-3pt)/3\relax]{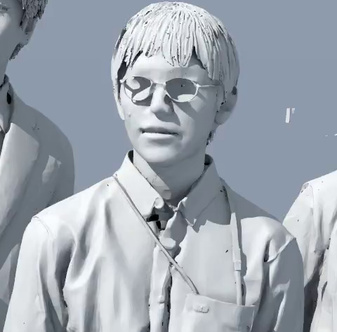} \\[-4.5pt]
    \makebox[\dimexpr(\linewidth-3pt)/3\relax][c]{\scriptsize Baseline} & \makebox[\dimexpr(\linewidth-3pt)/3\relax][c]{\scriptsize Voxel attn.} & \makebox[\dimexpr(\linewidth-3pt)/3\relax][c]{\scriptsize + ConvNeXt} \\
  \end{tabular}
    \\[0pt] {\footnotesize (b) Human capture, 56 input views}
  \end{minipage}%
  \caption{\textbf{Effect of image conditioning.} The baseline uses the GenRecon-style conditioner, our voxel-queried attention replaces it, and the ConvNeXt branch is then added, all at the finest cascade level. (a) Mean over the 25 largest ScanNet++ scenes, with the best results in bold. (b) A multi-view capture of people.}
  \label{fig:ablation-conditioning}
\end{figure*}

\textbf{Image conditioning.}
Figure~\ref{fig:ablation-conditioning} compares, at the finest level, the
GenRecon-style conditioner, our voxel-queried attention
(Section~\ref{sec:method:attention}), and the latter with the ConvNeXt branch, which adds pixel-level detail. The baseline
bilinearly samples DINO features at projected voxel
centers, which loses detail finer than a token. On ScanNet++, voxel-queried attention alone
is on par with the baseline, and the ConvNeXt branch substantially improves every
metric. On the human
capture, the baseline over-smooths the hair, the nose, and the shirt wrinkles.
Voxel-queried attention recovers strands of hair, and the ConvNeXt branch sharpens them
further and adds the shirt buttons. Appendix~\ref{app:conditioning-ablation} shows further qualitative results.

\textbf{Inference time.}
On large ScanNet++ scenes, a single H200 takes 17 (L1) and 89 (L2) minutes per scene, comparable to per-scene optimization but slower than feed-forward methods (Appendix~\ref{app:inference-time}).

\section{Conclusion and Limitations}
\label{sec:conclusion}

We have presented \method, a generative reconstruction framework that applies a single
fixed-resolution 3D generator across scene scales through an inference-time
coarse-to-fine cascade. The coarse level establishes a globally consistent layout, finer
levels recover surface detail, and a cell-independent multi-scale image conditioning
gives every cell images at the resolution it needs. On ScanNet++, Tanks and Temples, and
city blocks, \method outperforms the generative baseline and is competitive with or
better than feed-forward and per-scene optimization methods.

Since \method does not fit the input images pixel by pixel, its surfaces can be slightly
offset from the observations, which lowers the F-score at tight thresholds when views are
dense. Its cost grows with the number of cells and cascade levels, so sampling large
scenes at high resolution remains slow, and it assumes posed images. Moreover, it does not work on distant objects in
unbounded scenes (Appendix~\ref{app:failure-cases}). In future work, we
plan to reduce the sampling cost by distilling the flow models into few-step samplers and
to extend \method to unbounded scenes.

\subsection*{AI use statement}

In this work, we used generative AI tools to search for related work, to propose and refine hypotheses, to give
feedback on the design of the method and the experiments, and to implement the methods,
including the model, training, inference, and evaluation code. We also used them to
generate the synthetic training datasets, namely the procedural houses and grass scenes
(Appendix~\ref{app:general-training}), to clean and reformat datasets, and to interpret the
results, including quantitative analyses of the experimental outputs. Additionally, we used generative AI tools to draft and edit the
paper for readability and to create figures. We have reviewed all AI-assisted work. The authors specified the implementation, checked the
implemented code against these specifications, checked the qualitative and quantitative
results, and reviewed the AI-assisted text. We take responsibility for the final content of this work,
including text, claims, and artifacts produced with the aid of generative AI.

\subsection*{Ethics statement}

Our city-block images are taken on public streets and show passers-by and vehicles. In the
released data, people and license plates are masked, and the released
3D Gaussian Splatting models and BIM meshes contain no people
(Appendix~\ref{app:city-dataset}). The five people in our in-house human capture consented
to being recorded. All other datasets are public and used under their own terms.

\subsection*{Reproducibility statement}

Section~\ref{sec:method} and Appendices~\ref{app:method} and~\ref{app:conditioning} describe
the method, including the inference procedure, the cell sizes and extents, and the image
conditioning. Appendix~\ref{app:training} gives the training data, the curriculum, and the
optimization settings, including the number of steps and GPUs, and
Appendix~\ref{app:evaluation} gives the datasets, metrics, evaluation protocols, and baseline
settings. All datasets other than ours are public,
and Appendix~\ref{app:city-dataset} describes which parts of our city-block data we will
release.

\subsubsection*{Acknowledgments}
This paper is based on results obtained from a project, JPNP20017, commissioned by the New
Energy and Industrial Technology Development Organization (NEDO). We would like to thank members of Preferred
Networks for their helpful discussions. We also would like to thank cluster team members for
infrastructure support.

\bibliography{refs}
\bibliographystyle{preprint}
\clearpage

\appendix
\section{Additional Results}
\label{app:results}

This section gives the complete results on each benchmark of the main paper and
qualitative comparisons on captures outside the benchmarks. The metrics and evaluation protocols are defined in
Appendix~\ref{app:evaluation}.

\subsection{ScanNet++}
\label{app:scannetpp-results}

Tables~\ref{tab:app-scannetpp-2d} and~\ref{tab:app-scannetpp-3d} report results on
all 50 validation scenes with 32 and 256 input views, split into the 25 smallest and
the 25 largest by ground-truth bounding-box volume. The large split is the one in
Table~\ref{tab:main-scannetpp}. Table~\ref{tab:app-scannetpp-3d} also reports the
whole-mesh protocol of GenRecon. DA3 obtains the lowest MAE, AbsRel, and Chamfer
distance on the small scenes, while our finest level has the lowest normal error in
all four settings. Figures~\ref{fig:scannetpp-qualitative-fps32:1}
and~\ref{fig:scannetpp-qualitative-fps32:2} show four scenes with 32 input views, and
Figures~\ref{fig:scannetpp-qualitative-fps256:1} and~\ref{fig:scannetpp-qualitative-fps256:2}
with 256. Figure~\ref{fig:baseline-comparison-scannetpp} compares close-ups of our meshes
with those of GenRecon on \texttt{0d2ee665be} and
\texttt{578511c8a9} in the 256-view setting.
\begin{table}[H]
  \centering
  \caption{\textbf{Quantitative comparison on ScanNet++ (2D metrics).} Mean over the 25 smallest, the 25 largest, and all 50 scenes. Depth and normals are rendered from the meshes at held-out views. The top three results are highlighted.}
  \label{tab:app-scannetpp-2d}
  \scriptsize
  \setlength{\tabcolsep}{3pt}
  \definecolor{rank1}{HTML}{FFB3B3}
  \definecolor{rank2}{HTML}{FFD9B3}
  \definecolor{rank3}{HTML}{FFFFB3}
  \definecolor{emphrow}{HTML}{EDEDED}
  \ifdefined\paperautocellwidth\else\newlength{\paperautocellwidth}\fi
  \ifdefined\paperautoprobe\else\newlength{\paperautoprobe}\fi
  \settowidth{\paperautocellwidth}{\textbf{0.0000}}
  \settowidth{\paperautoprobe}{Small (25)}
  \ifdim\dimexpr\paperautoprobe-10\tabcolsep\relax>6\paperautocellwidth  \setlength{\paperautocellwidth}{\dimexpr(\paperautoprobe-10\tabcolsep)/6\relax}\fi
  \settowidth{\paperautoprobe}{Large (25)}
  \ifdim\dimexpr\paperautoprobe-10\tabcolsep\relax>6\paperautocellwidth  \setlength{\paperautocellwidth}{\dimexpr(\paperautoprobe-10\tabcolsep)/6\relax}\fi
  \settowidth{\paperautoprobe}{All (50)}
  \ifdim\dimexpr\paperautoprobe-10\tabcolsep\relax>6\paperautocellwidth  \setlength{\paperautocellwidth}{\dimexpr(\paperautoprobe-10\tabcolsep)/6\relax}\fi
  \settowidth{\paperautoprobe}{Depth}
  \ifdim\dimexpr\paperautoprobe-6\tabcolsep\relax>4\paperautocellwidth  \setlength{\paperautocellwidth}{\dimexpr(\paperautoprobe-6\tabcolsep)/4\relax}\fi
  \settowidth{\paperautoprobe}{Normal}
  \ifdim\dimexpr\paperautoprobe-0\tabcolsep\relax>1\paperautocellwidth  \setlength{\paperautocellwidth}{\dimexpr(\paperautoprobe-0\tabcolsep)/1\relax}\fi
  \settowidth{\paperautoprobe}{Coverage}
  \ifdim\dimexpr\paperautoprobe-0\tabcolsep\relax>1\paperautocellwidth  \setlength{\paperautocellwidth}{\dimexpr(\paperautoprobe-0\tabcolsep)/1\relax}\fi
  \settowidth{\paperautoprobe}{MAE$\downarrow$}
  \ifdim\dimexpr\paperautoprobe-0\tabcolsep\relax>1\paperautocellwidth  \setlength{\paperautocellwidth}{\dimexpr(\paperautoprobe-0\tabcolsep)/1\relax}\fi
  \settowidth{\paperautoprobe}{RMSE$\downarrow$}
  \ifdim\dimexpr\paperautoprobe-0\tabcolsep\relax>1\paperautocellwidth  \setlength{\paperautocellwidth}{\dimexpr(\paperautoprobe-0\tabcolsep)/1\relax}\fi
  \settowidth{\paperautoprobe}{AbsRel$\downarrow$}
  \ifdim\dimexpr\paperautoprobe-0\tabcolsep\relax>1\paperautocellwidth  \setlength{\paperautocellwidth}{\dimexpr(\paperautoprobe-0\tabcolsep)/1\relax}\fi
  \settowidth{\paperautoprobe}{SqRel$\downarrow$}
  \ifdim\dimexpr\paperautoprobe-0\tabcolsep\relax>1\paperautocellwidth  \setlength{\paperautocellwidth}{\dimexpr(\paperautoprobe-0\tabcolsep)/1\relax}\fi
  \settowidth{\paperautoprobe}{Ang$\downarrow$}
  \ifdim\dimexpr\paperautoprobe-0\tabcolsep\relax>1\paperautocellwidth  \setlength{\paperautocellwidth}{\dimexpr(\paperautoprobe-0\tabcolsep)/1\relax}\fi
  \settowidth{\paperautoprobe}{Comp$\uparrow$}
  \ifdim\dimexpr\paperautoprobe-0\tabcolsep\relax>1\paperautocellwidth  \setlength{\paperautocellwidth}{\dimexpr(\paperautoprobe-0\tabcolsep)/1\relax}\fi
  \providecommand{\paperautospan}[2]{\makebox[\dimexpr#1\paperautocellwidth+#1\tabcolsep+#1\tabcolsep-\tabcolsep-\tabcolsep\relax][c]{#2}}
  \providecommand{\paperautocell}[1]{\paperautospan{1}{#1}}
  \resizebox{\linewidth}{!}{%
  \begin{tabular}{lcccccccccccccccccc}
    \toprule
     & \multicolumn{6}{c}{\paperautospan{6}{Small (25)}} & \multicolumn{6}{c}{\paperautospan{6}{Large (25)}} & \multicolumn{6}{c}{\paperautospan{6}{All (50)}} \\
    \cmidrule(lr){2-7}\cmidrule(lr){8-13}\cmidrule(lr){14-19}
     & \multicolumn{4}{c}{\paperautospan{4}{Depth}} & \multicolumn{1}{c}{\paperautospan{1}{Normal}} & \multicolumn{1}{c}{\paperautospan{1}{Coverage}} & \multicolumn{4}{c}{\paperautospan{4}{Depth}} & \multicolumn{1}{c}{\paperautospan{1}{Normal}} & \multicolumn{1}{c}{\paperautospan{1}{Coverage}} & \multicolumn{4}{c}{\paperautospan{4}{Depth}} & \multicolumn{1}{c}{\paperautospan{1}{Normal}} & \multicolumn{1}{c}{\paperautospan{1}{Coverage}} \\
    \cmidrule(lr){2-5}\cmidrule(lr){6-6}\cmidrule(lr){7-7}\cmidrule(lr){8-11}\cmidrule(lr){12-12}\cmidrule(lr){13-13}\cmidrule(lr){14-17}\cmidrule(lr){18-18}\cmidrule(lr){19-19}
    Method & \multicolumn{1}{c}{\paperautospan{1}{MAE$\downarrow$}} & \multicolumn{1}{c}{\paperautospan{1}{RMSE$\downarrow$}} & \multicolumn{1}{c}{\paperautospan{1}{AbsRel$\downarrow$}} & \multicolumn{1}{c}{\paperautospan{1}{SqRel$\downarrow$}} & \multicolumn{1}{c}{\paperautospan{1}{Ang$\downarrow$}} & \multicolumn{1}{c}{\paperautospan{1}{Comp$\uparrow$}} & \multicolumn{1}{c}{\paperautospan{1}{MAE$\downarrow$}} & \multicolumn{1}{c}{\paperautospan{1}{RMSE$\downarrow$}} & \multicolumn{1}{c}{\paperautospan{1}{AbsRel$\downarrow$}} & \multicolumn{1}{c}{\paperautospan{1}{SqRel$\downarrow$}} & \multicolumn{1}{c}{\paperautospan{1}{Ang$\downarrow$}} & \multicolumn{1}{c}{\paperautospan{1}{Comp$\uparrow$}} & \multicolumn{1}{c}{\paperautospan{1}{MAE$\downarrow$}} & \multicolumn{1}{c}{\paperautospan{1}{RMSE$\downarrow$}} & \multicolumn{1}{c}{\paperautospan{1}{AbsRel$\downarrow$}} & \multicolumn{1}{c}{\paperautospan{1}{SqRel$\downarrow$}} & \multicolumn{1}{c}{\paperautospan{1}{Ang$\downarrow$}} & \multicolumn{1}{c}{\paperautospan{1}{Comp$\uparrow$}} \\
    \midrule
    \multicolumn{19}{l}{\textbf{\textit{32 Views}}} \\
    2DGS & \paperautocell{0.1261} & \paperautocell{0.2493} & \paperautocell{0.1329} & \paperautocell{0.0591} & \paperautocell{40.149} & \paperautocell{0.9542} & \paperautocell{0.1426} & \paperautocell{0.3412} & \paperautocell{0.1141} & \paperautocell{0.0799} & \paperautocell{38.376} & \paperautocell{0.9058} & \paperautocell{0.1344} & \paperautocell{0.2953} & \paperautocell{0.1235} & \paperautocell{0.0695} & \paperautocell{39.262} & \paperautocell{0.9300} \\
    MILo & \paperautocell{0.1748} & \paperautocell{0.3699} & \paperautocell{0.1855} & \paperautocell{0.1530} & \paperautocell{44.731} & \paperautocell{0.9913} & \paperautocell{0.2523} & \paperautocell{0.6171} & \paperautocell{0.1780} & \paperautocell{0.2623} & \paperautocell{42.156} & \paperautocell{0.9848} & \paperautocell{0.2136} & \paperautocell{0.4935} & \paperautocell{0.1817} & \paperautocell{0.2077} & \paperautocell{43.444} & \paperautocell{0.9881} \\
    GaussianWrapping & \paperautocell{0.4688} & \paperautocell{1.4378} & \paperautocell{0.6228} & \paperautocell{6.9407} & \paperautocell{40.150} & \paperautocell{0.9365} & \paperautocell{0.3467} & \paperautocell{1.3934} & \paperautocell{0.2762} & \paperautocell{2.8323} & \paperautocell{35.974} & \paperautocell{0.9648} & \paperautocell{0.4077} & \paperautocell{1.4156} & \paperautocell{0.4495} & \paperautocell{4.8865} & \paperautocell{38.062} & \paperautocell{0.9507} \\
    DA3 & \cellcolor{rank1}\paperautocell{0.0255} & \cellcolor{rank3}\paperautocell{0.0925} & \cellcolor{rank1}\paperautocell{0.0348} & \cellcolor{rank3}\paperautocell{0.0139} & \paperautocell{19.773} & \paperautocell{0.9888} & \paperautocell{0.0811} & \paperautocell{0.2849} & \paperautocell{0.0800} & \paperautocell{0.0900} & \paperautocell{28.000} & \paperautocell{0.9682} & \cellcolor{rank3}\paperautocell{0.0533} & \paperautocell{0.1887} & \cellcolor{rank3}\paperautocell{0.0574} & \paperautocell{0.0520} & \paperautocell{23.886} & \paperautocell{0.9785} \\
    MapAnything & \cellcolor{rank3}\paperautocell{0.0335} & \paperautocell{0.1143} & \paperautocell{0.0508} & \paperautocell{0.0249} & \paperautocell{20.971} & \paperautocell{0.9717} & \paperautocell{0.1019} & \paperautocell{0.3277} & \paperautocell{0.1213} & \paperautocell{0.1466} & \paperautocell{30.090} & \paperautocell{0.9275} & \paperautocell{0.0677} & \paperautocell{0.2210} & \paperautocell{0.0860} & \paperautocell{0.0857} & \paperautocell{25.530} & \paperautocell{0.9496} \\
    Murre & \paperautocell{0.0758} & \paperautocell{0.1779} & \paperautocell{0.1024} & \paperautocell{0.0501} & \paperautocell{32.075} & \paperautocell{0.9816} & \paperautocell{0.1990} & \paperautocell{0.4693} & \paperautocell{0.2041} & \paperautocell{0.2290} & \paperautocell{40.515} & \paperautocell{0.9431} & \paperautocell{0.1374} & \paperautocell{0.3236} & \paperautocell{0.1532} & \paperautocell{0.1396} & \paperautocell{36.295} & \paperautocell{0.9623} \\
    GenRecon & \paperautocell{0.0571} & \paperautocell{0.1311} & \paperautocell{0.0739} & \paperautocell{0.0224} & \cellcolor{rank3}\paperautocell{18.123} & \paperautocell{0.9983} & \paperautocell{0.1175} & \paperautocell{0.2939} & \paperautocell{0.1228} & \paperautocell{0.1004} & \paperautocell{20.928} & \paperautocell{0.9764} & \paperautocell{0.0873} & \paperautocell{0.2125} & \paperautocell{0.0984} & \paperautocell{0.0614} & \paperautocell{19.526} & \paperautocell{0.9873} \\
    \cellcolor{emphrow}\textbf{Ours (L0)} & \paperautocell{0.0438} & \paperautocell{0.1157} & \paperautocell{0.0585} & \paperautocell{0.0176} & \paperautocell{18.745} & \cellcolor{rank3}\paperautocell{0.9986} & \cellcolor{rank3}\paperautocell{0.0714} & \cellcolor{rank3}\paperautocell{0.2394} & \cellcolor{rank3}\paperautocell{0.0746} & \cellcolor{rank3}\paperautocell{0.0599} & \cellcolor{rank3}\paperautocell{19.207} & \cellcolor{rank1}\paperautocell{0.9984} & \paperautocell{0.0576} & \cellcolor{rank3}\paperautocell{0.1775} & \paperautocell{0.0666} & \cellcolor{rank3}\paperautocell{0.0387} & \cellcolor{rank3}\paperautocell{18.976} & \cellcolor{rank2}\paperautocell{0.9985} \\
    \cellcolor{emphrow}\textbf{Ours (L1)} & \paperautocell{0.0346} & \cellcolor{rank2}\paperautocell{0.0924} & \cellcolor{rank3}\paperautocell{0.0440} & \cellcolor{rank2}\paperautocell{0.0096} & \cellcolor{rank2}\paperautocell{16.351} & \cellcolor{rank1}\paperautocell{0.9994} & \cellcolor{rank2}\paperautocell{0.0569} & \cellcolor{rank1}\paperautocell{0.2046} & \cellcolor{rank2}\paperautocell{0.0575} & \cellcolor{rank2}\paperautocell{0.0431} & \cellcolor{rank2}\paperautocell{17.098} & \cellcolor{rank2}\paperautocell{0.9983} & \cellcolor{rank2}\paperautocell{0.0457} & \cellcolor{rank1}\paperautocell{0.1485} & \cellcolor{rank2}\paperautocell{0.0508} & \cellcolor{rank2}\paperautocell{0.0264} & \cellcolor{rank2}\paperautocell{16.724} & \cellcolor{rank1}\paperautocell{0.9988} \\
    \cellcolor{emphrow}\textbf{Ours (L2)} & \cellcolor{rank2}\paperautocell{0.0324} & \cellcolor{rank1}\paperautocell{0.0913} & \cellcolor{rank2}\paperautocell{0.0407} & \cellcolor{rank1}\paperautocell{0.0090} & \cellcolor{rank1}\paperautocell{15.312} & \cellcolor{rank2}\paperautocell{0.9990} & \cellcolor{rank1}\paperautocell{0.0565} & \cellcolor{rank2}\paperautocell{0.2132} & \cellcolor{rank1}\paperautocell{0.0536} & \cellcolor{rank1}\paperautocell{0.0408} & \cellcolor{rank1}\paperautocell{16.330} & \cellcolor{rank3}\paperautocell{0.9962} & \cellcolor{rank1}\paperautocell{0.0445} & \cellcolor{rank2}\paperautocell{0.1522} & \cellcolor{rank1}\paperautocell{0.0471} & \cellcolor{rank1}\paperautocell{0.0249} & \cellcolor{rank1}\paperautocell{15.821} & \cellcolor{rank3}\paperautocell{0.9976} \\
    \midrule
    \multicolumn{19}{l}{\textbf{\textit{256 Views}}} \\
    2DGS & \paperautocell{0.0962} & \paperautocell{0.1901} & \paperautocell{0.1117} & \paperautocell{0.0386} & \paperautocell{37.382} & \paperautocell{0.9794} & \paperautocell{0.0929} & \paperautocell{0.2391} & \paperautocell{0.0846} & \paperautocell{0.0537} & \paperautocell{32.208} & \paperautocell{0.9747} & \paperautocell{0.0946} & \paperautocell{0.2146} & \paperautocell{0.0981} & \paperautocell{0.0462} & \paperautocell{34.795} & \paperautocell{0.9771} \\
    MILo & \paperautocell{0.1535} & \paperautocell{0.5276} & \paperautocell{0.1718} & \paperautocell{0.5368} & \paperautocell{39.667} & \paperautocell{0.9978} & \paperautocell{0.1406} & \paperautocell{0.5426} & \paperautocell{0.1226} & \paperautocell{0.3784} & \paperautocell{33.612} & \cellcolor{rank1}\paperautocell{0.9998} & \paperautocell{0.1470} & \paperautocell{0.5351} & \paperautocell{0.1472} & \paperautocell{0.4576} & \paperautocell{36.639} & \cellcolor{rank2}\paperautocell{0.9988} \\
    GaussianWrapping & \paperautocell{0.4076} & \paperautocell{1.2380} & \paperautocell{0.5049} & \paperautocell{3.5345} & \paperautocell{35.282} & \paperautocell{0.9321} & \paperautocell{0.3197} & \paperautocell{1.4556} & \paperautocell{0.3071} & \paperautocell{5.0707} & \paperautocell{30.118} & \paperautocell{0.9761} & \paperautocell{0.3636} & \paperautocell{1.3468} & \paperautocell{0.4060} & \paperautocell{4.3026} & \paperautocell{32.700} & \paperautocell{0.9541} \\
    DA3 & \cellcolor{rank1}\paperautocell{0.0254} & \cellcolor{rank1}\paperautocell{0.0928} & \cellcolor{rank1}\paperautocell{0.0348} & \cellcolor{rank3}\paperautocell{0.0140} & \cellcolor{rank2}\paperautocell{16.238} & \paperautocell{0.9957} & \paperautocell{0.0818} & \paperautocell{0.2910} & \paperautocell{0.0792} & \paperautocell{0.0966} & \paperautocell{20.789} & \cellcolor{rank3}\paperautocell{0.9958} & \cellcolor{rank3}\paperautocell{0.0536} & \paperautocell{0.1919} & \cellcolor{rank3}\paperautocell{0.0570} & \paperautocell{0.0553} & \cellcolor{rank3}\paperautocell{18.513} & \cellcolor{rank3}\paperautocell{0.9958} \\
    MapAnything & \cellcolor{rank2}\paperautocell{0.0345} & \paperautocell{0.1080} & \paperautocell{0.0480} & \paperautocell{0.0187} & \paperautocell{18.299} & \paperautocell{0.9922} & \paperautocell{0.0907} & \paperautocell{0.2944} & \paperautocell{0.0969} & \paperautocell{0.1038} & \paperautocell{23.982} & \paperautocell{0.9905} & \paperautocell{0.0626} & \paperautocell{0.2012} & \paperautocell{0.0724} & \paperautocell{0.0613} & \paperautocell{21.140} & \paperautocell{0.9914} \\
    Murre & \paperautocell{0.0962} & \paperautocell{0.2185} & \paperautocell{0.1236} & \paperautocell{0.0713} & \paperautocell{29.964} & \paperautocell{0.9926} & \paperautocell{0.2759} & \paperautocell{0.5960} & \paperautocell{0.2792} & \paperautocell{0.3498} & \paperautocell{40.780} & \paperautocell{0.9803} & \paperautocell{0.1861} & \paperautocell{0.4073} & \paperautocell{0.2014} & \paperautocell{0.2106} & \paperautocell{35.372} & \paperautocell{0.9864} \\
    GenRecon & \paperautocell{0.0577} & \paperautocell{0.1337} & \paperautocell{0.0742} & \paperautocell{0.0237} & \paperautocell{17.690} & \paperautocell{0.9981} & \paperautocell{0.1049} & \paperautocell{0.2732} & \paperautocell{0.1099} & \paperautocell{0.0867} & \paperautocell{20.030} & \paperautocell{0.9782} & \paperautocell{0.0813} & \paperautocell{0.2035} & \paperautocell{0.0921} & \paperautocell{0.0552} & \paperautocell{18.860} & \paperautocell{0.9881} \\
    \cellcolor{emphrow}\textbf{Ours (L0)} & \paperautocell{0.0469} & \paperautocell{0.1167} & \paperautocell{0.0603} & \paperautocell{0.0155} & \paperautocell{18.980} & \cellcolor{rank3}\paperautocell{0.9988} & \cellcolor{rank3}\paperautocell{0.0683} & \cellcolor{rank3}\paperautocell{0.2358} & \cellcolor{rank3}\paperautocell{0.0677} & \cellcolor{rank3}\paperautocell{0.0533} & \cellcolor{rank3}\paperautocell{18.952} & \cellcolor{rank2}\paperautocell{0.9990} & \paperautocell{0.0576} & \cellcolor{rank3}\paperautocell{0.1762} & \paperautocell{0.0640} & \cellcolor{rank3}\paperautocell{0.0344} & \paperautocell{18.966} & \cellcolor{rank1}\paperautocell{0.9989} \\
    \cellcolor{emphrow}\textbf{Ours (L1)} & \paperautocell{0.0375} & \cellcolor{rank2}\paperautocell{0.0954} & \cellcolor{rank3}\paperautocell{0.0478} & \cellcolor{rank1}\paperautocell{0.0101} & \cellcolor{rank3}\paperautocell{16.440} & \cellcolor{rank1}\paperautocell{0.9995} & \cellcolor{rank2}\paperautocell{0.0571} & \cellcolor{rank1}\paperautocell{0.2102} & \cellcolor{rank2}\paperautocell{0.0535} & \cellcolor{rank2}\paperautocell{0.0426} & \cellcolor{rank2}\paperautocell{16.684} & \paperautocell{0.9892} & \cellcolor{rank2}\paperautocell{0.0473} & \cellcolor{rank1}\paperautocell{0.1528} & \cellcolor{rank2}\paperautocell{0.0506} & \cellcolor{rank1}\paperautocell{0.0264} & \cellcolor{rank2}\paperautocell{16.562} & \paperautocell{0.9943} \\
    \cellcolor{emphrow}\textbf{Ours (L2)} & \cellcolor{rank3}\paperautocell{0.0361} & \cellcolor{rank3}\paperautocell{0.0963} & \cellcolor{rank2}\paperautocell{0.0446} & \cellcolor{rank2}\paperautocell{0.0110} & \cellcolor{rank1}\paperautocell{15.493} & \cellcolor{rank2}\paperautocell{0.9991} & \cellcolor{rank1}\paperautocell{0.0557} & \cellcolor{rank2}\paperautocell{0.2131} & \cellcolor{rank1}\paperautocell{0.0499} & \cellcolor{rank1}\paperautocell{0.0418} & \cellcolor{rank1}\paperautocell{15.624} & \paperautocell{0.9902} & \cellcolor{rank1}\paperautocell{0.0459} & \cellcolor{rank2}\paperautocell{0.1547} & \cellcolor{rank1}\paperautocell{0.0472} & \cellcolor{rank1}\paperautocell{0.0264} & \cellcolor{rank1}\paperautocell{15.558} & \paperautocell{0.9947} \\
    \bottomrule
  \end{tabular}%
  }
\end{table}

\begin{table}[H]
  \centering
  \caption{\textbf{Quantitative comparison on ScanNet++ (3D metrics).} Mean over the 25 smallest, the 25 largest, and all 50 scenes. We report the whole-mesh protocol of GenRecon~\citep{schmid2026genrecon} and the virtual-scan protocol of GaussianWrapping~\citep{gomez2026gaussianwrapping}. The top three results are highlighted.}
  \label{tab:app-scannetpp-3d}
  \scriptsize
  \setlength{\tabcolsep}{3pt}
  \definecolor{rank1}{HTML}{FFB3B3}
  \definecolor{rank2}{HTML}{FFD9B3}
  \definecolor{rank3}{HTML}{FFFFB3}
  \definecolor{emphrow}{HTML}{EDEDED}
  \ifdefined\paperautocellwidth\else\newlength{\paperautocellwidth}\fi
  \ifdefined\paperautoprobe\else\newlength{\paperautoprobe}\fi
  \settowidth{\paperautocellwidth}{\textbf{0.0000}}
  \settowidth{\paperautoprobe}{Small (25)}
  \ifdim\dimexpr\paperautoprobe-12\tabcolsep\relax>7\paperautocellwidth  \setlength{\paperautocellwidth}{\dimexpr(\paperautoprobe-12\tabcolsep)/7\relax}\fi
  \settowidth{\paperautoprobe}{Large (25)}
  \ifdim\dimexpr\paperautoprobe-12\tabcolsep\relax>7\paperautocellwidth  \setlength{\paperautocellwidth}{\dimexpr(\paperautoprobe-12\tabcolsep)/7\relax}\fi
  \settowidth{\paperautoprobe}{All (50)}
  \ifdim\dimexpr\paperautoprobe-12\tabcolsep\relax>7\paperautocellwidth  \setlength{\paperautocellwidth}{\dimexpr(\paperautoprobe-12\tabcolsep)/7\relax}\fi
  \settowidth{\paperautoprobe}{Whole mesh}
  \ifdim\dimexpr\paperautoprobe-4\tabcolsep\relax>3\paperautocellwidth  \setlength{\paperautocellwidth}{\dimexpr(\paperautoprobe-4\tabcolsep)/3\relax}\fi
  \settowidth{\paperautoprobe}{Virtual scan}
  \ifdim\dimexpr\paperautoprobe-6\tabcolsep\relax>4\paperautocellwidth  \setlength{\paperautocellwidth}{\dimexpr(\paperautoprobe-6\tabcolsep)/4\relax}\fi
  \settowidth{\paperautoprobe}{CD$\downarrow$}
  \ifdim\dimexpr\paperautoprobe-0\tabcolsep\relax>1\paperautocellwidth  \setlength{\paperautocellwidth}{\dimexpr(\paperautoprobe-0\tabcolsep)/1\relax}\fi
  \settowidth{\paperautoprobe}{F@10$\uparrow$}
  \ifdim\dimexpr\paperautoprobe-0\tabcolsep\relax>1\paperautocellwidth  \setlength{\paperautocellwidth}{\dimexpr(\paperautoprobe-0\tabcolsep)/1\relax}\fi
  \settowidth{\paperautoprobe}{NC$\uparrow$}
  \ifdim\dimexpr\paperautoprobe-0\tabcolsep\relax>1\paperautocellwidth  \setlength{\paperautocellwidth}{\dimexpr(\paperautoprobe-0\tabcolsep)/1\relax}\fi
  \settowidth{\paperautoprobe}{F@2$\uparrow$}
  \ifdim\dimexpr\paperautoprobe-0\tabcolsep\relax>1\paperautocellwidth  \setlength{\paperautocellwidth}{\dimexpr(\paperautoprobe-0\tabcolsep)/1\relax}\fi
  \providecommand{\paperautospan}[2]{\makebox[\dimexpr#1\paperautocellwidth+#1\tabcolsep+#1\tabcolsep-\tabcolsep-\tabcolsep\relax][c]{#2}}
  \providecommand{\paperautocell}[1]{\paperautospan{1}{#1}}
  \resizebox{\linewidth}{!}{%
  \begin{tabular}{lccccccccccccccccccccc}
    \toprule
     & \multicolumn{7}{c}{\paperautospan{7}{Small (25)}} & \multicolumn{7}{c}{\paperautospan{7}{Large (25)}} & \multicolumn{7}{c}{\paperautospan{7}{All (50)}} \\
    \cmidrule(lr){2-8}\cmidrule(lr){9-15}\cmidrule(lr){16-22}
     & \multicolumn{3}{c}{\paperautospan{3}{Whole mesh}} & \multicolumn{4}{c}{\paperautospan{4}{Virtual scan}} & \multicolumn{3}{c}{\paperautospan{3}{Whole mesh}} & \multicolumn{4}{c}{\paperautospan{4}{Virtual scan}} & \multicolumn{3}{c}{\paperautospan{3}{Whole mesh}} & \multicolumn{4}{c}{\paperautospan{4}{Virtual scan}} \\
    \cmidrule(lr){2-4}\cmidrule(lr){5-8}\cmidrule(lr){9-11}\cmidrule(lr){12-15}\cmidrule(lr){16-18}\cmidrule(lr){19-22}
    Method & \multicolumn{1}{c}{\paperautospan{1}{CD$\downarrow$}} & \multicolumn{1}{c}{\paperautospan{1}{F@10$\uparrow$}} & \multicolumn{1}{c}{\paperautospan{1}{NC$\uparrow$}} & \multicolumn{1}{c}{\paperautospan{1}{CD$\downarrow$}} & \multicolumn{1}{c}{\paperautospan{1}{F@2$\uparrow$}} & \multicolumn{1}{c}{\paperautospan{1}{F@10$\uparrow$}} & \multicolumn{1}{c}{\paperautospan{1}{NC$\uparrow$}} & \multicolumn{1}{c}{\paperautospan{1}{CD$\downarrow$}} & \multicolumn{1}{c}{\paperautospan{1}{F@10$\uparrow$}} & \multicolumn{1}{c}{\paperautospan{1}{NC$\uparrow$}} & \multicolumn{1}{c}{\paperautospan{1}{CD$\downarrow$}} & \multicolumn{1}{c}{\paperautospan{1}{F@2$\uparrow$}} & \multicolumn{1}{c}{\paperautospan{1}{F@10$\uparrow$}} & \multicolumn{1}{c}{\paperautospan{1}{NC$\uparrow$}} & \multicolumn{1}{c}{\paperautospan{1}{CD$\downarrow$}} & \multicolumn{1}{c}{\paperautospan{1}{F@10$\uparrow$}} & \multicolumn{1}{c}{\paperautospan{1}{NC$\uparrow$}} & \multicolumn{1}{c}{\paperautospan{1}{CD$\downarrow$}} & \multicolumn{1}{c}{\paperautospan{1}{F@2$\uparrow$}} & \multicolumn{1}{c}{\paperautospan{1}{F@10$\uparrow$}} & \multicolumn{1}{c}{\paperautospan{1}{NC$\uparrow$}} \\
    \midrule
    \multicolumn{22}{l}{\textbf{\textit{32 Views}}} \\
    2DGS & \paperautocell{0.0682} & \paperautocell{0.7878} & \paperautocell{0.6587} & \paperautocell{0.0741} & \paperautocell{0.4453} & \paperautocell{0.7698} & \paperautocell{0.7065} & \paperautocell{0.1017} & \paperautocell{0.7748} & \paperautocell{0.6399} & \paperautocell{0.0723} & \paperautocell{0.4837} & \paperautocell{0.8070} & \paperautocell{0.7181} & \paperautocell{0.0850} & \paperautocell{0.7813} & \paperautocell{0.6493} & \paperautocell{0.0732} & \paperautocell{0.4645} & \paperautocell{0.7884} & \paperautocell{0.7123} \\
    MILo & \paperautocell{0.0605} & \paperautocell{0.8043} & \paperautocell{0.6696} & \paperautocell{0.1013} & \paperautocell{0.4264} & \paperautocell{0.7381} & \paperautocell{0.6779} & \paperautocell{0.0910} & \paperautocell{0.7708} & \paperautocell{0.6532} & \paperautocell{0.1127} & \paperautocell{0.4241} & \paperautocell{0.7382} & \paperautocell{0.6910} & \paperautocell{0.0757} & \paperautocell{0.7876} & \paperautocell{0.6614} & \paperautocell{0.1070} & \paperautocell{0.4252} & \paperautocell{0.7382} & \paperautocell{0.6845} \\
    GaussianWrapping & \paperautocell{0.0633} & \paperautocell{0.8085} & \paperautocell{0.6728} & \paperautocell{0.4135} & \paperautocell{0.4433} & \paperautocell{0.7501} & \paperautocell{0.6723} & \paperautocell{0.0766} & \paperautocell{0.7909} & \paperautocell{0.6963} & \paperautocell{0.2830} & \paperautocell{0.4413} & \paperautocell{0.8073} & \paperautocell{0.7505} & \paperautocell{0.0699} & \paperautocell{0.7997} & \paperautocell{0.6846} & \paperautocell{0.3483} & \paperautocell{0.4423} & \paperautocell{0.7787} & \paperautocell{0.7114} \\
    DA3 & \cellcolor{rank1}\paperautocell{0.0166} & \cellcolor{rank2}\paperautocell{0.9775} & \cellcolor{rank1}\paperautocell{0.8675} & \cellcolor{rank1}\paperautocell{0.0113} & \cellcolor{rank1}\paperautocell{0.8829} & \cellcolor{rank2}\paperautocell{0.9915} & \cellcolor{rank3}\paperautocell{0.9200} & \paperautocell{0.0455} & \cellcolor{rank1}\paperautocell{0.9452} & \paperautocell{0.7846} & \cellcolor{rank2}\paperautocell{0.0212} & \cellcolor{rank1}\paperautocell{0.7765} & \cellcolor{rank3}\paperautocell{0.9773} & \paperautocell{0.8658} & \cellcolor{rank1}\paperautocell{0.0311} & \cellcolor{rank1}\paperautocell{0.9614} & \cellcolor{rank3}\paperautocell{0.8260} & \cellcolor{rank1}\paperautocell{0.0162} & \cellcolor{rank1}\paperautocell{0.8297} & \cellcolor{rank3}\paperautocell{0.9844} & \paperautocell{0.8929} \\
    MapAnything & \cellcolor{rank2}\paperautocell{0.0186} & \cellcolor{rank1}\paperautocell{0.9777} & \cellcolor{rank2}\paperautocell{0.8594} & \cellcolor{rank2}\paperautocell{0.0132} & \cellcolor{rank2}\paperautocell{0.8350} & \cellcolor{rank1}\paperautocell{0.9927} & \paperautocell{0.9104} & \paperautocell{0.0482} & \paperautocell{0.9225} & \paperautocell{0.7607} & \paperautocell{0.0249} & \paperautocell{0.6716} & \paperautocell{0.9662} & \paperautocell{0.8458} & \cellcolor{rank2}\paperautocell{0.0334} & \cellcolor{rank2}\paperautocell{0.9501} & \paperautocell{0.8101} & \cellcolor{rank2}\paperautocell{0.0191} & \cellcolor{rank2}\paperautocell{0.7533} & \paperautocell{0.9794} & \paperautocell{0.8781} \\
    Murre & \paperautocell{0.0368} & \paperautocell{0.9186} & \paperautocell{0.7665} & \paperautocell{0.0377} & \paperautocell{0.5269} & \paperautocell{0.9135} & \paperautocell{0.8137} & \paperautocell{0.0699} & \paperautocell{0.8409} & \paperautocell{0.6822} & \paperautocell{0.0638} & \paperautocell{0.3867} & \paperautocell{0.8258} & \paperautocell{0.7180} & \paperautocell{0.0534} & \paperautocell{0.8798} & \paperautocell{0.7244} & \paperautocell{0.0507} & \paperautocell{0.4568} & \paperautocell{0.8697} & \paperautocell{0.7659} \\
    GenRecon & \paperautocell{0.0434} & \paperautocell{0.8736} & \paperautocell{0.8289} & \paperautocell{0.0343} & \paperautocell{0.3885} & \paperautocell{0.9574} & \paperautocell{0.8984} & \paperautocell{0.1007} & \paperautocell{0.8405} & \paperautocell{0.7863} & \paperautocell{0.0604} & \paperautocell{0.2846} & \paperautocell{0.8823} & \paperautocell{0.8652} & \paperautocell{0.0721} & \paperautocell{0.8570} & \paperautocell{0.8076} & \paperautocell{0.0473} & \paperautocell{0.3366} & \paperautocell{0.9198} & \paperautocell{0.8818} \\
    \cellcolor{emphrow}\textbf{Ours (L0)} & \paperautocell{0.0275} & \paperautocell{0.9485} & \paperautocell{0.8134} & \paperautocell{0.0230} & \paperautocell{0.5816} & \paperautocell{0.9848} & \paperautocell{0.9110} & \cellcolor{rank3}\paperautocell{0.0440} & \cellcolor{rank3}\paperautocell{0.9347} & \cellcolor{rank3}\paperautocell{0.8112} & \paperautocell{0.0235} & \paperautocell{0.6336} & \paperautocell{0.9768} & \cellcolor{rank3}\paperautocell{0.9090} & \paperautocell{0.0358} & \paperautocell{0.9416} & \paperautocell{0.8123} & \paperautocell{0.0233} & \paperautocell{0.6076} & \paperautocell{0.9808} & \cellcolor{rank3}\paperautocell{0.9100} \\
    \cellcolor{emphrow}\textbf{Ours (L1)} & \paperautocell{0.0259} & \paperautocell{0.9494} & \paperautocell{0.8295} & \paperautocell{0.0200} & \paperautocell{0.6359} & \cellcolor{rank2}\paperautocell{0.9915} & \cellcolor{rank2}\paperautocell{0.9240} & \cellcolor{rank1}\paperautocell{0.0433} & \cellcolor{rank2}\paperautocell{0.9361} & \cellcolor{rank2}\paperautocell{0.8284} & \cellcolor{rank1}\paperautocell{0.0199} & \cellcolor{rank3}\paperautocell{0.6959} & \cellcolor{rank1}\paperautocell{0.9830} & \cellcolor{rank2}\paperautocell{0.9232} & \paperautocell{0.0346} & \cellcolor{rank3}\paperautocell{0.9427} & \cellcolor{rank2}\paperautocell{0.8289} & \cellcolor{rank3}\paperautocell{0.0199} & \paperautocell{0.6659} & \cellcolor{rank1}\paperautocell{0.9873} & \cellcolor{rank2}\paperautocell{0.9236} \\
    \cellcolor{emphrow}\textbf{Ours (L2)} & \cellcolor{rank3}\paperautocell{0.0251} & \cellcolor{rank3}\paperautocell{0.9502} & \cellcolor{rank3}\paperautocell{0.8371} & \cellcolor{rank3}\paperautocell{0.0190} & \cellcolor{rank3}\paperautocell{0.6718} & \paperautocell{0.9914} & \cellcolor{rank1}\paperautocell{0.9296} & \cellcolor{rank2}\paperautocell{0.0439} & \paperautocell{0.9329} & \cellcolor{rank1}\paperautocell{0.8360} & \cellcolor{rank2}\paperautocell{0.0212} & \cellcolor{rank2}\paperautocell{0.7155} & \cellcolor{rank2}\paperautocell{0.9805} & \cellcolor{rank1}\paperautocell{0.9247} & \cellcolor{rank3}\paperautocell{0.0345} & \paperautocell{0.9416} & \cellcolor{rank1}\paperautocell{0.8365} & \paperautocell{0.0201} & \cellcolor{rank3}\paperautocell{0.6936} & \cellcolor{rank2}\paperautocell{0.9859} & \cellcolor{rank1}\paperautocell{0.9272} \\
    \midrule
    \multicolumn{22}{l}{\textbf{\textit{256 Views}}} \\
    2DGS & \paperautocell{0.0583} & \paperautocell{0.8216} & \paperautocell{0.6711} & \paperautocell{0.0631} & \paperautocell{0.4776} & \paperautocell{0.8021} & \paperautocell{0.7468} & \paperautocell{0.0762} & \paperautocell{0.8463} & \paperautocell{0.6962} & \paperautocell{0.0513} & \paperautocell{0.5345} & \paperautocell{0.8668} & \paperautocell{0.8006} & \paperautocell{0.0672} & \paperautocell{0.8340} & \paperautocell{0.6836} & \paperautocell{0.0572} & \paperautocell{0.5060} & \paperautocell{0.8345} & \paperautocell{0.7737} \\
    MILo & \paperautocell{0.0511} & \paperautocell{0.8423} & \paperautocell{0.7070} & \paperautocell{0.1292} & \paperautocell{0.4963} & \paperautocell{0.8140} & \paperautocell{0.7491} & \paperautocell{0.0611} & \paperautocell{0.8428} & \paperautocell{0.7260} & \paperautocell{0.0817} & \paperautocell{0.5508} & \paperautocell{0.8508} & \paperautocell{0.7985} & \paperautocell{0.0561} & \paperautocell{0.8425} & \paperautocell{0.7165} & \paperautocell{0.1054} & \paperautocell{0.5235} & \paperautocell{0.8324} & \paperautocell{0.7738} \\
    GaussianWrapping & \paperautocell{0.0496} & \paperautocell{0.8490} & \paperautocell{0.7273} & \paperautocell{0.3414} & \paperautocell{0.5675} & \paperautocell{0.7924} & \paperautocell{0.7324} & \paperautocell{0.0607} & \paperautocell{0.8511} & \paperautocell{0.7563} & \paperautocell{0.2136} & \paperautocell{0.6075} & \paperautocell{0.8562} & \paperautocell{0.8113} & \paperautocell{0.0552} & \paperautocell{0.8501} & \paperautocell{0.7418} & \paperautocell{0.2775} & \paperautocell{0.5875} & \paperautocell{0.8243} & \paperautocell{0.7718} \\
    DA3 & \cellcolor{rank1}\paperautocell{0.0167} & \cellcolor{rank2}\paperautocell{0.9779} & \cellcolor{rank1}\paperautocell{0.8911} & \cellcolor{rank1}\paperautocell{0.0119} & \cellcolor{rank1}\paperautocell{0.8784} & \cellcolor{rank2}\paperautocell{0.9898} & \cellcolor{rank1}\paperautocell{0.9368} & \cellcolor{rank1}\paperautocell{0.0381} & \cellcolor{rank1}\paperautocell{0.9591} & \cellcolor{rank1}\paperautocell{0.8438} & \cellcolor{rank1}\paperautocell{0.0193} & \cellcolor{rank1}\paperautocell{0.7882} & \cellcolor{rank1}\paperautocell{0.9825} & \cellcolor{rank3}\paperautocell{0.9128} & \cellcolor{rank1}\paperautocell{0.0274} & \cellcolor{rank1}\paperautocell{0.9685} & \cellcolor{rank1}\paperautocell{0.8675} & \cellcolor{rank1}\paperautocell{0.0156} & \cellcolor{rank1}\paperautocell{0.8333} & \cellcolor{rank1}\paperautocell{0.9861} & \cellcolor{rank2}\paperautocell{0.9248} \\
    MapAnything & \cellcolor{rank2}\paperautocell{0.0209} & \cellcolor{rank1}\paperautocell{0.9785} & \cellcolor{rank2}\paperautocell{0.8784} & \cellcolor{rank2}\paperautocell{0.0165} & \cellcolor{rank2}\paperautocell{0.7386} & \cellcolor{rank1}\paperautocell{0.9924} & \cellcolor{rank3}\paperautocell{0.9246} & \paperautocell{0.0417} & \cellcolor{rank2}\paperautocell{0.9460} & \paperautocell{0.8181} & \paperautocell{0.0255} & \paperautocell{0.5761} & \paperautocell{0.9746} & \paperautocell{0.8916} & \cellcolor{rank2}\paperautocell{0.0313} & \cellcolor{rank2}\paperautocell{0.9623} & \cellcolor{rank2}\paperautocell{0.8482} & \cellcolor{rank2}\paperautocell{0.0210} & \cellcolor{rank3}\paperautocell{0.6573} & \cellcolor{rank2}\paperautocell{0.9835} & \paperautocell{0.9081} \\
    Murre & \paperautocell{0.0420} & \paperautocell{0.8778} & \paperautocell{0.7754} & \paperautocell{0.0434} & \paperautocell{0.5441} & \paperautocell{0.8909} & \paperautocell{0.8301} & \paperautocell{0.0776} & \paperautocell{0.7827} & \paperautocell{0.6667} & \paperautocell{0.0916} & \paperautocell{0.3471} & \paperautocell{0.7447} & \paperautocell{0.6952} & \paperautocell{0.0598} & \paperautocell{0.8302} & \paperautocell{0.7210} & \paperautocell{0.0675} & \paperautocell{0.4456} & \paperautocell{0.8178} & \paperautocell{0.7626} \\
    GenRecon & \paperautocell{0.0427} & \paperautocell{0.8822} & \paperautocell{0.8303} & \paperautocell{0.0343} & \paperautocell{0.3628} & \paperautocell{0.9680} & \paperautocell{0.8997} & \paperautocell{0.0975} & \paperautocell{0.8602} & \paperautocell{0.7982} & \paperautocell{0.0611} & \paperautocell{0.2834} & \paperautocell{0.8997} & \paperautocell{0.8723} & \paperautocell{0.0701} & \paperautocell{0.8712} & \paperautocell{0.8142} & \paperautocell{0.0477} & \paperautocell{0.3231} & \paperautocell{0.9339} & \paperautocell{0.8860} \\
    \cellcolor{emphrow}\textbf{Ours (L0)} & \paperautocell{0.0287} & \paperautocell{0.9519} & \paperautocell{0.8093} & \paperautocell{0.0279} & \paperautocell{0.4897} & \paperautocell{0.9777} & \paperautocell{0.9086} & \cellcolor{rank2}\paperautocell{0.0382} & \cellcolor{rank3}\paperautocell{0.9392} & \paperautocell{0.8136} & \paperautocell{0.0233} & \paperautocell{0.6251} & \paperautocell{0.9787} & \paperautocell{0.9125} & \paperautocell{0.0334} & \paperautocell{0.9456} & \paperautocell{0.8114} & \paperautocell{0.0256} & \paperautocell{0.5574} & \paperautocell{0.9782} & \paperautocell{0.9106} \\
    \cellcolor{emphrow}\textbf{Ours (L1)} & \paperautocell{0.0268} & \cellcolor{rank3}\paperautocell{0.9533} & \paperautocell{0.8233} & \paperautocell{0.0238} & \paperautocell{0.5726} & \cellcolor{rank3}\paperautocell{0.9859} & \paperautocell{0.9211} & \paperautocell{0.0402} & \paperautocell{0.9388} & \cellcolor{rank3}\paperautocell{0.8304} & \cellcolor{rank3}\paperautocell{0.0217} & \cellcolor{rank3}\paperautocell{0.6874} & \cellcolor{rank3}\paperautocell{0.9811} & \cellcolor{rank2}\paperautocell{0.9224} & \paperautocell{0.0335} & \cellcolor{rank3}\paperautocell{0.9460} & \paperautocell{0.8268} & \paperautocell{0.0227} & \paperautocell{0.6300} & \cellcolor{rank2}\paperautocell{0.9835} & \cellcolor{rank3}\paperautocell{0.9217} \\
    \cellcolor{emphrow}\textbf{Ours (L2)} & \cellcolor{rank3}\paperautocell{0.0260} & \cellcolor{rank3}\paperautocell{0.9533} & \cellcolor{rank3}\paperautocell{0.8324} & \cellcolor{rank3}\paperautocell{0.0234} & \cellcolor{rank3}\paperautocell{0.6167} & \paperautocell{0.9843} & \cellcolor{rank2}\paperautocell{0.9260} & \cellcolor{rank3}\paperautocell{0.0401} & \paperautocell{0.9377} & \cellcolor{rank2}\paperautocell{0.8419} & \cellcolor{rank2}\paperautocell{0.0202} & \cellcolor{rank2}\paperautocell{0.7176} & \cellcolor{rank2}\paperautocell{0.9821} & \cellcolor{rank1}\paperautocell{0.9283} & \cellcolor{rank3}\paperautocell{0.0330} & \paperautocell{0.9455} & \cellcolor{rank3}\paperautocell{0.8372} & \cellcolor{rank3}\paperautocell{0.0218} & \cellcolor{rank2}\paperautocell{0.6672} & \paperautocell{0.9832} & \cellcolor{rank1}\paperautocell{0.9271} \\
    \bottomrule
  \end{tabular}%
  }
\end{table}

\begin{figure}[p]
  \centering
  \setlength{\tabcolsep}{0.5pt}
  \renewcommand{\arraystretch}{1.0}
  \begin{tabular}{cccc}
    \includegraphics[width=\dimexpr(\textwidth-4pt)/4\relax]{imgs/scannetpp_qualitative_fps32/7b6477cb95/gt_rgb_v0.jpg} & \includegraphics[width=\dimexpr(\textwidth-4pt)/4\relax]{imgs/scannetpp_qualitative_fps32/7b6477cb95/2dgs_v0.jpg} & \includegraphics[width=\dimexpr(\textwidth-4pt)/4\relax]{imgs/scannetpp_qualitative_fps32/7b6477cb95/milo_v0.jpg} & \includegraphics[width=\dimexpr(\textwidth-4pt)/4\relax]{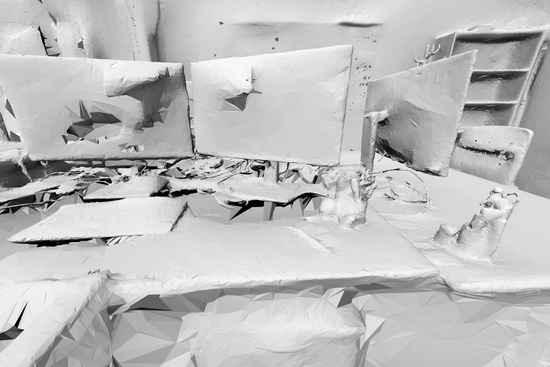} \\[-4.5pt]
    \makebox[\dimexpr(\textwidth-4pt)/4\relax][c]{\scriptsize Image} & \makebox[\dimexpr(\textwidth-4pt)/4\relax][c]{\scriptsize 2DGS} & \makebox[\dimexpr(\textwidth-4pt)/4\relax][c]{\scriptsize MILo} & \makebox[\dimexpr(\textwidth-4pt)/4\relax][c]{\scriptsize GaussianWrapping} \\[3pt]
    \includegraphics[width=\dimexpr(\textwidth-4pt)/4\relax]{imgs/scannetpp_qualitative_fps32/7b6477cb95/da3_v0.jpg} & \includegraphics[width=\dimexpr(\textwidth-4pt)/4\relax]{imgs/scannetpp_qualitative_fps32/7b6477cb95/mapanything_v0.jpg} & \includegraphics[width=\dimexpr(\textwidth-4pt)/4\relax]{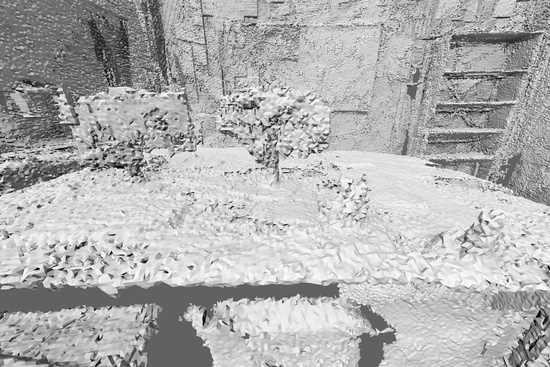} & \includegraphics[width=\dimexpr(\textwidth-4pt)/4\relax]{imgs/scannetpp_qualitative_fps32/7b6477cb95/genrecon_v0.jpg} \\[-4.5pt]
    \makebox[\dimexpr(\textwidth-4pt)/4\relax][c]{\scriptsize DA3} & \makebox[\dimexpr(\textwidth-4pt)/4\relax][c]{\scriptsize MapAnything} & \makebox[\dimexpr(\textwidth-4pt)/4\relax][c]{\scriptsize Murre} & \makebox[\dimexpr(\textwidth-4pt)/4\relax][c]{\scriptsize GenRecon} \\[3pt]
    \includegraphics[width=\dimexpr(\textwidth-4pt)/4\relax]{imgs/scannetpp_qualitative_fps32/7b6477cb95/ours_g004_v0.jpg} & \includegraphics[width=\dimexpr(\textwidth-4pt)/4\relax]{imgs/scannetpp_qualitative_fps32/7b6477cb95/ours_g002_v0.jpg} & \includegraphics[width=\dimexpr(\textwidth-4pt)/4\relax]{imgs/scannetpp_qualitative_fps32/7b6477cb95/ours_g001_v0.jpg} & \includegraphics[width=\dimexpr(\textwidth-4pt)/4\relax]{imgs/scannetpp_qualitative_fps32/7b6477cb95/gt_v0.jpg} \\[-4.5pt]
    \makebox[\dimexpr(\textwidth-4pt)/4\relax][c]{\scriptsize Ours (L0)} & \makebox[\dimexpr(\textwidth-4pt)/4\relax][c]{\scriptsize Ours (L1)} & \makebox[\dimexpr(\textwidth-4pt)/4\relax][c]{\scriptsize Ours (L2)} & \makebox[\dimexpr(\textwidth-4pt)/4\relax][c]{\scriptsize Ground truth} \\[8pt]
    \includegraphics[width=\dimexpr(\textwidth-4pt)/4\relax]{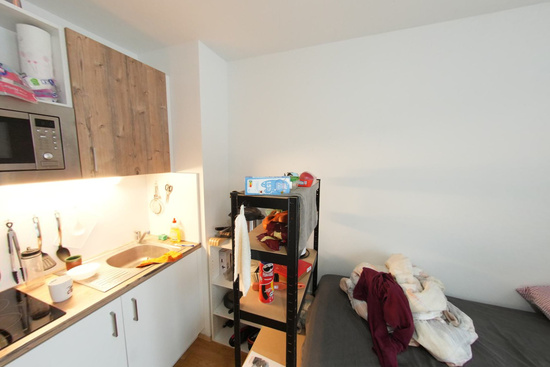} & \includegraphics[width=\dimexpr(\textwidth-4pt)/4\relax]{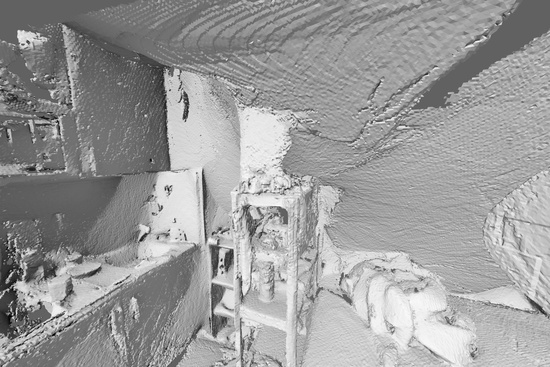} & \includegraphics[width=\dimexpr(\textwidth-4pt)/4\relax]{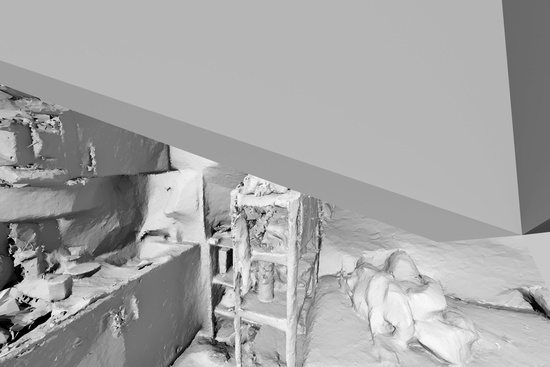} & \includegraphics[width=\dimexpr(\textwidth-4pt)/4\relax]{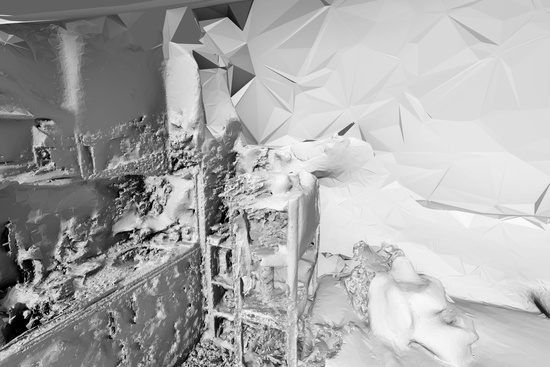} \\[-4.5pt]
    \makebox[\dimexpr(\textwidth-4pt)/4\relax][c]{\scriptsize Image} & \makebox[\dimexpr(\textwidth-4pt)/4\relax][c]{\scriptsize 2DGS} & \makebox[\dimexpr(\textwidth-4pt)/4\relax][c]{\scriptsize MILo} & \makebox[\dimexpr(\textwidth-4pt)/4\relax][c]{\scriptsize GaussianWrapping} \\[3pt]
    \includegraphics[width=\dimexpr(\textwidth-4pt)/4\relax]{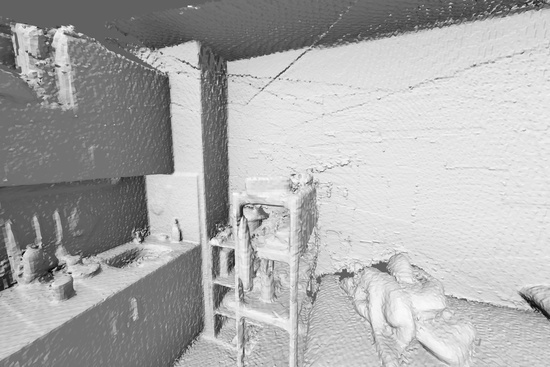} & \includegraphics[width=\dimexpr(\textwidth-4pt)/4\relax]{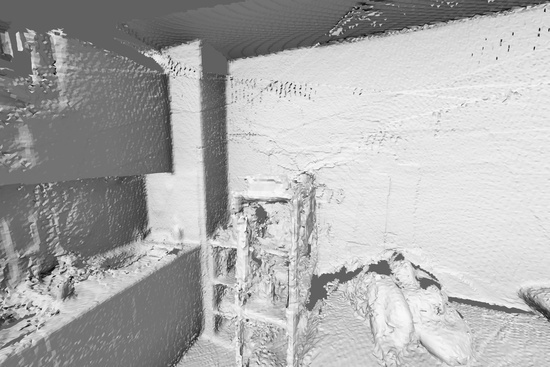} & \includegraphics[width=\dimexpr(\textwidth-4pt)/4\relax]{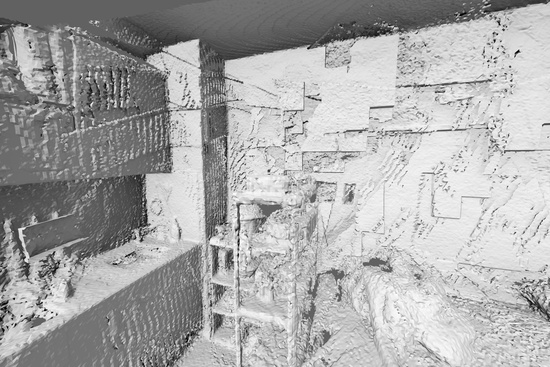} & \includegraphics[width=\dimexpr(\textwidth-4pt)/4\relax]{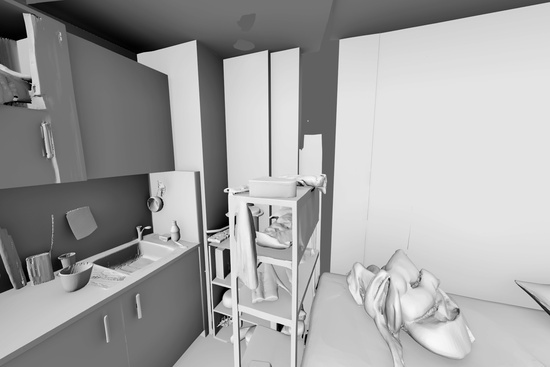} \\[-4.5pt]
    \makebox[\dimexpr(\textwidth-4pt)/4\relax][c]{\scriptsize DA3} & \makebox[\dimexpr(\textwidth-4pt)/4\relax][c]{\scriptsize MapAnything} & \makebox[\dimexpr(\textwidth-4pt)/4\relax][c]{\scriptsize Murre} & \makebox[\dimexpr(\textwidth-4pt)/4\relax][c]{\scriptsize GenRecon} \\[3pt]
    \includegraphics[width=\dimexpr(\textwidth-4pt)/4\relax]{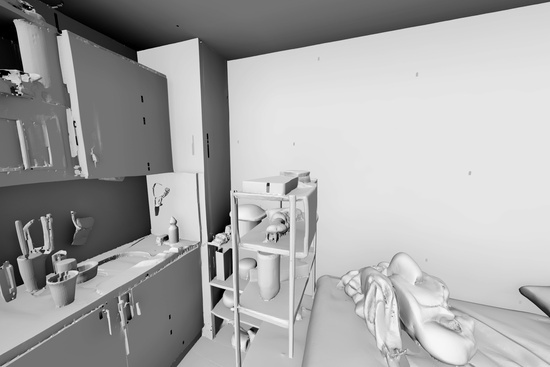} & \includegraphics[width=\dimexpr(\textwidth-4pt)/4\relax]{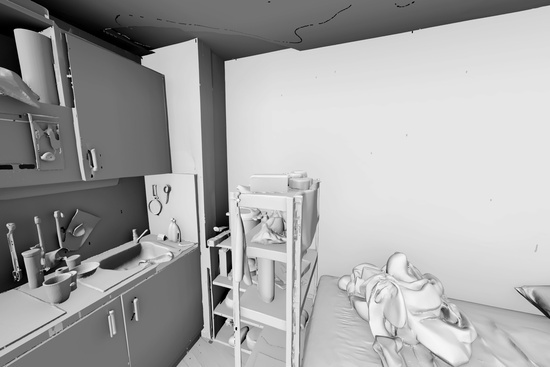} & \includegraphics[width=\dimexpr(\textwidth-4pt)/4\relax]{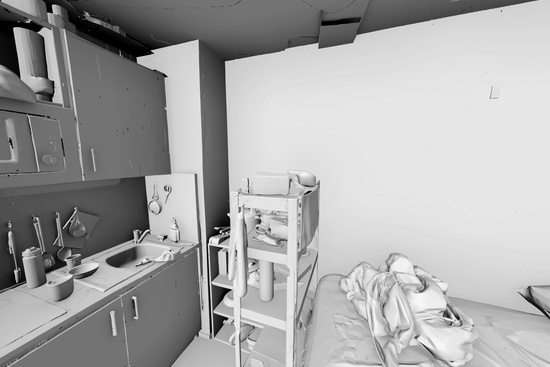} & \includegraphics[width=\dimexpr(\textwidth-4pt)/4\relax]{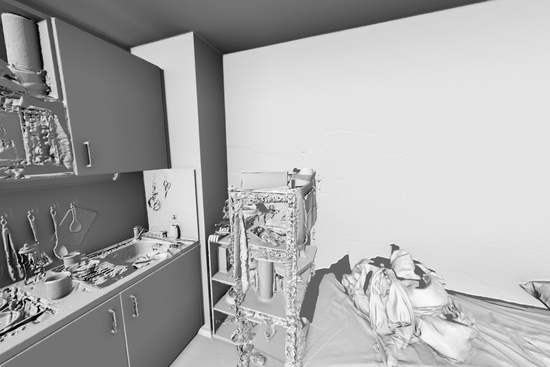} \\[-4.5pt]
    \makebox[\dimexpr(\textwidth-4pt)/4\relax][c]{\scriptsize Ours (L0)} & \makebox[\dimexpr(\textwidth-4pt)/4\relax][c]{\scriptsize Ours (L1)} & \makebox[\dimexpr(\textwidth-4pt)/4\relax][c]{\scriptsize Ours (L2)} & \makebox[\dimexpr(\textwidth-4pt)/4\relax][c]{\scriptsize Ground truth} \\
  \end{tabular}
  \caption{\textbf{Qualitative comparison on ScanNet++ with 32 input views.} All meshes are rendered from a held-out view of each scene.}
  \label{fig:scannetpp-qualitative-fps32:1}
\end{figure}

\begin{figure}[p]
  \centering
  \setlength{\tabcolsep}{0.5pt}
  \renewcommand{\arraystretch}{1.0}
  \begin{tabular}{cccc}
    \includegraphics[width=\dimexpr(\textwidth-4pt)/4\relax]{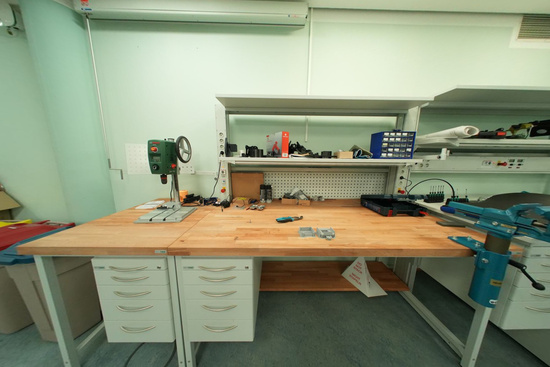} & \includegraphics[width=\dimexpr(\textwidth-4pt)/4\relax]{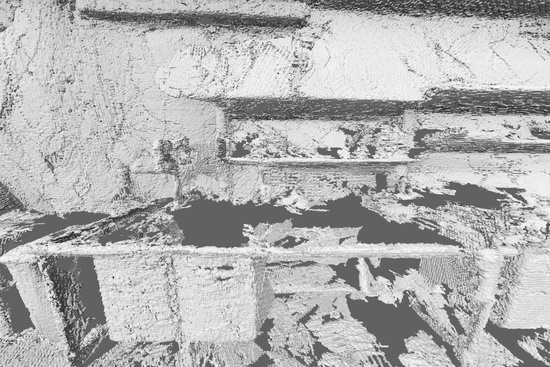} & \includegraphics[width=\dimexpr(\textwidth-4pt)/4\relax]{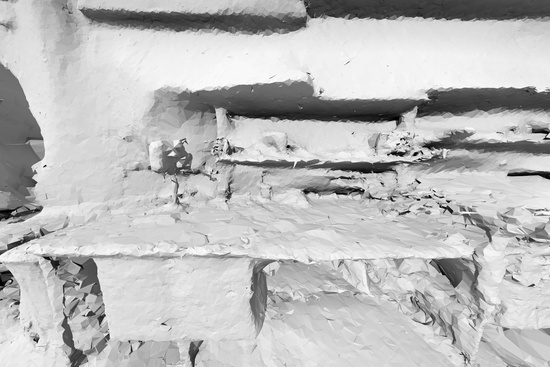} & \includegraphics[width=\dimexpr(\textwidth-4pt)/4\relax]{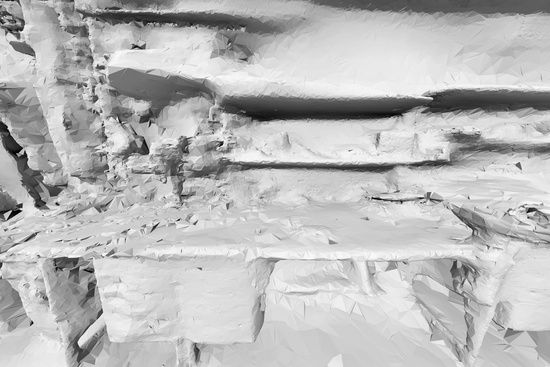} \\[-4.5pt]
    \makebox[\dimexpr(\textwidth-4pt)/4\relax][c]{\scriptsize Image} & \makebox[\dimexpr(\textwidth-4pt)/4\relax][c]{\scriptsize 2DGS} & \makebox[\dimexpr(\textwidth-4pt)/4\relax][c]{\scriptsize MILo} & \makebox[\dimexpr(\textwidth-4pt)/4\relax][c]{\scriptsize GaussianWrapping} \\[3pt]
    \includegraphics[width=\dimexpr(\textwidth-4pt)/4\relax]{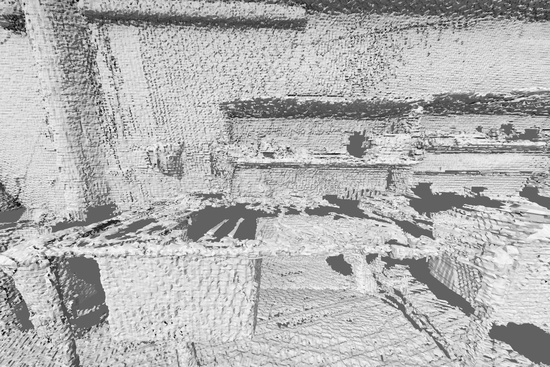} & \includegraphics[width=\dimexpr(\textwidth-4pt)/4\relax]{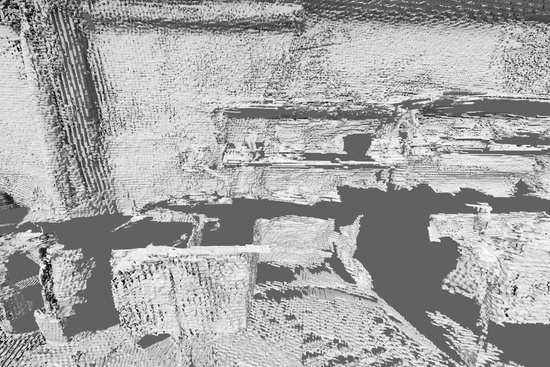} & \includegraphics[width=\dimexpr(\textwidth-4pt)/4\relax]{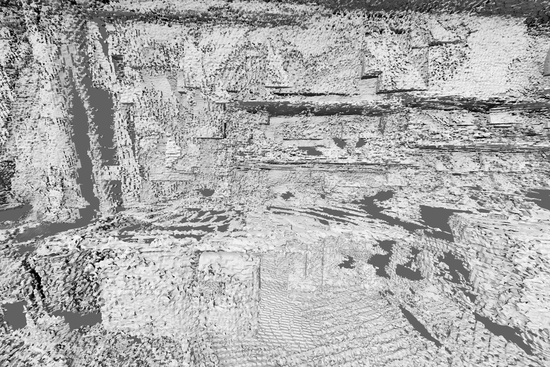} & \includegraphics[width=\dimexpr(\textwidth-4pt)/4\relax]{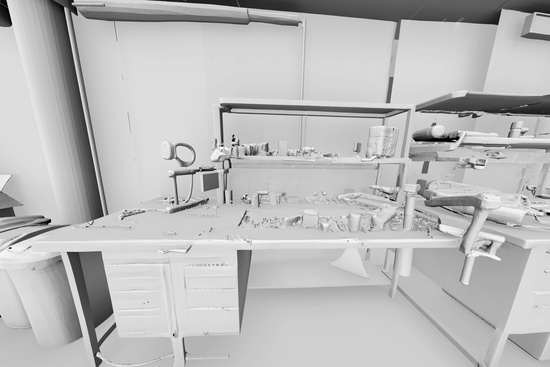} \\[-4.5pt]
    \makebox[\dimexpr(\textwidth-4pt)/4\relax][c]{\scriptsize DA3} & \makebox[\dimexpr(\textwidth-4pt)/4\relax][c]{\scriptsize MapAnything} & \makebox[\dimexpr(\textwidth-4pt)/4\relax][c]{\scriptsize Murre} & \makebox[\dimexpr(\textwidth-4pt)/4\relax][c]{\scriptsize GenRecon} \\[3pt]
    \includegraphics[width=\dimexpr(\textwidth-4pt)/4\relax]{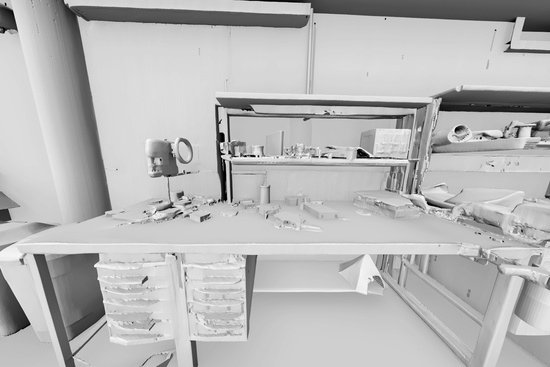} & \includegraphics[width=\dimexpr(\textwidth-4pt)/4\relax]{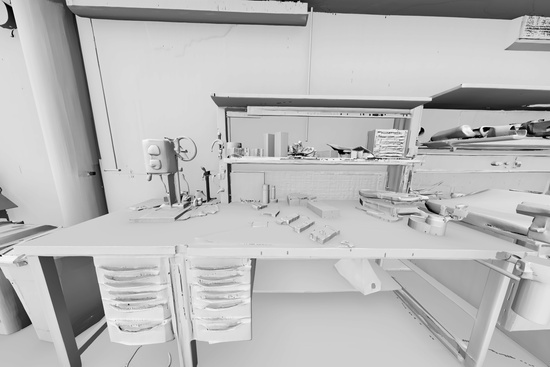} & \includegraphics[width=\dimexpr(\textwidth-4pt)/4\relax]{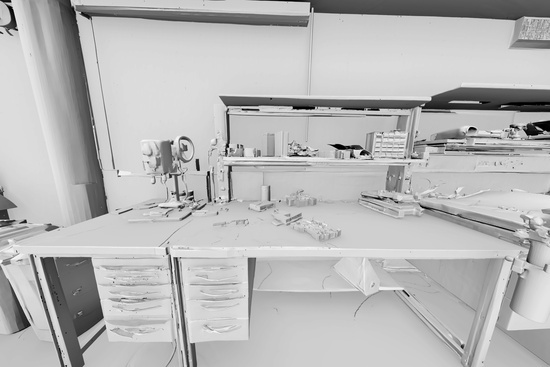} & \includegraphics[width=\dimexpr(\textwidth-4pt)/4\relax]{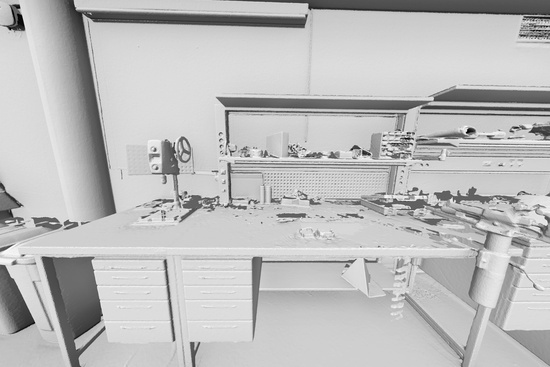} \\[-4.5pt]
    \makebox[\dimexpr(\textwidth-4pt)/4\relax][c]{\scriptsize Ours (L0)} & \makebox[\dimexpr(\textwidth-4pt)/4\relax][c]{\scriptsize Ours (L1)} & \makebox[\dimexpr(\textwidth-4pt)/4\relax][c]{\scriptsize Ours (L2)} & \makebox[\dimexpr(\textwidth-4pt)/4\relax][c]{\scriptsize Ground truth} \\[8pt]
    \includegraphics[width=\dimexpr(\textwidth-4pt)/4\relax]{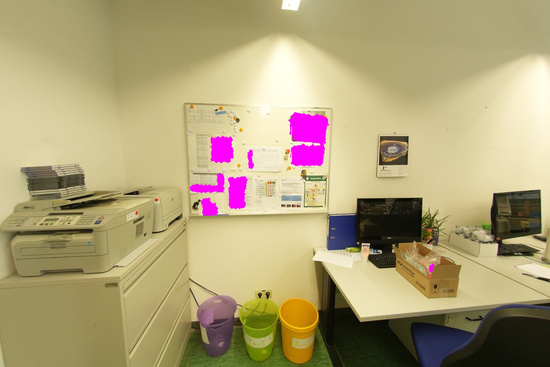} & \includegraphics[width=\dimexpr(\textwidth-4pt)/4\relax]{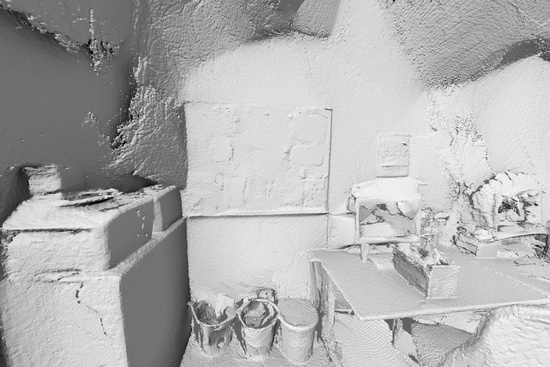} & \includegraphics[width=\dimexpr(\textwidth-4pt)/4\relax]{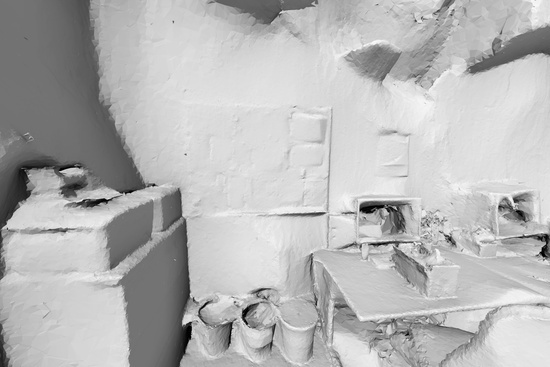} & \includegraphics[width=\dimexpr(\textwidth-4pt)/4\relax]{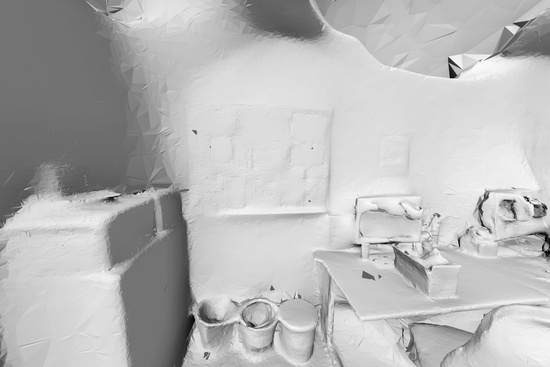} \\[-4.5pt]
    \makebox[\dimexpr(\textwidth-4pt)/4\relax][c]{\scriptsize Image} & \makebox[\dimexpr(\textwidth-4pt)/4\relax][c]{\scriptsize 2DGS} & \makebox[\dimexpr(\textwidth-4pt)/4\relax][c]{\scriptsize MILo} & \makebox[\dimexpr(\textwidth-4pt)/4\relax][c]{\scriptsize GaussianWrapping} \\[3pt]
    \includegraphics[width=\dimexpr(\textwidth-4pt)/4\relax]{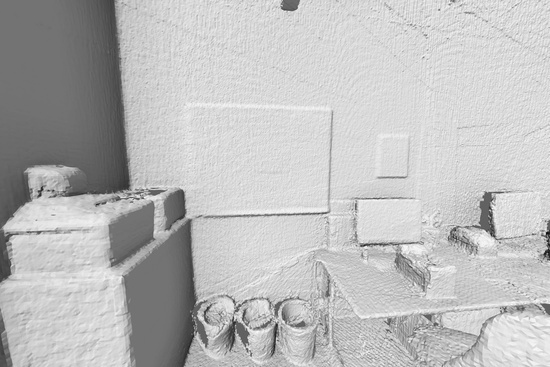} & \includegraphics[width=\dimexpr(\textwidth-4pt)/4\relax]{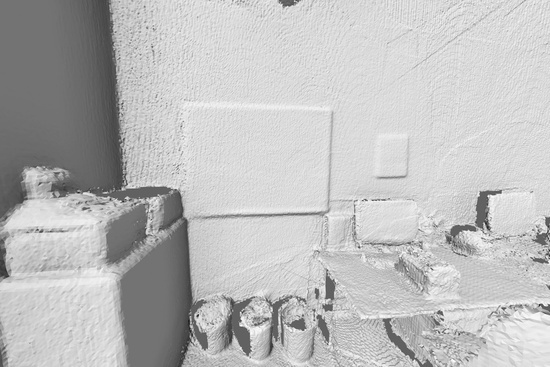} & \includegraphics[width=\dimexpr(\textwidth-4pt)/4\relax]{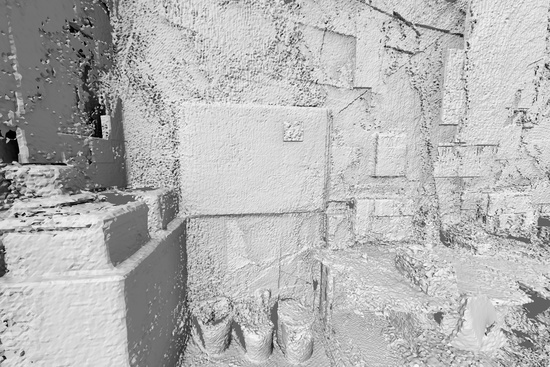} & \includegraphics[width=\dimexpr(\textwidth-4pt)/4\relax]{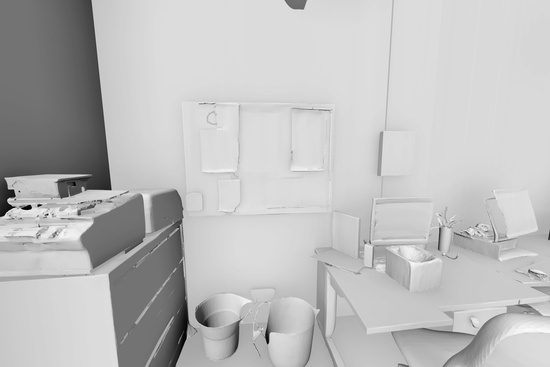} \\[-4.5pt]
    \makebox[\dimexpr(\textwidth-4pt)/4\relax][c]{\scriptsize DA3} & \makebox[\dimexpr(\textwidth-4pt)/4\relax][c]{\scriptsize MapAnything} & \makebox[\dimexpr(\textwidth-4pt)/4\relax][c]{\scriptsize Murre} & \makebox[\dimexpr(\textwidth-4pt)/4\relax][c]{\scriptsize GenRecon} \\[3pt]
    \includegraphics[width=\dimexpr(\textwidth-4pt)/4\relax]{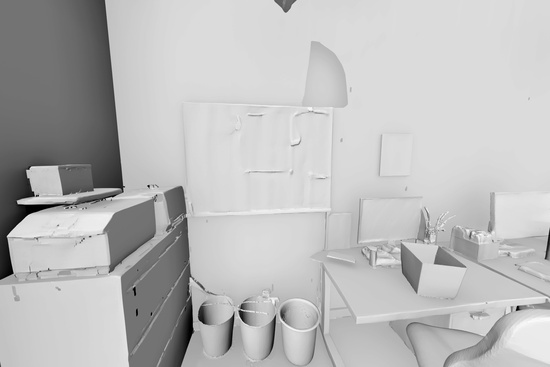} & \includegraphics[width=\dimexpr(\textwidth-4pt)/4\relax]{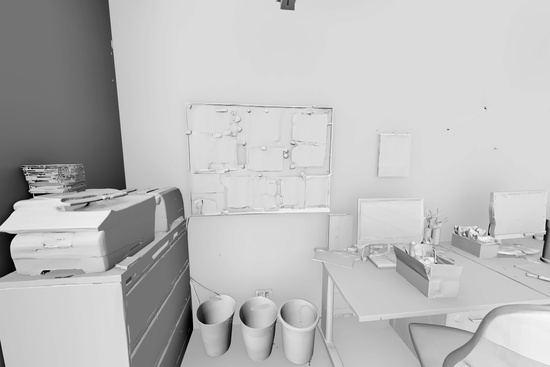} & \includegraphics[width=\dimexpr(\textwidth-4pt)/4\relax]{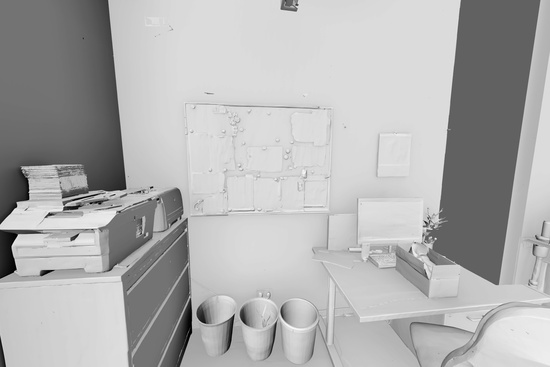} & \includegraphics[width=\dimexpr(\textwidth-4pt)/4\relax]{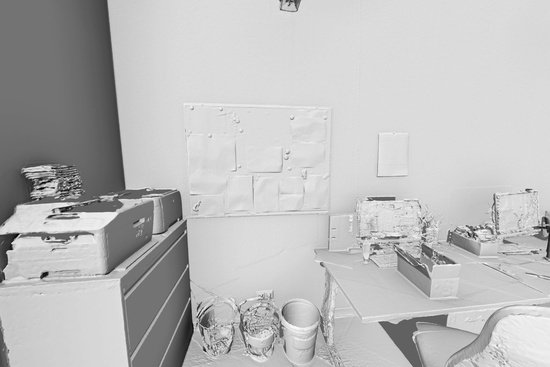} \\[-4.5pt]
    \makebox[\dimexpr(\textwidth-4pt)/4\relax][c]{\scriptsize Ours (L0)} & \makebox[\dimexpr(\textwidth-4pt)/4\relax][c]{\scriptsize Ours (L1)} & \makebox[\dimexpr(\textwidth-4pt)/4\relax][c]{\scriptsize Ours (L2)} & \makebox[\dimexpr(\textwidth-4pt)/4\relax][c]{\scriptsize Ground truth} \\
  \end{tabular}
  \caption{\textbf{Qualitative comparison on ScanNet++ with 32 input views.} All meshes are rendered from a held-out view of each scene.}
  \label{fig:scannetpp-qualitative-fps32:2}
\end{figure}

\begin{figure}[p]
  \centering
  \setlength{\tabcolsep}{0.5pt}
  \renewcommand{\arraystretch}{1.0}
  \begin{tabular}{cccc}
    \includegraphics[width=\dimexpr(\textwidth-4pt)/4\relax]{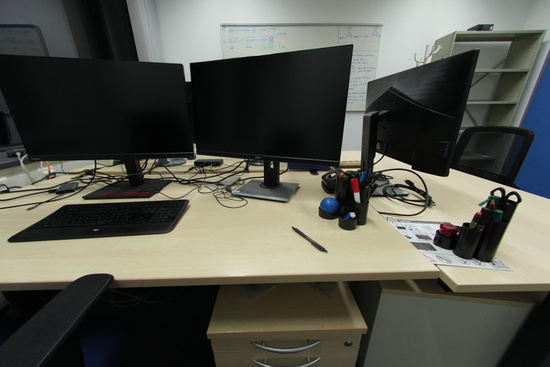} & \includegraphics[width=\dimexpr(\textwidth-4pt)/4\relax]{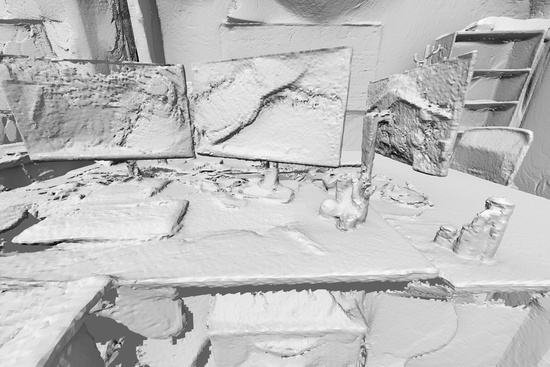} & \includegraphics[width=\dimexpr(\textwidth-4pt)/4\relax]{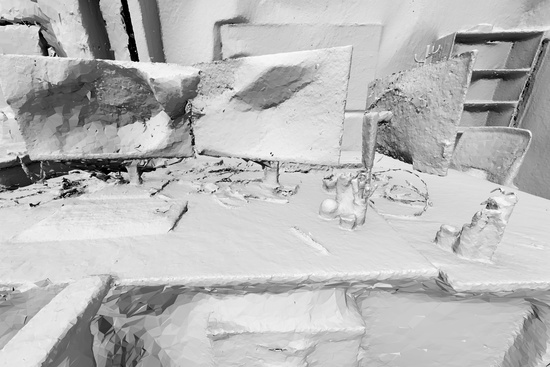} & \includegraphics[width=\dimexpr(\textwidth-4pt)/4\relax]{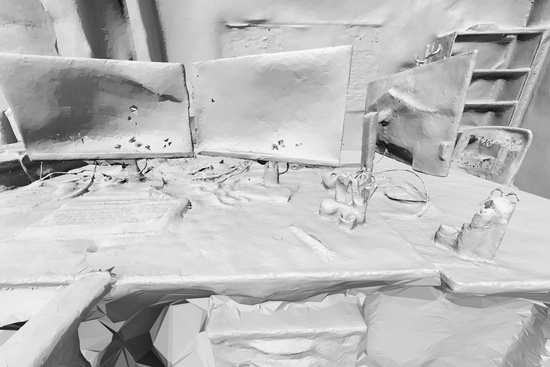} \\[-4.5pt]
    \makebox[\dimexpr(\textwidth-4pt)/4\relax][c]{\scriptsize Image} & \makebox[\dimexpr(\textwidth-4pt)/4\relax][c]{\scriptsize 2DGS} & \makebox[\dimexpr(\textwidth-4pt)/4\relax][c]{\scriptsize MILo} & \makebox[\dimexpr(\textwidth-4pt)/4\relax][c]{\scriptsize GaussianWrapping} \\[3pt]
    \includegraphics[width=\dimexpr(\textwidth-4pt)/4\relax]{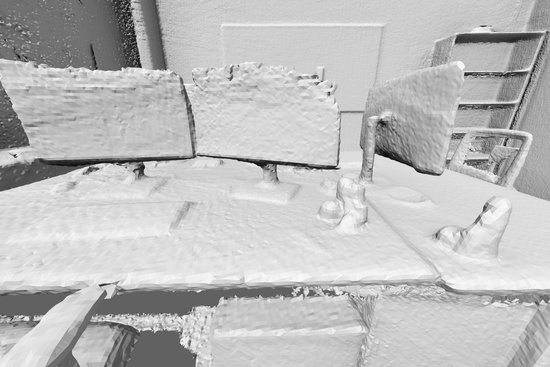} & \includegraphics[width=\dimexpr(\textwidth-4pt)/4\relax]{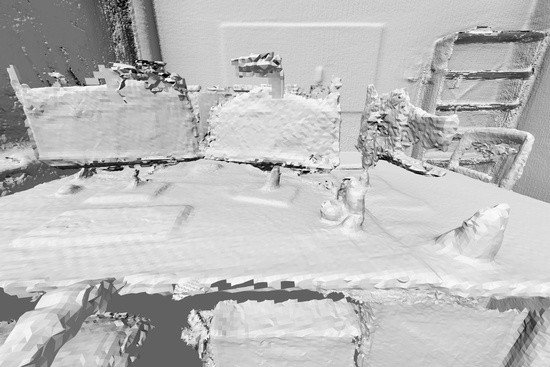} & \includegraphics[width=\dimexpr(\textwidth-4pt)/4\relax]{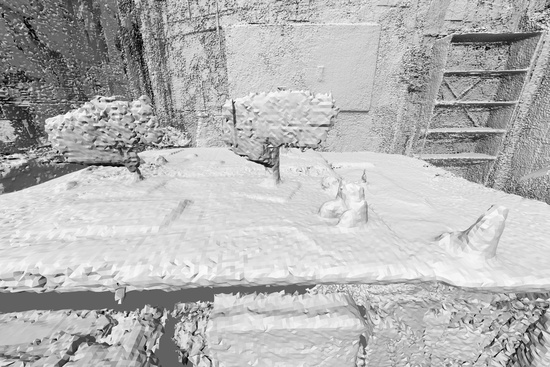} & \includegraphics[width=\dimexpr(\textwidth-4pt)/4\relax]{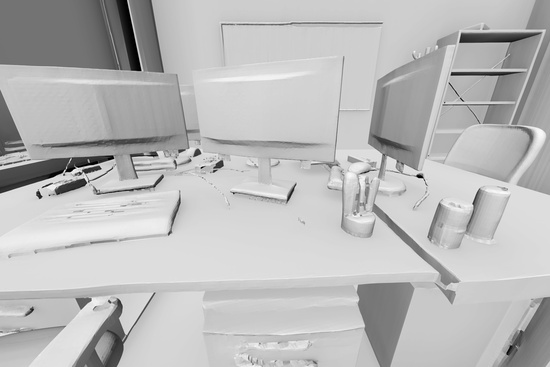} \\[-4.5pt]
    \makebox[\dimexpr(\textwidth-4pt)/4\relax][c]{\scriptsize DA3} & \makebox[\dimexpr(\textwidth-4pt)/4\relax][c]{\scriptsize MapAnything} & \makebox[\dimexpr(\textwidth-4pt)/4\relax][c]{\scriptsize Murre} & \makebox[\dimexpr(\textwidth-4pt)/4\relax][c]{\scriptsize GenRecon} \\[3pt]
    \includegraphics[width=\dimexpr(\textwidth-4pt)/4\relax]{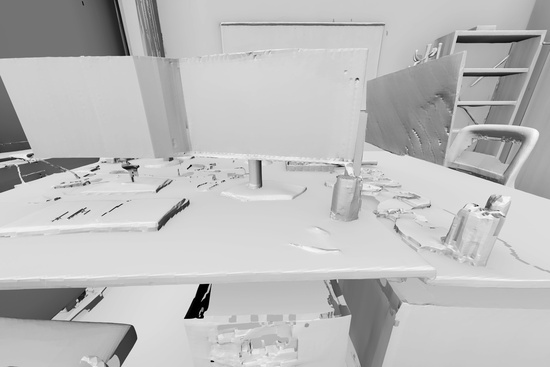} & \includegraphics[width=\dimexpr(\textwidth-4pt)/4\relax]{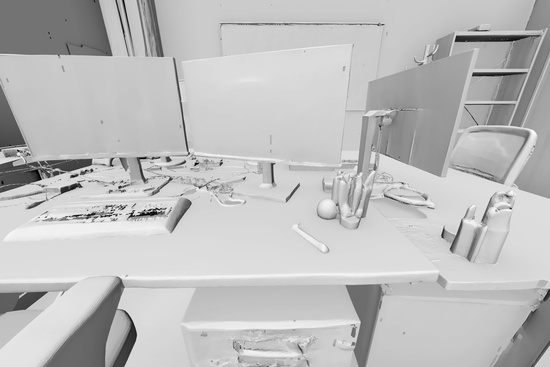} & \includegraphics[width=\dimexpr(\textwidth-4pt)/4\relax]{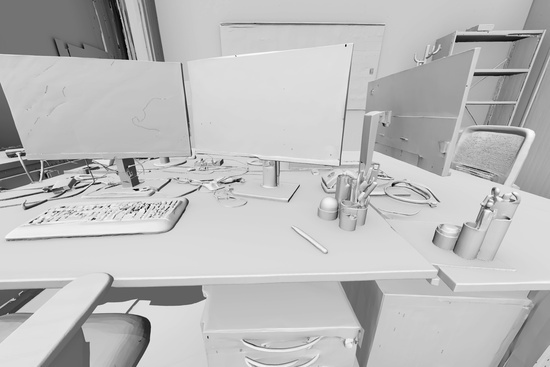} & \includegraphics[width=\dimexpr(\textwidth-4pt)/4\relax]{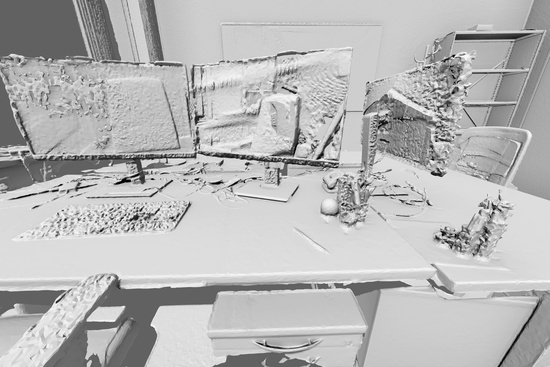} \\[-4.5pt]
    \makebox[\dimexpr(\textwidth-4pt)/4\relax][c]{\scriptsize Ours (L0)} & \makebox[\dimexpr(\textwidth-4pt)/4\relax][c]{\scriptsize Ours (L1)} & \makebox[\dimexpr(\textwidth-4pt)/4\relax][c]{\scriptsize Ours (L2)} & \makebox[\dimexpr(\textwidth-4pt)/4\relax][c]{\scriptsize Ground truth} \\[8pt]
    \includegraphics[width=\dimexpr(\textwidth-4pt)/4\relax]{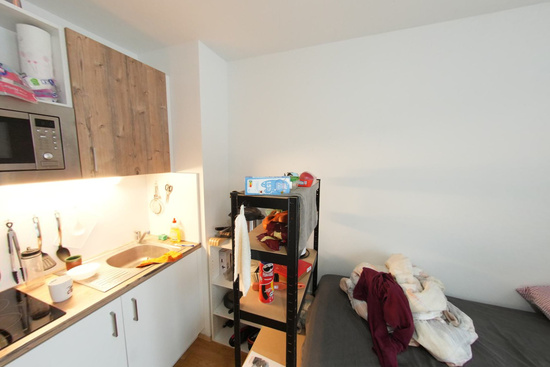} & \includegraphics[width=\dimexpr(\textwidth-4pt)/4\relax]{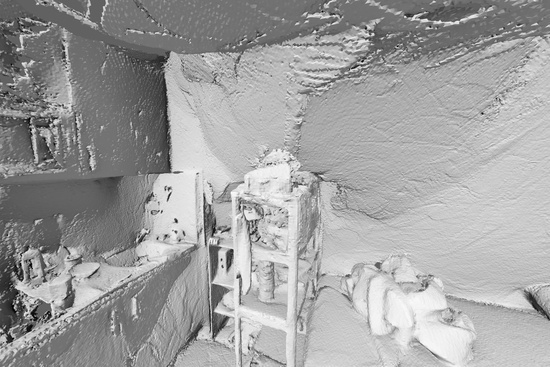} & \includegraphics[width=\dimexpr(\textwidth-4pt)/4\relax]{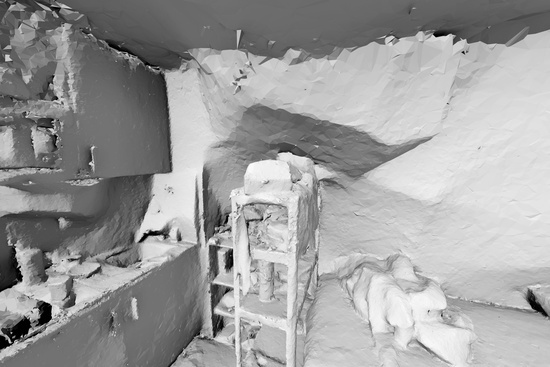} & \includegraphics[width=\dimexpr(\textwidth-4pt)/4\relax]{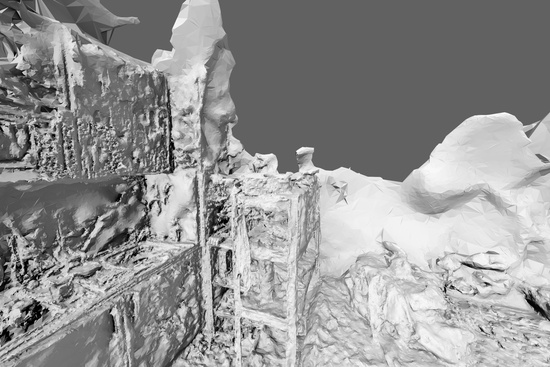} \\[-4.5pt]
    \makebox[\dimexpr(\textwidth-4pt)/4\relax][c]{\scriptsize Image} & \makebox[\dimexpr(\textwidth-4pt)/4\relax][c]{\scriptsize 2DGS} & \makebox[\dimexpr(\textwidth-4pt)/4\relax][c]{\scriptsize MILo} & \makebox[\dimexpr(\textwidth-4pt)/4\relax][c]{\scriptsize GaussianWrapping} \\[3pt]
    \includegraphics[width=\dimexpr(\textwidth-4pt)/4\relax]{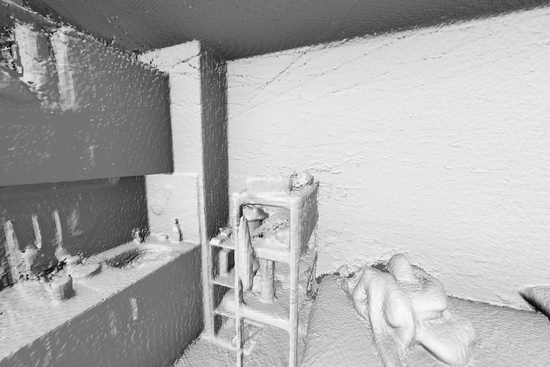} & \includegraphics[width=\dimexpr(\textwidth-4pt)/4\relax]{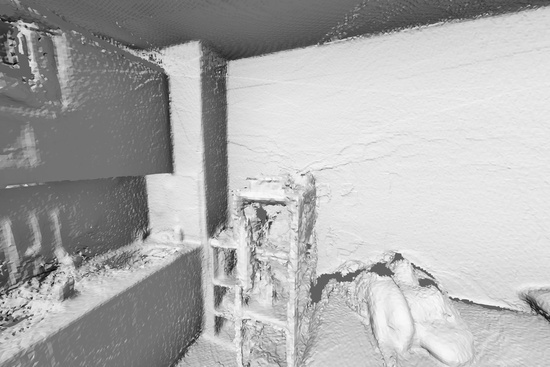} & \includegraphics[width=\dimexpr(\textwidth-4pt)/4\relax]{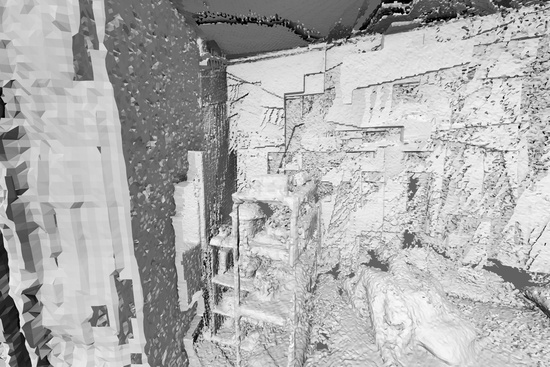} & \includegraphics[width=\dimexpr(\textwidth-4pt)/4\relax]{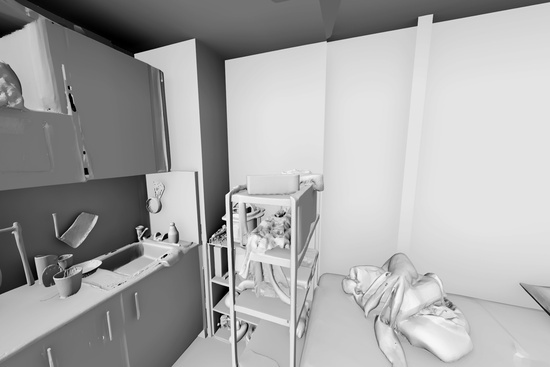} \\[-4.5pt]
    \makebox[\dimexpr(\textwidth-4pt)/4\relax][c]{\scriptsize DA3} & \makebox[\dimexpr(\textwidth-4pt)/4\relax][c]{\scriptsize MapAnything} & \makebox[\dimexpr(\textwidth-4pt)/4\relax][c]{\scriptsize Murre} & \makebox[\dimexpr(\textwidth-4pt)/4\relax][c]{\scriptsize GenRecon} \\[3pt]
    \includegraphics[width=\dimexpr(\textwidth-4pt)/4\relax]{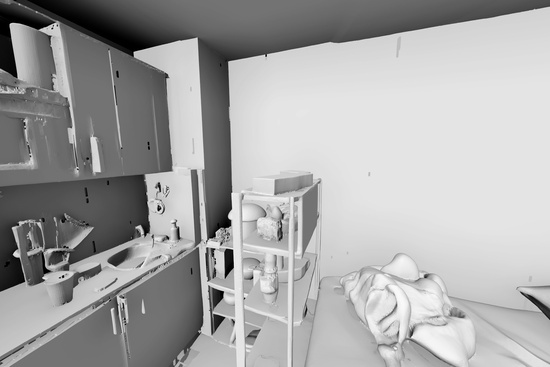} & \includegraphics[width=\dimexpr(\textwidth-4pt)/4\relax]{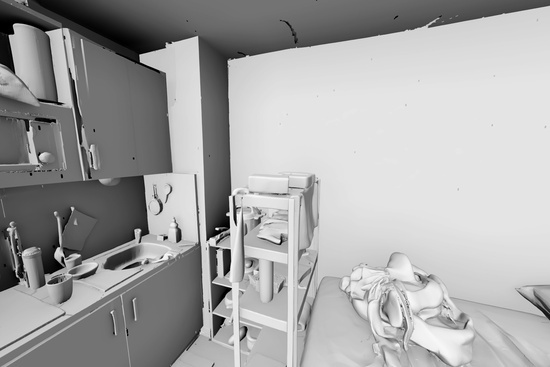} & \includegraphics[width=\dimexpr(\textwidth-4pt)/4\relax]{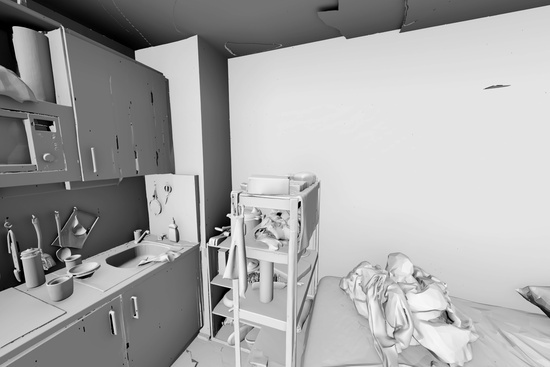} & \includegraphics[width=\dimexpr(\textwidth-4pt)/4\relax]{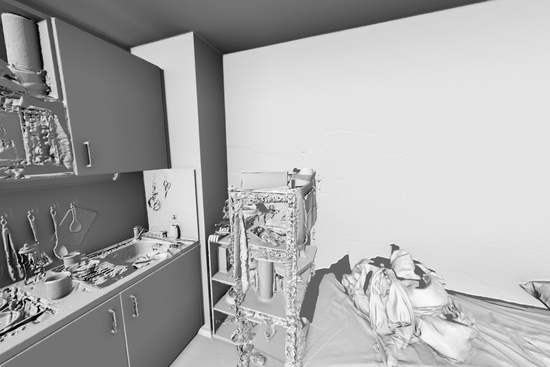} \\[-4.5pt]
    \makebox[\dimexpr(\textwidth-4pt)/4\relax][c]{\scriptsize Ours (L0)} & \makebox[\dimexpr(\textwidth-4pt)/4\relax][c]{\scriptsize Ours (L1)} & \makebox[\dimexpr(\textwidth-4pt)/4\relax][c]{\scriptsize Ours (L2)} & \makebox[\dimexpr(\textwidth-4pt)/4\relax][c]{\scriptsize Ground truth} \\
  \end{tabular}
  \caption{\textbf{Qualitative comparison on ScanNet++ with 256 input views.} All meshes are rendered from a held-out view of each scene.}
  \label{fig:scannetpp-qualitative-fps256:1}
\end{figure}

\begin{figure}[p]
  \centering
  \setlength{\tabcolsep}{0.5pt}
  \renewcommand{\arraystretch}{1.0}
  \begin{tabular}{cccc}
    \includegraphics[width=\dimexpr(\textwidth-4pt)/4\relax]{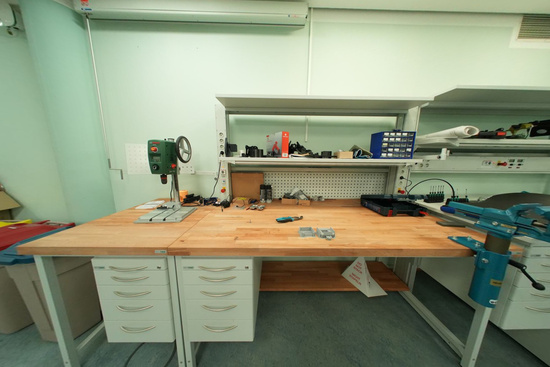} & \includegraphics[width=\dimexpr(\textwidth-4pt)/4\relax]{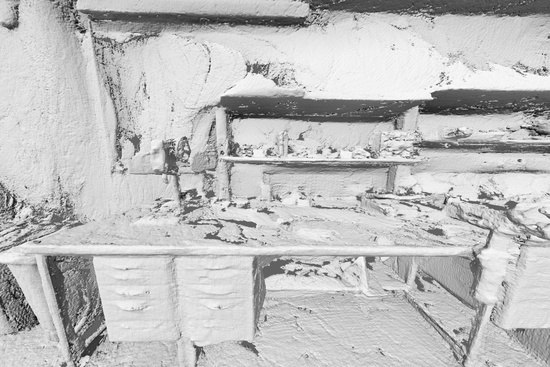} & \includegraphics[width=\dimexpr(\textwidth-4pt)/4\relax]{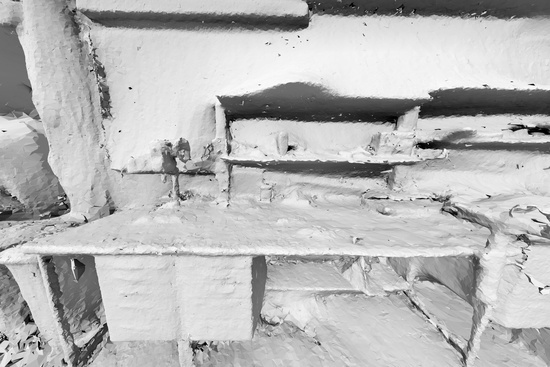} & \includegraphics[width=\dimexpr(\textwidth-4pt)/4\relax]{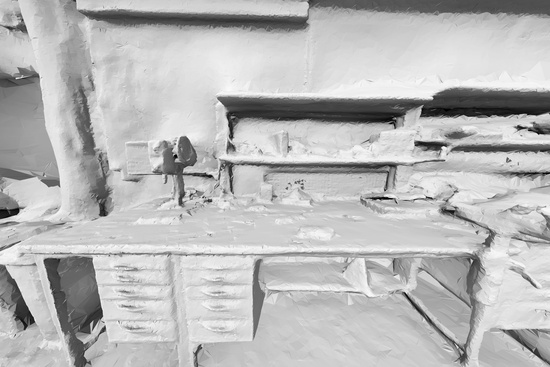} \\[-4.5pt]
    \makebox[\dimexpr(\textwidth-4pt)/4\relax][c]{\scriptsize Image} & \makebox[\dimexpr(\textwidth-4pt)/4\relax][c]{\scriptsize 2DGS} & \makebox[\dimexpr(\textwidth-4pt)/4\relax][c]{\scriptsize MILo} & \makebox[\dimexpr(\textwidth-4pt)/4\relax][c]{\scriptsize GaussianWrapping} \\[3pt]
    \includegraphics[width=\dimexpr(\textwidth-4pt)/4\relax]{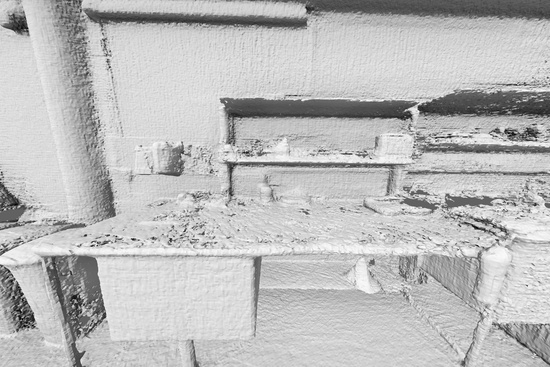} & \includegraphics[width=\dimexpr(\textwidth-4pt)/4\relax]{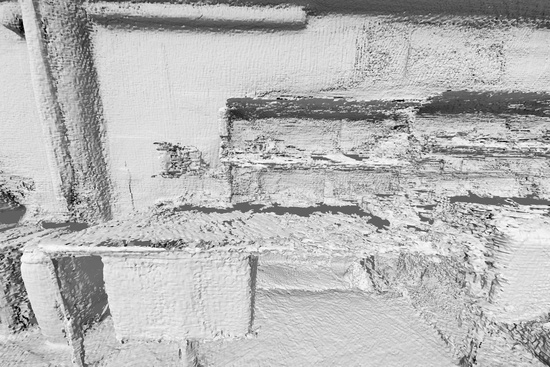} & \includegraphics[width=\dimexpr(\textwidth-4pt)/4\relax]{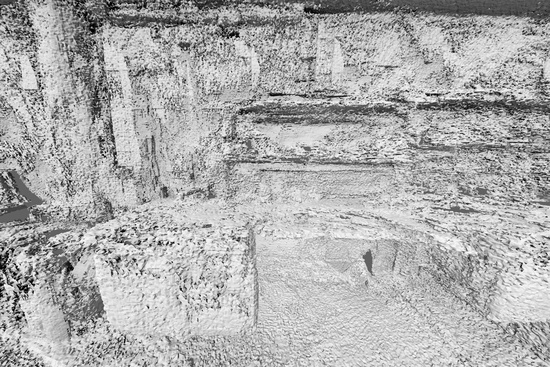} & \includegraphics[width=\dimexpr(\textwidth-4pt)/4\relax]{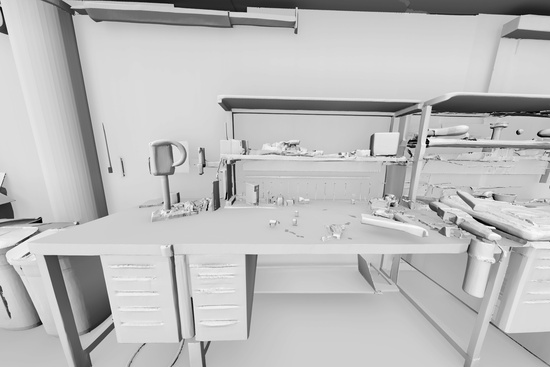} \\[-4.5pt]
    \makebox[\dimexpr(\textwidth-4pt)/4\relax][c]{\scriptsize DA3} & \makebox[\dimexpr(\textwidth-4pt)/4\relax][c]{\scriptsize MapAnything} & \makebox[\dimexpr(\textwidth-4pt)/4\relax][c]{\scriptsize Murre} & \makebox[\dimexpr(\textwidth-4pt)/4\relax][c]{\scriptsize GenRecon} \\[3pt]
    \includegraphics[width=\dimexpr(\textwidth-4pt)/4\relax]{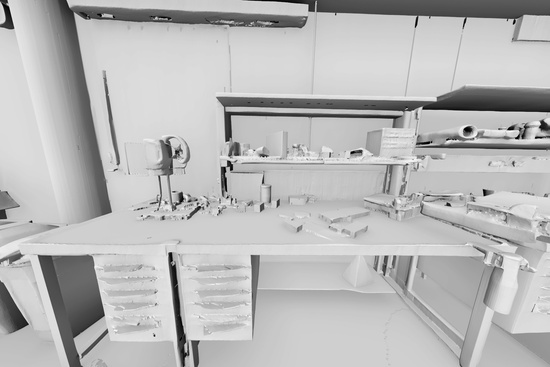} & \includegraphics[width=\dimexpr(\textwidth-4pt)/4\relax]{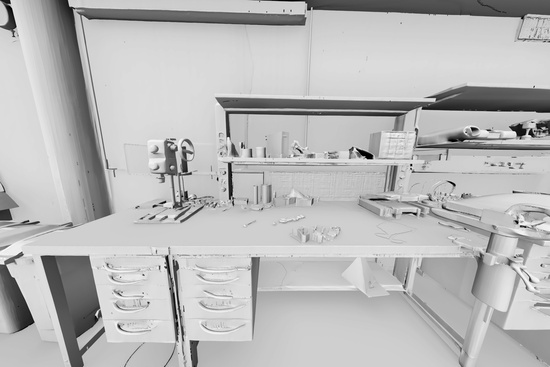} & \includegraphics[width=\dimexpr(\textwidth-4pt)/4\relax]{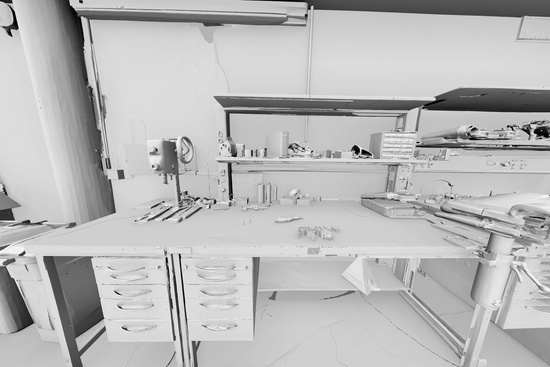} & \includegraphics[width=\dimexpr(\textwidth-4pt)/4\relax]{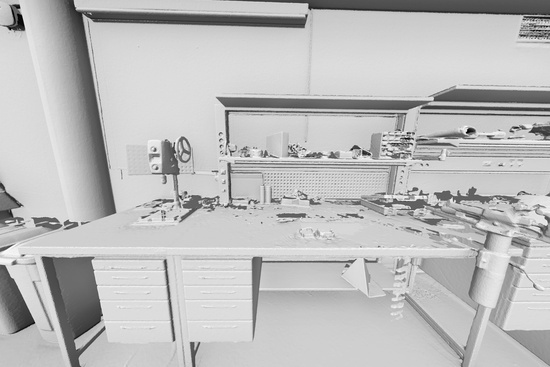} \\[-4.5pt]
    \makebox[\dimexpr(\textwidth-4pt)/4\relax][c]{\scriptsize Ours (L0)} & \makebox[\dimexpr(\textwidth-4pt)/4\relax][c]{\scriptsize Ours (L1)} & \makebox[\dimexpr(\textwidth-4pt)/4\relax][c]{\scriptsize Ours (L2)} & \makebox[\dimexpr(\textwidth-4pt)/4\relax][c]{\scriptsize Ground truth} \\[8pt]
    \includegraphics[width=\dimexpr(\textwidth-4pt)/4\relax]{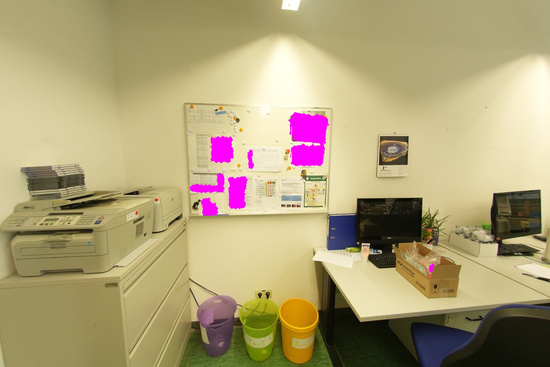} & \includegraphics[width=\dimexpr(\textwidth-4pt)/4\relax]{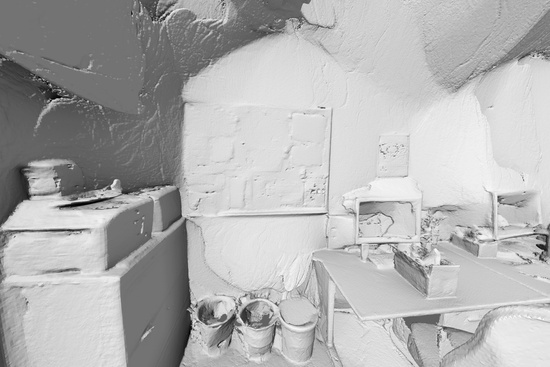} & \includegraphics[width=\dimexpr(\textwidth-4pt)/4\relax]{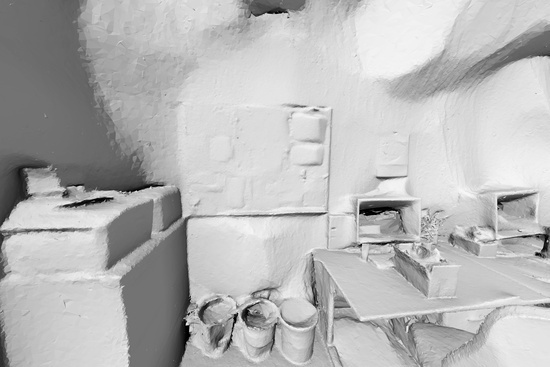} & \includegraphics[width=\dimexpr(\textwidth-4pt)/4\relax]{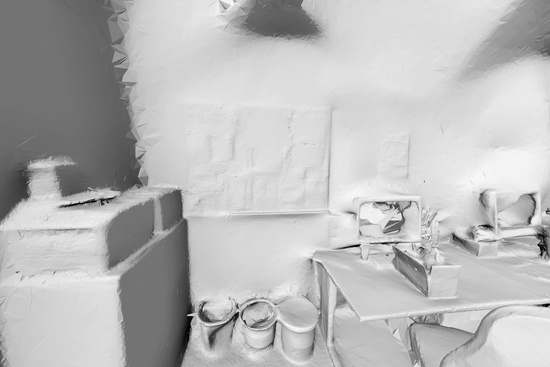} \\[-4.5pt]
    \makebox[\dimexpr(\textwidth-4pt)/4\relax][c]{\scriptsize Image} & \makebox[\dimexpr(\textwidth-4pt)/4\relax][c]{\scriptsize 2DGS} & \makebox[\dimexpr(\textwidth-4pt)/4\relax][c]{\scriptsize MILo} & \makebox[\dimexpr(\textwidth-4pt)/4\relax][c]{\scriptsize GaussianWrapping} \\[3pt]
    \includegraphics[width=\dimexpr(\textwidth-4pt)/4\relax]{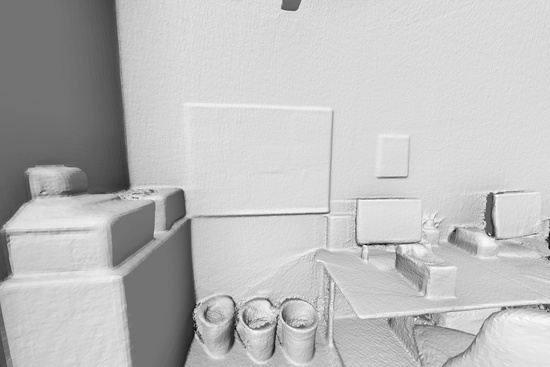} & \includegraphics[width=\dimexpr(\textwidth-4pt)/4\relax]{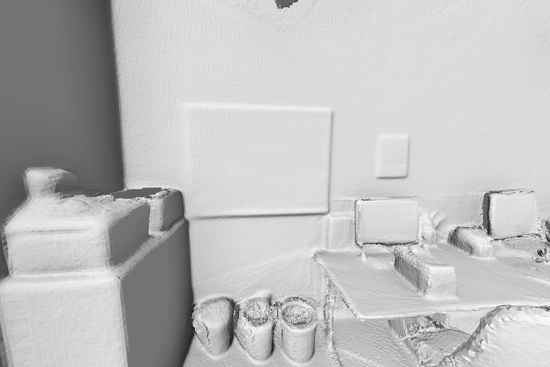} & \includegraphics[width=\dimexpr(\textwidth-4pt)/4\relax]{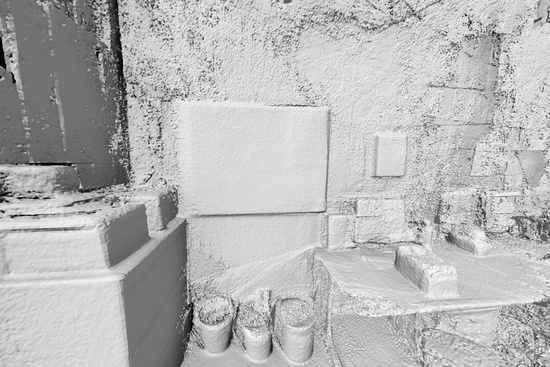} & \includegraphics[width=\dimexpr(\textwidth-4pt)/4\relax]{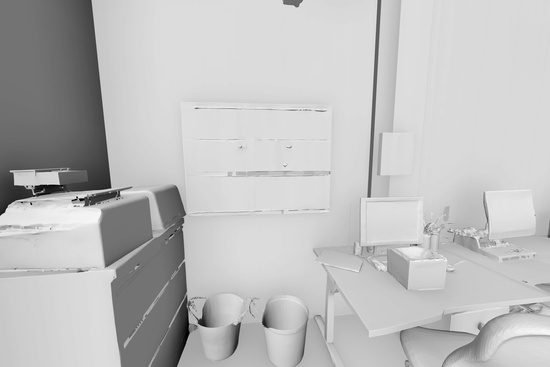} \\[-4.5pt]
    \makebox[\dimexpr(\textwidth-4pt)/4\relax][c]{\scriptsize DA3} & \makebox[\dimexpr(\textwidth-4pt)/4\relax][c]{\scriptsize MapAnything} & \makebox[\dimexpr(\textwidth-4pt)/4\relax][c]{\scriptsize Murre} & \makebox[\dimexpr(\textwidth-4pt)/4\relax][c]{\scriptsize GenRecon} \\[3pt]
    \includegraphics[width=\dimexpr(\textwidth-4pt)/4\relax]{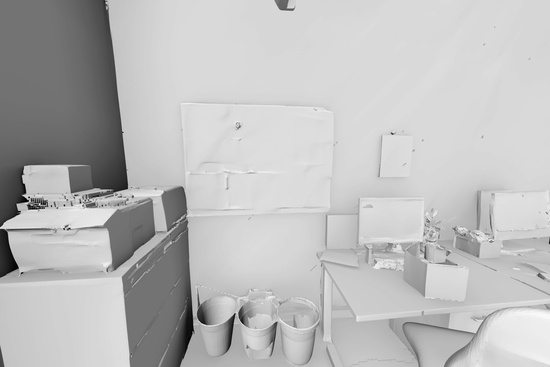} & \includegraphics[width=\dimexpr(\textwidth-4pt)/4\relax]{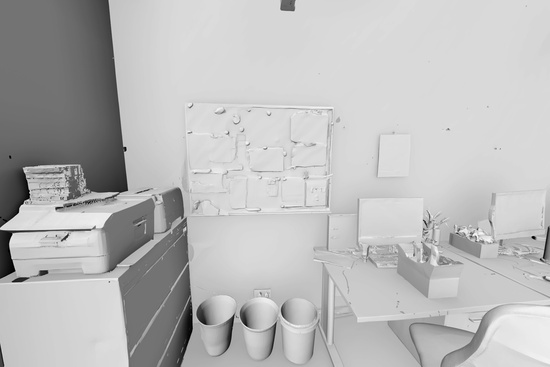} & \includegraphics[width=\dimexpr(\textwidth-4pt)/4\relax]{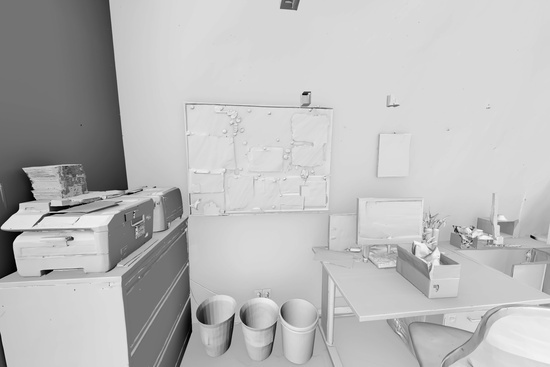} & \includegraphics[width=\dimexpr(\textwidth-4pt)/4\relax]{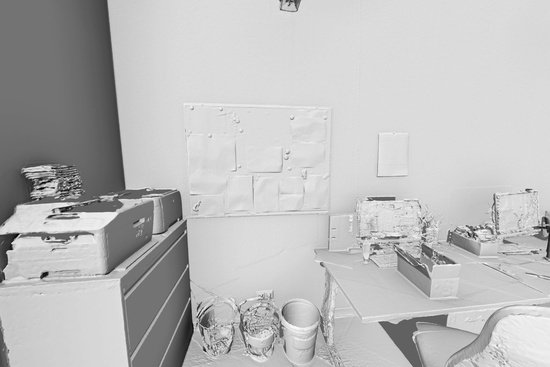} \\[-4.5pt]
    \makebox[\dimexpr(\textwidth-4pt)/4\relax][c]{\scriptsize Ours (L0)} & \makebox[\dimexpr(\textwidth-4pt)/4\relax][c]{\scriptsize Ours (L1)} & \makebox[\dimexpr(\textwidth-4pt)/4\relax][c]{\scriptsize Ours (L2)} & \makebox[\dimexpr(\textwidth-4pt)/4\relax][c]{\scriptsize Ground truth} \\
  \end{tabular}
  \caption{\textbf{Qualitative comparison on ScanNet++ with 256 input views.} All meshes are rendered from a held-out view of each scene.}
  \label{fig:scannetpp-qualitative-fps256:2}
\end{figure}

\ifdefined\bcScene \fi
\newlength{\bcGap}\setlength{\bcGap}{4pt}
\newlength{\bcRender}\setlength{\bcRender}{\dimexpr(\textwidth-\bcGap-1pt)/2\relax}
\newlength{\bcCrop}\setlength{\bcCrop}{\dimexpr(\bcRender-2pt)/3\relax}
\newlength{\bcInput}\setlength{\bcInput}{\dimexpr(\textwidth-5pt)/5\relax}

\newcommand{\bcNamegw}{GaussianWrapping}
\newcommand{\bcNamegenrecon}{GenRecon}

\newcommand{\bcInputs}[1]{%
  \begin{tabular}{@{}ccccc@{}}
    \includegraphics[width=\bcInput]{imgs/baseline_comparison/#1_input1.jpg} &
    \includegraphics[width=\bcInput]{imgs/baseline_comparison/#1_input2.jpg} &
    \includegraphics[width=\bcInput]{imgs/baseline_comparison/#1_input3.jpg} &
    \includegraphics[width=\bcInput]{imgs/baseline_comparison/#1_input4.jpg} &
    \includegraphics[width=\bcInput]{imgs/baseline_comparison/#1_input5.jpg}
  \end{tabular}%
}

\newcommand{\bcScene}[3][gw]{%
  \setlength{\tabcolsep}{0.5pt}%
  \renewcommand{\arraystretch}{1.0}%
  \begin{tabular}{@{}ccc@{\hspace{\bcGap}}ccc@{}}
    \multicolumn{6}{c}{\bcInputs{#2}} \\[1pt]
    \multicolumn{6}{c}{\scriptsize #3} \\[2pt]
    \multicolumn{3}{c}{\includegraphics[width=\bcRender]{imgs/baseline_comparison/#2_#1_render.jpg}} &
    \multicolumn{3}{c}{\includegraphics[width=\bcRender]{imgs/baseline_comparison/#2_ours_render.jpg}} \\[1pt]
    \includegraphics[width=\bcCrop]{imgs/baseline_comparison/#2_#1_crop1.jpg} &
    \includegraphics[width=\bcCrop]{imgs/baseline_comparison/#2_#1_crop2.jpg} &
    \includegraphics[width=\bcCrop]{imgs/baseline_comparison/#2_#1_crop3.jpg} &
    \includegraphics[width=\bcCrop]{imgs/baseline_comparison/#2_ours_crop1.jpg} &
    \includegraphics[width=\bcCrop]{imgs/baseline_comparison/#2_ours_crop2.jpg} &
    \includegraphics[width=\bcCrop]{imgs/baseline_comparison/#2_ours_crop3.jpg} \\[1pt]
    \multicolumn{3}{c}{\scriptsize \csname bcName#1\endcsname} &
    \multicolumn{3}{c}{\scriptsize Ours} \\
  \end{tabular}%
}

\begin{figure}[p]
  \centering
  \bcScene[genrecon]{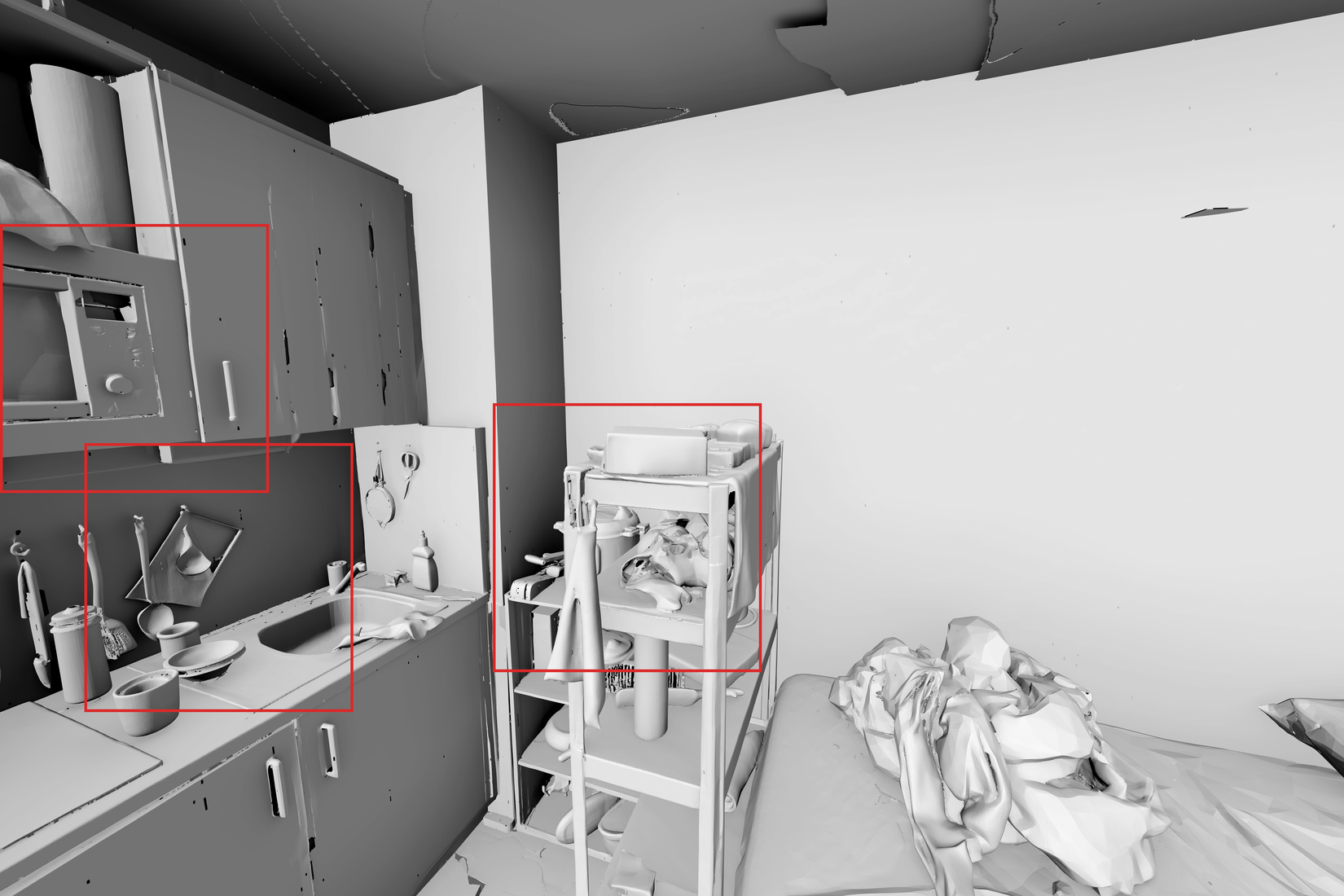}{\texttt{0d2ee665be} (ScanNet++) -- five of the 79 input views at $1752\times1168$ resolution} \\[8pt]
  \bcScene[genrecon]{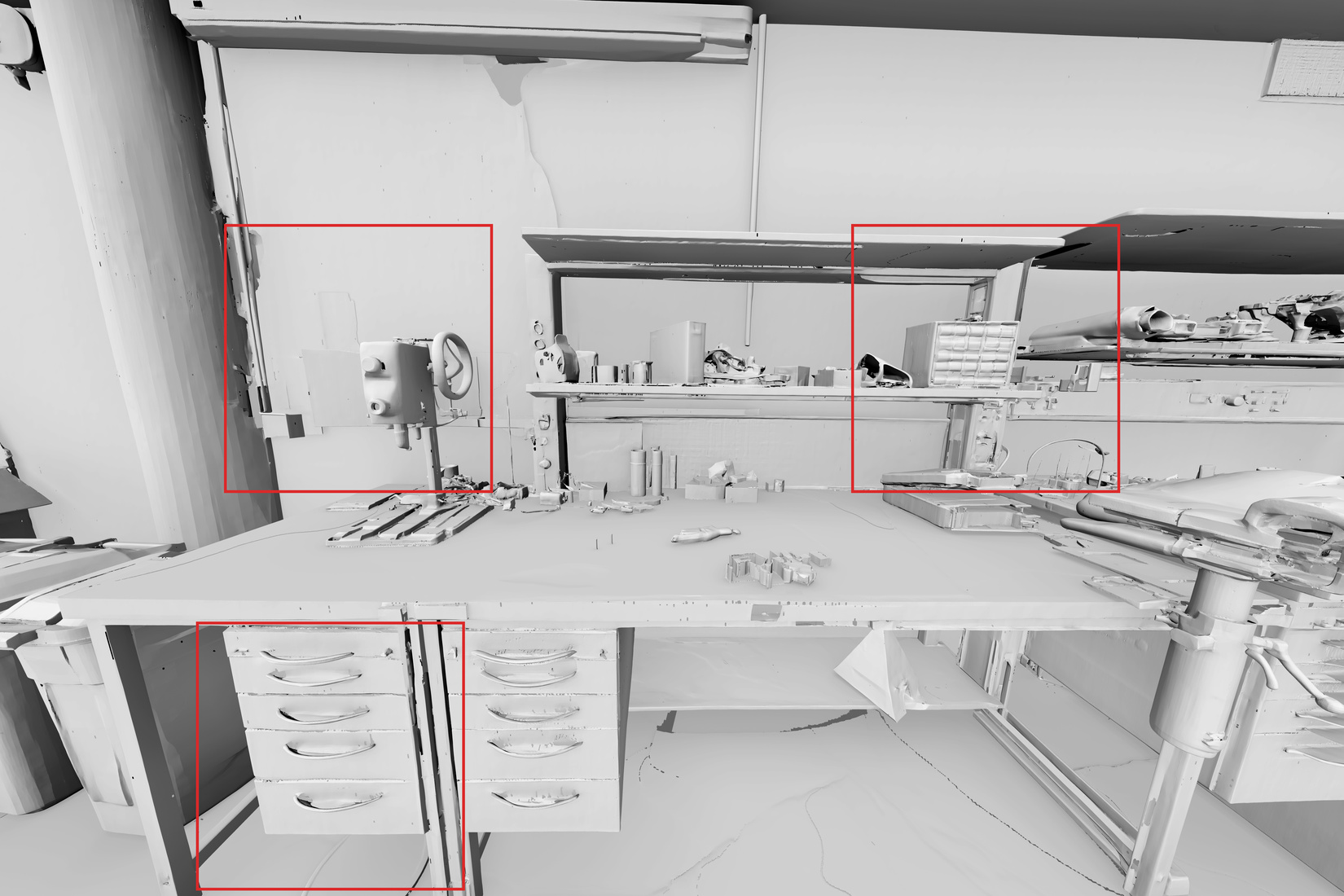}{\texttt{578511c8a9} (ScanNet++) -- five of the 253 input views at $1752\times1168$ resolution}
  \caption{\textbf{Qualitative comparison with GenRecon on ScanNet++ in the 256-view setting.} Both meshes are rendered from the same held-out viewpoint, with close-ups of the boxed regions below.}
  \label{fig:baseline-comparison-scannetpp}
\end{figure}

\clearpage

\subsection{Tanks and Temples}
\label{app:tnt-results}

\begin{figure}[H]
  \centering
  \bcScene{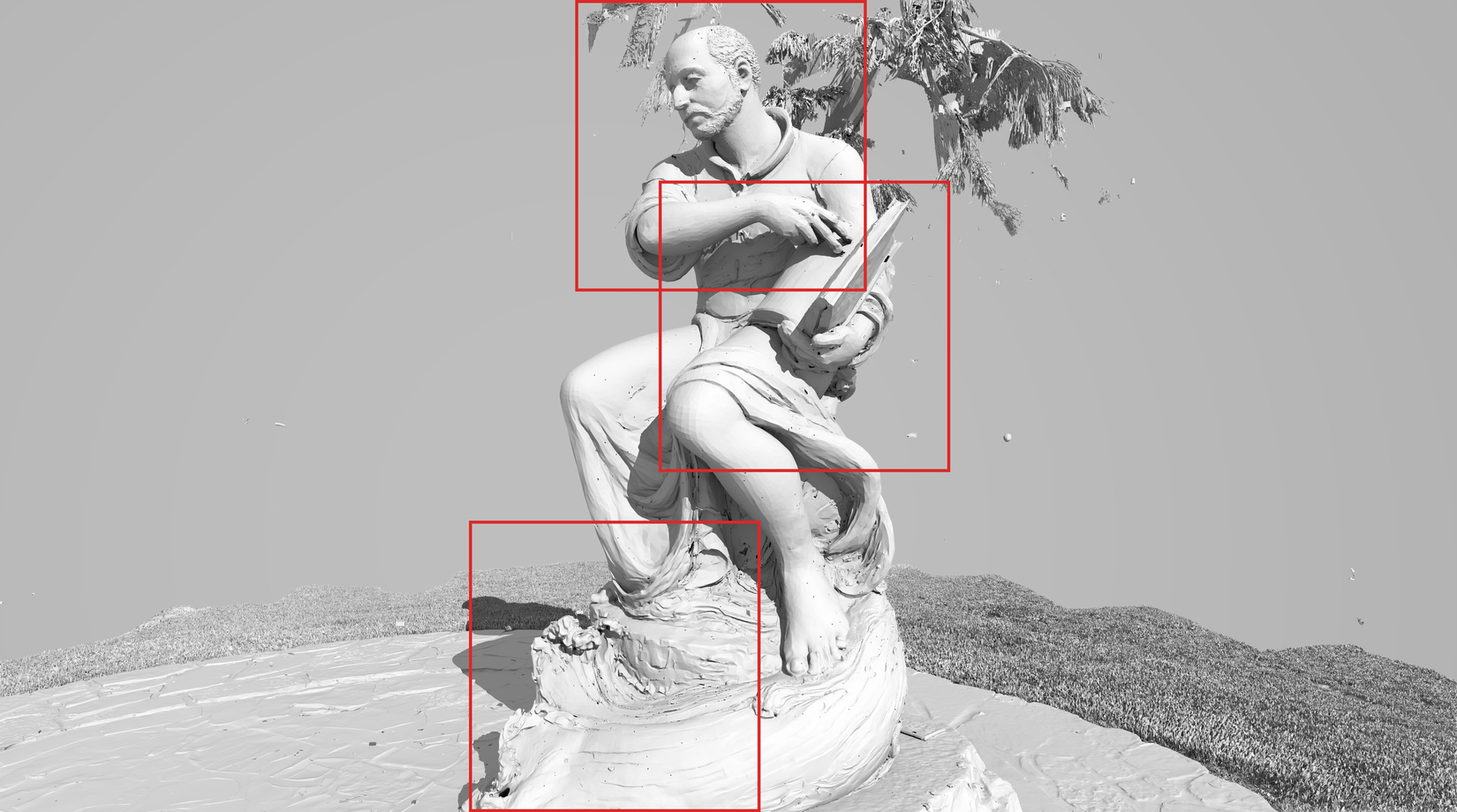}{\texttt{ignatius} (Tanks and Temples) -- five of the 256 input views at $3914\times2183$ resolution} \\[8pt]
  \bcScene{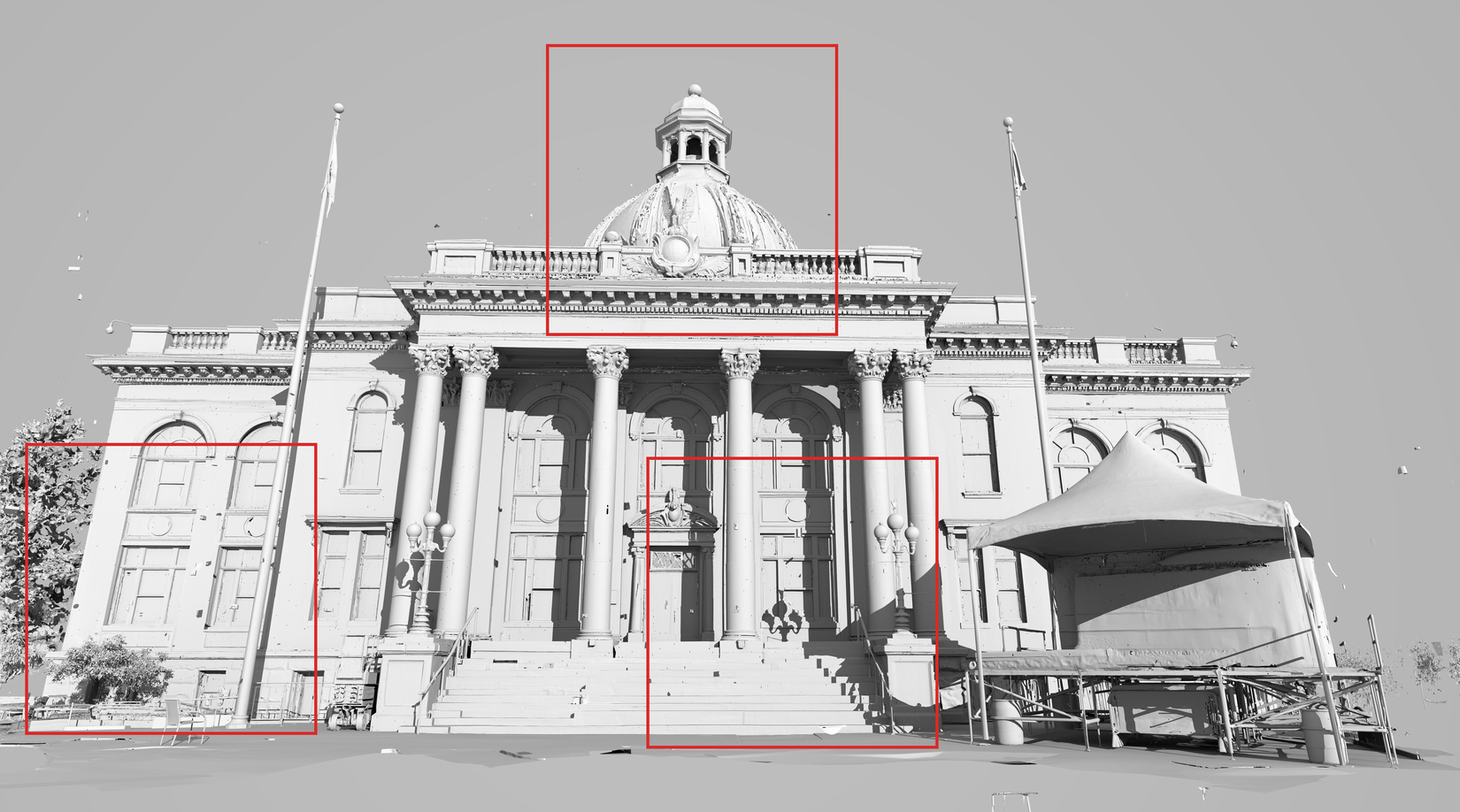}{\texttt{courthouse} (Tanks and Temples) -- five of the 256 input views at $3920\times2183$ resolution}
  \caption{\textbf{Qualitative comparison with GaussianWrapping on Tanks and Temples with 256 input views.} Both meshes are rendered from the same viewpoint, with close-ups of the boxed regions below.}
  \label{fig:baseline-comparison-tnt}
\end{figure}

Table~\ref{tab:tnt-per-scene} reports per-scene results with 16, 32, and 256 input
views. Since each scene is listed separately, the Chamfer distance is given in meters
rather than divided by $\tau$. The rows marked 4K use the 4K images instead of the
2K ones (Appendix~\ref{app:datasets}). With the 4K images, the finest level reaches a higher
mean F@$\tau$ in every view setting (e.g., $0.3344$ vs.\ $0.3009$ with 256 views), and with
256 views also a lower CD$/\tau$. The gain is concentrated at the finer levels, because only
their cells read the full-resolution tiles. Figures~\ref{fig:tnt-appendix-fps16:1} and~\ref{fig:tnt-appendix-fps16:2} show
all six scenes with 16 input views, and Figures~\ref{fig:tnt-appendix:1}
and~\ref{fig:tnt-appendix:2} with 256. Figure~\ref{fig:baseline-comparison-tnt} compares
close-ups of our meshes with those of GaussianWrapping, the per-scene optimization
baseline with the highest F@$\tau$, on \texttt{ignatius} and \texttt{courthouse} in the
256-view setting.

\begin{table}[H]
  \centering
  \caption{\textbf{Per-scene quantitative comparison on Tanks and Temples.} F@$\tau$ uses the official per-scene threshold $\tau$, and CD is in meters. Rows marked 4K use the 4K input images instead of the 2K ones. The top three results are highlighted.}
  \label{tab:tnt-per-scene}
  \scriptsize
  \setlength{\tabcolsep}{3pt}
  \definecolor{rank1}{HTML}{FFB3B3}
  \definecolor{rank2}{HTML}{FFD9B3}
  \definecolor{rank3}{HTML}{FFFFB3}
  \definecolor{emphrow}{HTML}{EDEDED}
  \ifdefined\paperautocellwidth\else\newlength{\paperautocellwidth}\fi
  \ifdefined\paperautoprobe\else\newlength{\paperautoprobe}\fi
  \settowidth{\paperautocellwidth}{\textbf{0.0000}}
  \settowidth{\paperautoprobe}{Barn}
  \ifdim\dimexpr\paperautoprobe-2\tabcolsep\relax>2\paperautocellwidth  \setlength{\paperautocellwidth}{\dimexpr(\paperautoprobe-2\tabcolsep)/2\relax}\fi
  \settowidth{\paperautoprobe}{Caterpillar}
  \ifdim\dimexpr\paperautoprobe-2\tabcolsep\relax>2\paperautocellwidth  \setlength{\paperautocellwidth}{\dimexpr(\paperautoprobe-2\tabcolsep)/2\relax}\fi
  \settowidth{\paperautoprobe}{Courthouse}
  \ifdim\dimexpr\paperautoprobe-2\tabcolsep\relax>2\paperautocellwidth  \setlength{\paperautocellwidth}{\dimexpr(\paperautoprobe-2\tabcolsep)/2\relax}\fi
  \settowidth{\paperautoprobe}{Ignatius}
  \ifdim\dimexpr\paperautoprobe-2\tabcolsep\relax>2\paperautocellwidth  \setlength{\paperautocellwidth}{\dimexpr(\paperautoprobe-2\tabcolsep)/2\relax}\fi
  \settowidth{\paperautoprobe}{Meetingroom}
  \ifdim\dimexpr\paperautoprobe-2\tabcolsep\relax>2\paperautocellwidth  \setlength{\paperautocellwidth}{\dimexpr(\paperautoprobe-2\tabcolsep)/2\relax}\fi
  \settowidth{\paperautoprobe}{Truck}
  \ifdim\dimexpr\paperautoprobe-2\tabcolsep\relax>2\paperautocellwidth  \setlength{\paperautocellwidth}{\dimexpr(\paperautoprobe-2\tabcolsep)/2\relax}\fi
  \settowidth{\paperautoprobe}{F@$\tau$$\uparrow$}
  \ifdim\dimexpr\paperautoprobe-0\tabcolsep\relax>1\paperautocellwidth  \setlength{\paperautocellwidth}{\dimexpr(\paperautoprobe-0\tabcolsep)/1\relax}\fi
  \settowidth{\paperautoprobe}{CD$\downarrow$}
  \ifdim\dimexpr\paperautoprobe-0\tabcolsep\relax>1\paperautocellwidth  \setlength{\paperautocellwidth}{\dimexpr(\paperautoprobe-0\tabcolsep)/1\relax}\fi
  \providecommand{\paperautospan}[2]{\makebox[\dimexpr#1\paperautocellwidth+#1\tabcolsep+#1\tabcolsep-\tabcolsep-\tabcolsep\relax][c]{#2}}
  \providecommand{\paperautocell}[1]{\paperautospan{1}{#1}}
  \resizebox{\linewidth}{!}{%
  \begin{tabular}{lcccccccccccc}
    \toprule
     & \multicolumn{2}{c}{\paperautospan{2}{Barn}} & \multicolumn{2}{c}{\paperautospan{2}{Caterpillar}} & \multicolumn{2}{c}{\paperautospan{2}{Courthouse}} & \multicolumn{2}{c}{\paperautospan{2}{Ignatius}} & \multicolumn{2}{c}{\paperautospan{2}{Meetingroom}} & \multicolumn{2}{c}{\paperautospan{2}{Truck}} \\
    \cmidrule(lr){2-3}\cmidrule(lr){4-5}\cmidrule(lr){6-7}\cmidrule(lr){8-9}\cmidrule(lr){10-11}\cmidrule(lr){12-13}
    Method & \multicolumn{1}{c}{\paperautospan{1}{F@$\tau$$\uparrow$}} & \multicolumn{1}{c}{\paperautospan{1}{CD$\downarrow$}} & \multicolumn{1}{c}{\paperautospan{1}{F@$\tau$$\uparrow$}} & \multicolumn{1}{c}{\paperautospan{1}{CD$\downarrow$}} & \multicolumn{1}{c}{\paperautospan{1}{F@$\tau$$\uparrow$}} & \multicolumn{1}{c}{\paperautospan{1}{CD$\downarrow$}} & \multicolumn{1}{c}{\paperautospan{1}{F@$\tau$$\uparrow$}} & \multicolumn{1}{c}{\paperautospan{1}{CD$\downarrow$}} & \multicolumn{1}{c}{\paperautospan{1}{F@$\tau$$\uparrow$}} & \multicolumn{1}{c}{\paperautospan{1}{CD$\downarrow$}} & \multicolumn{1}{c}{\paperautospan{1}{F@$\tau$$\uparrow$}} & \multicolumn{1}{c}{\paperautospan{1}{CD$\downarrow$}} \\
    \midrule
    \multicolumn{13}{l}{\textbf{\textit{16 Views}}} \\
    2DGS & \paperautocell{0.1680} & \paperautocell{0.4530} & \cellcolor{rank1}\paperautocell{0.2467} & \cellcolor{rank2}\paperautocell{0.0667} & \paperautocell{0.0773} & \paperautocell{1.1360} & \paperautocell{0.3975} & \paperautocell{0.0260} & \paperautocell{0.0566} & \paperautocell{0.4412} & \cellcolor{rank1}\paperautocell{0.4213} & \paperautocell{0.0314} \\
    MILo & \paperautocell{0.0622} & \paperautocell{0.5128} & \paperautocell{0.1522} & \paperautocell{0.1275} & \paperautocell{0.0235} & \paperautocell{2.3359} & \cellcolor{rank1}\paperautocell{0.5351} & \paperautocell{0.0210} & \paperautocell{0.0740} & \paperautocell{0.1741} & \cellcolor{rank3}\paperautocell{0.3510} & \paperautocell{0.0594} \\
    GaussianWrapping & \paperautocell{0.1520} & \paperautocell{0.1269} & \cellcolor{rank2}\paperautocell{0.2098} & \paperautocell{0.0777} & \paperautocell{0.0457} & \paperautocell{2.0951} & \cellcolor{rank2}\paperautocell{0.5061} & \paperautocell{0.0081} & \paperautocell{0.0840} & \paperautocell{0.1635} & \paperautocell{0.3192} & \paperautocell{0.0409} \\
    DA3 & \paperautocell{0.1123} & \paperautocell{0.2928} & \paperautocell{0.1088} & \paperautocell{0.0876} & \paperautocell{0.0418} & \paperautocell{1.3350} & \paperautocell{0.1417} & \paperautocell{0.0349} & \paperautocell{0.1878} & \paperautocell{0.2334} & \paperautocell{0.1926} & \paperautocell{0.0414} \\
    MapAnything & \paperautocell{0.0055} & \paperautocell{2.9563} & \paperautocell{0.0445} & \paperautocell{0.1260} & \paperautocell{0.0010} & \paperautocell{6.2998} & \paperautocell{0.0870} & \paperautocell{0.0451} & \paperautocell{0.0352} & \paperautocell{0.3645} & \paperautocell{0.0815} & \paperautocell{0.0873} \\
    Murre & \paperautocell{0.0652} & \paperautocell{0.2628} & \paperautocell{0.0283} & \paperautocell{0.2581} & \paperautocell{0.0467} & \paperautocell{1.4180} & \paperautocell{0.0263} & \paperautocell{0.2148} & \paperautocell{0.0408} & \paperautocell{0.4338} & \paperautocell{0.0998} & \paperautocell{0.1005} \\
    \cellcolor{emphrow}\textbf{Ours (L0)} & \paperautocell{0.1655} & \cellcolor{rank2}\paperautocell{0.0600} & \paperautocell{0.1230} & \paperautocell{0.0767} & \paperautocell{0.0869} & \paperautocell{0.8989} & \paperautocell{0.2633} & \paperautocell{0.0091} & \paperautocell{0.2729} & \paperautocell{0.0519} & \paperautocell{0.2699} & \paperautocell{0.0311} \\
    \cellcolor{emphrow}\textbf{Ours (L1)} & \cellcolor{rank3}\paperautocell{0.1985} & \paperautocell{0.0764} & \paperautocell{0.1569} & \paperautocell{0.0710} & \paperautocell{0.1057} & \cellcolor{rank3}\paperautocell{0.6151} & \paperautocell{0.3453} & \paperautocell{0.0070} & \paperautocell{0.2797} & \paperautocell{0.0513} & \paperautocell{0.3316} & \paperautocell{0.0309} \\
    \cellcolor{emphrow}\textbf{Ours (L2)} & \paperautocell{0.1926} & \paperautocell{0.0789} & \paperautocell{0.1631} & \paperautocell{0.0679} & \cellcolor{rank3}\paperautocell{0.1079} & \cellcolor{rank1}\paperautocell{0.4984} & \paperautocell{0.3898} & \cellcolor{rank3}\paperautocell{0.0067} & \paperautocell{0.2675} & \paperautocell{0.0553} & \paperautocell{0.3430} & \cellcolor{rank1}\paperautocell{0.0302} \\
    \cellcolor{emphrow}\textbf{Ours (L0), 4K input} & \paperautocell{0.1894} & \cellcolor{rank3}\paperautocell{0.0630} & \paperautocell{0.1140} & \paperautocell{0.0773} & \paperautocell{0.0874} & \paperautocell{0.8770} & \paperautocell{0.2691} & \paperautocell{0.0090} & \cellcolor{rank3}\paperautocell{0.2958} & \cellcolor{rank3}\paperautocell{0.0508} & \paperautocell{0.2577} & \cellcolor{rank2}\paperautocell{0.0307} \\
    \cellcolor{emphrow}\textbf{Ours (L1), 4K input} & \cellcolor{rank1}\paperautocell{0.2081} & \cellcolor{rank1}\paperautocell{0.0523} & \paperautocell{0.1711} & \cellcolor{rank3}\paperautocell{0.0669} & \cellcolor{rank2}\paperautocell{0.1087} & \paperautocell{0.7036} & \paperautocell{0.3987} & \cellcolor{rank2}\paperautocell{0.0063} & \cellcolor{rank2}\paperautocell{0.2965} & \cellcolor{rank2}\paperautocell{0.0484} & \paperautocell{0.3221} & \cellcolor{rank3}\paperautocell{0.0308} \\
    \cellcolor{emphrow}\textbf{Ours (L2), 4K input} & \cellcolor{rank2}\paperautocell{0.2080} & \paperautocell{0.0895} & \cellcolor{rank3}\paperautocell{0.1949} & \cellcolor{rank1}\paperautocell{0.0637} & \cellcolor{rank1}\paperautocell{0.1159} & \cellcolor{rank2}\paperautocell{0.5905} & \cellcolor{rank3}\paperautocell{0.4704} & \cellcolor{rank1}\paperautocell{0.0055} & \cellcolor{rank1}\paperautocell{0.3084} & \cellcolor{rank1}\paperautocell{0.0476} & \cellcolor{rank2}\paperautocell{0.3563} & \paperautocell{0.0311} \\
    \midrule
    \multicolumn{13}{l}{\textbf{\textit{32 Views}}} \\
    2DGS & \cellcolor{rank1}\paperautocell{0.2741} & \paperautocell{0.2013} & \cellcolor{rank2}\paperautocell{0.2771} & \cellcolor{rank1}\paperautocell{0.0526} & \paperautocell{0.0998} & \paperautocell{1.1360} & \cellcolor{rank2}\paperautocell{0.6163} & \paperautocell{0.0074} & \paperautocell{0.1083} & \paperautocell{0.3033} & \cellcolor{rank2}\paperautocell{0.4419} & \cellcolor{rank1}\paperautocell{0.0292} \\
    MILo & \cellcolor{rank3}\paperautocell{0.2216} & \paperautocell{0.6706} & \cellcolor{rank3}\paperautocell{0.2154} & \paperautocell{0.1273} & \paperautocell{0.0459} & \paperautocell{1.8242} & \cellcolor{rank1}\paperautocell{0.6513} & \paperautocell{0.0478} & \paperautocell{0.1363} & \paperautocell{0.1274} & \cellcolor{rank1}\paperautocell{0.4545} & \paperautocell{0.0447} \\
    GaussianWrapping & \cellcolor{rank2}\paperautocell{0.2633} & \paperautocell{0.0688} & \cellcolor{rank1}\paperautocell{0.2882} & \paperautocell{0.0732} & \paperautocell{0.1167} & \paperautocell{1.6180} & \cellcolor{rank3}\paperautocell{0.6007} & \cellcolor{rank1}\paperautocell{0.0050} & \paperautocell{0.1309} & \paperautocell{0.1154} & \cellcolor{rank3}\paperautocell{0.3909} & \paperautocell{0.0375} \\
    DA3 & \paperautocell{0.1371} & \paperautocell{0.2588} & \paperautocell{0.1387} & \cellcolor{rank2}\paperautocell{0.0650} & \paperautocell{0.0394} & \paperautocell{1.0687} & \paperautocell{0.1809} & \paperautocell{0.0162} & \paperautocell{0.2832} & \paperautocell{0.1259} & \paperautocell{0.2244} & \paperautocell{0.0382} \\
    MapAnything & \paperautocell{0.0117} & \paperautocell{0.7994} & \paperautocell{0.0672} & \paperautocell{0.0842} & \paperautocell{0.0021} & \paperautocell{5.4052} & \paperautocell{0.1000} & \paperautocell{0.0271} & \paperautocell{0.0647} & \paperautocell{0.2432} & \paperautocell{0.1072} & \paperautocell{0.0718} \\
    Murre & \paperautocell{0.0637} & \paperautocell{0.2332} & \paperautocell{0.0295} & \paperautocell{0.2111} & \paperautocell{0.0513} & \paperautocell{1.1645} & \paperautocell{0.0199} & \paperautocell{0.1984} & \paperautocell{0.0664} & \paperautocell{0.2941} & \paperautocell{0.1085} & \paperautocell{0.0810} \\
    \cellcolor{emphrow}\textbf{Ours (L0)} & \paperautocell{0.1854} & \cellcolor{rank3}\paperautocell{0.0671} & \paperautocell{0.1367} & \paperautocell{0.0781} & \paperautocell{0.1008} & \paperautocell{0.3191} & \paperautocell{0.3211} & \paperautocell{0.0077} & \paperautocell{0.3402} & \paperautocell{0.0344} & \paperautocell{0.3010} & \paperautocell{0.0322} \\
    \cellcolor{emphrow}\textbf{Ours (L1)} & \paperautocell{0.2113} & \cellcolor{rank1}\paperautocell{0.0528} & \paperautocell{0.1795} & \paperautocell{0.0713} & \cellcolor{rank3}\paperautocell{0.1190} & \cellcolor{rank3}\paperautocell{0.2639} & \paperautocell{0.3897} & \paperautocell{0.0069} & \paperautocell{0.3491} & \cellcolor{rank3}\paperautocell{0.0332} & \paperautocell{0.3542} & \paperautocell{0.0310} \\
    \cellcolor{emphrow}\textbf{Ours (L2)} & \paperautocell{0.2069} & \paperautocell{0.0778} & \paperautocell{0.1890} & \cellcolor{rank3}\paperautocell{0.0672} & \cellcolor{rank2}\paperautocell{0.1231} & \cellcolor{rank1}\paperautocell{0.2298} & \paperautocell{0.4244} & \paperautocell{0.0062} & \paperautocell{0.3396} & \paperautocell{0.0343} & \paperautocell{0.3751} & \paperautocell{0.0306} \\
    \cellcolor{emphrow}\textbf{Ours (L0), 4K input} & \paperautocell{0.1463} & \paperautocell{0.0787} & \paperautocell{0.1328} & \paperautocell{0.0805} & \paperautocell{0.0965} & \paperautocell{0.3619} & \paperautocell{0.3223} & \paperautocell{0.0073} & \cellcolor{rank3}\paperautocell{0.3561} & \paperautocell{0.0349} & \paperautocell{0.2955} & \paperautocell{0.0317} \\
    \cellcolor{emphrow}\textbf{Ours (L1), 4K input} & \paperautocell{0.1926} & \cellcolor{rank2}\paperautocell{0.0641} & \paperautocell{0.1752} & \paperautocell{0.0727} & \paperautocell{0.1181} & \paperautocell{0.3355} & \paperautocell{0.4194} & \cellcolor{rank3}\paperautocell{0.0058} & \cellcolor{rank2}\paperautocell{0.3721} & \cellcolor{rank2}\paperautocell{0.0318} & \paperautocell{0.3523} & \cellcolor{rank3}\paperautocell{0.0305} \\
    \cellcolor{emphrow}\textbf{Ours (L2), 4K input} & \paperautocell{0.1938} & \paperautocell{0.0866} & \paperautocell{0.1945} & \paperautocell{0.0740} & \cellcolor{rank1}\paperautocell{0.1324} & \cellcolor{rank2}\paperautocell{0.2553} & \paperautocell{0.4904} & \cellcolor{rank2}\paperautocell{0.0051} & \cellcolor{rank1}\paperautocell{0.3806} & \cellcolor{rank1}\paperautocell{0.0308} & \paperautocell{0.3705} & \cellcolor{rank2}\paperautocell{0.0304} \\
    \midrule
    \multicolumn{13}{l}{\textbf{\textit{256 Views}}} \\
    2DGS & \cellcolor{rank3}\paperautocell{0.4156} & \paperautocell{0.0602} & \cellcolor{rank3}\paperautocell{0.2566} & \paperautocell{0.0725} & \paperautocell{0.1000} & \paperautocell{1.6849} & \cellcolor{rank3}\paperautocell{0.6432} & \paperautocell{0.0232} & \paperautocell{0.1841} & \paperautocell{0.1082} & \cellcolor{rank3}\paperautocell{0.4427} & \paperautocell{0.0334} \\
    MILo & \cellcolor{rank2}\paperautocell{0.4617} & \paperautocell{0.7530} & \cellcolor{rank2}\paperautocell{0.3048} & \paperautocell{0.1064} & \paperautocell{0.1172} & \paperautocell{1.6766} & \cellcolor{rank2}\paperautocell{0.7452} & \paperautocell{0.0193} & \paperautocell{0.2250} & \paperautocell{0.1151} & \cellcolor{rank2}\paperautocell{0.5314} & \paperautocell{0.0359} \\
    GaussianWrapping & \cellcolor{rank1}\paperautocell{0.6366} & \cellcolor{rank1}\paperautocell{0.0285} & \cellcolor{rank1}\paperautocell{0.4850} & \cellcolor{rank1}\paperautocell{0.0549} & \cellcolor{rank1}\paperautocell{0.1796} & \paperautocell{2.9165} & \cellcolor{rank1}\paperautocell{0.8413} & \cellcolor{rank1}\paperautocell{0.0022} & \paperautocell{0.2945} & \paperautocell{0.0802} & \cellcolor{rank1}\paperautocell{0.6226} & \cellcolor{rank1}\paperautocell{0.0245} \\
    DA3 & \paperautocell{0.1256} & \paperautocell{0.3737} & \paperautocell{0.1320} & \paperautocell{0.0750} & \paperautocell{0.0634} & \paperautocell{0.7696} & \paperautocell{0.2420} & \paperautocell{0.0096} & \paperautocell{0.3521} & \paperautocell{0.0462} & \paperautocell{0.2232} & \paperautocell{0.0478} \\
    MapAnything & \paperautocell{0.0080} & \paperautocell{1.0894} & \paperautocell{0.0303} & \paperautocell{0.2213} & \paperautocell{0.0043} & \paperautocell{3.2754} & \paperautocell{0.0959} & \paperautocell{0.0263} & \paperautocell{0.1045} & \paperautocell{0.1936} & \paperautocell{0.1033} & \paperautocell{0.1533} \\
    Murre & \paperautocell{0.0503} & \paperautocell{0.3124} & \paperautocell{0.0391} & \paperautocell{0.2477} & \paperautocell{0.0308} & \paperautocell{2.5481} & \paperautocell{0.0410} & \paperautocell{0.0896} & \paperautocell{0.0977} & \paperautocell{0.1601} & \paperautocell{0.0941} & \paperautocell{0.0806} \\
    \cellcolor{emphrow}\textbf{Ours (L0)} & \paperautocell{0.2004} & \paperautocell{0.0672} & \paperautocell{0.1305} & \paperautocell{0.0775} & \paperautocell{0.1077} & \paperautocell{0.3000} & \paperautocell{0.3083} & \paperautocell{0.0076} & \cellcolor{rank3}\paperautocell{0.3885} & \cellcolor{rank1}\paperautocell{0.0253} & \paperautocell{0.3081} & \paperautocell{0.0305} \\
    \cellcolor{emphrow}\textbf{Ours (L1)} & \paperautocell{0.2070} & \paperautocell{0.0624} & \paperautocell{0.1846} & \paperautocell{0.0694} & \paperautocell{0.1283} & \paperautocell{0.2926} & \paperautocell{0.4109} & \paperautocell{0.0058} & \paperautocell{0.3833} & \paperautocell{0.0277} & \paperautocell{0.3419} & \paperautocell{0.0304} \\
    \cellcolor{emphrow}\textbf{Ours (L2)} & \paperautocell{0.2238} & \paperautocell{0.0601} & \paperautocell{0.2036} & \cellcolor{rank3}\paperautocell{0.0654} & \cellcolor{rank3}\paperautocell{0.1352} & \paperautocell{0.3052} & \paperautocell{0.4739} & \cellcolor{rank3}\paperautocell{0.0053} & \paperautocell{0.3839} & \paperautocell{0.0280} & \paperautocell{0.3849} & \cellcolor{rank3}\paperautocell{0.0302} \\
    \cellcolor{emphrow}\textbf{Ours (L0), 4K input} & \paperautocell{0.2303} & \paperautocell{0.0576} & \paperautocell{0.1347} & \paperautocell{0.0773} & \paperautocell{0.1127} & \cellcolor{rank3}\paperautocell{0.2764} & \paperautocell{0.3178} & \paperautocell{0.0077} & \paperautocell{0.3832} & \cellcolor{rank2}\paperautocell{0.0258} & \paperautocell{0.2813} & \paperautocell{0.0316} \\
    \cellcolor{emphrow}\textbf{Ours (L1), 4K input} & \paperautocell{0.2570} & \cellcolor{rank2}\paperautocell{0.0514} & \paperautocell{0.1867} & \paperautocell{0.0672} & \paperautocell{0.1292} & \cellcolor{rank2}\paperautocell{0.2547} & \paperautocell{0.4331} & \paperautocell{0.0056} & \cellcolor{rank2}\paperautocell{0.4049} & \paperautocell{0.0294} & \paperautocell{0.3601} & \paperautocell{0.0305} \\
    \cellcolor{emphrow}\textbf{Ours (L2), 4K input} & \paperautocell{0.2791} & \cellcolor{rank3}\paperautocell{0.0516} & \paperautocell{0.2243} & \cellcolor{rank2}\paperautocell{0.0632} & \cellcolor{rank2}\paperautocell{0.1472} & \cellcolor{rank1}\paperautocell{0.2377} & \paperautocell{0.5365} & \cellcolor{rank2}\paperautocell{0.0046} & \cellcolor{rank1}\paperautocell{0.4185} & \cellcolor{rank3}\paperautocell{0.0269} & \paperautocell{0.4008} & \cellcolor{rank2}\paperautocell{0.0299} \\
    \bottomrule
  \end{tabular}%
  }
\end{table}

\begin{figure}[p]
  \centering
  \setlength{\tabcolsep}{0.5pt}
  \renewcommand{\arraystretch}{1.0}
  \begin{tabular}{ccccc}
    \multicolumn{5}{c}{\scriptsize (a) \texttt{barn}} \\[1.5pt]
    \includegraphics[width=\dimexpr(\textwidth-5pt)/5\relax]{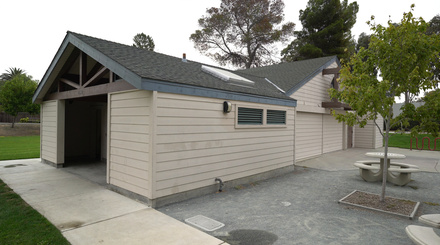} & \includegraphics[width=\dimexpr(\textwidth-5pt)/5\relax]{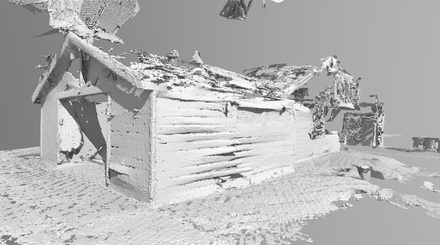} & \includegraphics[width=\dimexpr(\textwidth-5pt)/5\relax]{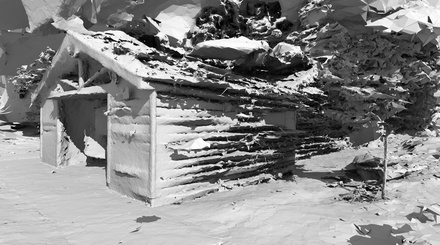} & \includegraphics[width=\dimexpr(\textwidth-5pt)/5\relax]{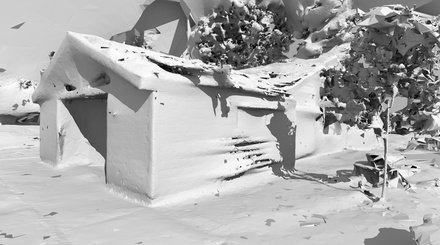} & \includegraphics[width=\dimexpr(\textwidth-5pt)/5\relax]{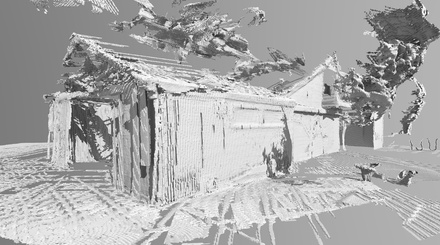} \\[-4.5pt]
    \makebox[\dimexpr(\textwidth-5pt)/5\relax][c]{\scriptsize Image} & \makebox[\dimexpr(\textwidth-5pt)/5\relax][c]{\scriptsize 2DGS} & \makebox[\dimexpr(\textwidth-5pt)/5\relax][c]{\scriptsize MILo} & \makebox[\dimexpr(\textwidth-5pt)/5\relax][c]{\scriptsize GaussianWrapping} & \makebox[\dimexpr(\textwidth-5pt)/5\relax][c]{\scriptsize DA3} \\[3pt]
    \includegraphics[width=\dimexpr(\textwidth-5pt)/5\relax]{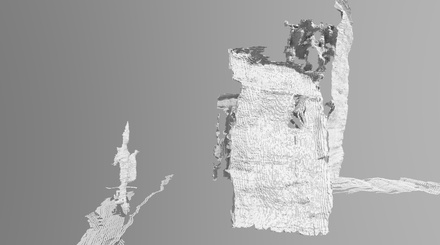} & \includegraphics[width=\dimexpr(\textwidth-5pt)/5\relax]{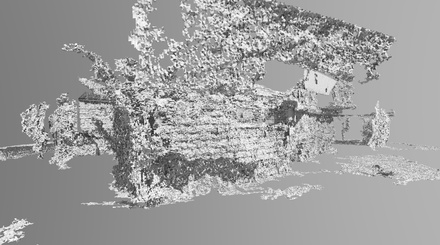} & \includegraphics[width=\dimexpr(\textwidth-5pt)/5\relax]{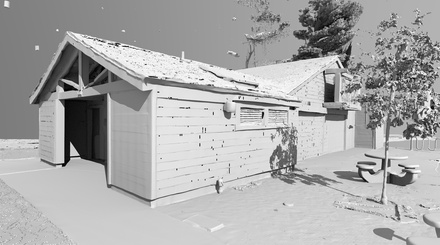} & \includegraphics[width=\dimexpr(\textwidth-5pt)/5\relax]{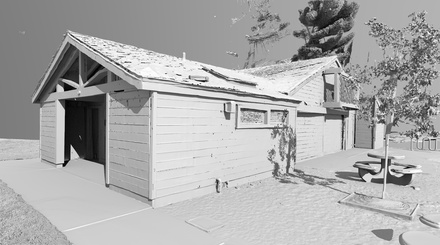} & \includegraphics[width=\dimexpr(\textwidth-5pt)/5\relax]{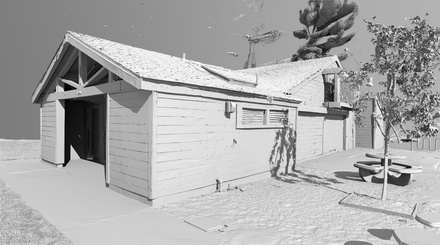} \\[-4.5pt]
    \makebox[\dimexpr(\textwidth-5pt)/5\relax][c]{\scriptsize MapAnything} & \makebox[\dimexpr(\textwidth-5pt)/5\relax][c]{\scriptsize Murre} & \makebox[\dimexpr(\textwidth-5pt)/5\relax][c]{\scriptsize Ours (L0)} & \makebox[\dimexpr(\textwidth-5pt)/5\relax][c]{\scriptsize Ours (L1)} & \makebox[\dimexpr(\textwidth-5pt)/5\relax][c]{\scriptsize Ours (L2)} \\[8pt]
    \multicolumn{5}{c}{\scriptsize (b) \texttt{caterpillar}} \\[1.5pt]
    \includegraphics[width=\dimexpr(\textwidth-5pt)/5\relax]{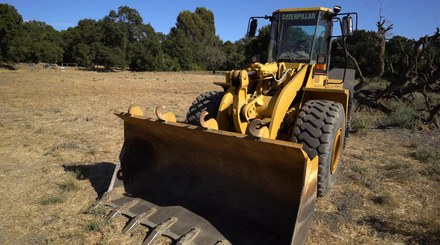} & \includegraphics[width=\dimexpr(\textwidth-5pt)/5\relax]{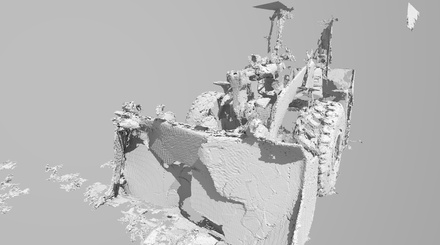} & \includegraphics[width=\dimexpr(\textwidth-5pt)/5\relax]{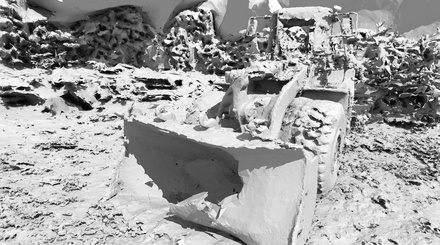} & \includegraphics[width=\dimexpr(\textwidth-5pt)/5\relax]{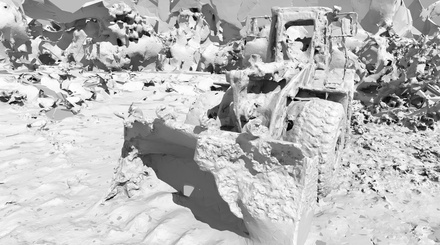} & \includegraphics[width=\dimexpr(\textwidth-5pt)/5\relax]{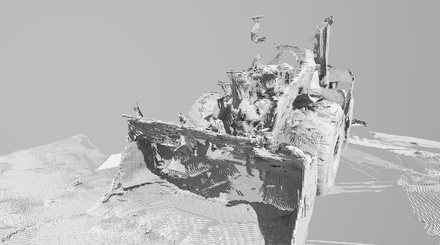} \\[-4.5pt]
    \makebox[\dimexpr(\textwidth-5pt)/5\relax][c]{\scriptsize Image} & \makebox[\dimexpr(\textwidth-5pt)/5\relax][c]{\scriptsize 2DGS} & \makebox[\dimexpr(\textwidth-5pt)/5\relax][c]{\scriptsize MILo} & \makebox[\dimexpr(\textwidth-5pt)/5\relax][c]{\scriptsize GaussianWrapping} & \makebox[\dimexpr(\textwidth-5pt)/5\relax][c]{\scriptsize DA3} \\[3pt]
    \includegraphics[width=\dimexpr(\textwidth-5pt)/5\relax]{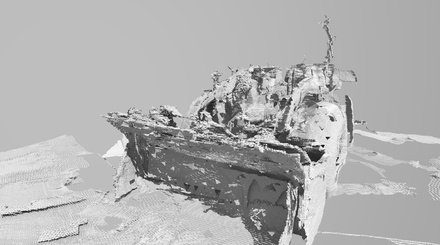} & \includegraphics[width=\dimexpr(\textwidth-5pt)/5\relax]{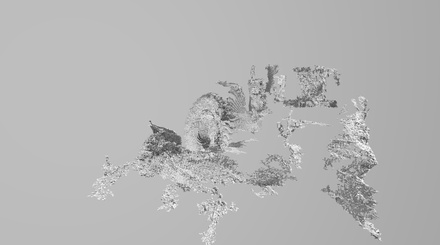} & \includegraphics[width=\dimexpr(\textwidth-5pt)/5\relax]{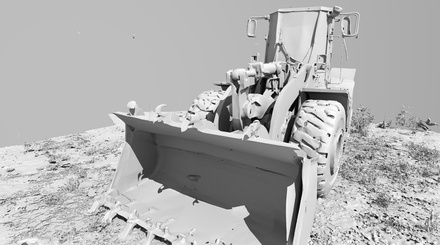} & \includegraphics[width=\dimexpr(\textwidth-5pt)/5\relax]{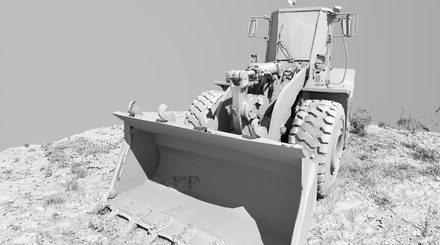} & \includegraphics[width=\dimexpr(\textwidth-5pt)/5\relax]{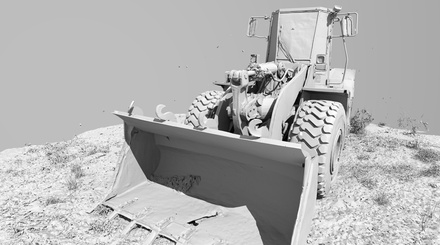} \\[-4.5pt]
    \makebox[\dimexpr(\textwidth-5pt)/5\relax][c]{\scriptsize MapAnything} & \makebox[\dimexpr(\textwidth-5pt)/5\relax][c]{\scriptsize Murre} & \makebox[\dimexpr(\textwidth-5pt)/5\relax][c]{\scriptsize Ours (L0)} & \makebox[\dimexpr(\textwidth-5pt)/5\relax][c]{\scriptsize Ours (L1)} & \makebox[\dimexpr(\textwidth-5pt)/5\relax][c]{\scriptsize Ours (L2)} \\[8pt]
    \multicolumn{5}{c}{\scriptsize (c) \texttt{courthouse}} \\[1.5pt]
    \includegraphics[width=\dimexpr(\textwidth-5pt)/5\relax]{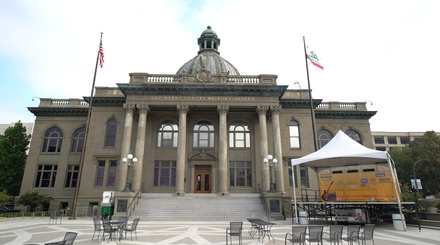} & \includegraphics[width=\dimexpr(\textwidth-5pt)/5\relax]{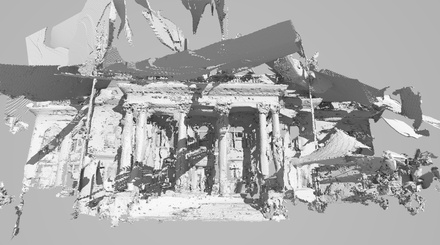} & \includegraphics[width=\dimexpr(\textwidth-5pt)/5\relax]{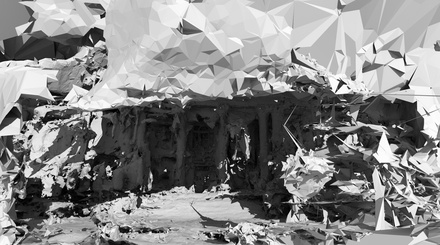} & \includegraphics[width=\dimexpr(\textwidth-5pt)/5\relax]{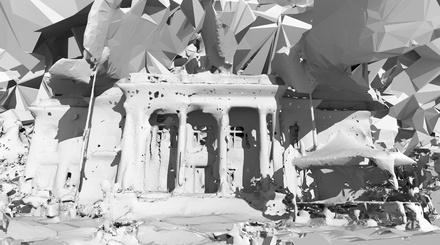} & \includegraphics[width=\dimexpr(\textwidth-5pt)/5\relax]{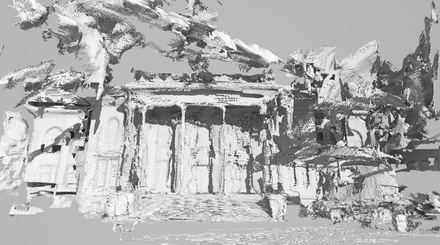} \\[-4.5pt]
    \makebox[\dimexpr(\textwidth-5pt)/5\relax][c]{\scriptsize Image} & \makebox[\dimexpr(\textwidth-5pt)/5\relax][c]{\scriptsize 2DGS} & \makebox[\dimexpr(\textwidth-5pt)/5\relax][c]{\scriptsize MILo} & \makebox[\dimexpr(\textwidth-5pt)/5\relax][c]{\scriptsize GaussianWrapping} & \makebox[\dimexpr(\textwidth-5pt)/5\relax][c]{\scriptsize DA3} \\[3pt]
    \includegraphics[width=\dimexpr(\textwidth-5pt)/5\relax]{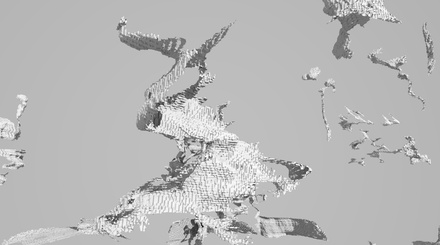} & \includegraphics[width=\dimexpr(\textwidth-5pt)/5\relax]{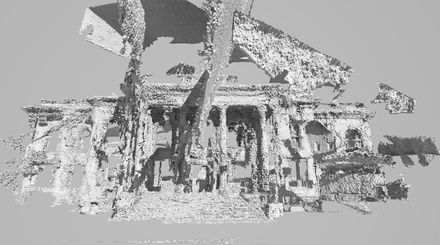} & \includegraphics[width=\dimexpr(\textwidth-5pt)/5\relax]{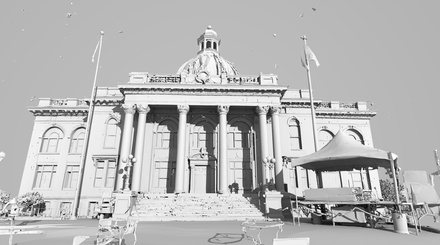} & \includegraphics[width=\dimexpr(\textwidth-5pt)/5\relax]{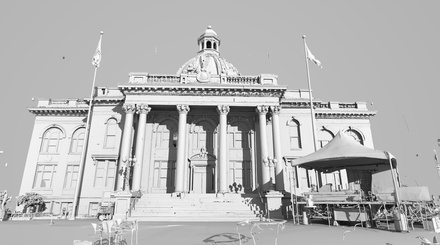} & \includegraphics[width=\dimexpr(\textwidth-5pt)/5\relax]{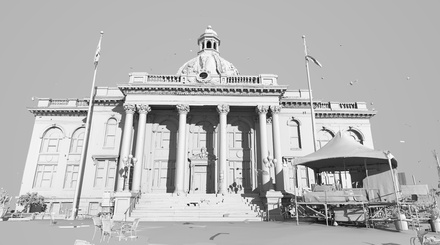} \\[-4.5pt]
    \makebox[\dimexpr(\textwidth-5pt)/5\relax][c]{\scriptsize MapAnything} & \makebox[\dimexpr(\textwidth-5pt)/5\relax][c]{\scriptsize Murre} & \makebox[\dimexpr(\textwidth-5pt)/5\relax][c]{\scriptsize Ours (L0)} & \makebox[\dimexpr(\textwidth-5pt)/5\relax][c]{\scriptsize Ours (L1)} & \makebox[\dimexpr(\textwidth-5pt)/5\relax][c]{\scriptsize Ours (L2)} \\
  \end{tabular}
  \caption{\textbf{Qualitative comparison on Tanks and Temples with 16 input views.} All meshes of a scene are rendered from the same viewpoint.}
  \label{fig:tnt-appendix-fps16:1}
\end{figure}

\begin{figure}[p]
  \centering
  \setlength{\tabcolsep}{0.5pt}
  \renewcommand{\arraystretch}{1.0}
  \begin{tabular}{ccccc}
    \multicolumn{5}{c}{\scriptsize (a) \texttt{ignatius}} \\[1.5pt]
    \includegraphics[width=\dimexpr(\textwidth-5pt)/5\relax]{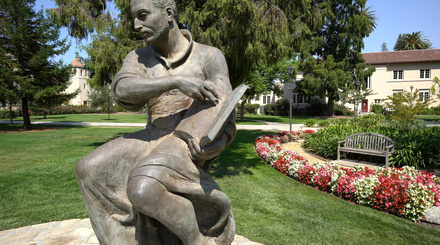} & \includegraphics[width=\dimexpr(\textwidth-5pt)/5\relax]{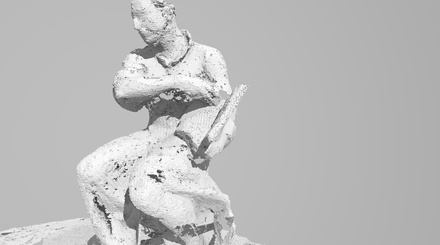} & \includegraphics[width=\dimexpr(\textwidth-5pt)/5\relax]{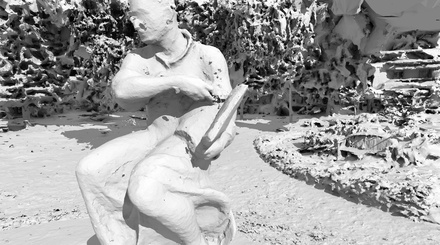} & \includegraphics[width=\dimexpr(\textwidth-5pt)/5\relax]{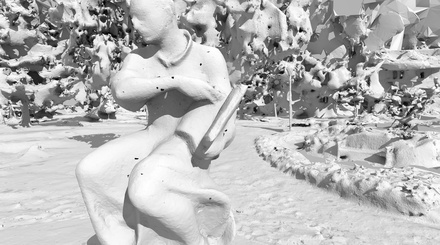} & \includegraphics[width=\dimexpr(\textwidth-5pt)/5\relax]{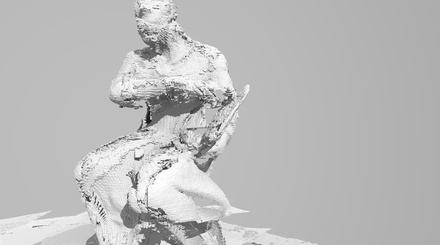} \\[-4.5pt]
    \makebox[\dimexpr(\textwidth-5pt)/5\relax][c]{\scriptsize Image} & \makebox[\dimexpr(\textwidth-5pt)/5\relax][c]{\scriptsize 2DGS} & \makebox[\dimexpr(\textwidth-5pt)/5\relax][c]{\scriptsize MILo} & \makebox[\dimexpr(\textwidth-5pt)/5\relax][c]{\scriptsize GaussianWrapping} & \makebox[\dimexpr(\textwidth-5pt)/5\relax][c]{\scriptsize DA3} \\[3pt]
    \includegraphics[width=\dimexpr(\textwidth-5pt)/5\relax]{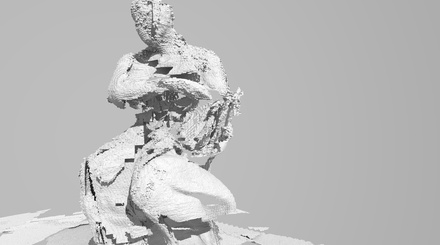} & \includegraphics[width=\dimexpr(\textwidth-5pt)/5\relax]{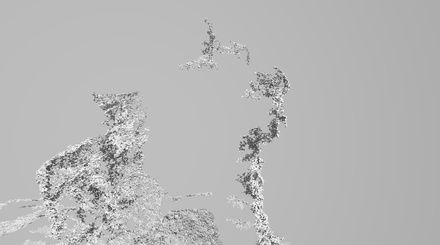} & \includegraphics[width=\dimexpr(\textwidth-5pt)/5\relax]{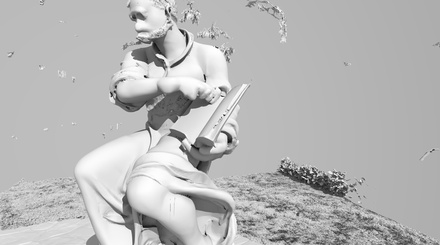} & \includegraphics[width=\dimexpr(\textwidth-5pt)/5\relax]{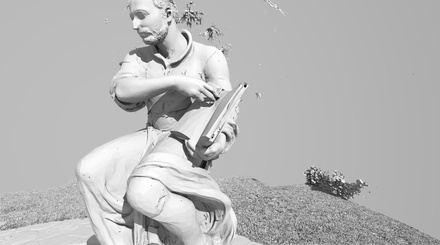} & \includegraphics[width=\dimexpr(\textwidth-5pt)/5\relax]{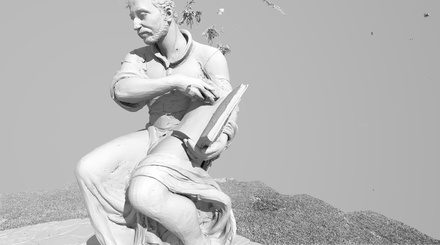} \\[-4.5pt]
    \makebox[\dimexpr(\textwidth-5pt)/5\relax][c]{\scriptsize MapAnything} & \makebox[\dimexpr(\textwidth-5pt)/5\relax][c]{\scriptsize Murre} & \makebox[\dimexpr(\textwidth-5pt)/5\relax][c]{\scriptsize Ours (L0)} & \makebox[\dimexpr(\textwidth-5pt)/5\relax][c]{\scriptsize Ours (L1)} & \makebox[\dimexpr(\textwidth-5pt)/5\relax][c]{\scriptsize Ours (L2)} \\[8pt]
    \multicolumn{5}{c}{\scriptsize (b) \texttt{meetingroom}} \\[1.5pt]
    \includegraphics[width=\dimexpr(\textwidth-5pt)/5\relax]{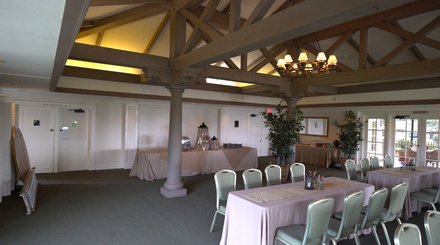} & \includegraphics[width=\dimexpr(\textwidth-5pt)/5\relax]{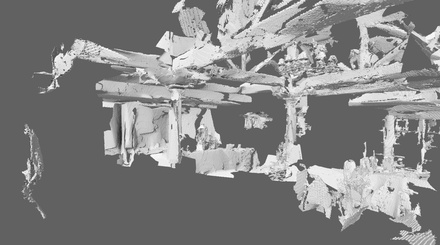} & \includegraphics[width=\dimexpr(\textwidth-5pt)/5\relax]{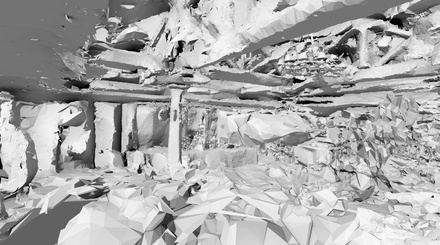} & \includegraphics[width=\dimexpr(\textwidth-5pt)/5\relax]{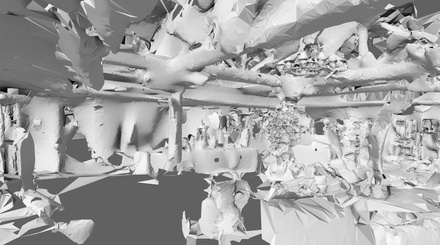} & \includegraphics[width=\dimexpr(\textwidth-5pt)/5\relax]{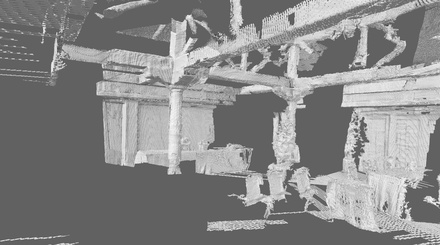} \\[-4.5pt]
    \makebox[\dimexpr(\textwidth-5pt)/5\relax][c]{\scriptsize Image} & \makebox[\dimexpr(\textwidth-5pt)/5\relax][c]{\scriptsize 2DGS} & \makebox[\dimexpr(\textwidth-5pt)/5\relax][c]{\scriptsize MILo} & \makebox[\dimexpr(\textwidth-5pt)/5\relax][c]{\scriptsize GaussianWrapping} & \makebox[\dimexpr(\textwidth-5pt)/5\relax][c]{\scriptsize DA3} \\[3pt]
    \includegraphics[width=\dimexpr(\textwidth-5pt)/5\relax]{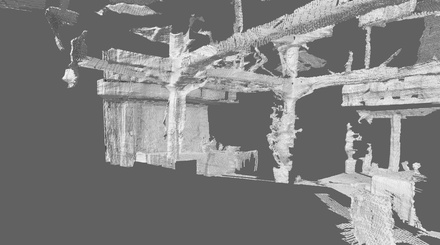} & \includegraphics[width=\dimexpr(\textwidth-5pt)/5\relax]{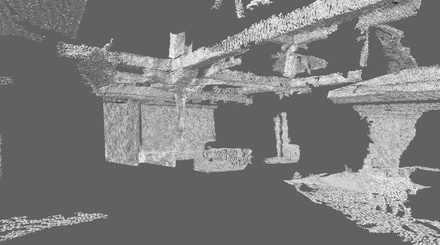} & \includegraphics[width=\dimexpr(\textwidth-5pt)/5\relax]{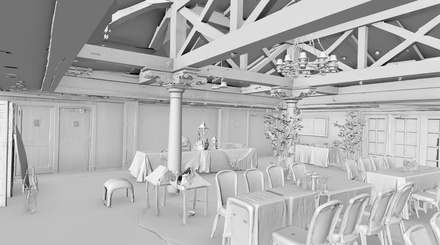} & \includegraphics[width=\dimexpr(\textwidth-5pt)/5\relax]{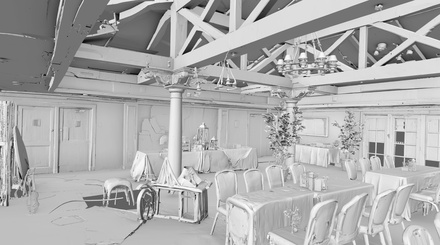} & \includegraphics[width=\dimexpr(\textwidth-5pt)/5\relax]{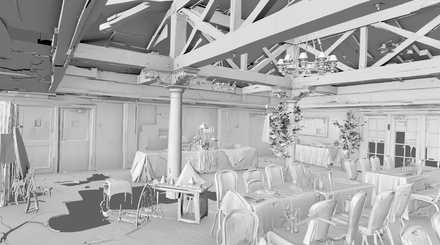} \\[-4.5pt]
    \makebox[\dimexpr(\textwidth-5pt)/5\relax][c]{\scriptsize MapAnything} & \makebox[\dimexpr(\textwidth-5pt)/5\relax][c]{\scriptsize Murre} & \makebox[\dimexpr(\textwidth-5pt)/5\relax][c]{\scriptsize Ours (L0)} & \makebox[\dimexpr(\textwidth-5pt)/5\relax][c]{\scriptsize Ours (L1)} & \makebox[\dimexpr(\textwidth-5pt)/5\relax][c]{\scriptsize Ours (L2)} \\[8pt]
    \multicolumn{5}{c}{\scriptsize (c) \texttt{truck}} \\[1.5pt]
    \includegraphics[width=\dimexpr(\textwidth-5pt)/5\relax]{imgs/tnt_appendix_fps16/truck/gt_rgb.jpg} & \includegraphics[width=\dimexpr(\textwidth-5pt)/5\relax]{imgs/tnt_appendix_fps16/truck/2dgs.jpg} & \includegraphics[width=\dimexpr(\textwidth-5pt)/5\relax]{imgs/tnt_appendix_fps16/truck/milo.jpg} & \includegraphics[width=\dimexpr(\textwidth-5pt)/5\relax]{imgs/tnt_appendix_fps16/truck/gaussianwrapping.jpg} & \includegraphics[width=\dimexpr(\textwidth-5pt)/5\relax]{imgs/tnt_appendix_fps16/truck/da3.jpg} \\[-4.5pt]
    \makebox[\dimexpr(\textwidth-5pt)/5\relax][c]{\scriptsize Image} & \makebox[\dimexpr(\textwidth-5pt)/5\relax][c]{\scriptsize 2DGS} & \makebox[\dimexpr(\textwidth-5pt)/5\relax][c]{\scriptsize MILo} & \makebox[\dimexpr(\textwidth-5pt)/5\relax][c]{\scriptsize GaussianWrapping} & \makebox[\dimexpr(\textwidth-5pt)/5\relax][c]{\scriptsize DA3} \\[3pt]
    \includegraphics[width=\dimexpr(\textwidth-5pt)/5\relax]{imgs/tnt_appendix_fps16/truck/mapanything.jpg} & \includegraphics[width=\dimexpr(\textwidth-5pt)/5\relax]{imgs/tnt_appendix_fps16/truck/murre.jpg} & \includegraphics[width=\dimexpr(\textwidth-5pt)/5\relax]{imgs/tnt_appendix_fps16/truck/ours_g004.jpg} & \includegraphics[width=\dimexpr(\textwidth-5pt)/5\relax]{imgs/tnt_appendix_fps16/truck/ours_g002.jpg} & \includegraphics[width=\dimexpr(\textwidth-5pt)/5\relax]{imgs/tnt_appendix_fps16/truck/ours_g001.jpg} \\[-4.5pt]
    \makebox[\dimexpr(\textwidth-5pt)/5\relax][c]{\scriptsize MapAnything} & \makebox[\dimexpr(\textwidth-5pt)/5\relax][c]{\scriptsize Murre} & \makebox[\dimexpr(\textwidth-5pt)/5\relax][c]{\scriptsize Ours (L0)} & \makebox[\dimexpr(\textwidth-5pt)/5\relax][c]{\scriptsize Ours (L1)} & \makebox[\dimexpr(\textwidth-5pt)/5\relax][c]{\scriptsize Ours (L2)} \\
  \end{tabular}
  \caption{\textbf{Qualitative comparison on Tanks and Temples with 16 input views.} All meshes of a scene are rendered from the same viewpoint.}
  \label{fig:tnt-appendix-fps16:2}
\end{figure}

\begin{figure}[p]
  \centering
  \setlength{\tabcolsep}{0.5pt}
  \renewcommand{\arraystretch}{1.0}
  \begin{tabular}{ccccc}
    \multicolumn{5}{c}{\scriptsize (a) \texttt{barn}} \\[1.5pt]
    \includegraphics[width=\dimexpr(\textwidth-5pt)/5\relax]{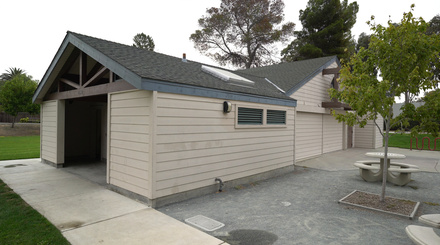} & \includegraphics[width=\dimexpr(\textwidth-5pt)/5\relax]{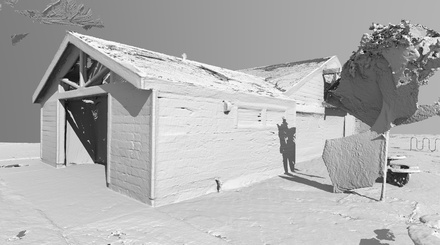} & \includegraphics[width=\dimexpr(\textwidth-5pt)/5\relax]{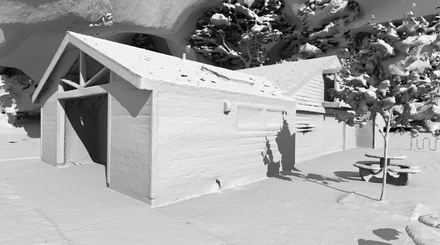} & \includegraphics[width=\dimexpr(\textwidth-5pt)/5\relax]{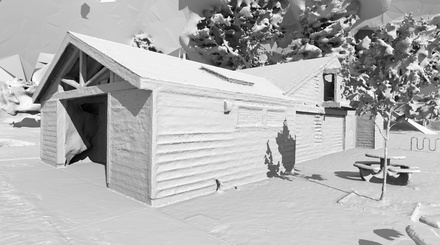} & \includegraphics[width=\dimexpr(\textwidth-5pt)/5\relax]{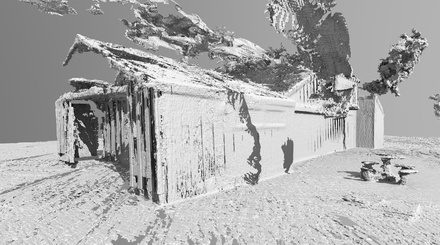} \\[-4.5pt]
    \makebox[\dimexpr(\textwidth-5pt)/5\relax][c]{\scriptsize Image} & \makebox[\dimexpr(\textwidth-5pt)/5\relax][c]{\scriptsize 2DGS} & \makebox[\dimexpr(\textwidth-5pt)/5\relax][c]{\scriptsize MILo} & \makebox[\dimexpr(\textwidth-5pt)/5\relax][c]{\scriptsize GaussianWrapping} & \makebox[\dimexpr(\textwidth-5pt)/5\relax][c]{\scriptsize DA3} \\[3pt]
    \includegraphics[width=\dimexpr(\textwidth-5pt)/5\relax]{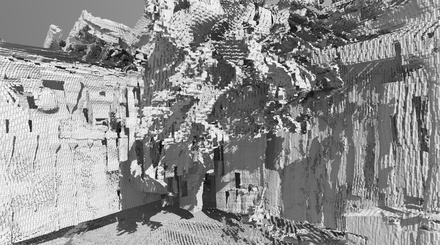} & \includegraphics[width=\dimexpr(\textwidth-5pt)/5\relax]{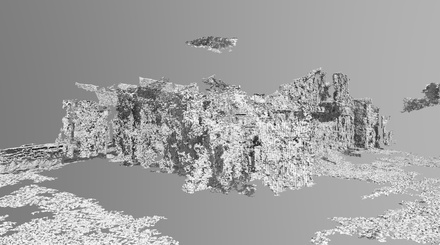} & \includegraphics[width=\dimexpr(\textwidth-5pt)/5\relax]{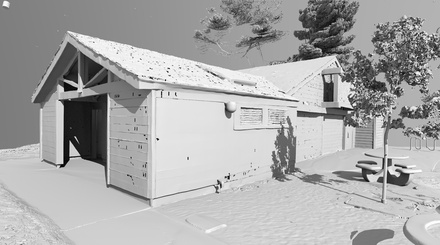} & \includegraphics[width=\dimexpr(\textwidth-5pt)/5\relax]{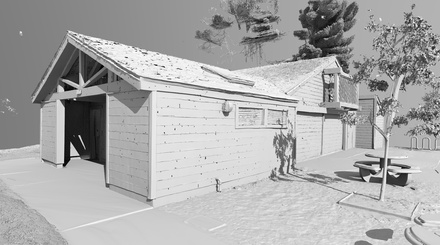} & \includegraphics[width=\dimexpr(\textwidth-5pt)/5\relax]{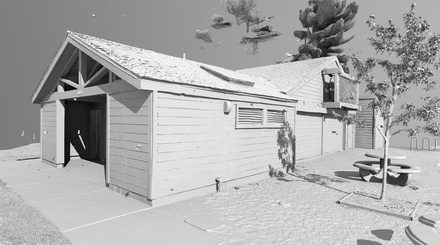} \\[-4.5pt]
    \makebox[\dimexpr(\textwidth-5pt)/5\relax][c]{\scriptsize MapAnything} & \makebox[\dimexpr(\textwidth-5pt)/5\relax][c]{\scriptsize Murre} & \makebox[\dimexpr(\textwidth-5pt)/5\relax][c]{\scriptsize Ours (L0)} & \makebox[\dimexpr(\textwidth-5pt)/5\relax][c]{\scriptsize Ours (L1)} & \makebox[\dimexpr(\textwidth-5pt)/5\relax][c]{\scriptsize Ours (L2)} \\[8pt]
    \multicolumn{5}{c}{\scriptsize (b) \texttt{caterpillar}} \\[1.5pt]
    \includegraphics[width=\dimexpr(\textwidth-5pt)/5\relax]{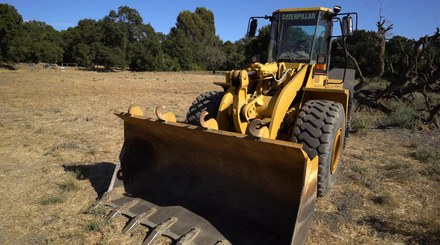} & \includegraphics[width=\dimexpr(\textwidth-5pt)/5\relax]{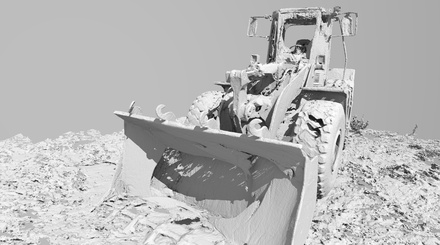} & \includegraphics[width=\dimexpr(\textwidth-5pt)/5\relax]{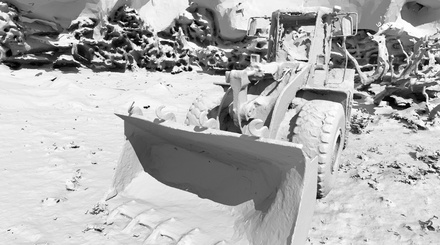} & \includegraphics[width=\dimexpr(\textwidth-5pt)/5\relax]{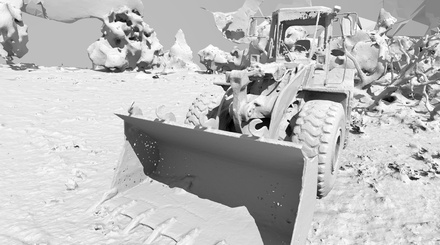} & \includegraphics[width=\dimexpr(\textwidth-5pt)/5\relax]{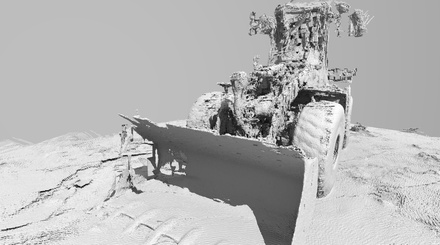} \\[-4.5pt]
    \makebox[\dimexpr(\textwidth-5pt)/5\relax][c]{\scriptsize Image} & \makebox[\dimexpr(\textwidth-5pt)/5\relax][c]{\scriptsize 2DGS} & \makebox[\dimexpr(\textwidth-5pt)/5\relax][c]{\scriptsize MILo} & \makebox[\dimexpr(\textwidth-5pt)/5\relax][c]{\scriptsize GaussianWrapping} & \makebox[\dimexpr(\textwidth-5pt)/5\relax][c]{\scriptsize DA3} \\[3pt]
    \includegraphics[width=\dimexpr(\textwidth-5pt)/5\relax]{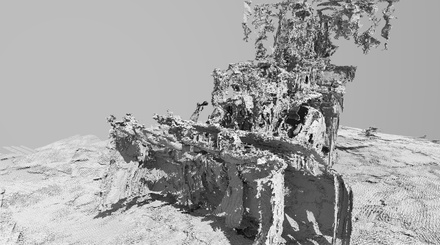} & \includegraphics[width=\dimexpr(\textwidth-5pt)/5\relax]{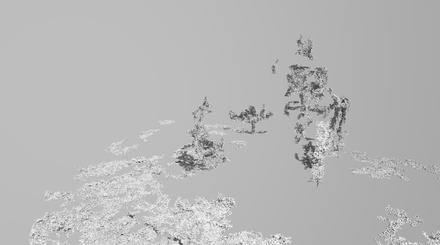} & \includegraphics[width=\dimexpr(\textwidth-5pt)/5\relax]{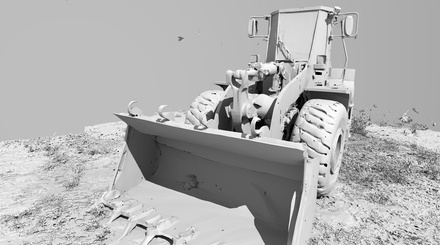} & \includegraphics[width=\dimexpr(\textwidth-5pt)/5\relax]{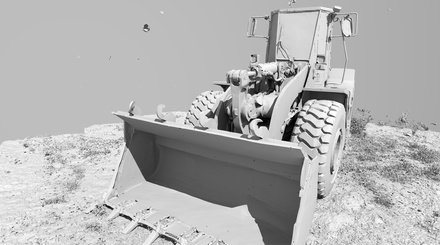} & \includegraphics[width=\dimexpr(\textwidth-5pt)/5\relax]{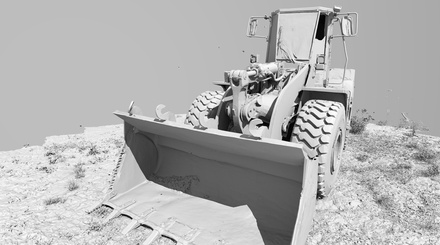} \\[-4.5pt]
    \makebox[\dimexpr(\textwidth-5pt)/5\relax][c]{\scriptsize MapAnything} & \makebox[\dimexpr(\textwidth-5pt)/5\relax][c]{\scriptsize Murre} & \makebox[\dimexpr(\textwidth-5pt)/5\relax][c]{\scriptsize Ours (L0)} & \makebox[\dimexpr(\textwidth-5pt)/5\relax][c]{\scriptsize Ours (L1)} & \makebox[\dimexpr(\textwidth-5pt)/5\relax][c]{\scriptsize Ours (L2)} \\[8pt]
    \multicolumn{5}{c}{\scriptsize (c) \texttt{courthouse}} \\[1.5pt]
    \includegraphics[width=\dimexpr(\textwidth-5pt)/5\relax]{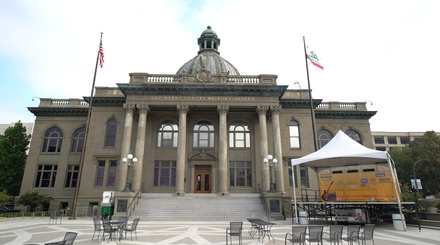} & \includegraphics[width=\dimexpr(\textwidth-5pt)/5\relax]{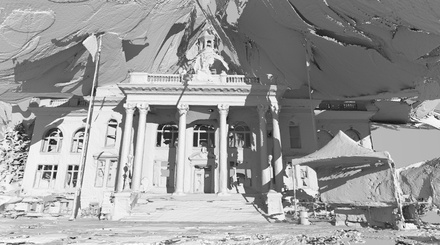} & \includegraphics[width=\dimexpr(\textwidth-5pt)/5\relax]{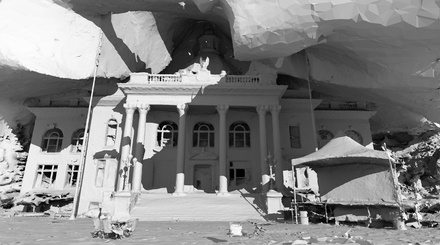} & \includegraphics[width=\dimexpr(\textwidth-5pt)/5\relax]{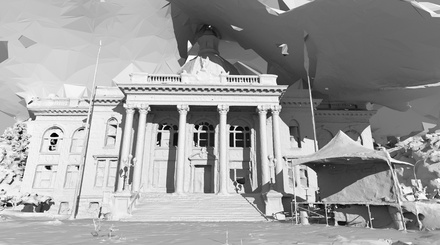} & \includegraphics[width=\dimexpr(\textwidth-5pt)/5\relax]{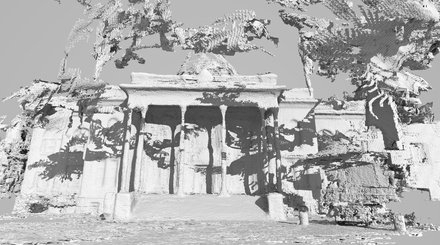} \\[-4.5pt]
    \makebox[\dimexpr(\textwidth-5pt)/5\relax][c]{\scriptsize Image} & \makebox[\dimexpr(\textwidth-5pt)/5\relax][c]{\scriptsize 2DGS} & \makebox[\dimexpr(\textwidth-5pt)/5\relax][c]{\scriptsize MILo} & \makebox[\dimexpr(\textwidth-5pt)/5\relax][c]{\scriptsize GaussianWrapping} & \makebox[\dimexpr(\textwidth-5pt)/5\relax][c]{\scriptsize DA3} \\[3pt]
    \includegraphics[width=\dimexpr(\textwidth-5pt)/5\relax]{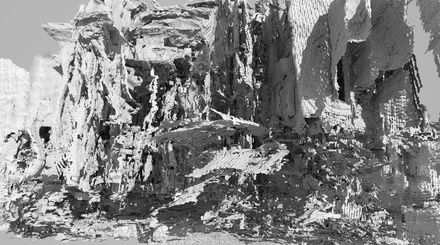} & \includegraphics[width=\dimexpr(\textwidth-5pt)/5\relax]{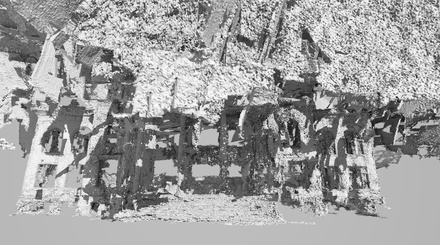} & \includegraphics[width=\dimexpr(\textwidth-5pt)/5\relax]{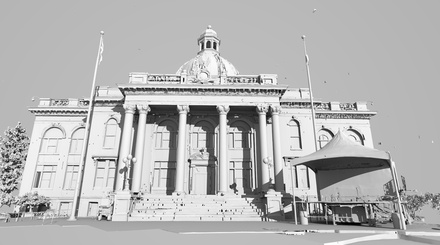} & \includegraphics[width=\dimexpr(\textwidth-5pt)/5\relax]{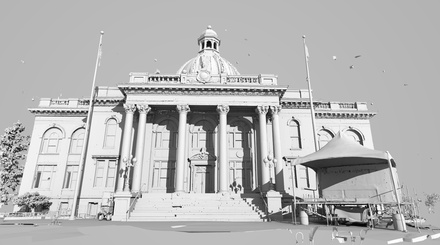} & \includegraphics[width=\dimexpr(\textwidth-5pt)/5\relax]{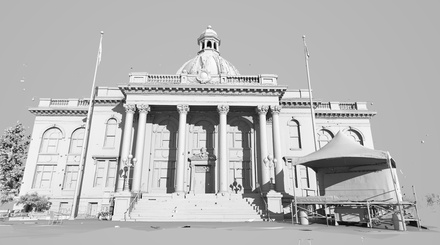} \\[-4.5pt]
    \makebox[\dimexpr(\textwidth-5pt)/5\relax][c]{\scriptsize MapAnything} & \makebox[\dimexpr(\textwidth-5pt)/5\relax][c]{\scriptsize Murre} & \makebox[\dimexpr(\textwidth-5pt)/5\relax][c]{\scriptsize Ours (L0)} & \makebox[\dimexpr(\textwidth-5pt)/5\relax][c]{\scriptsize Ours (L1)} & \makebox[\dimexpr(\textwidth-5pt)/5\relax][c]{\scriptsize Ours (L2)} \\
  \end{tabular}
  \caption{\textbf{Qualitative comparison on Tanks and Temples with 256 input views.} All meshes of a scene are rendered from the same viewpoint.}
  \label{fig:tnt-appendix:1}
\end{figure}

\begin{figure}[p]
  \centering
  \setlength{\tabcolsep}{0.5pt}
  \renewcommand{\arraystretch}{1.0}
  \begin{tabular}{ccccc}
    \multicolumn{5}{c}{\scriptsize (a) \texttt{ignatius}} \\[1.5pt]
    \includegraphics[width=\dimexpr(\textwidth-5pt)/5\relax]{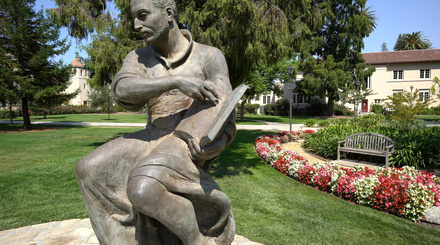} & \includegraphics[width=\dimexpr(\textwidth-5pt)/5\relax]{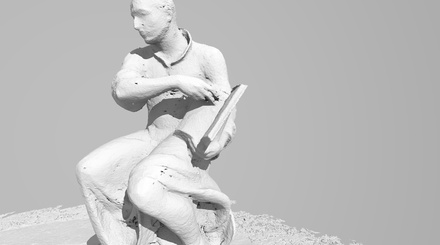} & \includegraphics[width=\dimexpr(\textwidth-5pt)/5\relax]{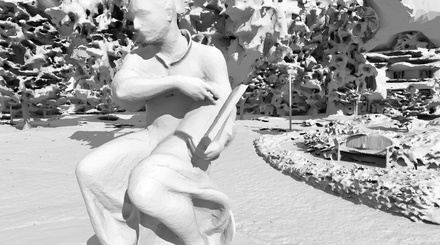} & \includegraphics[width=\dimexpr(\textwidth-5pt)/5\relax]{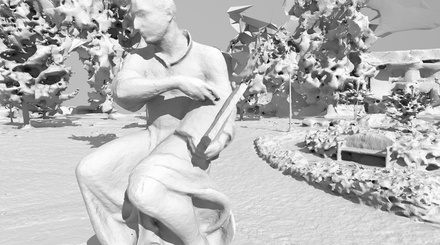} & \includegraphics[width=\dimexpr(\textwidth-5pt)/5\relax]{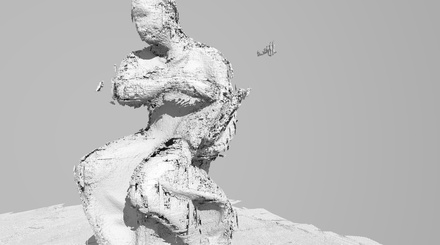} \\[-4.5pt]
    \makebox[\dimexpr(\textwidth-5pt)/5\relax][c]{\scriptsize Image} & \makebox[\dimexpr(\textwidth-5pt)/5\relax][c]{\scriptsize 2DGS} & \makebox[\dimexpr(\textwidth-5pt)/5\relax][c]{\scriptsize MILo} & \makebox[\dimexpr(\textwidth-5pt)/5\relax][c]{\scriptsize GaussianWrapping} & \makebox[\dimexpr(\textwidth-5pt)/5\relax][c]{\scriptsize DA3} \\[3pt]
    \includegraphics[width=\dimexpr(\textwidth-5pt)/5\relax]{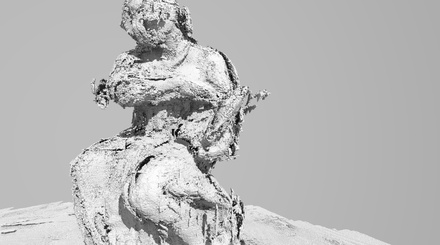} & \includegraphics[width=\dimexpr(\textwidth-5pt)/5\relax]{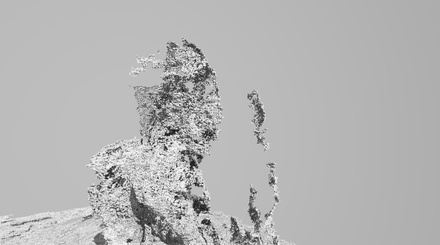} & \includegraphics[width=\dimexpr(\textwidth-5pt)/5\relax]{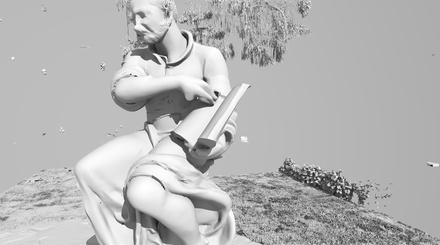} & \includegraphics[width=\dimexpr(\textwidth-5pt)/5\relax]{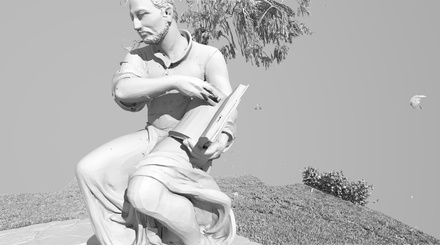} & \includegraphics[width=\dimexpr(\textwidth-5pt)/5\relax]{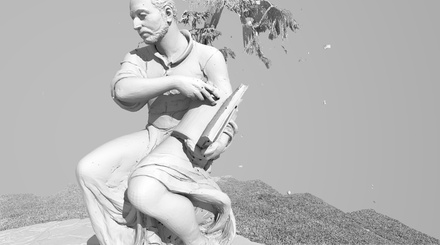} \\[-4.5pt]
    \makebox[\dimexpr(\textwidth-5pt)/5\relax][c]{\scriptsize MapAnything} & \makebox[\dimexpr(\textwidth-5pt)/5\relax][c]{\scriptsize Murre} & \makebox[\dimexpr(\textwidth-5pt)/5\relax][c]{\scriptsize Ours (L0)} & \makebox[\dimexpr(\textwidth-5pt)/5\relax][c]{\scriptsize Ours (L1)} & \makebox[\dimexpr(\textwidth-5pt)/5\relax][c]{\scriptsize Ours (L2)} \\[8pt]
    \multicolumn{5}{c}{\scriptsize (b) \texttt{meetingroom}} \\[1.5pt]
    \includegraphics[width=\dimexpr(\textwidth-5pt)/5\relax]{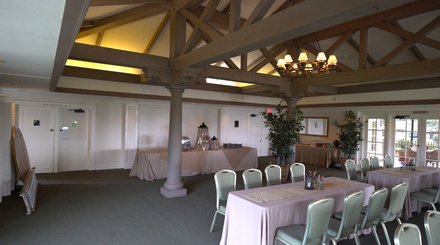} & \includegraphics[width=\dimexpr(\textwidth-5pt)/5\relax]{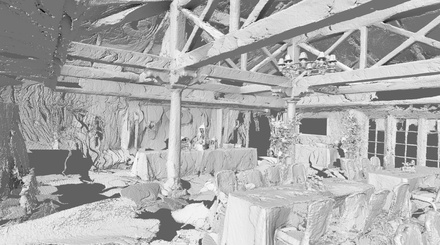} & \includegraphics[width=\dimexpr(\textwidth-5pt)/5\relax]{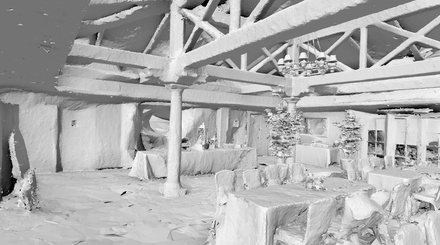} & \includegraphics[width=\dimexpr(\textwidth-5pt)/5\relax]{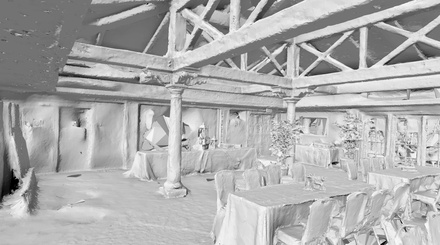} & \includegraphics[width=\dimexpr(\textwidth-5pt)/5\relax]{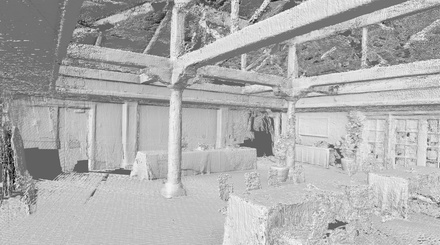} \\[-4.5pt]
    \makebox[\dimexpr(\textwidth-5pt)/5\relax][c]{\scriptsize Image} & \makebox[\dimexpr(\textwidth-5pt)/5\relax][c]{\scriptsize 2DGS} & \makebox[\dimexpr(\textwidth-5pt)/5\relax][c]{\scriptsize MILo} & \makebox[\dimexpr(\textwidth-5pt)/5\relax][c]{\scriptsize GaussianWrapping} & \makebox[\dimexpr(\textwidth-5pt)/5\relax][c]{\scriptsize DA3} \\[3pt]
    \includegraphics[width=\dimexpr(\textwidth-5pt)/5\relax]{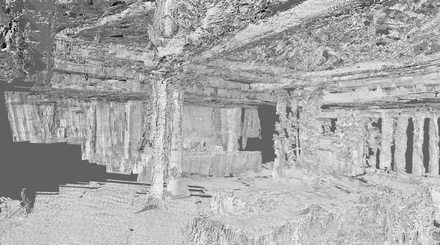} & \includegraphics[width=\dimexpr(\textwidth-5pt)/5\relax]{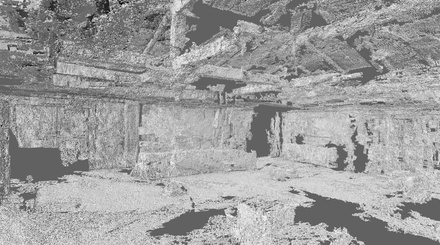} & \includegraphics[width=\dimexpr(\textwidth-5pt)/5\relax]{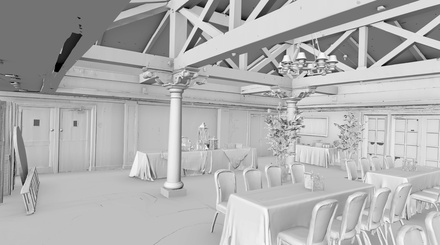} & \includegraphics[width=\dimexpr(\textwidth-5pt)/5\relax]{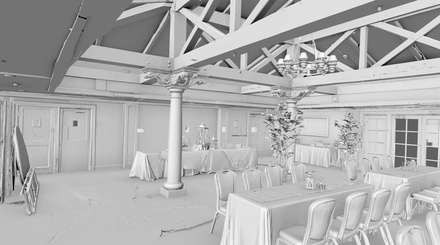} & \includegraphics[width=\dimexpr(\textwidth-5pt)/5\relax]{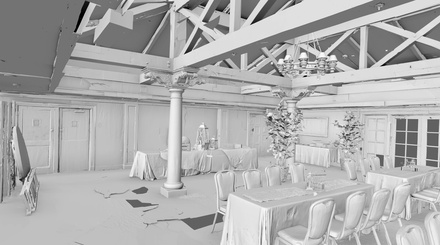} \\[-4.5pt]
    \makebox[\dimexpr(\textwidth-5pt)/5\relax][c]{\scriptsize MapAnything} & \makebox[\dimexpr(\textwidth-5pt)/5\relax][c]{\scriptsize Murre} & \makebox[\dimexpr(\textwidth-5pt)/5\relax][c]{\scriptsize Ours (L0)} & \makebox[\dimexpr(\textwidth-5pt)/5\relax][c]{\scriptsize Ours (L1)} & \makebox[\dimexpr(\textwidth-5pt)/5\relax][c]{\scriptsize Ours (L2)} \\[8pt]
    \multicolumn{5}{c}{\scriptsize (c) \texttt{truck}} \\[1.5pt]
    \includegraphics[width=\dimexpr(\textwidth-5pt)/5\relax]{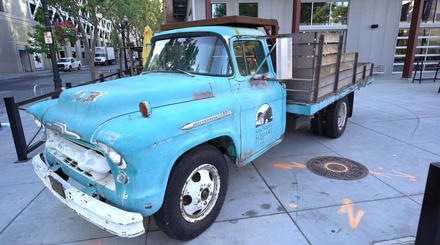} & \includegraphics[width=\dimexpr(\textwidth-5pt)/5\relax]{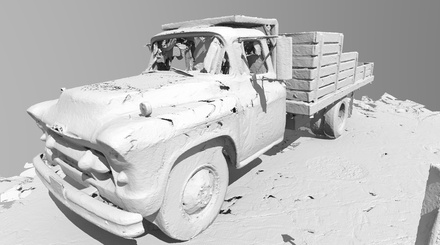} & \includegraphics[width=\dimexpr(\textwidth-5pt)/5\relax]{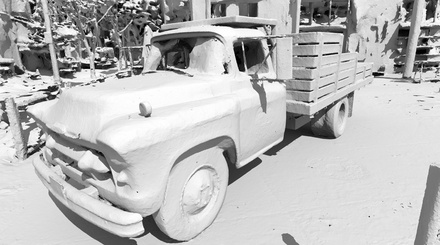} & \includegraphics[width=\dimexpr(\textwidth-5pt)/5\relax]{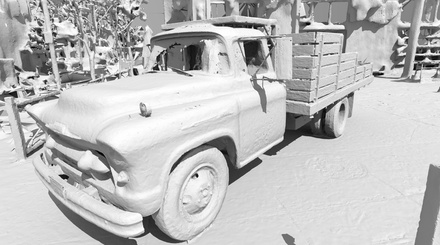} & \includegraphics[width=\dimexpr(\textwidth-5pt)/5\relax]{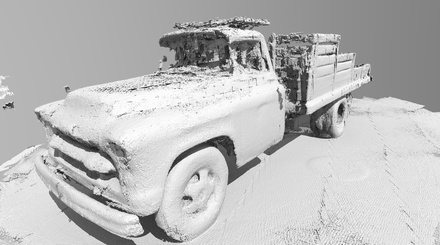} \\[-4.5pt]
    \makebox[\dimexpr(\textwidth-5pt)/5\relax][c]{\scriptsize Image} & \makebox[\dimexpr(\textwidth-5pt)/5\relax][c]{\scriptsize 2DGS} & \makebox[\dimexpr(\textwidth-5pt)/5\relax][c]{\scriptsize MILo} & \makebox[\dimexpr(\textwidth-5pt)/5\relax][c]{\scriptsize GaussianWrapping} & \makebox[\dimexpr(\textwidth-5pt)/5\relax][c]{\scriptsize DA3} \\[3pt]
    \includegraphics[width=\dimexpr(\textwidth-5pt)/5\relax]{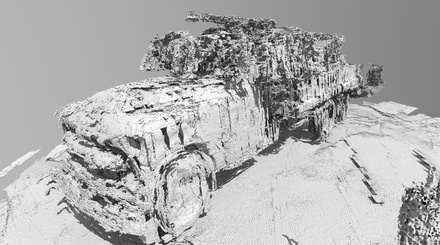} & \includegraphics[width=\dimexpr(\textwidth-5pt)/5\relax]{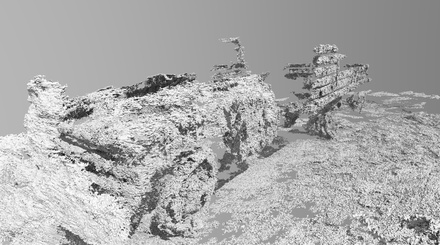} & \includegraphics[width=\dimexpr(\textwidth-5pt)/5\relax]{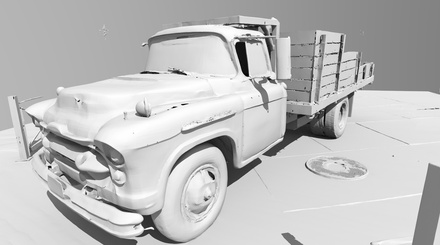} & \includegraphics[width=\dimexpr(\textwidth-5pt)/5\relax]{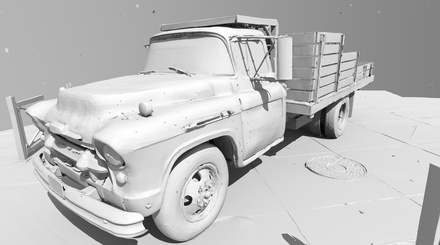} & \includegraphics[width=\dimexpr(\textwidth-5pt)/5\relax]{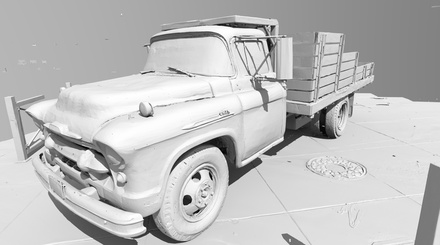} \\[-4.5pt]
    \makebox[\dimexpr(\textwidth-5pt)/5\relax][c]{\scriptsize MapAnything} & \makebox[\dimexpr(\textwidth-5pt)/5\relax][c]{\scriptsize Murre} & \makebox[\dimexpr(\textwidth-5pt)/5\relax][c]{\scriptsize Ours (L0)} & \makebox[\dimexpr(\textwidth-5pt)/5\relax][c]{\scriptsize Ours (L1)} & \makebox[\dimexpr(\textwidth-5pt)/5\relax][c]{\scriptsize Ours (L2)} \\
  \end{tabular}
  \caption{\textbf{Qualitative comparison on Tanks and Temples with 256 input views.} All meshes of a scene are rendered from the same viewpoint.}
  \label{fig:tnt-appendix:2}
\end{figure}

\clearpage

\subsection{City blocks}
\label{app:city-results}

Table~\ref{tab:bim-per-scene} reports per-block results with 256 input views. Each
finer cascade level improves all three metrics on every block.

\begin{table}[H]
  \centering
  \caption{\textbf{Per-block quantitative comparison on the city-block dataset.} All methods use 256 input views. The cell size of each level of our method is shown in parentheses. CD is in meters. The top three results are highlighted.}
  \label{tab:bim-per-scene}
  \scriptsize
  \setlength{\tabcolsep}{3pt}
  \definecolor{rank1}{HTML}{FFB3B3}
  \definecolor{rank2}{HTML}{FFD9B3}
  \definecolor{rank3}{HTML}{FFFFB3}
  \definecolor{emphrow}{HTML}{EDEDED}
  \resizebox{\linewidth}{!}{%
  \begin{tabular}{lcccccccccccc}
    \toprule
     & \multicolumn{3}{c}{Scene 025} & \multicolumn{3}{c}{Scene 033} & \multicolumn{3}{c}{Scene 034} & \multicolumn{3}{c}{Mean} \\
    \cmidrule(lr){2-4}\cmidrule(lr){5-7}\cmidrule(lr){8-10}\cmidrule(lr){11-13}
    Method & CD $\downarrow$ & F@10 $\uparrow$ & NC $\uparrow$ & CD $\downarrow$ & F@10 $\uparrow$ & NC $\uparrow$ & CD $\downarrow$ & F@10 $\uparrow$ & NC $\uparrow$ & CD $\downarrow$ & F@10 $\uparrow$ & NC $\uparrow$ \\
    \midrule
    \multicolumn{13}{l}{\textbf{\textit{w/ per-scene optimization}}} \\
    2DGS & 4.701 & 0.101 & 0.231 & 5.011 & 0.088 & 0.157 & 4.863 & 0.039 & 0.134 & 4.858 & 0.076 & 0.174 \\
    CityGaussianV2 & 6.278 & 0.119 & 0.216 & 4.262 & 0.123 & 0.217 & 4.926 & 0.094 & 0.252 & 5.155 & 0.112 & 0.228 \\
    \midrule
    \multicolumn{13}{l}{\textbf{\textit{w/o per-scene optimization}}} \\
    DA3 & 4.514 & 0.014 & 0.039 & 4.754 & 0.010 & 0.027 & 6.285 & 0.022 & 0.079 & 5.184 & 0.015 & 0.048 \\
    Murre & 10.744 & 0.001 & 0.008 & 10.189 & 0.001 & 0.021 & 9.061 & 0.000 & 0.002 & 9.998 & 0.001 & 0.010 \\
    \cellcolor{emphrow}\textbf{Ours (L0, 80\,m)} & 0.967 & 0.226 & 0.239 & 0.795 & 0.202 & 0.251 & 0.855 & 0.307 & 0.317 & 0.872 & 0.245 & 0.269 \\
    \cellcolor{emphrow}\textbf{Ours (L1, 40\,m)} & \cellcolor{rank3}0.792 & \cellcolor{rank3}0.282 & \cellcolor{rank3}0.297 & \cellcolor{rank3}0.673 & \cellcolor{rank3}0.272 & \cellcolor{rank3}0.304 & \cellcolor{rank3}0.676 & \cellcolor{rank3}0.335 & \cellcolor{rank3}0.345 & \cellcolor{rank3}0.714 & \cellcolor{rank3}0.296 & \cellcolor{rank3}0.315 \\
    \cellcolor{emphrow}\textbf{Ours (L2, 20\,m)} & \cellcolor{rank2}0.775 & \cellcolor{rank2}0.323 & \cellcolor{rank2}0.330 & \cellcolor{rank2}0.603 & \cellcolor{rank2}0.325 & \cellcolor{rank2}0.337 & \cellcolor{rank2}0.607 & \cellcolor{rank2}0.404 & \cellcolor{rank2}0.382 & \cellcolor{rank2}0.662 & \cellcolor{rank2}0.351 & \cellcolor{rank2}0.350 \\
    \cellcolor{emphrow}\textbf{Ours (L3, 10\,m)} & \cellcolor{rank1}0.747 & \cellcolor{rank1}0.339 & \cellcolor{rank1}0.339 & \cellcolor{rank1}0.596 & \cellcolor{rank1}0.338 & \cellcolor{rank1}0.349 & \cellcolor{rank1}0.591 & \cellcolor{rank1}0.423 & \cellcolor{rank1}0.396 & \cellcolor{rank1}0.645 & \cellcolor{rank1}0.367 & \cellcolor{rank1}0.361 \\
    \bottomrule
  \end{tabular}%
  }
\end{table}

\subsection{Comparison with GaussianWrapping on real-world captures}
\label{app:inthewild-results}

We compare our method with GaussianWrapping, which achieves the highest F@$\tau$
among the per-scene optimization baselines on Tanks and Temples with 256 views.
Figure~\ref{fig:baseline-comparison} shows \texttt{garden} from
Mip-NeRF~360~\citep{barron2022mipnerf360} and an in-house capture of five people. We show the number and resolution of our input views in the figure. Both methods use the same input views, but GaussianWrapping takes them
at 2K resolution. Their meshes are rendered from the same camera with the same shading.

\clearpage
\ifdefined\bcScene \fi
\newlength{\bcGap}\setlength{\bcGap}{4pt}
\newlength{\bcRender}\setlength{\bcRender}{\dimexpr(\textwidth-\bcGap-1pt)/2\relax}
\newlength{\bcCrop}\setlength{\bcCrop}{\dimexpr(\bcRender-2pt)/3\relax}
\newlength{\bcInput}\setlength{\bcInput}{\dimexpr(\textwidth-5pt)/5\relax}

\newcommand{\bcNamegw}{GaussianWrapping}
\newcommand{\bcNamegenrecon}{GenRecon}

\newcommand{\bcInputs}[1]{%
  \begin{tabular}{@{}ccccc@{}}
    \includegraphics[width=\bcInput]{imgs/baseline_comparison/#1_input1.jpg} &
    \includegraphics[width=\bcInput]{imgs/baseline_comparison/#1_input2.jpg} &
    \includegraphics[width=\bcInput]{imgs/baseline_comparison/#1_input3.jpg} &
    \includegraphics[width=\bcInput]{imgs/baseline_comparison/#1_input4.jpg} &
    \includegraphics[width=\bcInput]{imgs/baseline_comparison/#1_input5.jpg}
  \end{tabular}%
}

\newcommand{\bcScene}[3][gw]{%
  \setlength{\tabcolsep}{0.5pt}%
  \renewcommand{\arraystretch}{1.0}%
  \begin{tabular}{@{}ccc@{\hspace{\bcGap}}ccc@{}}
    \multicolumn{6}{c}{\bcInputs{#2}} \\[1pt]
    \multicolumn{6}{c}{\scriptsize #3} \\[2pt]
    \multicolumn{3}{c}{\includegraphics[width=\bcRender]{imgs/baseline_comparison/#2_#1_render.jpg}} &
    \multicolumn{3}{c}{\includegraphics[width=\bcRender]{imgs/baseline_comparison/#2_ours_render.jpg}} \\[1pt]
    \includegraphics[width=\bcCrop]{imgs/baseline_comparison/#2_#1_crop1.jpg} &
    \includegraphics[width=\bcCrop]{imgs/baseline_comparison/#2_#1_crop2.jpg} &
    \includegraphics[width=\bcCrop]{imgs/baseline_comparison/#2_#1_crop3.jpg} &
    \includegraphics[width=\bcCrop]{imgs/baseline_comparison/#2_ours_crop1.jpg} &
    \includegraphics[width=\bcCrop]{imgs/baseline_comparison/#2_ours_crop2.jpg} &
    \includegraphics[width=\bcCrop]{imgs/baseline_comparison/#2_ours_crop3.jpg} \\[1pt]
    \multicolumn{3}{c}{\scriptsize \csname bcName#1\endcsname} &
    \multicolumn{3}{c}{\scriptsize Ours} \\
  \end{tabular}%
}

\begin{figure}[H]
  \centering
  \bcScene{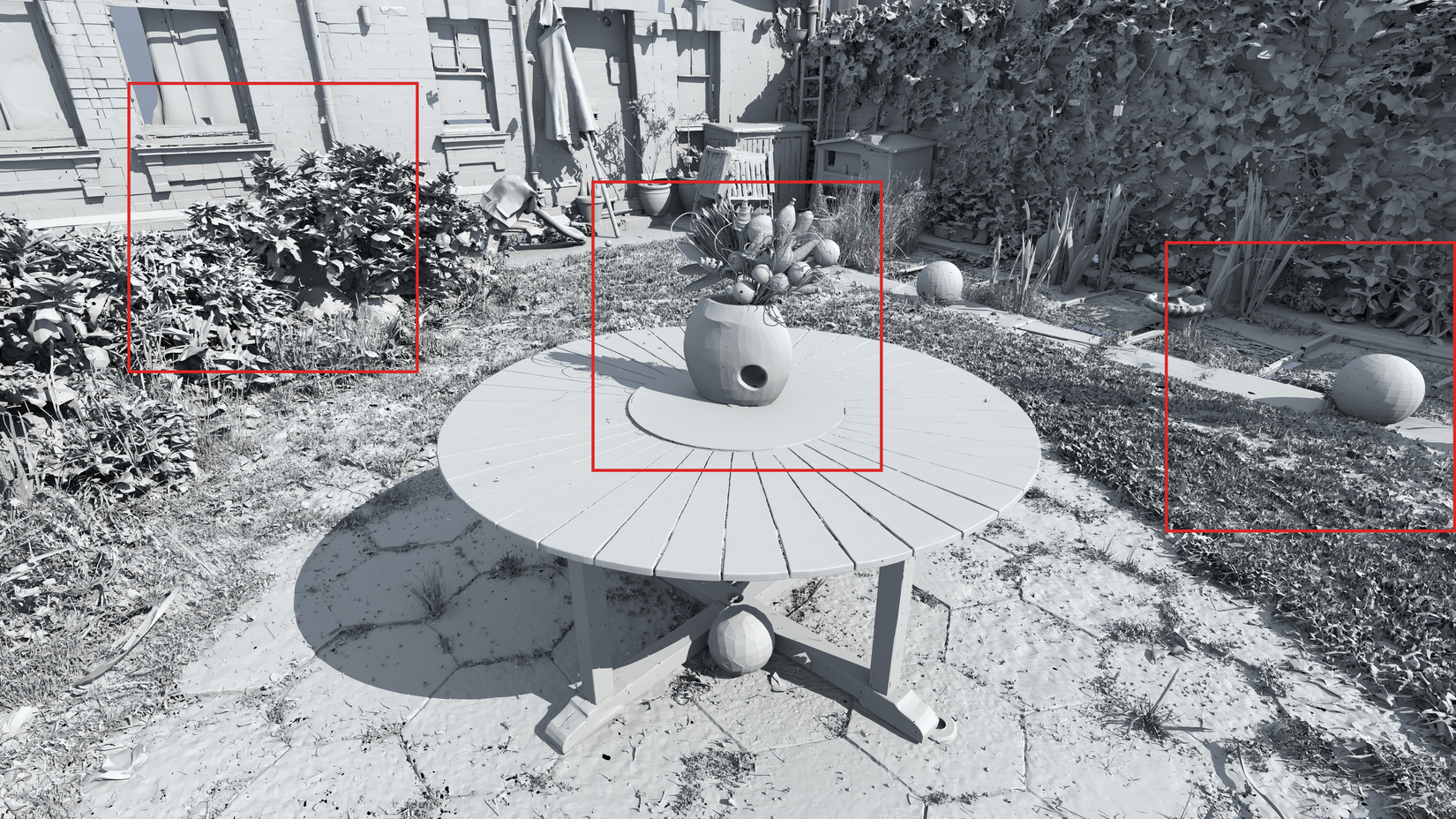}{\texttt{garden} (Mip-NeRF 360, \citealp{barron2022mipnerf360}) -- five of the 185 input views at $5187\times3361$ resolution} \\[8pt]
  \bcScene{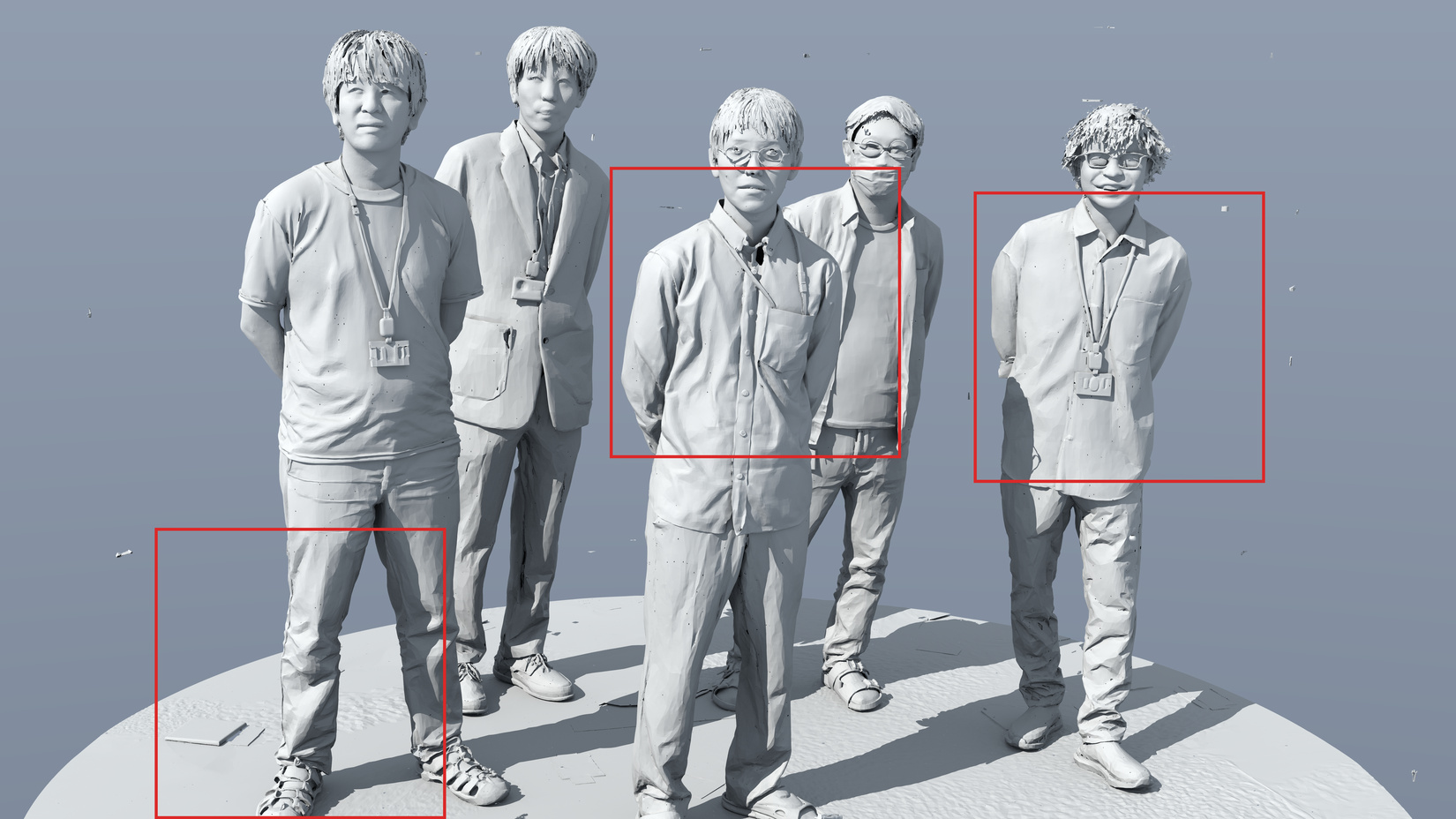}{in-house person capture -- five of the 56 input views at $3840\times2160$ resolution}
  \caption{\textbf{Qualitative comparison with GaussianWrapping on real-world captures.} Both
  meshes are rendered from the same viewpoint, with close-ups of the boxed regions below.}
  \label{fig:baseline-comparison}
\end{figure}

\clearpage
\subsection{Failure case in unbounded scenes}
\label{app:failure-cases}

\begin{figure}[H]
  \centering
  \setlength{\tabcolsep}{0.5pt}
  \renewcommand{\arraystretch}{1.0}
  \begin{tabular}{ccc}
    \multicolumn{3}{c}{\scriptsize (a) \texttt{ignatius}} \\[1.5pt]
    \includegraphics[width=\dimexpr(\textwidth-3pt)*17944/45888\relax]{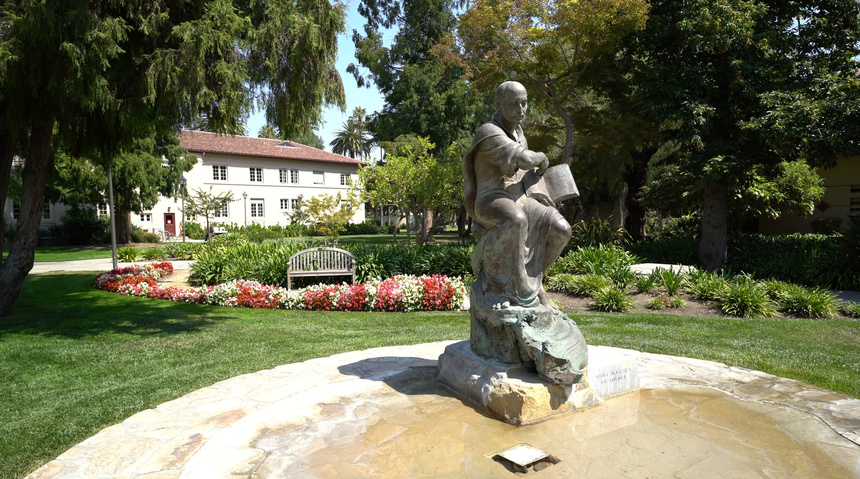} & \includegraphics[width=\dimexpr(\textwidth-3pt)*17944/45888\relax]{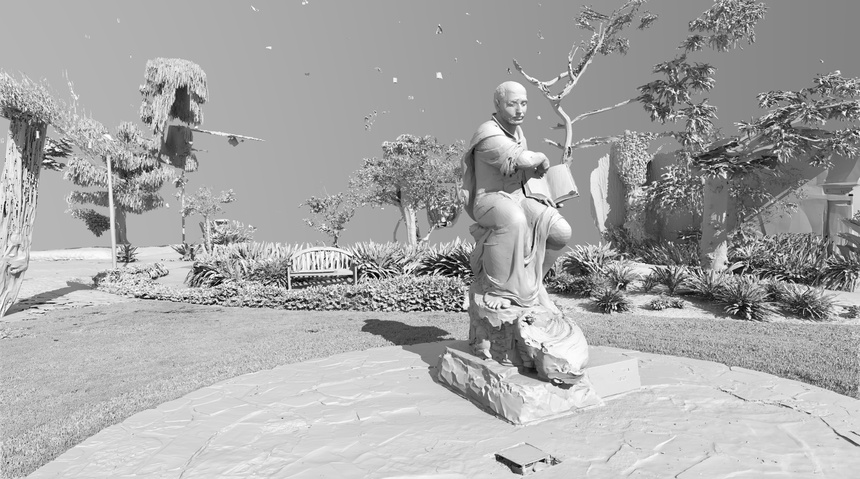} & \includegraphics[width=\dimexpr(\textwidth-3pt)*10000/45888\relax]{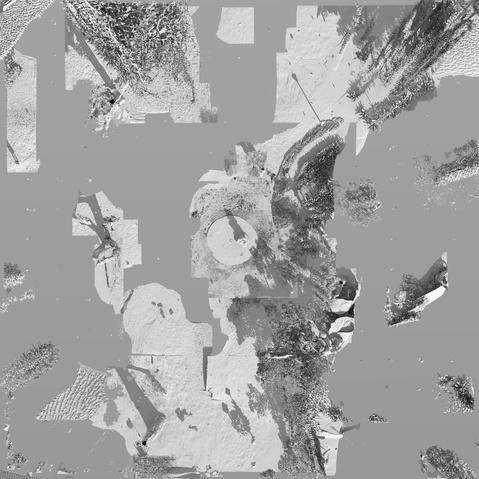} \\[-4.5pt]
    \makebox[\dimexpr(\textwidth-3pt)*17944/45888\relax][c]{\scriptsize Input image} & \makebox[\dimexpr(\textwidth-3pt)*17944/45888\relax][c]{\scriptsize Ours} & \makebox[\dimexpr(\textwidth-3pt)*10000/45888\relax][c]{\scriptsize Ours (top view)} \\[8pt]
    \multicolumn{3}{c}{\scriptsize (b) \texttt{caterpillar}} \\[1.5pt]
    \includegraphics[width=\dimexpr(\textwidth-3pt)*17961/45922\relax]{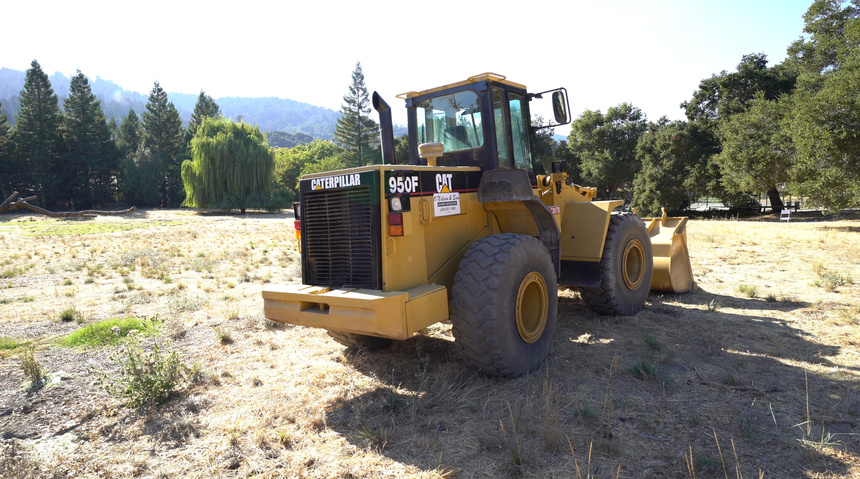} & \includegraphics[width=\dimexpr(\textwidth-3pt)*17961/45922\relax]{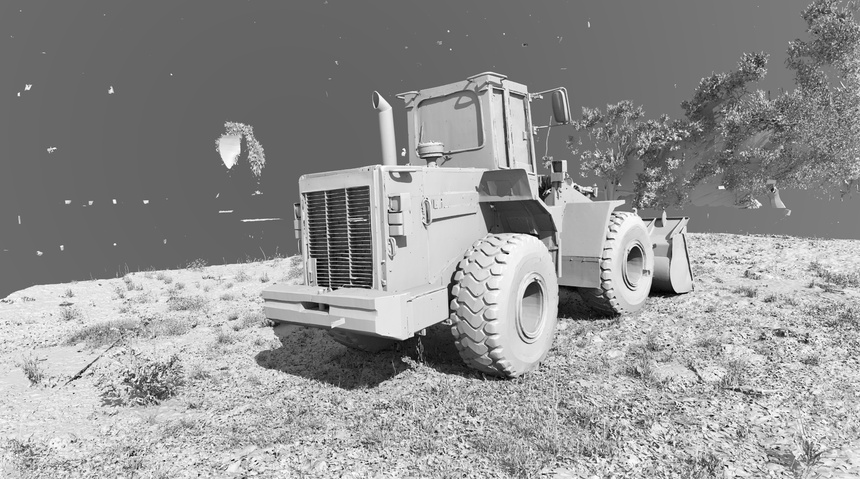} & \includegraphics[width=\dimexpr(\textwidth-3pt)*10000/45922\relax]{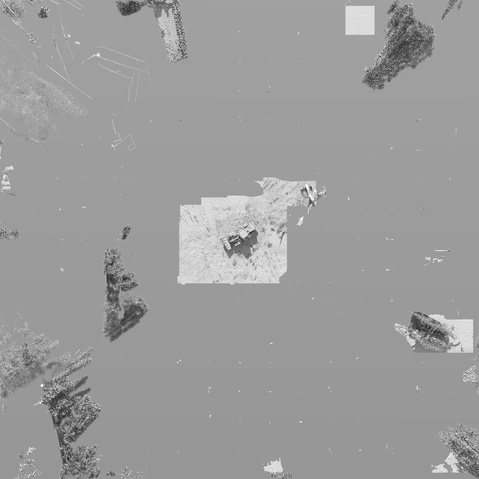} \\[-4.5pt]
    \makebox[\dimexpr(\textwidth-3pt)*17961/45922\relax][c]{\scriptsize Input image} & \makebox[\dimexpr(\textwidth-3pt)*17961/45922\relax][c]{\scriptsize Ours} & \makebox[\dimexpr(\textwidth-3pt)*10000/45922\relax][c]{\scriptsize Ours (top view)} \\
  \end{tabular}
  \caption{\textbf{Failure cases with distant objects.} Input images containing distant objects, and our meshes from 256 input views at 4K, rendered from the same cameras and from the top. The distant objects are not reconstructed well.}
  \label{fig:failure-unbounded}
\end{figure}

Since our method sets the scale and placement of cells freely, it can in principle reconstruct
distant regions by placing large cells far from the cameras.
We reconstruct \texttt{ignatius} and \texttt{caterpillar} from Tanks and Temples with 256
input views using such far and coarse cells. We use four cascade levels and place cells up to
eight times farther from the scene center than the cameras. We render the meshes from an input
camera and from the top in Figure~\ref{fig:failure-unbounded}. Despite these far cells, our method does not work well on distant objects such
as the ground and the trees behind the central object.
This might be caused by the limited diversity of the training data. Moreover, by construction, our method cannot handle objects at infinite distance.

\clearpage

\section{Experimental Setup Details}
\label{app:evaluation}

\subsection{Datasets}
\label{app:datasets}

\paragraph{ScanNet++.}
We evaluate the 50 validation scenes of ScanNet++~\citep{yeshwanth2023scannetpp} and
divide them by the volume of the ground-truth bounding box into a small and a large
split of 25 scenes each. We use the undistorted DSLR images at $1752\times1168$ as inputs and choose the 32- and
256-view sets by farthest-point sampling of the camera centers among the training frames
not flagged as bad. For the scenes
with fewer than 256 frames, 12 in the small split and 9 in the large split, we use all
of them. The ground truth is the mesh reconstructed from the laser scans, with its
anonymized regions removed.

\paragraph{Tanks and Temples.}
We evaluate six scenes of Tanks and Temples~\citep{knapitsch2017tanks}. As for
ScanNet++, the 16-, 32-, and 256-view sets are chosen by farthest-point sampling of
the camera centers. \texttt{truck} has only 251 images and uses all of them in the 256-view
setting. We extract the frames from the 4K videos provided by the dataset and undistort them
to match the official COLMAP reconstruction, which gives images of approximately
$3920\times2180$ (4K). For a fair comparison with the baselines, the main paper uses these
images downsampled by two to approximately $1960\times1090$ (2K), and
Appendix~\ref{app:tnt-results} reports additional results with the 4K images. The ground
truth is the
official laser scan with its crop volume and alignment.

\paragraph{City blocks.}
\label{app:city-dataset}
Our city-block dataset contains 41 blocks with 248 buildings and 139{,}893 posed
images at $9504\times6336$. We split it geographically into 38 training blocks with
220 buildings and 126{,}068 images, and three held-out evaluation blocks with 28
buildings and 13{,}825 images. The blocks range from $42\times22$\,m to
$156\times137$\,m in horizontal extent, and the buildings are $23$ to $172$\,m tall.
All images are captured with a handheld camera at walking height, about $1$ to $2$\,m
above the ground.

We register the images using RealityScan and an in-house SfM pipeline, and align the
reconstruction to a laser scan of the block to obtain metric scale. The exterior BIM
of each building is modeled in the coordinate frame of the same scan. The cameras and
the BIM are therefore aligned without any further registration. We use the mesh
derived from the BIM as the ground truth. It represents building exteriors,
vegetation, and street furniture down to signs and traffic cones, but not people or
vehicles. Each image is accompanied by masks of the sky, people, and vehicles.
We will release the evaluation split, including the images, a 3D Gaussian Splatting
model of each block, and the BIM meshes. For evaluation, each held-out block is
reconstructed from 256 of its images, sampled at a uniform stride over the capture
sequence.

\subsection{Metrics}
\label{app:metrics}

\paragraph{3D metrics.}
Let $P$ and $G$ denote the points sampled from the predicted mesh and the ground
truth, respectively, following each protocol (Appendix~\ref{app:protocols}). The
Chamfer distance (CD) is the average of the mean nearest-neighbor distances from $P$
to $G$ and from $G$ to $P$, in meters and without truncation. Precision and recall at
a threshold $\tau$ are the fractions of points in $P$ and in $G$ whose nearest neighbor
in the other set lies within $\tau$, and the F-score F@$\tau$ is their harmonic mean, with
$\tau$ in centimeters. Normal consistency (NC) is the mean absolute cosine similarity
between the normal of each point in $P$ and $G$ and that of its nearest neighbor in
the other set, with pairs more than $20$\,cm apart counted as zero.

\paragraph{2D metrics.}
We follow the 2D metrics of GenRecon~\citep{schmid2026genrecon}. We render depth and normal maps of the predicted and ground-truth meshes and compute
the errors on the pixels covered by both. We report the mean absolute error (MAE) and
root mean squared error (RMSE) of depth in meters, the absolute relative error
(AbsRel) $\operatorname{mean}|\hat z-z|/z$, and the squared relative error (SqRel)
$\operatorname{mean}(\hat z-z)^2/z$, where $\hat z$ and $z$ are the predicted and
ground-truth depths. Ang is the mean angular error in degrees between the rendered
normals, which are interpolated from area-weighted vertex normals. Completeness (Comp)
is the fraction of pixels covered by the ground truth that are also covered by the
prediction. For depth, the main paper reports MAE and AbsRel, and
Table~\ref{tab:app-scannetpp-2d} additionally reports RMSE and SqRel.

\subsection{Evaluation Protocols}
\label{app:protocols}

\paragraph{ScanNet++.}
Although ScanNet++ registers the camera poses to the ground-truth mesh, the
registration leaves residual errors on some scenes that are not negligible at the F@2
threshold. We therefore rigidly align the ground-truth mesh to the COLMAP sparse
points of each scene with ICP before evaluation. This alignment is shared by all
methods.

We compute the 2D metrics following GenRecon~\citep{schmid2026genrecon}, rendering
both meshes from the cameras of the official test split, invalidating masked pixels,
and pooling the errors over all pixels of all frames. As its evaluation code is not
released, we reimplement it based on the paper.

For the 3D metrics, the main paper uses the virtual-scan protocol of
GaussianWrapping~\citep{gomez2026gaussianwrapping}, which scores only the surfaces
that the cameras see and the ground truth contains. Unlike the original protocol,
which uses all cameras of the capture, we use up to 300 held-out cameras chosen by
farthest-point sampling among the frames that are neither input views of the run nor
flagged as bad. We cast only the rays that hit the ground truth against the prediction, so a missing surface lowers recall and a surface outside the ground truth
is never scored.

Table~\ref{tab:app-scannetpp-3d} also reports the 3D evaluation protocol of GenRecon,
which samples points uniformly over the complete meshes after discarding predicted
triangles farther than $15$\,cm from the ground truth. We run its released code with
four million points per mesh instead of 200{,}000 to reduce sampling noise. Since the
ground truth misses windows, mirrors, ceilings, and surfaces behind furniture, this
protocol cannot distinguish a hole in the ground truth from a hole in the prediction.

\paragraph{Tanks and Temples.}
We use the virtual-scan protocol of GaussianWrapping. The protocol takes the points
seen from all cameras of the official trajectory, crops them to the official volume,
and aligns them to the ground truth by the released transform refined with ICP.
F@$\tau$ uses the official per-scene threshold $\tau$ of $3$\,mm for \texttt{ignatius}, $5$\,mm
for \texttt{caterpillar} and \texttt{truck}, $10$\,mm for \texttt{barn} and \texttt{meetingroom}, and $25$\,mm for
\texttt{courthouse}. The main paper reports CD divided by $\tau$ to average over scenes, and
Table~\ref{tab:tnt-per-scene} reports the per-scene CD in meters. On both ScanNet++
and Tanks and Temples, all methods are scored by the same implementation. Since our
meshes can reach hundreds of millions of triangles, we implement a custom rendering and
evaluation script that handles such meshes without decimation.

\paragraph{City blocks.}
As the BIM ground truth covers complete building exteriors, we compare the complete
meshes rather than only the surfaces seen by the cameras. We crop only the
reconstructions to the ground-truth footprint, the ground-plane projection of the
ground-truth mesh, to remove their surroundings outside the block. Since NC counts
only pairs within $20$\,cm, on this dataset it mainly reflects how much of the surface
is recovered.

\subsection{Baselines}
\label{app:baselines}

All baselines receive the same input views, camera poses, and intrinsics, and are
evaluated on the meshes they produce. We use the official implementations and
checkpoints. On ScanNet++ and Tanks and Temples, our method and the per-scene
optimization methods take the input images at full resolution on ScanNet++ and at 2K on
Tanks and Temples. The feed-forward
methods and GenRecon resize them to the resolution expected by each model. The rows
marked 4K in Table~\ref{tab:tnt-per-scene} instead give our method the
higher-resolution images of Tanks and Temples. On city blocks, our method takes the
original $9504\times6336$ images, whereas the baselines take undistorted images
downsampled to $1920\times1280$. All methods are also given the extent of the block
from the ground truth.

\textbf{2DGS}~\citep{huang20242dgs} uses a depth ratio of 1 and a distortion weight of
10 on ScanNet++ and city blocks, and the released settings on Tanks and Temples, with
the masks applied to the photometric loss on city blocks.
\textbf{MILo}~\citep{guedon2025milo} uses the released fast configuration, with its
preset for indoor scenes on ScanNet++ and the released evaluation settings on Tanks
and Temples. \textbf{GaussianWrapping}~\citep{gomez2026gaussianwrapping} uses the
released training and mesh extraction settings.
\textbf{CityGaussianV2}~\citep{liu2025citygaussianv2} uses the released street-scene
configuration, which trains a coarse model and fine-tunes $2\times2$ partitions before
merging them, with depth supervision from Depth Anything
V2~\citep{yang2024depthanythingv2} and the masks.

\textbf{DA3}~\citep{lin2025da3} uses DA3-GIANT-1.1 conditioned on the input poses and
intrinsics and processes all views in a single forward pass, so that one scale
converts the depth to meters. On ScanNet++, each image is split into four virtual
cameras with narrower fields of view.
\textbf{MapAnything}~\citep{keetha2026mapanything} uses the released model conditioned
on the input poses and intrinsics. \textbf{Murre}~\citep{guo2025murre} uses the
released settings, with sparse depth from the COLMAP points, or from the RealityScan
export on city blocks.

\textbf{GenRecon}~\citep{schmid2026genrecon} partitions the scene into overlapping
chunks placed from the COLMAP sparse points, generates each chunk with a
TRELLIS.2-based generative model conditioned on the input images, and aggregates the
overlapping latents into a single mesh. We run its released code in the ScanNet++ mode
with the default settings, conditioning on all input views.

\textbf{Mesh extraction.} MILo, GaussianWrapping, and GenRecon output meshes directly.
The other methods produce depth maps, which we fuse by TSDF fusion in
Open3D~\citep{zhou2018open3d} with the ground-truth poses, keeping the 50 largest
connected components. The voxel size, truncation, and maximum depth are $1$\,cm,
$4$\,cm, and $4.5$\,m on ScanNet++, following GenRecon, and $5$\,cm, $8$\,cm, and
$200$\,m on city blocks. On Tanks and Temples, we use the original per-scene settings
of 2DGS. For 2DGS on ScanNet++, we use the voxel size of $0.6$\,cm and
truncation of $2.4$\,cm that its authors use for large Tanks and Temples scenes. For CityGaussianV2, pixels with opacity below $0.5$ are discarded before
fusion. On city blocks, we mask the depth and crop the volume to the
ground-truth extent enlarged by $10$\,m. Since Murre aligns each view to its own
sparse depth, overlapping views are not made consistent, and on city blocks its fused
mesh covers only a small part of the ground truth (Table~\ref{tab:bim-per-scene}).

\section{Additional Method Details}
\label{app:method}

The full inference procedure for one scene is as follows. The subsections below describe the
generator (Appendix~\ref{app:backbone}), its training (Appendix~\ref{app:multiscale-training}),
the scene-wide blending (Appendix~\ref{app:multidiffusion}), the transfer between levels
(Appendix~\ref{app:cascade-transfer}), and the cell sizes and extents
(Appendix~\ref{app:cell-extents}).
\begin{enumerate}
  \item For each level $l=0,\dots,L-1$, with cell edge length $s_l=s_0/2^l$:
  \begin{enumerate}
    \item Place the cells. At level 0, they cover the scene extent; at finer
      levels, they cover the surface predicted by the parent
      (Appendices~\ref{app:cell-extents} and~\ref{app:cascade-transfer}).
    \item Select the conditioning views and tiles for each cell
      (Appendix~\ref{app:da3-selection}).
    \item Encode the tiles selected at this level into the tile bank, encoding each tile
      only once even if selected by multiple cells (Appendix~\ref{app:tile-bank}).
    \item Run the SS stage on the scene-wide SS latent grid with MultiDiffusion, starting
      from noise at level 0 and from the noised parent latent at finer levels, and decode
      the predicted occupancy $\widehat O^{(l)}$.
    \item Run the mesh stage on the active voxels of $\widehat O^{(l)}$ with MultiDiffusion.
    \item Unless $l$ is the finest level, decode the shape latent up to resolution $2R$
      and encode the blocks of the child cells as their inherited SS latents.
  \end{enumerate}
  \item Decode the shape latent of the finest level, and optionally of other levels
    (Appendix~\ref{app:ring-composition}), into a mesh with margin-based chunk decoding
    (Appendix~\ref{app:halo-stitching}).
\end{enumerate}

\subsection{Two-stage 3D flow backbone}
\label{app:backbone}

Our generator follows the two-stage design of TRELLIS.2~\citep{xiang2025trellis2}.
It operates on one cell in canonical cell coordinates
(Appendix~\ref{app:multiscale-training}) and is conditioned on the posed images
$\mathcal I$ that observe the cell (Section~\ref{sec:method:singlelevel}). The SS stage
generates the occupancy $O\in\{0,1\}^{R^3}$ of the cell with $R=64$, where active
voxels ($O=1$) contain surface geometry. The mesh stage then generates a shape latent on the active voxels, which the shape VAE
decoder converts into a triangle mesh at $512^3$ resolution. Both stages are latent flow-matching models. The SS latent of an
occupancy is $x_0=A_{\mathrm{SS}}(E_{\mathrm{SS}}(O))$, where $E_{\mathrm{SS}}$ is the SS
VAE encoder and $A_{\mathrm{SS}}$ normalizes latents using the mean and standard deviation
of the training latents. The shape latent is obtained from the ground-truth mesh by the
shape VAE encoder. We keep both pretrained VAEs frozen (Appendix~\ref{app:training-initialization}) and train
only the two flow models.

\textbf{Training.}
For either stage $q\in\{\mathrm{SS},\mathrm{mesh}\}$, a clean latent $x_0$ is moved to
time $t\in[0,1]$ along the linear path
\begin{equation}
  x_t = (1-t)\,x_0 + \sigma_t\,\epsilon,
  \qquad \sigma_t=\sigma_{\min}+(1-\sigma_{\min})\,t,
  \qquad \epsilon\sim\mathcal N(0,I),
  \label{eq:app-flow-path}
\end{equation}
interpolating between data at $t=0$ and noise at $t=1$, with $\sigma_{\min}=10^{-5}$
(Appendix~\ref{app:training-optimization}). The velocity network $v_{\theta_q}$ is trained
to predict the time derivative of this path with the conditional flow-matching objective
\begin{equation}
  \mathcal L_{\mathrm{FM}}(\theta_q)
  = \mathbb E_{x_0,t,\epsilon}
  \left[
    \left\|v_{\theta_q}(x_t,t;\mathcal I)
    -\left((1-\sigma_{\min})\,\epsilon-x_0\right)\right\|_2^2
  \right].
  \label{eq:app-flow-loss}
\end{equation}
The mesh flow is additionally conditioned on the occupancy $O$. The two flows are trained
separately. Appendix~\ref{app:multiscale-training} describes how cells of different sizes
are sampled for this objective, and Appendix~\ref{app:conditioning} describes how
$\mathcal I$ enters the network.

\textbf{Inference.}
Each stage starts from Gaussian noise at $t=1$ and integrates
$\mathrm dx_t/\mathrm dt=v_{\theta_q}(x_t,t;\mathcal I)$ to $t=0$ with Euler steps and
classifier-free guidance. The SS stage applies the inverse of $A_{\mathrm{SS}}$ to the
resulting latent and decodes it with the SS VAE decoder into the predicted occupancy
$\widehat O$. The mesh stage samples a shape latent on the active voxels of $\widehat O$
and decodes it into a mesh. Every cell at every cascade level goes through this sequence
with the same SS flow, mesh flow, VAEs, and image conditioner. In the full pipeline, overlapping cells share each Euler step through scene-wide
MultiDiffusion (Appendix~\ref{app:multidiffusion}). At finer levels, the SS stage starts
from the noised parent latent at $t_s$ rather than $t=1$ (Appendix~\ref{app:cascade-transfer}).

\subsection{Multi-scale training with shared weights}
\label{app:multiscale-training}

For a cubic cell with center $o_c$ and edge length $s$, we map a world-space point $p$ to
canonical cell coordinates $(p-o_c)/s$ and transform the ground-truth geometry and camera
poses in the same way. We therefore encode every cell at the same SS and shape resolutions
regardless of its size, and $s$ changes only the world-space voxel spacing
$\Delta=s/R$. Small cells represent fine geometry, whereas large cells cover more context
with the same latent budget.

The general-scene model is trained with cell edge lengths from two ladders:
$\{1,2,4\}$\,m and $\{1.5,3,6\}$\,m (Appendix~\ref{app:training-augmentation}).
To expose the model evenly to different scene scales, we sample the cell size explicitly:
for each training example, we first choose a training data source with the sampling share of
Appendix~\ref{app:general-training}, then one of its edge lengths with equal probability,
and finally a cell of that size from one of its scenes.
This avoids biasing training toward smaller cells, which are more numerous under spatial
sampling. Each stage uses a single set of weights shared across all cell sizes.
Each training example consists of only one cell; we do not train on parent--child pairs
or multi-level trajectories.

\subsection{Scene-wide MultiDiffusion}
\label{app:multidiffusion}

At each cascade level, we maintain a single scene-wide latent grid rather than separate
latents per cell. We start sampling from Gaussian noise on this grid and update the grid in place at every
flow step; at finer levels, we start the SS grid from the noised SS latent encoded from the
parent's subdivision output (Appendix~\ref{app:cascade-transfer}). We place cells on a
lattice with a stride of half the cell edge length, giving a 50\% overlap between neighbors.
At every flow step, each cell predicts the latent inside
its window from the current grid, and we update the grid at a voxel $p$ with the weighted average of the predictions of all
cells $c$ that contain $p$,
\begin{equation}
  \widehat z(p)=
  \frac{\sum_{c:p\in c}w_c(p)\,\widehat z_c(p)}
       {\sum_{c:p\in c}w_c(p)},
  \label{eq:app-multidiffusion}
\end{equation}
where $\widehat z_c(p)$ is the prediction of cell $c$ at $p$ and $w_c(p)$ is its blending
weight. We use a separable Hann window,
\begin{equation}
  w_c(p)=\prod_{d\in\{x,y,z\}}\sin^2\!\left(\frac{\pi(j_d+1/2)}{R_z}\right),
  \label{eq:app-hann}
\end{equation}
where $R_z$ is the latent resolution of the flow ($16$ for the SS flow and $32$ for the mesh
flow) and $j_d\in\{0,\dots,R_z-1\}$ is the voxel index of $p$ along axis $d$ within cell $c$.
The weight peaks at the cell center and approaches zero at the cell boundary, reducing
the influence of predictions near the boundary. With a half-cell stride, the weights of overlapping cells sum to one
along each axis. The SS latent grid is dense, whereas the shape latent grid is sparse and
holds only the active voxels; we apply the same averaging to both. We apply classifier-free guidance within each cell
before blending: each cell combines its conditional and unconditional velocities and takes
an Euler step, and we blend the updated latents of the cells with
Equation~\ref{eq:app-multidiffusion}. Since all cells start from the same
grid, this is equivalent to blending the guided velocities.

We use the overlapping half-stride cells only for blending. For mesh decoding, we use a
non-overlapping base lattice to decode each region once without duplicated surfaces
(Appendix~\ref{app:halo-stitching}). Cell ordering, random seeds, and view
selection are deterministic within a run.

\subsection{Parent-to-child geometry transfer}
\label{app:cascade-transfer}

The shape VAE decoder upsamples a shape latent through subdivision stages that
each double the resolution. For the parent level, we run the decoder only up to
resolution $2R$. Because a child cell has half the edge length of its parent, it
spans exactly an $R^3$ block of this grid, which we crop without spatial resampling.
We encode the block with the SS VAE encoder to obtain the inherited SS latent
$\bar x_0^{(l+1)}$ of Section~\ref{sec:method:cascade}. For a general power-of-two ratio $r$
between the parent and child edge lengths, we stop at the stage of resolution $rR$
instead. Because the child inherits its SS latent directly from the parent's subdivision output
rather than from the parent mesh, we extract intermediate meshes only when needed (e.g.,
when combining multiple levels, Appendix~\ref{app:ring-composition}) and skip them otherwise.

Because the SS flow starts from $t_s$ rather than $t=1$, it requires fewer Euler steps.

The parent also determines child cell placement: we keep a child cell only if it
intersects the parent's predicted occupancy $\widehat O^{(l)}$, dilated by one parent
SS voxel so that regions narrowly missed at the coarse level can still be recovered.
Without this guidance, the location of the surface would be unknown, and fine cells would
have to fill the entire scene volume at a much higher cost. The DA3 points used to place
the coarsest cells cannot take this role at finer levels: DA3 predicts points only for regions
observed in the images, so in wide scenes with sparse views it misses surfaces that the
coarse level can still generate. Parent-guided placement is used even at $t_s=1$, where noise
replaces the inherited latent.

\subsection{Cell sizes and extents}
\label{app:cell-extents}

ScanNet++ and the city blocks use metric coordinates: in both, the camera poses are
aligned to laser scans (Appendix~\ref{app:datasets}).
On these datasets, cell edge lengths are specified in meters: 4, 2, and 1\,m on ScanNet++,
and 80, 40, 20, and 10\,m on city blocks. The poses for Tanks and Temples and the three
real-world captures (\texttt{garden} from Mip-NeRF~360, \texttt{coffee\_martini} from N3DV~\citep{li2022n3dv},
and our in-house human capture) come from SfM and have no metric scale. We rescale camera
translations so that the median distance from the cameras to their centroid equals a target
value: 6 or 8 on Tanks and Temples, 1 on \texttt{coffee\_martini}, and 4 on \texttt{garden}
and the human capture. Cell sizes are defined in these normalized units: 4, 2, and 1 units
on Tanks and Temples, \texttt{garden}, and \texttt{coffee\_martini}, and 2.8, 1.4, and 0.7
units on the human capture.

The coarsest cells cover a horizontal extent and a height range. The height range is
computed from DA3 points for each scene (Appendix~\ref{app:da3-selection}) and is capped
at 8 and 4 units on \texttt{garden} and the human capture, respectively. The horizontal extent is
also computed from DA3 points on ScanNet++, \texttt{meetingroom} (Tanks and Temples), and
\texttt{coffee\_martini}.
City blocks use the footprint of the ground-truth mesh, while the remaining Tanks and Temples
scenes and the human capture use the convex hull of the camera trajectory. On
\texttt{garden}, cascade levels cover nested squares with half-widths of 16, 8, and 4
units around the camera centroid, and their meshes are combined as described in
Appendix~\ref{app:ring-composition}.

\section{Image Conditioning and Scene Preparation}
\label{app:conditioning}

\subsection{Voxel-queried image attention}
\label{app:relative-attention}

This section details the voxel-queried image attention introduced in
Section~\ref{sec:method:attention}. It replaces the image cross-attention in every
transformer block of both flows.
In the mesh flow, each voxel attends to two types of image features: DINOv3 tokens
around its projection in each view, and ConvNeXt features sampled at the projection.
Below, we first contrast this design with pixel-aligned conditioning, and then
describe the DINO branch (projection, geometric encoding, and attention), the ConvNeXt
branch, and the variant used in the SS flow.

\paragraph{Difference from pixel-aligned conditioning.}
Pixel-aligned generators sample one feature per view at the voxel projection and
combine views before the 3D network. This design has two limitations in our setting.

The first is resolution. A DINO token covers a $16\times16$-pixel patch; thus,
features sampled from the token grid, even with bilinear interpolation, lack direct
information finer than 16 pixels, which is insufficient for reconstructing fine mesh
details. Pixal3D~\citep{li2026pixal3d} upsamples DINO features to image resolution
before sampling. For a $512\times512$ tile, the upsampled 1,024-channel BF16 feature
map requires
\begin{equation}
  512^2\cdot1024\cdot2\ \text{bytes}=0.5\ \text{GiB per view},
  \label{eq:app-dino-map-memory}
\end{equation}
and 64 views require 32~GiB. MultiDiffusion evaluates every cell at every flow step, so
features exceeding GPU memory must be recomputed at every evaluation.

The second is view aggregation. Pixal3D averages sampled features over views, and
GenRecon~\citep{schmid2026genrecon} replaces the average with a learned aggregator
conditioned solely on image features. In both approaches, the view weights are
independent of the voxel feature.

Our attention addresses both limitations. For resolution, the attention score receives
the offset from the voxel projection to each token center, preserving the sub-token
position without upsampling. Shallow ConvNeXt feature maps provide image detail finer
than 16 pixels at about 1\% of the memory of upsampled DINO features. Both features are small enough to keep in memory for all views, so we compute them once and
reuse them across cells and flow steps. For aggregation, we keep the tokens of all views
separate until the attention step and use the current voxel feature as the query, so that
the view weights depend on the voxel feature. The attention score also incorporates the apparent
voxel size and the viewing ray.

\paragraph{Projection and token gathering.}
Let $p\in[-\tfrac12,\tfrac12]^3$ be a voxel center in cell coordinates, and let
$h_p\in\mathbb R^C$ be its feature after pre-attention layer normalization.
View $i$ has rotation $R_i$, translation $t_i$, and intrinsics $K_i$ in
cell coordinates, where $K_i$ includes the scale and crop origin of the tile
(Appendix~\ref{app:tile-bank}). The voxel projects to
\begin{equation}
  (x_i,y_i,z_i)^\top = R_i p+t_i,
  \qquad
  (u_i,v_i)^\top=\pi_i(p)=
  \begin{pmatrix}
    f_{x,i}x_i/z_i+c_{x,i}\\
    f_{y,i}y_i/z_i+c_{y,i}
  \end{pmatrix}
  \label{eq:app-voxel-projection}
\end{equation}
in pixel coordinates of the $512\times512$ tile. A view is invalid for $p$ if $z_i\leq0$
or $\pi_i(p)$ lies outside the tile. In each valid view, we gather the
$w\times w$ DINOv3 tokens $\mathcal G_i(p)$ whose centers $(u_{ir},v_{ir})$ are nearest to
$\pi_i(p)$. We use
$w=2$ (Appendix~\ref{app:window-ablation}), which selects the four tokens around the
projection. Each of the $H=12$ heads computes
\begin{equation}
  q_p=\operatorname{RMSNorm}(W_qh_p),
  \qquad
  k_{ir}=\operatorname{RMSNorm}(W_kf_{ir}),
  \qquad
  v_{ir}=W_vf_{ir},
  \label{eq:app-qkv}
\end{equation}
with head dimension $d=C/H=128$, where $f_{ir}\in\mathbb R^{1024}$ is the DINOv3 feature of
token $r$ in view $i$.

\paragraph{Relative geometric encoding.}
We aim to run a single model across arbitrary cell layouts, where cells of different
sizes can be placed anywhere, cameras can be at very different distances and directions, and output resolution is
chosen at inference. The attention should therefore receive identical geometric inputs
whenever a cell appears the same in the images, regardless of where it is placed or how
large it is. Standard camera encodings, such as camera-origin coordinates or
Pl\"ucker coordinates~\citep{sitzmann2021lfn,zhang2024raydiffusion}, are measured from the origin of a coordinate frame: their values grow when a camera is far
from the origin or looks at a region far from it, so two neighboring cells that see the
same image region receive different encodings. We therefore encode only relative
quantities. The attention score receives three such inputs:
(i) the viewing ray direction of each token in the cell frame,
(ii) the pixel offset from the voxel projection to each token center, and
(iii) the apparent voxel size in the image, in pixels.
Because these represent local directions or pixel lengths, they depend only on how the
cell appears in the images, without referencing absolute cell positions, cell indices,
cascade levels, or output resolution. Changing the cell layout or enlarging the scene therefore changes none of the inputs
beyond what the images themselves show, allowing the same attention mechanism to
operate across all scales.

Each quantity is encoded with RoPE~\citep{su2024roformer} on its own subspace of the
head. The head is split into subspaces for the ray direction, horizontal offset, vertical
offset, and scale, plus a remainder without encoding. Their dimensions are
\begin{equation}
  d_{\mathrm{ray}}=3\left\lfloor\frac{d}{6}\right\rfloor,
  \qquad
  d_{xy}=2\left\lfloor\frac{d-d_{\mathrm{ray}}}{6}\right\rfloor,
  \qquad
  d_{\rho}=2\left\lfloor
    \frac{d-d_{\mathrm{ray}}-2d_{xy}}{2}
  \right\rfloor,
  \label{eq:app-geometry-dims}
\end{equation}
which for $d=128$ gives 63, 20, 20, 24, and 1. A subspace of dimension $d_b$ uses
the RoPE frequencies $\omega_m=10{,}000^{-m/(d_b/2)}$ for $m=0,\ldots,d_b/2-1$.

\textit{(i) Ray direction.} In cell coordinates, the viewing ray through the center of
token $r$ is
\begin{equation}
  d_{ir}=\operatorname{normalize}\!\left(
    R_i^{\top}K_i^{-1}
    [u_{ir},v_{ir},1]^\top
  \right).
  \label{eq:app-token-ray}
\end{equation}
As in PRoPE~\citep{li2025prope}, which encodes cameras as relative positional
encodings in multi-view attention, camera geometry enters the score as a rotation
of the key. Let $Q_{ir}$ be the rotation that takes the canonical optical axis to
$d_{ir}$, applied to each of the 21 three-dimensional blocks of the ray subspace of
$k_{ir}$. The voxel query has no camera and is not rotated. The score therefore
depends on the direction of the token ray in cell coordinates.

\textit{(ii) Image offset.} The query is assigned the voxel projection $(u_i,v_i)$ as its
position, and the key is assigned the token center $(u_{ir},v_{ir})$. With RoPE on
each image axis, the score depends only on the offset
\begin{equation}
  \delta u_{ir}=u_i-u_{ir},
  \qquad
  \delta v_{ir}=v_i-v_{ir}
  \label{eq:app-relative-projection}
\end{equation}
in pixels. The attention score thus captures where the voxel projects relative to each
token center at sub-token precision. With $w=2$, the four gathered tokens correspond to
those used in bilinear interpolation. Whereas bilinear sampling applies fixed weights
based purely on offset, our attention weights depend jointly on offsets and features,
varying across heads, blocks, and flow steps.

\textit{(iii) Projected scale.} Let $s_{\mathrm{vox}}$ be the voxel width in cell
coordinates, $1/32$ in the mesh flow and $1/16$ in the SS flow. The apparent voxel
size in view $i$, as the logarithm of its width in pixels, is
\begin{equation}
  \rho_i(p)=\log\!\left(
    \frac{s_{\mathrm{vox}}\sqrt{f_{x,i}f_{y,i}}}{z_i}
  \right).
  \label{eq:app-projected-scale}
\end{equation}
The query is assigned position $\rho_i(p)$ and the key is assigned zero, making the
score sensitive to apparent voxel scale, which varies with cascade level and camera
distance.

\paragraph{Attention over tokens and views.}
The three encodings rotate the query and the key. Let $\Theta(x)$ denote the RoPE
rotation of a subspace for a scalar position $x$. With the subspaces ordered as ray,
horizontal offset, vertical offset, scale, and remainder, the encoded query and key are
\begin{equation}
  \hat q_{pi}=\operatorname{diag}\!\big(I,\ \Theta(u_i),\ \Theta(v_i),\ \Theta(\rho_i(p)),\ 1\big)\,q_p,
  \qquad
  \hat k_{ir}=\operatorname{diag}\!\big(Q_{ir},\ \Theta(u_{ir}),\ \Theta(v_{ir}),\ I,\ 1\big)\,k_{ir}.
  \label{eq:app-encoded-qk}
\end{equation}
The query depends on the view through the projection and the apparent size, and the
key depends only on the token. The attention weights are a single softmax over the
gathered tokens of all valid views, so the views are weighted against each other rather
than averaged,
\begin{equation}
  a_{pir}=
  \frac{
    \exp\!\left(\hat q_{pi}^\top\hat k_{ir}/\sqrt d\right)
  }{
    \displaystyle
    \sum_{j}\sum_{s\in\mathcal G_j(p)}
    \exp\!\left(\hat q_{pj}^\top\hat k_{js}/\sqrt d\right)
  }.
  \label{eq:app-relative-attention-weights}
\end{equation}
The output of the
head is $o_p=\sum_i\sum_{r\in\mathcal G_i(p)}a_{pir}v_{ir}$.

\paragraph{ConvNeXt detail branch.}
The DINO tokens contain no direct information finer than 16 pixels. The mesh flow
therefore also uses the stride-4, 8, and 16 feature maps of a frozen
ConvNeXt-T~\citep{liu2022convnext}, with 96, 192, and 384 channels. Per view, they take
\begin{equation}
  2\left(
    128^2\cdot96+64^2\cdot192+32^2\cdot384
  \right)\ \text{bytes}
  \simeq 5.25\ \text{MiB},
  \label{eq:app-convnext-map-memory}
\end{equation}
or 0.33~GiB for 64 views, compared with 0.5~GiB per view for the upsampled DINO
features (Equation~\ref{eq:app-dino-map-memory}). The ConvNeXt features of all views
are therefore kept in the feature bank and reused across cells and flow steps.

In this branch, we sample each feature map bilinearly at the
voxel projection. Let $F_i^{(s)}$ be the stride-$s$ feature map of view $i$, and let
$P_s$ be a linear projection to 96 channels shared by all blocks. The feature of view
$i$ is
\begin{equation}
  g_{pi}=
  \operatorname*{Concat}_{s\in\{4,8,16\}}
  P_s\,\operatorname{Bilinear}
  \!\left(F_i^{(s)},\pi_i(p)/s\right)\in\mathbb R^{288}.
  \label{eq:app-convnext-sampling}
\end{equation}
The branch uses four heads of dimension $d_{\mathrm d}=64$,
\begin{equation}
  q_p^{\mathrm d}=\operatorname{RMSNorm}(W_q^{\mathrm d}h_p),
  \qquad
  k_{pi}^{\mathrm d}=\operatorname{RMSNorm}(W_k^{\mathrm d}g_{pi}),
  \qquad
  v_{pi}^{\mathrm d}=W_v^{\mathrm d}g_{pi}.
  \label{eq:app-convnext-qkv}
\end{equation}
Because the feature is sampled at the projection, there is no offset to encode, and
the head has only the ray and scale subspaces. With $Q_{pi}$ the ray rotation for the
ray through $\pi_i(p)$, the encoded query and key are
$\hat q^{\mathrm d}_{pi}=\operatorname{diag}(I,\Theta(\rho_i(p)))\,q^{\mathrm d}_p$ and
$\hat k^{\mathrm d}_{pi}=\operatorname{diag}(Q_{pi},I)\,k^{\mathrm d}_{pi}$. The weights over
views and the output are
\begin{equation}
  b_{pi}=
  \operatorname*{softmax}_{i\in\mathcal V(p)}
  \left(
    \hat q^{\mathrm d\,\top}_{pi}\hat k^{\mathrm d}_{pi}/\sqrt{d_{\mathrm d}}
  \right),
  \qquad
  o_p^{\mathrm d}=\sum_{i\in\mathcal V(p)}b_{pi}v_{pi}^{\mathrm d},
  \label{eq:app-convnext-view-attention}
\end{equation}
where $\mathcal V(p)$ is the set of valid views. The DINO branch thus attends over
tokens and views jointly, and the ConvNeXt branch attends over views only.

Both outputs are added to the voxel feature $x_p$ before the layer normalization,
\begin{equation}
  \widetilde x_p
  = x_p+\gamma_{\mathrm{out}}\odot\left(
    W_o\,o_p+\gamma_{\mathrm d}\odot W_o^{\mathrm d}o_p^{\mathrm d}
  \right),
  \label{eq:app-relative-attention-output}
\end{equation}
where $o_p$ and $o_p^{\mathrm d}$ concatenate the heads, and $\gamma_{\mathrm{out}}$
and $\gamma_{\mathrm d}$ are learned per-channel gates initialized to 1 and 0.1.

\paragraph{Attention in the SS flow.}
We use the attention described above in the mesh flow and a simplified version in the SS
flow.
The SS flow follows the conditioning of Pixal3D~\citep{li2026pixal3d} and adds the
ray and scale encodings to it. Each voxel of the dense $16^3$ latent attends to the
CLS and register tokens of all views.
It also receives the DINO patch features sampled bilinearly at its projection in each
view, at their native $32\times32$ resolution. These sampled features are combined
over views by attention with the voxel feature as the query, using the ray and scale
encodings of (i) and (iii).

\subsection{Multi-scale tile feature bank}
\label{app:tile-bank}

Each cell needs images at a pixel spacing matched to its voxel spacing
(Section~\ref{sec:method:attention}). Cropping and encoding a separate image for
each cell would repeatedly invoke the image encoder for shared image regions and tie
the crops to specific cell layouts. We therefore partition images into tiles that are
defined independently of cells, encode each tile once, and allow cells to retrieve
their required tiles. This section describes the tiles, the feature bank that stores
their features at inference, and the use of the tiles in training.

\paragraph{Tile pyramid.}
For each source image $i$, level $\ell$ of the pyramid is the image resized by
$1/2^\ell$ from the original with Lanczos filtering. The deepest level $L_{\max}$ is the last one whose shorter side is at least 64 pixels. We cover each level with $512\times512$ tiles at a stride of 512 pixels, shifting the last
tile along each axis inward to keep it inside the image. An axis shorter than 512 pixels
gets one centered tile padded with black. A tile
\begin{equation}
  \tau=(i,\ell,a,b)
  \label{eq:app-tile-index}
\end{equation}
is identified by its source image, level, and crop origin $(a,b)$, and does not
depend on any cell. It is treated as a virtual camera with intrinsics
\begin{equation}
  K_\tau=
  \begin{pmatrix}
    f_{x,i}/2^\ell & 0 & c_{x,i}/2^\ell-a\\
    0 & f_{y,i}/2^\ell & c_{y,i}/2^\ell-b\\
    0 & 0 & 1
  \end{pmatrix}
  \label{eq:app-tile-intrinsics}
\end{equation}
and the pose of its source image. For each cell, only this pose is transformed into
the cell coordinates. The attention of Appendix~\ref{app:relative-attention} treats
every selected tile as one view, so a photograph can contribute several tiles to the
same cell, at different crop origins or levels. At inference, a cell selects its
tiles as described in Appendix~\ref{app:da3-selection}.

\paragraph{Feature bank at inference.}
Before sampling a cascade level, we identify the set of tiles required by its cells.
Each photograph is cropped at the required pyramid levels and processed
by the frozen DINOv3 and ConvNeXt encoders. The bank stores, in BF16,
\begin{equation}
  \mathcal B[\tau]
  =\left(
    F_\tau^{\mathrm{DINO}},
    F_{\tau,4}^{\mathrm{ConvNeXt}},
    F_{\tau,8}^{\mathrm{ConvNeXt}},
    F_{\tau,16}^{\mathrm{ConvNeXt}}
  \right),
  \label{eq:app-tile-bank-entry}
\end{equation}
where $F_\tau^{\mathrm{DINO}}$ is the DINO token sequence and the other entries are
the ConvNeXt feature maps at strides 4, 8, and 16. The SS and mesh flows use the same
DINOv3 checkpoint, so DINO runs once per tile for both.
The condition of a cell references the stored features $\mathcal B[\tau]$ of its
selected tiles instead of copying them, so overlapping cells and all flow steps of both
stages share a single encoding per tile. The bank is built per cascade level, as
finer levels instantiate different cells and require different tiles.

\paragraph{Tiles in training.}
Because tiles are defined independently of cells, we share a single set of tiles across all
cell layouts in training. Cells of every physical size and position, as well as
the cell-frame augmentations of Appendix~\ref{app:training-augmentation}, read the same
tiles, so the stored tiles remain unchanged as the number of 3D training variants grows. For each cell, we take the tiles at the level closest to 8 pixels per SS voxel and draw 16
of them at random for each example.

\subsection{DA3-based layout and view selection}
\label{app:da3-selection}

Running the generator on a new capture requires two decisions outside the model:
where to place the coarsest cells, and which tiles each cell reads.
We make both decisions using depth from
Depth Anything 3 (DA3)~\citep{lin2025da3}, run with input poses and intrinsics as in
the DA3 baseline (Appendix~\ref{app:baselines}). We use DA3 solely for these decisions: we do not feed its depth or features to the flow
models, fuse them into the output mesh, or use them to optimize it.

\paragraph{Cell placement.}
At the coarsest level, we discard the 10\% of DA3 depths with lowest confidence,
unproject the remainder, and place cells over the volume that these points occupy.
For some scenes, we additionally use the camera hull or ground-truth
footprint as in Appendix~\ref{app:cell-extents}.
The volume is covered by cells on a half-cell-stride lattice. At
finer levels, candidate cells are derived from the SS occupancy predicted at the parent
level in world coordinates. We dilate this occupancy by one parent SS voxel before
intersecting it with the child lattice, so that a small miss at the coarse level does not remove the region from all finer
levels.

\paragraph{Visibility gate.}
We determine which views observe each voxel center $p$ of the $16^3$ SS latent grid of a
cell. A view is considered to observe $p$ if $p$ projects within the image frame and is
not occluded in the DA3 depth map. The occlusion test tolerates larger depth discrepancies
where DA3 confidence is low. On city blocks, we ignore sky, people, and vehicles in
this test, using masks predicted by an off-the-shelf segmentation model. The generator
still receives the unmasked images.

\paragraph{Pyramid level.}
Let $\nu_{pi}$ be the width of one SS voxel in pixels of the full-resolution image,
\begin{equation}
  \nu_{pi}
  =\frac{f_{x,i}\,(s/64)}{z_i(p)}.
  \label{eq:app-pixels-per-voxel}
\end{equation}
We target $\nu_0=8$ pixels per voxel, so that a $512\times512$ tile spans the
$64^3$ SS grid, and use the same target when selecting tiles in training
(Appendix~\ref{app:tile-bank}). The preferred and available levels are
\begin{equation}
  \ell^*_{pi}=\log_2\!\left(\frac{\nu_{pi}}{\nu_0}\right),
  \qquad
  \ell_{pi}=\operatorname{clip}
  \!\left(\operatorname{round}(\ell^*_{pi}),0,L_{\max}\right).
  \label{eq:app-preferred-pyramid-level}
\end{equation}
Inside the pyramid, level $\ell_{pi}$ shows the voxel at approximately 8 pixels.
The residual $|\ell^*_{pi}-\ell_{pi}|$ measures how far the available level deviates
from this scale.

\paragraph{Tile ranking and selection.}
We select tiles by how well they observe the cell. For every voxel center $p$ and every
view $i$ that observes it, we compute the visibility score
\begin{equation}
  S(p,i)=
  |\cos\theta_{pi}|-|\ell^*_{pi}-\ell_{pi}|,
  \label{eq:app-view-score}
\end{equation}
where $\theta_{pi}$ is the angle between the viewing ray and the surface normal
estimated from the DA3 depth. The first term favors views facing the surface, and
the second penalizes pyramid levels that deviate from the preferred scale. For each
voxel center, we take its 16 highest-scoring views and add each score to the tile
containing the projection of $p$ in that view. The score of a tile is the sum of
these contributions over the cell.

Each cell retains its highest-scoring tiles, up to $\max(64,N)$, where $N$ is the number of input views. Because the network reads at most 64 tiles per cell, we sample 64 of them at random when
more are kept and use this fixed set of tiles for all flow steps.
A cell unobserved by any view receives an empty condition, identical
to the unconditional pass in classifier-free guidance.

\section{Training Details}
\label{app:training}

We train two models sharing the same architecture: a general-scene model evaluated on
ScanNet++, Tanks and Temples, and real-world captures, and a city-scale model specialized
for our city-block dataset. Both models start from the same initialization
(Appendix~\ref{app:training-initialization}) and shared multi-view pretraining on TexVerse
(Appendix~\ref{app:shared-pretraining}), differing only in their final scene-level fine-tuning
data (Appendices~\ref{app:general-training} and~\ref{app:city-training-data}). Both employ the
same inference procedure (Appendix~\ref{app:method}). The remaining subsections detail scale
and rotation augmentations (Appendix~\ref{app:training-augmentation}) and optimization
settings (Appendix~\ref{app:training-optimization}).

\subsection{Model initialization}
\label{app:training-initialization}

Both flows are initialized from the official TRELLIS.2 weights. The SS flow retains the
original 30-block architecture. For the mesh flow, we prune the 30-block backbone down to
15 transformer blocks by removing the blocks that affect its output the least, reducing the denoiser from
approximately 1.3B to 655M parameters while preserving the hidden dimension, attention heads,
and latent resolution of the retained blocks. To recover from pruning, we fine-tune the pruned
mesh flow with the official single-image conditioning for 50k steps on 16 H200 GPUs with a
learning rate of $3\times10^{-5}$. The voxel-queried image attention and the ConvNeXt branch are
subsequently trained during image-conditioned multi-view training.

We freeze the pretrained SS and shape VAEs, which encode the clean SS and shape targets and
decode generated latents.
We update only the two flow networks and their conditioning layers during training.

\subsection{Shared multi-view pretraining}
\label{app:shared-pretraining}

After initialization, we train both flows on multi-view renderings of
TexVerse~\citep{zhang2025texverse}. This stage adapts the single-object generators, alongside
the new conditioning modules, to posed multi-view inputs. The resulting checkpoint serves as
the starting point for scene-level fine-tuning
(Appendices~\ref{app:general-training} and~\ref{app:city-training-data}). Training follows a
curriculum over mesh resolution and the number of randomly sampled conditioning views. The
mesh flow is trained first with 4 views at mesh resolution $256^3$, then with 4 views at
$512^3$, and finally with 16 views at $512^3$. The SS flow, whose occupancy targets have a resolution of
$64^3$, is trained with 4 and subsequently 16 views. Conditioning images are rendered or
cropped to $512\times512$.

\subsection{General-scene fine-tuning}
\label{app:general-training}

Starting from the TexVerse checkpoint (Appendix~\ref{app:shared-pretraining}), we fine-tune
both flows of the general-scene model on multi-scale scene data. In addition to the single-room scene datasets
SAGE-10K~\citep{sage10k} and 3D-FRONT~\citep{fu2021front3d}, we construct two procedural
datasets: connected multi-room houses and outdoor ground with detailed geometry such as
grass.

\begin{figure*}[t]
  \centering
  \begin{minipage}{0.242\textwidth}
    \includegraphics[width=\linewidth]{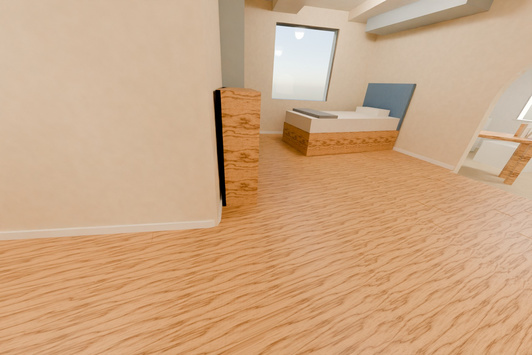}
  \end{minipage}\hfill
  \begin{minipage}{0.242\textwidth}
    \includegraphics[width=\linewidth]{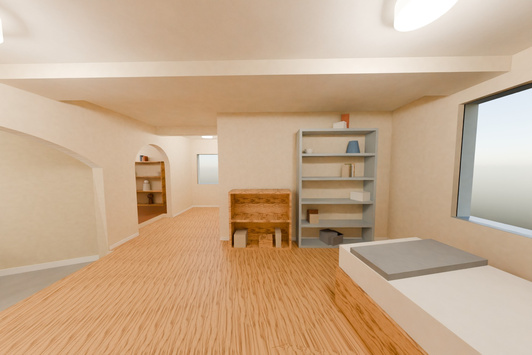}
  \end{minipage}\hfill
  \begin{minipage}{0.242\textwidth}
    \includegraphics[width=\linewidth]{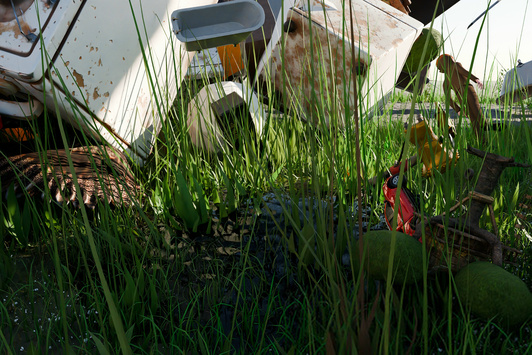}
  \end{minipage}\hfill
  \begin{minipage}{0.242\textwidth}
    \includegraphics[width=\linewidth]{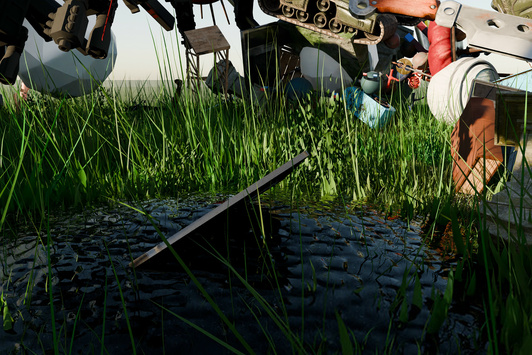}
  \end{minipage}
  \caption{Examples of our procedural scene generation. The first two images
  show a synthetic multi-room house, and the last two show procedural grass
  scenes.}
  \label{fig:app-synthetic-training-data}
\end{figure*}

\paragraph{Synthetic multi-room houses.}
Fine-tuning solely on single-room datasets caused missing walls and ceilings on multi-room
ScanNet++ scenes. We therefore generate 200 synthetic houses
(Figure~\ref{fig:app-synthetic-training-data}). For each house, we generate a floor plan with three to six connected rooms, some of them
non-rectangular, and build walls, floors, ceilings, doors, windows, and furnishings from
it. We allocate cameras to rooms in proportion to floor area and sample their height, pitch,
and roll from the empirical distributions of ScanNet++ DSLR captures.

\paragraph{Procedural grass scenes.}
Existing room datasets contain little outdoor ground cover, causing models trained without it
to over-smooth grass geometry. We generate 100 scenes of 32\,m extent
with procedural vegetation over displaced terrain (Figure~\ref{fig:app-synthetic-training-data}).
We model the vegetation as mesh geometry rather than image textures and vary its height,
density, tilt, and coloration across the scene with spatial noise. We also place TexVerse
objects at random positions with varied scales and orientations in each scene, and capture the scene
with cameras on spiral trajectories.

\paragraph{Rendering.}
Every scene is rendered from 256 views. SAGE-10K, 3D-FRONT, and synthetic houses use
$1752\times1168$ images with horizontal fields of view from $108.7^\circ$ to $124.3^\circ$,
matching ScanNet++ DSLR parameters. Grass scenes use $3840\times2560$ images with a $56.0^\circ$
horizontal field of view, matching Mip-NeRF~360 captures. Renderings are stored as
$512\times512$ tiles for conditioning (Appendix~\ref{app:tile-bank}).

\paragraph{Data mixture.}
The sampling shares among training sources are:
\begin{center}
\begin{tabular}{lcc}
\toprule
Training source & Scenes & Sampling share \\
\midrule
SAGE-10K & 2,000 & 0.40 \\
3D-FRONT & 997 & 0.20 \\
Synthetic houses & 200 & 0.25 \\
Procedural grass & 100 & 0.15 \\
\bottomrule
\end{tabular}
\end{center}
Within each source, the sampling share is divided equally among its cell sizes. Room and
house scenes use the six cell sizes $\{1,1.5,2,3,4,6\}$\,m, and grass scenes use
$\{1,2,4\}$\,m.

\subsection{City-scale fine-tuning}
\label{app:city-training-data}

The city-scale model is fine-tuned from the TexVerse checkpoint
(Appendix~\ref{app:shared-pretraining}) on our city-block dataset
(Appendix~\ref{app:city-dataset}), which pairs ground-level photographs of urban blocks with
mesh targets derived from BIMs. It uses the same representations, voxel-queried image conditioning,
multi-scale cell training, and inference procedure as the general-scene model, differing only
in the fine-tuning data to accommodate urban blocks and building exteriors.

\subsection{Scale and rotation augmentation of cells}
\label{app:training-augmentation}

We augment training cells in both scale and orientation.

\paragraph{Scale.}
We assign each room or house scene one of the two scale ladders $\{1,2,4\}$\,m and
$\{1.5,3,6\}$\,m and partition it into cells of those sizes, which gives six cell sizes
across the dataset. Because a scene yields far more small cells than large ones, we sample
cells with the scale-balanced scheme of Appendix~\ref{app:multiscale-training} to avoid over-representing
small cells.

\paragraph{Rotation.}
We rotate each cell independently of its scale in two ways. First, we rotate the cell about
the vertical axis by a uniformly sampled angle and tilt the vertical axis by up to $10^\circ$.
We sample the tilt angle from a distribution concentrated near zero, which keeps most cells
nearly upright while making the model robust to camera poses that are not perfectly level.
Second, we apply rotations by multiples of $90^\circ$ about the vertical axis alongside
reflection flips, yielding eight discrete orientations per cell. The same transform is
applied to the target geometry and the camera poses, and reflections also flip the images. Because these
eight discrete transforms only permute the voxel grid, we apply them directly to the
dual-grid representation of TRELLIS.2~\citep{xiang2025trellis2} instead of recomputing it
from the mesh. We still re-encode the SS and shape latents of each transformed cell,
because neither VAE encoder is equivariant to these transforms.
Because the tile bank is independent of cells
(Appendix~\ref{app:tile-bank}), these augmentations leave the image processing unchanged:
all cell variants access the same tiles and differ only in their relative camera transforms.

\subsection{Optimization settings}
\label{app:training-optimization}

Both flows are optimized using AdamW with a constant learning rate, weight decay of 0.01,
$\beta=(0.9,0.95)$, and $\epsilon=10^{-8}$, with BF16 mixed precision and a per-GPU batch size
of 8. The learning rate is $10^{-4}$ during the 4-view phases of the mesh flow, $5\times10^{-5}$
in other pretraining phases, and $10^{-5}$ during scene-level fine-tuning. For classifier-free
guidance, the conditioning input is dropped with probability 0.1. We set $\sigma_{\min}=10^{-5}$
in the flow path of Equation~\ref{eq:app-flow-path}. TexVerse pretraining
(Appendix~\ref{app:shared-pretraining}) runs on 16 to 32 NVIDIA H200 GPUs for approximately
80k steps for the SS flow and 200k steps for the mesh flow in total. Scene-level fine-tuning
(Appendices~\ref{app:general-training} and~\ref{app:city-training-data}) runs for 25,000 steps
per flow on 8 H200 GPUs.

\section{Scalable Mesh Decoding}
\label{app:mesh-decoding}

\subsection{Margin-based chunk decoding and stitching}
\label{app:halo-stitching}

We decode each cascade level into a unified scene mesh. Decoding cells independently
would truncate the shape VAE receptive field at cell boundaries, producing visible seams.
Instead, we assemble the shape tokens across all cells into a single sparse tensor for
the level. Because cells overlap by 50\% during MultiDiffusion
(Appendix~\ref{app:multidiffusion}), we select every other cell along each axis to
partition the volume without overlap.

Because decoding the entire level at once exceeds GPU memory, we process it in blocks
of $4^3$ cells. Each block is padded with a 4-voxel latent margin from its neighbors,
so that voxels near the block boundary are decoded with their surroundings. Each block outputs only the mesh
faces belonging to its interior region, preventing duplicate faces between adjacent
blocks. We merge the copies of each boundary vertex produced by adjacent blocks into a single
vertex, identifying them by their integer voxel coordinates. The
blocks thus join into one mesh without duplicate faces or open seams.

\subsection{Combining cascade levels}
\label{app:ring-composition}

In object-centered captures such as Mip-NeRF~360~\citep{barron2022mipnerf360}, cameras
orbit a central subject observed closely from many angles, while the background is seen
only from afar. Because the background appears too small in the images to show fine detail, we do not
reconstruct it at the finest resolution: we run the coarsest level over the full bounding volume and restrict successive refinement levels to
concentric squares centered on the subject.
For the \texttt{garden} scene in
Figure~\ref{fig:teaser}, we employ three cascade levels, with each finer level covering
half the horizontal extent of its predecessor.

Each level yields an independent mesh, which we merge hierarchically. Because finer
meshes provide higher fidelity in the central region, we remove coarse faces whose
centroids fall within the boundary of the next finer level, and combine the remaining
outer ring with the finer mesh. This ring composition is used only for the nonuniform qualitative reconstructions; on ScanNet++ and Tanks and Temples, all levels share identical
spatial extents, and the finest level is evaluated directly.

\section{Additional Ablations}
\label{app:ablations}

\subsection{Parent initialization}
\label{app:cascade-ablation}

Table~\ref{tab:app-cascade-ablation} compares $t_s\in\{0.8,0.9,1\}$ on all 50 ScanNet++ scenes
with 32 and 256 input views, extending Figure~\ref{fig:ablation-parent}(a), which shows
only the 25 large scenes with 32 views.
We change $t_s$ at every finer level. At $t_s=1$, every finer level starts from pure noise,
inherits only the cell placement, and runs the full schedule of 12 flow steps instead of
$K=10$. All other experiments use $t_s=0.8$.

Parent initialization is most effective in wide scenes with sparse views. On the large
scenes with 32 views, $t_s=0.8$ improves every metric over $t_s=1$: it raises F@2 from
$0.6306$ to $0.7155$ and reduces the Chamfer distance by $5.0\times$.

In the other three settings, where the views cover the scene more densely, $t_s=1$ reaches
a higher F@2. Without the parent, the finest level does not inherit the errors of the
coarser level, so it can place the surface more accurately where the views constrain it.
The parent, however, still makes the reconstruction more consistent as a whole. With
$t_s=0.8$, the Chamfer distance drops by $2.9\times$ on the small scenes with 32 views and
by $1.6\times$ on the large scenes with 256 views. NC is higher with $t_s=0.8$ in every setting.

\begin{table}[H]
  \centering
  \caption{\textbf{Effect of parent initialization on ScanNet++.} Every finer cascade level starts from a latent noised to $t_s$, where $t_s=1$ generates from pure noise. Mean over the 25 smallest and the 25 largest scenes, with the best results in bold.}
  \label{tab:app-cascade-ablation}
  \scriptsize
  \setlength{\tabcolsep}{3pt}
  \definecolor{rank1}{HTML}{FFB3B3}
  \definecolor{rank2}{HTML}{FFD9B3}
  \definecolor{rank3}{HTML}{FFFFB3}
  \ifdefined\paperautocellwidth\else\newlength{\paperautocellwidth}\fi
  \ifdefined\paperautoprobe\else\newlength{\paperautoprobe}\fi
  \settowidth{\paperautocellwidth}{\textbf{0.0000}}
  \settowidth{\paperautoprobe}{Small (25)}
  \ifdim\dimexpr\paperautoprobe-10\tabcolsep\relax>6\paperautocellwidth  \setlength{\paperautocellwidth}{\dimexpr(\paperautoprobe-10\tabcolsep)/6\relax}\fi
  \settowidth{\paperautoprobe}{Large (25)}
  \ifdim\dimexpr\paperautoprobe-10\tabcolsep\relax>6\paperautocellwidth  \setlength{\paperautocellwidth}{\dimexpr(\paperautoprobe-10\tabcolsep)/6\relax}\fi
  \settowidth{\paperautoprobe}{2D}
  \ifdim\dimexpr\paperautoprobe-4\tabcolsep\relax>3\paperautocellwidth  \setlength{\paperautocellwidth}{\dimexpr(\paperautoprobe-4\tabcolsep)/3\relax}\fi
  \settowidth{\paperautoprobe}{3D virtual scan}
  \ifdim\dimexpr\paperautoprobe-4\tabcolsep\relax>3\paperautocellwidth  \setlength{\paperautocellwidth}{\dimexpr(\paperautoprobe-4\tabcolsep)/3\relax}\fi
  \settowidth{\paperautoprobe}{MAE$\downarrow$}
  \ifdim\dimexpr\paperautoprobe-0\tabcolsep\relax>1\paperautocellwidth  \setlength{\paperautocellwidth}{\dimexpr(\paperautoprobe-0\tabcolsep)/1\relax}\fi
  \settowidth{\paperautoprobe}{Ang$\downarrow$}
  \ifdim\dimexpr\paperautoprobe-0\tabcolsep\relax>1\paperautocellwidth  \setlength{\paperautocellwidth}{\dimexpr(\paperautoprobe-0\tabcolsep)/1\relax}\fi
  \settowidth{\paperautoprobe}{Comp$\uparrow$}
  \ifdim\dimexpr\paperautoprobe-0\tabcolsep\relax>1\paperautocellwidth  \setlength{\paperautocellwidth}{\dimexpr(\paperautoprobe-0\tabcolsep)/1\relax}\fi
  \settowidth{\paperautoprobe}{CD$\downarrow$}
  \ifdim\dimexpr\paperautoprobe-0\tabcolsep\relax>1\paperautocellwidth  \setlength{\paperautocellwidth}{\dimexpr(\paperautoprobe-0\tabcolsep)/1\relax}\fi
  \settowidth{\paperautoprobe}{F@2$\uparrow$}
  \ifdim\dimexpr\paperautoprobe-0\tabcolsep\relax>1\paperautocellwidth  \setlength{\paperautocellwidth}{\dimexpr(\paperautoprobe-0\tabcolsep)/1\relax}\fi
  \settowidth{\paperautoprobe}{NC$\uparrow$}
  \ifdim\dimexpr\paperautoprobe-0\tabcolsep\relax>1\paperautocellwidth  \setlength{\paperautocellwidth}{\dimexpr(\paperautoprobe-0\tabcolsep)/1\relax}\fi
  \providecommand{\paperautospan}[2]{\makebox[\dimexpr#1\paperautocellwidth+#1\tabcolsep+#1\tabcolsep-\tabcolsep-\tabcolsep\relax][c]{#2}}
  \providecommand{\paperautocell}[1]{\paperautospan{1}{#1}}
  \resizebox{\linewidth}{!}{%
  \begin{tabular}{lcccccccccccc}
    \toprule
     & \multicolumn{6}{c}{\paperautospan{6}{Small (25)}} & \multicolumn{6}{c}{\paperautospan{6}{Large (25)}} \\
    \cmidrule(lr){2-7}\cmidrule(lr){8-13}
     & \multicolumn{3}{c}{\paperautospan{3}{2D}} & \multicolumn{3}{c}{\paperautospan{3}{3D virtual scan}} & \multicolumn{3}{c}{\paperautospan{3}{2D}} & \multicolumn{3}{c}{\paperautospan{3}{3D virtual scan}} \\
    \cmidrule(lr){2-4}\cmidrule(lr){5-7}\cmidrule(lr){8-10}\cmidrule(lr){11-13}
    $t_s$ & \multicolumn{1}{c}{\paperautospan{1}{MAE$\downarrow$}} & \multicolumn{1}{c}{\paperautospan{1}{Ang$\downarrow$}} & \multicolumn{1}{c}{\paperautospan{1}{Comp$\uparrow$}} & \multicolumn{1}{c}{\paperautospan{1}{CD$\downarrow$}} & \multicolumn{1}{c}{\paperautospan{1}{F@2$\uparrow$}} & \multicolumn{1}{c}{\paperautospan{1}{NC$\uparrow$}} & \multicolumn{1}{c}{\paperautospan{1}{MAE$\downarrow$}} & \multicolumn{1}{c}{\paperautospan{1}{Ang$\downarrow$}} & \multicolumn{1}{c}{\paperautospan{1}{Comp$\uparrow$}} & \multicolumn{1}{c}{\paperautospan{1}{CD$\downarrow$}} & \multicolumn{1}{c}{\paperautospan{1}{F@2$\uparrow$}} & \multicolumn{1}{c}{\paperautospan{1}{NC$\uparrow$}} \\
    \midrule
    \multicolumn{13}{l}{\textbf{\textit{32 Views}}} \\
    0.8 & \paperautocell{\textbf{0.0324}} & \paperautocell{\textbf{15.31}} & \paperautocell{\textbf{0.9990}} & \paperautocell{\textbf{0.0190}} & \paperautocell{0.6718} & \paperautocell{\textbf{0.9296}} & \paperautocell{\textbf{0.0565}} & \paperautocell{\textbf{16.33}} & \paperautocell{\textbf{0.9962}} & \paperautocell{\textbf{0.0212}} & \paperautocell{\textbf{0.7155}} & \paperautocell{\textbf{0.9247}} \\
    0.9 & \paperautocell{0.0715} & \paperautocell{16.97} & \paperautocell{0.9909} & \paperautocell{0.0394} & \paperautocell{0.6730} & \paperautocell{0.8945} & \paperautocell{0.1450} & \paperautocell{19.16} & \paperautocell{0.9773} & \paperautocell{0.0621} & \paperautocell{0.6779} & \paperautocell{0.8580} \\
    1.0 & \paperautocell{0.0998} & \paperautocell{18.31} & \paperautocell{0.9823} & \paperautocell{0.0548} & \paperautocell{\textbf{0.7255}} & \paperautocell{0.8646} & \paperautocell{0.2170} & \paperautocell{22.15} & \paperautocell{0.9535} & \paperautocell{0.1051} & \paperautocell{0.6306} & \paperautocell{0.7881} \\
    \midrule
    \multicolumn{13}{l}{\textbf{\textit{256 Views}}} \\
    0.8 & \paperautocell{\textbf{0.0361}} & \paperautocell{\textbf{15.49}} & \paperautocell{\textbf{0.9991}} & \paperautocell{\textbf{0.0234}} & \paperautocell{0.6167} & \paperautocell{\textbf{0.9260}} & \paperautocell{\textbf{0.0557}} & \paperautocell{\textbf{15.62}} & \paperautocell{0.9902} & \paperautocell{\textbf{0.0202}} & \paperautocell{0.7176} & \paperautocell{\textbf{0.9283}} \\
    0.9 & \paperautocell{0.0396} & \paperautocell{15.60} & \paperautocell{0.9967} & \paperautocell{0.0240} & \paperautocell{0.6487} & \paperautocell{0.9241} & \paperautocell{0.0619} & \paperautocell{16.12} & \paperautocell{\textbf{0.9955}} & \paperautocell{0.0268} & \paperautocell{0.7515} & \paperautocell{0.9190} \\
    1.0 & \paperautocell{0.0371} & \paperautocell{15.53} & \paperautocell{0.9913} & \paperautocell{0.0235} & \paperautocell{\textbf{0.7849}} & \paperautocell{0.9201} & \paperautocell{0.0684} & \paperautocell{16.54} & \paperautocell{0.9898} & \paperautocell{0.0315} & \paperautocell{\textbf{0.7820}} & \paperautocell{0.9081} \\
    \bottomrule
  \end{tabular}%
  }
\end{table}

Tanks and Temples with 16 input views at 4K shows the same trade-off as the denser ScanNet++
settings (Table~\ref{tab:app-cascade-ablation-tnt}): on average, $t_s=0.8$ gives a slightly
lower F@$\tau$ than $t_s=1$ ($0.2756$ vs.\ $0.2830$) but a clearly lower CD$/\tau$
($9.69$ vs.\ $12.01$). Figure~\ref{fig:app-cascade-ablation-tnt} shows that at $t_s=1$ the reconstruction loses
consistency: floaters appear and the ground has holes.

\begin{table}[H]
  \centering
  \caption{\textbf{Effect of parent initialization on Tanks and Temples with 16 input views.} Every finer cascade level starts from a latent noised to $t_s$, where $t_s=1$ generates from pure noise. Inputs are the 4K images. F@$\tau$ and CD$/\tau$ use the official per-scene threshold $\tau$, and CD is in meters. The best results are in bold and the second best underlined.}
  \label{tab:app-cascade-ablation-tnt}
  \scriptsize
  \setlength{\tabcolsep}{3pt}
  \definecolor{rank1}{HTML}{FFB3B3}
  \definecolor{rank2}{HTML}{FFD9B3}
  \definecolor{rank3}{HTML}{FFFFB3}
  \definecolor{emphrow}{HTML}{EDEDED}
  \ifdefined\paperautocellwidth\else\newlength{\paperautocellwidth}\fi
  \ifdefined\paperautoprobe\else\newlength{\paperautoprobe}\fi
  \settowidth{\paperautocellwidth}{\textbf{00.0000}}
  \settowidth{\paperautoprobe}{Barn}
  \ifdim\dimexpr\paperautoprobe-2\tabcolsep\relax>2\paperautocellwidth  \setlength{\paperautocellwidth}{\dimexpr(\paperautoprobe-2\tabcolsep)/2\relax}\fi
  \settowidth{\paperautoprobe}{Caterpillar}
  \ifdim\dimexpr\paperautoprobe-2\tabcolsep\relax>2\paperautocellwidth  \setlength{\paperautocellwidth}{\dimexpr(\paperautoprobe-2\tabcolsep)/2\relax}\fi
  \settowidth{\paperautoprobe}{Courthouse}
  \ifdim\dimexpr\paperautoprobe-2\tabcolsep\relax>2\paperautocellwidth  \setlength{\paperautocellwidth}{\dimexpr(\paperautoprobe-2\tabcolsep)/2\relax}\fi
  \settowidth{\paperautoprobe}{Ignatius}
  \ifdim\dimexpr\paperautoprobe-2\tabcolsep\relax>2\paperautocellwidth  \setlength{\paperautocellwidth}{\dimexpr(\paperautoprobe-2\tabcolsep)/2\relax}\fi
  \settowidth{\paperautoprobe}{Meetingroom}
  \ifdim\dimexpr\paperautoprobe-2\tabcolsep\relax>2\paperautocellwidth  \setlength{\paperautocellwidth}{\dimexpr(\paperautoprobe-2\tabcolsep)/2\relax}\fi
  \settowidth{\paperautoprobe}{Truck}
  \ifdim\dimexpr\paperautoprobe-2\tabcolsep\relax>2\paperautocellwidth  \setlength{\paperautocellwidth}{\dimexpr(\paperautoprobe-2\tabcolsep)/2\relax}\fi
  \settowidth{\paperautoprobe}{Mean}
  \ifdim\dimexpr\paperautoprobe-2\tabcolsep\relax>2\paperautocellwidth  \setlength{\paperautocellwidth}{\dimexpr(\paperautoprobe-2\tabcolsep)/2\relax}\fi
  \settowidth{\paperautoprobe}{F@$\tau$$\uparrow$}
  \ifdim\dimexpr\paperautoprobe-0\tabcolsep\relax>1\paperautocellwidth  \setlength{\paperautocellwidth}{\dimexpr(\paperautoprobe-0\tabcolsep)/1\relax}\fi
  \settowidth{\paperautoprobe}{CD$\downarrow$}
  \ifdim\dimexpr\paperautoprobe-0\tabcolsep\relax>1\paperautocellwidth  \setlength{\paperautocellwidth}{\dimexpr(\paperautoprobe-0\tabcolsep)/1\relax}\fi
  \settowidth{\paperautoprobe}{CD$/\tau$$\downarrow$}
  \ifdim\dimexpr\paperautoprobe-0\tabcolsep\relax>1\paperautocellwidth  \setlength{\paperautocellwidth}{\dimexpr(\paperautoprobe-0\tabcolsep)/1\relax}\fi
  \providecommand{\paperautospan}[2]{\makebox[\dimexpr#1\paperautocellwidth+#1\tabcolsep+#1\tabcolsep-\tabcolsep-\tabcolsep\relax][c]{#2}}
  \providecommand{\paperautocell}[1]{\paperautospan{1}{#1}}
  \resizebox{\linewidth}{!}{%
  \begin{tabular}{lcccccccccccccc}
    \toprule
     & \multicolumn{2}{c}{\paperautospan{2}{Barn}} & \multicolumn{2}{c}{\paperautospan{2}{Caterpillar}} & \multicolumn{2}{c}{\paperautospan{2}{Courthouse}} & \multicolumn{2}{c}{\paperautospan{2}{Ignatius}} & \multicolumn{2}{c}{\paperautospan{2}{Meetingroom}} & \multicolumn{2}{c}{\paperautospan{2}{Truck}} & \multicolumn{2}{c}{\paperautospan{2}{Mean}} \\
    \cmidrule(lr){2-3}\cmidrule(lr){4-5}\cmidrule(lr){6-7}\cmidrule(lr){8-9}\cmidrule(lr){10-11}\cmidrule(lr){12-13}\cmidrule(lr){14-15}
    $t_s$ & \multicolumn{1}{c}{\paperautospan{1}{F@$\tau$$\uparrow$}} & \multicolumn{1}{c}{\paperautospan{1}{CD$\downarrow$}} & \multicolumn{1}{c}{\paperautospan{1}{F@$\tau$$\uparrow$}} & \multicolumn{1}{c}{\paperautospan{1}{CD$\downarrow$}} & \multicolumn{1}{c}{\paperautospan{1}{F@$\tau$$\uparrow$}} & \multicolumn{1}{c}{\paperautospan{1}{CD$\downarrow$}} & \multicolumn{1}{c}{\paperautospan{1}{F@$\tau$$\uparrow$}} & \multicolumn{1}{c}{\paperautospan{1}{CD$\downarrow$}} & \multicolumn{1}{c}{\paperautospan{1}{F@$\tau$$\uparrow$}} & \multicolumn{1}{c}{\paperautospan{1}{CD$\downarrow$}} & \multicolumn{1}{c}{\paperautospan{1}{F@$\tau$$\uparrow$}} & \multicolumn{1}{c}{\paperautospan{1}{CD$\downarrow$}} & \multicolumn{1}{c}{\paperautospan{1}{F@$\tau$$\uparrow$}} & \multicolumn{1}{c}{\paperautospan{1}{CD$/\tau$$\downarrow$}} \\
    \midrule
    \cellcolor{emphrow}\textbf{0.8} & \paperautocell{0.2080} & \paperautocell{\underline{0.0895}} & \paperautocell{0.1949} & \paperautocell{0.0637} & \paperautocell{\underline{0.1159}} & \paperautocell{\textbf{0.5905}} & \paperautocell{\textbf{0.4704}} & \paperautocell{\textbf{0.0055}} & \paperautocell{\textbf{0.3084}} & \paperautocell{\textbf{0.0476}} & \paperautocell{0.3563} & \paperautocell{\underline{0.0311}} & \paperautocell{0.2756} & \paperautocell{\textbf{9.6855}} \\
    0.9 & \paperautocell{\textbf{0.2336}} & \paperautocell{\textbf{0.0831}} & \paperautocell{\underline{0.1960}} & \paperautocell{\underline{0.0575}} & \paperautocell{\textbf{0.1169}} & \paperautocell{\underline{0.6717}} & \paperautocell{0.4436} & \paperautocell{\underline{0.0058}} & \paperautocell{\underline{0.3071}} & \paperautocell{\underline{0.0884}} & \paperautocell{\underline{0.3850}} & \paperautocell{\textbf{0.0304}} & \paperautocell{\underline{0.2804}} & \paperautocell{\underline{10.5876}} \\
    1.0 & \paperautocell{\underline{0.2304}} & \paperautocell{0.1030} & \paperautocell{\textbf{0.2057}} & \paperautocell{\textbf{0.0555}} & \paperautocell{0.1084} & \paperautocell{0.6974} & \paperautocell{\underline{0.4584}} & \paperautocell{0.0084} & \paperautocell{0.2654} & \paperautocell{0.1313} & \paperautocell{\textbf{0.4300}} & \paperautocell{0.0343} & \paperautocell{\textbf{0.2830}} & \paperautocell{12.0138} \\
    \bottomrule
  \end{tabular}%
  }
\end{table}

\begin{figure}[H]
  \centering
  \setlength{\tabcolsep}{0.5pt}
  \renewcommand{\arraystretch}{1.0}
  \begin{tabular}{ccc}
    \multicolumn{3}{c}{\scriptsize (a) \texttt{ignatius}} \\[1.5pt]
    \includegraphics[width=\dimexpr(\textwidth-3pt)/3\relax]{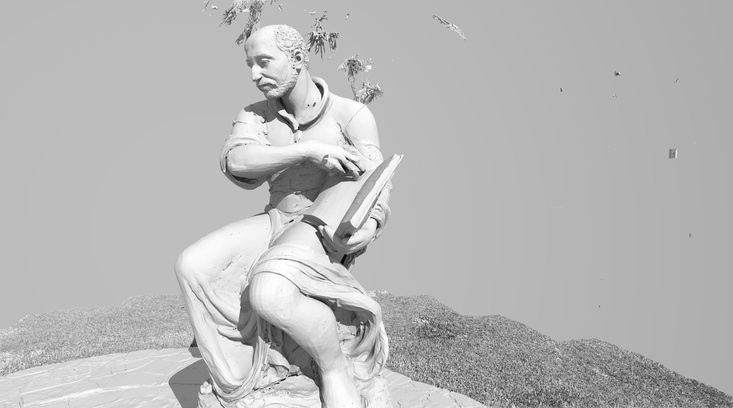} & \includegraphics[width=\dimexpr(\textwidth-3pt)/3\relax]{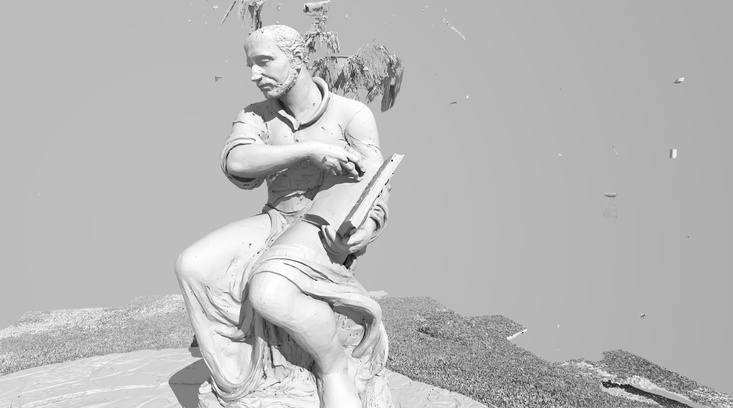} & \includegraphics[width=\dimexpr(\textwidth-3pt)/3\relax]{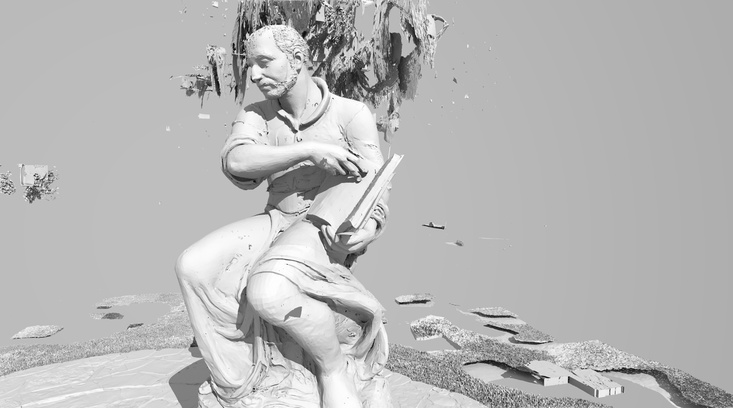} \\[-4.5pt]
    \makebox[\dimexpr(\textwidth-3pt)/3\relax][c]{\scriptsize $t_s=0.8$} & \makebox[\dimexpr(\textwidth-3pt)/3\relax][c]{\scriptsize $t_s=0.9$} & \makebox[\dimexpr(\textwidth-3pt)/3\relax][c]{\scriptsize $t_s=1$} \\[8pt]
    \multicolumn{3}{c}{\scriptsize (b) \texttt{caterpillar}} \\[1.5pt]
    \includegraphics[width=\dimexpr(\textwidth-3pt)/3\relax]{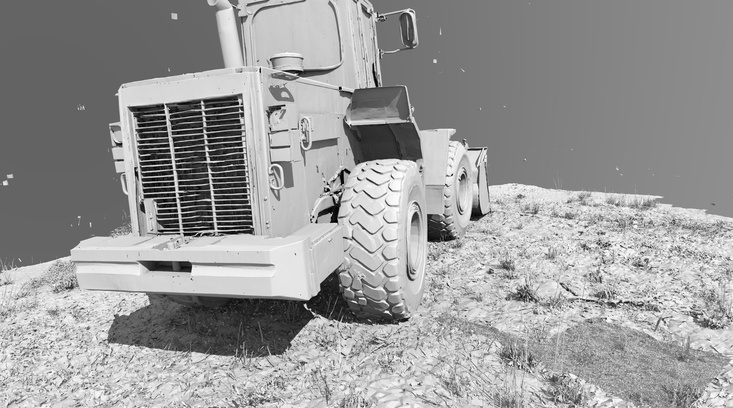} & \includegraphics[width=\dimexpr(\textwidth-3pt)/3\relax]{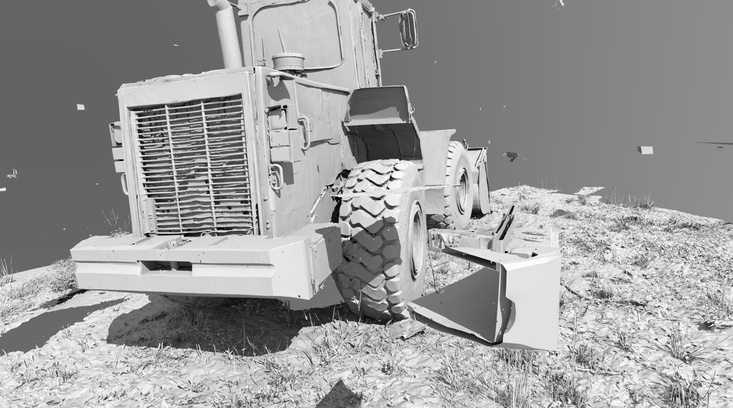} & \includegraphics[width=\dimexpr(\textwidth-3pt)/3\relax]{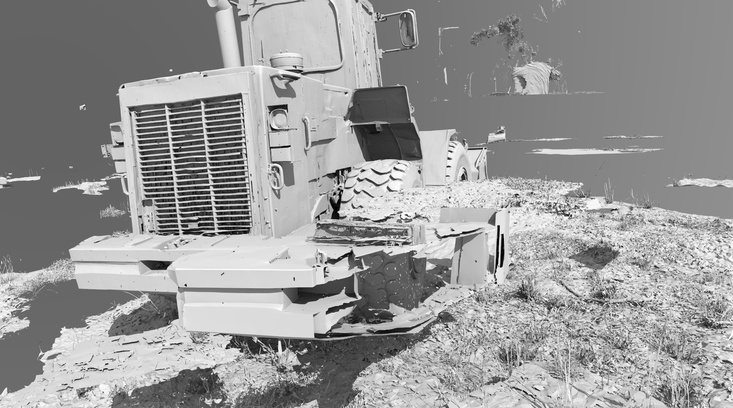} \\[-4.5pt]
    \makebox[\dimexpr(\textwidth-3pt)/3\relax][c]{\scriptsize $t_s=0.8$} & \makebox[\dimexpr(\textwidth-3pt)/3\relax][c]{\scriptsize $t_s=0.9$} & \makebox[\dimexpr(\textwidth-3pt)/3\relax][c]{\scriptsize $t_s=1$} \\
  \end{tabular}
  \caption{\textbf{Effect of parent initialization on Tanks and Temples.} Meshes at the finest cascade level from 16 input views at 4K, rendered from the same viewpoint per scene.}
  \label{fig:app-cascade-ablation-tnt}
\end{figure}

\subsection{Image conditioning}
\label{app:conditioning-ablation}

Table~\ref{tab:app-conditioning-tnt} extends the comparison in
Figure~\ref{fig:ablation-conditioning} to Tanks and Temples with 32 input views across
all cascade levels. Voxel-queried attention slightly improves the mean F@$\tau$ over the
baseline at every cascade level, while the ConvNeXt branch achieves the highest mean
F@$\tau$ across all levels. Figure~\ref{fig:app-conditioning-tnt} compares the three
conditioning variants on \texttt{ignatius} with 32 input views and on \texttt{caterpillar}
with 256 input views. The baseline produces overly smooth geometry: it flattens the face and
clothing folds of the statue, and merges the bars of the loader's rear grille
into a continuous wavy surface. Voxel-queried attention better recovers facial features and
clothing folds while separating several grille bars; the ConvNeXt branch sharpens these details
further, cleanly resolving individual bars.

\begin{table}[H]
  \centering
  \caption{\textbf{Effect of image conditioning on Tanks and Temples with 32 input views.} The baseline uses the GenRecon-style conditioner, our voxel-queried attention replaces it, and the ConvNeXt branch is then added. Inputs are the 4K images. F@$\tau$ and CD$/\tau$ use the official per-scene threshold $\tau$, and CD is in meters. The best results are in bold and the second best underlined.}
  \label{tab:app-conditioning-tnt}
  \scriptsize
  \setlength{\tabcolsep}{3pt}
  \definecolor{rank1}{HTML}{FFB3B3}
  \definecolor{rank2}{HTML}{FFD9B3}
  \definecolor{rank3}{HTML}{FFFFB3}
  \definecolor{emphrow}{HTML}{EDEDED}
  \ifdefined\paperautocellwidth\else\newlength{\paperautocellwidth}\fi
  \ifdefined\paperautoprobe\else\newlength{\paperautoprobe}\fi
  \settowidth{\paperautocellwidth}{\textbf{0.0000}}
  \settowidth{\paperautoprobe}{Barn}
  \ifdim\dimexpr\paperautoprobe-2\tabcolsep\relax>2\paperautocellwidth  \setlength{\paperautocellwidth}{\dimexpr(\paperautoprobe-2\tabcolsep)/2\relax}\fi
  \settowidth{\paperautoprobe}{Caterpillar}
  \ifdim\dimexpr\paperautoprobe-2\tabcolsep\relax>2\paperautocellwidth  \setlength{\paperautocellwidth}{\dimexpr(\paperautoprobe-2\tabcolsep)/2\relax}\fi
  \settowidth{\paperautoprobe}{Courthouse}
  \ifdim\dimexpr\paperautoprobe-2\tabcolsep\relax>2\paperautocellwidth  \setlength{\paperautocellwidth}{\dimexpr(\paperautoprobe-2\tabcolsep)/2\relax}\fi
  \settowidth{\paperautoprobe}{Ignatius}
  \ifdim\dimexpr\paperautoprobe-2\tabcolsep\relax>2\paperautocellwidth  \setlength{\paperautocellwidth}{\dimexpr(\paperautoprobe-2\tabcolsep)/2\relax}\fi
  \settowidth{\paperautoprobe}{Meetingroom}
  \ifdim\dimexpr\paperautoprobe-2\tabcolsep\relax>2\paperautocellwidth  \setlength{\paperautocellwidth}{\dimexpr(\paperautoprobe-2\tabcolsep)/2\relax}\fi
  \settowidth{\paperautoprobe}{Truck}
  \ifdim\dimexpr\paperautoprobe-2\tabcolsep\relax>2\paperautocellwidth  \setlength{\paperautocellwidth}{\dimexpr(\paperautoprobe-2\tabcolsep)/2\relax}\fi
  \settowidth{\paperautoprobe}{Mean}
  \ifdim\dimexpr\paperautoprobe-2\tabcolsep\relax>2\paperautocellwidth  \setlength{\paperautocellwidth}{\dimexpr(\paperautoprobe-2\tabcolsep)/2\relax}\fi
  \settowidth{\paperautoprobe}{F@$\tau$$\uparrow$}
  \ifdim\dimexpr\paperautoprobe-0\tabcolsep\relax>1\paperautocellwidth  \setlength{\paperautocellwidth}{\dimexpr(\paperautoprobe-0\tabcolsep)/1\relax}\fi
  \settowidth{\paperautoprobe}{CD$\downarrow$}
  \ifdim\dimexpr\paperautoprobe-0\tabcolsep\relax>1\paperautocellwidth  \setlength{\paperautocellwidth}{\dimexpr(\paperautoprobe-0\tabcolsep)/1\relax}\fi
  \settowidth{\paperautoprobe}{CD$/\tau$$\downarrow$}
  \ifdim\dimexpr\paperautoprobe-0\tabcolsep\relax>1\paperautocellwidth  \setlength{\paperautocellwidth}{\dimexpr(\paperautoprobe-0\tabcolsep)/1\relax}\fi
  \providecommand{\paperautospan}[2]{\makebox[\dimexpr#1\paperautocellwidth+#1\tabcolsep+#1\tabcolsep-\tabcolsep-\tabcolsep\relax][c]{#2}}
  \providecommand{\paperautocell}[1]{\paperautospan{1}{#1}}
  \resizebox{\linewidth}{!}{%
  \begin{tabular}{lcccccccccccccc}
    \toprule
     & \multicolumn{2}{c}{\paperautospan{2}{Barn}} & \multicolumn{2}{c}{\paperautospan{2}{Caterpillar}} & \multicolumn{2}{c}{\paperautospan{2}{Courthouse}} & \multicolumn{2}{c}{\paperautospan{2}{Ignatius}} & \multicolumn{2}{c}{\paperautospan{2}{Meetingroom}} & \multicolumn{2}{c}{\paperautospan{2}{Truck}} & \multicolumn{2}{c}{\paperautospan{2}{Mean}} \\
    \cmidrule(lr){2-3}\cmidrule(lr){4-5}\cmidrule(lr){6-7}\cmidrule(lr){8-9}\cmidrule(lr){10-11}\cmidrule(lr){12-13}\cmidrule(lr){14-15}
    Conditioning & \multicolumn{1}{c}{\paperautospan{1}{F@$\tau$$\uparrow$}} & \multicolumn{1}{c}{\paperautospan{1}{CD$\downarrow$}} & \multicolumn{1}{c}{\paperautospan{1}{F@$\tau$$\uparrow$}} & \multicolumn{1}{c}{\paperautospan{1}{CD$\downarrow$}} & \multicolumn{1}{c}{\paperautospan{1}{F@$\tau$$\uparrow$}} & \multicolumn{1}{c}{\paperautospan{1}{CD$\downarrow$}} & \multicolumn{1}{c}{\paperautospan{1}{F@$\tau$$\uparrow$}} & \multicolumn{1}{c}{\paperautospan{1}{CD$\downarrow$}} & \multicolumn{1}{c}{\paperautospan{1}{F@$\tau$$\uparrow$}} & \multicolumn{1}{c}{\paperautospan{1}{CD$\downarrow$}} & \multicolumn{1}{c}{\paperautospan{1}{F@$\tau$$\uparrow$}} & \multicolumn{1}{c}{\paperautospan{1}{CD$\downarrow$}} & \multicolumn{1}{c}{\paperautospan{1}{F@$\tau$$\uparrow$}} & \multicolumn{1}{c}{\paperautospan{1}{CD$/\tau$$\downarrow$}} \\
    \midrule
    \multicolumn{15}{l}{\textbf{\textit{L0}}} \\
    Baseline & \paperautocell{\textbf{0.1781}} & \paperautocell{\underline{0.0691}} & \paperautocell{0.1124} & \paperautocell{0.0839} & \paperautocell{0.0836} & \paperautocell{0.4428} & \paperautocell{0.2092} & \paperautocell{0.0113} & \paperautocell{0.3383} & \paperautocell{0.0360} & \paperautocell{0.2472} & \paperautocell{\underline{0.0344}} & \paperautocell{0.1948} & \paperautocell{9.2747} \\
    Voxel attn. & \paperautocell{\underline{0.1704}} & \paperautocell{\textbf{0.0645}} & \paperautocell{\underline{0.1223}} & \paperautocell{\underline{0.0814}} & \paperautocell{\underline{0.0875}} & \paperautocell{\underline{0.3684}} & \paperautocell{\underline{0.2381}} & \paperautocell{\underline{0.0100}} & \paperautocell{\underline{0.3531}} & \paperautocell{\underline{0.0359}} & \paperautocell{\underline{0.2482}} & \paperautocell{0.0347} & \paperautocell{\underline{0.2033}} & \paperautocell{\underline{8.5584}} \\
    \cellcolor{emphrow}\textbf{+ ConvNeXt} & \paperautocell{0.1463} & \paperautocell{0.0787} & \paperautocell{\textbf{0.1328}} & \paperautocell{\textbf{0.0805}} & \paperautocell{\textbf{0.0965}} & \paperautocell{\textbf{0.3619}} & \paperautocell{\textbf{0.3223}} & \paperautocell{\textbf{0.0073}} & \paperautocell{\textbf{0.3561}} & \paperautocell{\textbf{0.0349}} & \paperautocell{\textbf{0.2955}} & \paperautocell{\textbf{0.0317}} & \paperautocell{\textbf{0.2249}} & \paperautocell{\textbf{8.4522}} \\
    \midrule
    \multicolumn{15}{l}{\textbf{\textit{L1}}} \\
    Baseline & \paperautocell{0.1734} & \paperautocell{\underline{0.0762}} & \paperautocell{0.1439} & \paperautocell{0.0761} & \paperautocell{\underline{0.1146}} & \paperautocell{\underline{0.3388}} & \paperautocell{0.2918} & \paperautocell{0.0076} & \paperautocell{0.3579} & \paperautocell{0.0329} & \paperautocell{0.2659} & \paperautocell{\underline{0.0316}} & \paperautocell{0.2246} & \paperautocell{\underline{8.0897}} \\
    Voxel attn. & \paperautocell{\underline{0.1819}} & \paperautocell{0.0770} & \paperautocell{\underline{0.1524}} & \paperautocell{\underline{0.0732}} & \paperautocell{0.1038} & \paperautocell{0.3645} & \paperautocell{\underline{0.3175}} & \paperautocell{\underline{0.0072}} & \paperautocell{\underline{0.3647}} & \paperautocell{\underline{0.0321}} & \paperautocell{\underline{0.2668}} & \paperautocell{0.0323} & \paperautocell{\underline{0.2312}} & \paperautocell{8.1666} \\
    \cellcolor{emphrow}\textbf{+ ConvNeXt} & \paperautocell{\textbf{0.1926}} & \paperautocell{\textbf{0.0641}} & \paperautocell{\textbf{0.1752}} & \paperautocell{\textbf{0.0727}} & \paperautocell{\textbf{0.1181}} & \paperautocell{\textbf{0.3355}} & \paperautocell{\textbf{0.4194}} & \paperautocell{\textbf{0.0058}} & \paperautocell{\textbf{0.3721}} & \paperautocell{\textbf{0.0318}} & \paperautocell{\textbf{0.3523}} & \paperautocell{\textbf{0.0305}} & \paperautocell{\textbf{0.2716}} & \paperautocell{\textbf{7.5968}} \\
    \midrule
    \multicolumn{15}{l}{\textbf{\textit{L2}}} \\
    Baseline & \paperautocell{0.1623} & \paperautocell{\textbf{0.0798}} & \paperautocell{\underline{0.1778}} & \paperautocell{\underline{0.0720}} & \paperautocell{\underline{0.1179}} & \paperautocell{0.2642} & \paperautocell{0.3784} & \paperautocell{0.0063} & \paperautocell{0.3617} & \paperautocell{0.0320} & \paperautocell{\underline{0.2909}} & \paperautocell{0.0314} & \paperautocell{0.2482} & \paperautocell{\underline{7.4231}} \\
    Voxel attn. & \paperautocell{\textbf{0.1944}} & \paperautocell{0.0887} & \paperautocell{0.1762} & \paperautocell{\textbf{0.0695}} & \paperautocell{\underline{0.1179}} & \paperautocell{\underline{0.2607}} & \paperautocell{\underline{0.3904}} & \paperautocell{\underline{0.0061}} & \paperautocell{\underline{0.3636}} & \paperautocell{\underline{0.0319}} & \paperautocell{0.2733} & \paperautocell{\underline{0.0311}} & \paperautocell{\underline{0.2527}} & \paperautocell{7.4387} \\
    \cellcolor{emphrow}\textbf{+ ConvNeXt} & \paperautocell{\underline{0.1938}} & \paperautocell{\underline{0.0866}} & \paperautocell{\textbf{0.1945}} & \paperautocell{0.0740} & \paperautocell{\textbf{0.1324}} & \paperautocell{\textbf{0.2553}} & \paperautocell{\textbf{0.4904}} & \paperautocell{\textbf{0.0051}} & \paperautocell{\textbf{0.3806}} & \paperautocell{\textbf{0.0308}} & \paperautocell{\textbf{0.3705}} & \paperautocell{\textbf{0.0304}} & \paperautocell{\textbf{0.2937}} & \paperautocell{\textbf{7.4218}} \\
    \bottomrule
  \end{tabular}%
  }
\end{table}

\begin{figure}[H]
  \centering
  \setlength{\tabcolsep}{0.5pt}
  \renewcommand{\arraystretch}{1.0}
  \begin{tabular}{cccc}
    \multicolumn{4}{c}{\scriptsize (a) \texttt{ignatius}, 32 input views} \\[1.5pt]
    \includegraphics[width=\dimexpr(\textwidth-4pt)/4\relax]{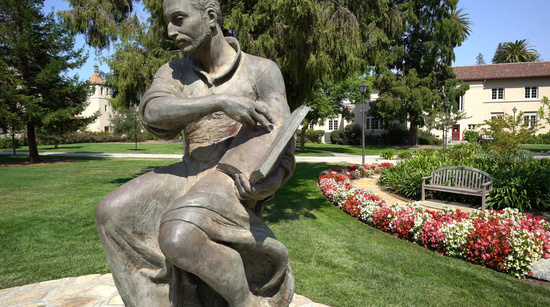} & \includegraphics[width=\dimexpr(\textwidth-4pt)/4\relax]{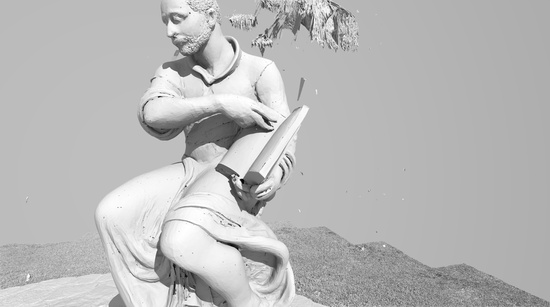} & \includegraphics[width=\dimexpr(\textwidth-4pt)/4\relax]{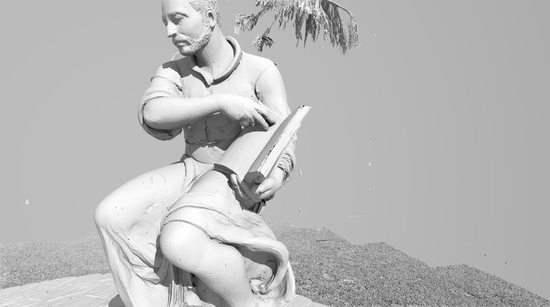} & \includegraphics[width=\dimexpr(\textwidth-4pt)/4\relax]{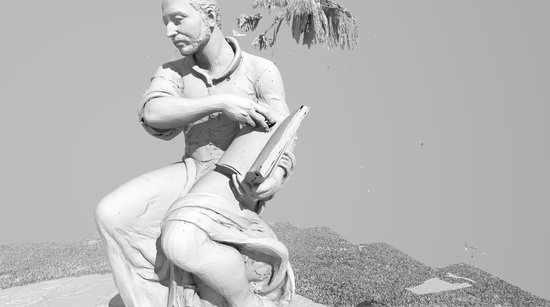} \\[-4.5pt]
    \makebox[\dimexpr(\textwidth-4pt)/4\relax][c]{\scriptsize Image} & \makebox[\dimexpr(\textwidth-4pt)/4\relax][c]{\scriptsize Baseline} & \makebox[\dimexpr(\textwidth-4pt)/4\relax][c]{\scriptsize Voxel attn.} & \makebox[\dimexpr(\textwidth-4pt)/4\relax][c]{\scriptsize + ConvNeXt} \\[8pt]
    \multicolumn{4}{c}{\scriptsize (b) \texttt{caterpillar}, 256 input views} \\[1.5pt]
    \includegraphics[width=\dimexpr(\textwidth-4pt)/4\relax]{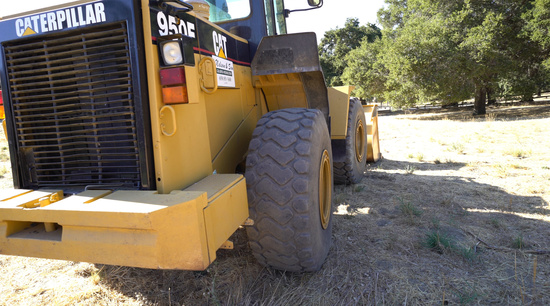} & \includegraphics[width=\dimexpr(\textwidth-4pt)/4\relax]{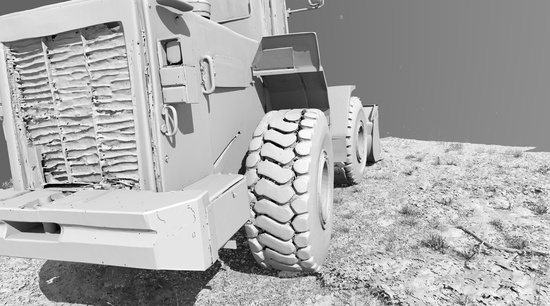} & \includegraphics[width=\dimexpr(\textwidth-4pt)/4\relax]{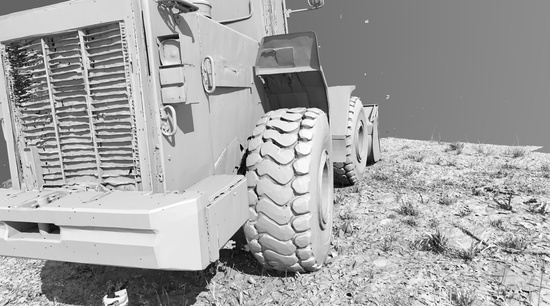} & \includegraphics[width=\dimexpr(\textwidth-4pt)/4\relax]{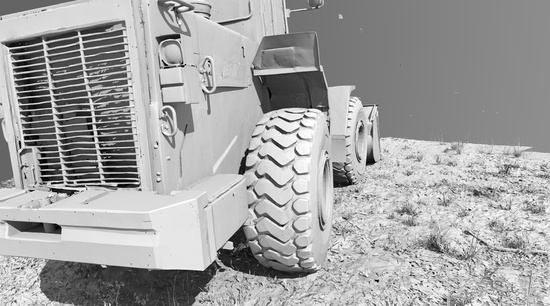} \\[-4.5pt]
    \makebox[\dimexpr(\textwidth-4pt)/4\relax][c]{\scriptsize Image} & \makebox[\dimexpr(\textwidth-4pt)/4\relax][c]{\scriptsize Baseline} & \makebox[\dimexpr(\textwidth-4pt)/4\relax][c]{\scriptsize Voxel attn.} & \makebox[\dimexpr(\textwidth-4pt)/4\relax][c]{\scriptsize + ConvNeXt} \\
  \end{tabular}
  \caption{\textbf{Effect of image conditioning on Tanks and Temples.} Meshes at the finest cascade level from the 4K images, all rendered from the same viewpoint per scene. The baseline uses the GenRecon-style conditioner, our voxel-queried attention replaces it, and the ConvNeXt branch is then added.}
  \label{fig:app-conditioning-tnt}
\end{figure}

\subsection{Tile pyramid level selection}
\label{app:tile-level-ablation}

For each view, a cell attends to the pyramid level matching its voxel spacing
(Section~\ref{sec:method:attention}). As detailed in Appendix~\ref{app:da3-selection}, we choose this level so that one SS voxel covers approximately 8 pixels
(Equation~\ref{eq:app-preferred-pyramid-level}). We compare this adaptive selection with
two fixed levels shared across all cells:
\textit{Coarse} selects the coarsest level whose image height is at least 512 pixels
(584 pixels on ScanNet++ and approximately 546 pixels on Tanks and Temples), whereas
\textit{Fine} uses the full-resolution images. Each configuration uses a separate
model trained with the corresponding selection rule under the same schedule.
We report every cascade level with 32 input views, on ScanNet++ in
Table~\ref{tab:app-tile-level-scannetpp} and on Tanks and Temples in
Table~\ref{tab:app-tile-level-tnt}.

On ScanNet++ (Table~\ref{tab:app-tile-level-scannetpp}), the images are 1168 pixels
high, so \textit{Coarse} halves them only once, to 584 pixels, and keeps most of their
resolution. \textit{Coarse} is therefore competitive on this dataset: it achieves a
slightly higher F@2 and a similar CD at levels 1 and 2. The adaptive selection still
yields the lowest MAE across all cascade levels and scene groups, and \textit{Fine}
underperforms the adaptive selection in nearly every column.

In contrast, on Tanks and Temples (Table~\ref{tab:app-tile-level-tnt}), the inputs are
4K images about 2180 pixels high, so \textit{Coarse} downsamples them by a factor of four,
discarding high-frequency details needed by finer cells. As a result, \textit{Coarse}
falls behind at level 2, with a mean F@$\tau$ of $0.2305$ compared to $0.2937$. While
\textit{Fine} performs best on certain scenes (e.g., Barn), the adaptive selection
achieves the best mean F@$\tau$ and CD$/\tau$ across all cascade levels.

\begin{table}[H]
  \centering
  \caption{\textbf{Effect of tile pyramid level selection on ScanNet++ with 32 input views.} Coarse fixes every cell to the coarsest pyramid level whose image height is at least 512 pixels, and Fine to the full-resolution level; ours matches the level to the cell's voxel spacing. Mean over the 25 smallest, the 25 largest, and all 50 scenes, with the best results in bold.}
  \label{tab:app-tile-level-scannetpp}
  \scriptsize
  \setlength{\tabcolsep}{3pt}
  \definecolor{rank1}{HTML}{FFB3B3}
  \definecolor{rank2}{HTML}{FFD9B3}
  \definecolor{rank3}{HTML}{FFFFB3}
  \definecolor{emphrow}{HTML}{EDEDED}
  \ifdefined\paperautocellwidth\else\newlength{\paperautocellwidth}\fi
  \ifdefined\paperautoprobe\else\newlength{\paperautoprobe}\fi
  \settowidth{\paperautocellwidth}{\textbf{0.0000}}
  \settowidth{\paperautoprobe}{Small (25)}
  \ifdim\dimexpr\paperautoprobe-10\tabcolsep\relax>6\paperautocellwidth  \setlength{\paperautocellwidth}{\dimexpr(\paperautoprobe-10\tabcolsep)/6\relax}\fi
  \settowidth{\paperautoprobe}{Large (25)}
  \ifdim\dimexpr\paperautoprobe-10\tabcolsep\relax>6\paperautocellwidth  \setlength{\paperautocellwidth}{\dimexpr(\paperautoprobe-10\tabcolsep)/6\relax}\fi
  \settowidth{\paperautoprobe}{All (50)}
  \ifdim\dimexpr\paperautoprobe-10\tabcolsep\relax>6\paperautocellwidth  \setlength{\paperautocellwidth}{\dimexpr(\paperautoprobe-10\tabcolsep)/6\relax}\fi
  \settowidth{\paperautoprobe}{2D}
  \ifdim\dimexpr\paperautoprobe-4\tabcolsep\relax>3\paperautocellwidth  \setlength{\paperautocellwidth}{\dimexpr(\paperautoprobe-4\tabcolsep)/3\relax}\fi
  \settowidth{\paperautoprobe}{3D virtual scan}
  \ifdim\dimexpr\paperautoprobe-4\tabcolsep\relax>3\paperautocellwidth  \setlength{\paperautocellwidth}{\dimexpr(\paperautoprobe-4\tabcolsep)/3\relax}\fi
  \settowidth{\paperautoprobe}{MAE$\downarrow$}
  \ifdim\dimexpr\paperautoprobe-0\tabcolsep\relax>1\paperautocellwidth  \setlength{\paperautocellwidth}{\dimexpr(\paperautoprobe-0\tabcolsep)/1\relax}\fi
  \settowidth{\paperautoprobe}{Ang$\downarrow$}
  \ifdim\dimexpr\paperautoprobe-0\tabcolsep\relax>1\paperautocellwidth  \setlength{\paperautocellwidth}{\dimexpr(\paperautoprobe-0\tabcolsep)/1\relax}\fi
  \settowidth{\paperautoprobe}{Comp$\uparrow$}
  \ifdim\dimexpr\paperautoprobe-0\tabcolsep\relax>1\paperautocellwidth  \setlength{\paperautocellwidth}{\dimexpr(\paperautoprobe-0\tabcolsep)/1\relax}\fi
  \settowidth{\paperautoprobe}{CD$\downarrow$}
  \ifdim\dimexpr\paperautoprobe-0\tabcolsep\relax>1\paperautocellwidth  \setlength{\paperautocellwidth}{\dimexpr(\paperautoprobe-0\tabcolsep)/1\relax}\fi
  \settowidth{\paperautoprobe}{F@2$\uparrow$}
  \ifdim\dimexpr\paperautoprobe-0\tabcolsep\relax>1\paperautocellwidth  \setlength{\paperautocellwidth}{\dimexpr(\paperautoprobe-0\tabcolsep)/1\relax}\fi
  \settowidth{\paperautoprobe}{NC$\uparrow$}
  \ifdim\dimexpr\paperautoprobe-0\tabcolsep\relax>1\paperautocellwidth  \setlength{\paperautocellwidth}{\dimexpr(\paperautoprobe-0\tabcolsep)/1\relax}\fi
  \providecommand{\paperautospan}[2]{\makebox[\dimexpr#1\paperautocellwidth+#1\tabcolsep+#1\tabcolsep-\tabcolsep-\tabcolsep\relax][c]{#2}}
  \providecommand{\paperautocell}[1]{\paperautospan{1}{#1}}
  \resizebox{\linewidth}{!}{%
  \begin{tabular}{lcccccccccccccccccc}
    \toprule
     & \multicolumn{6}{c}{\paperautospan{6}{Small (25)}} & \multicolumn{6}{c}{\paperautospan{6}{Large (25)}} & \multicolumn{6}{c}{\paperautospan{6}{All (50)}} \\
    \cmidrule(lr){2-7}\cmidrule(lr){8-13}\cmidrule(lr){14-19}
     & \multicolumn{3}{c}{\paperautospan{3}{2D}} & \multicolumn{3}{c}{\paperautospan{3}{3D virtual scan}} & \multicolumn{3}{c}{\paperautospan{3}{2D}} & \multicolumn{3}{c}{\paperautospan{3}{3D virtual scan}} & \multicolumn{3}{c}{\paperautospan{3}{2D}} & \multicolumn{3}{c}{\paperautospan{3}{3D virtual scan}} \\
    \cmidrule(lr){2-4}\cmidrule(lr){5-7}\cmidrule(lr){8-10}\cmidrule(lr){11-13}\cmidrule(lr){14-16}\cmidrule(lr){17-19}
    Tile level & \multicolumn{1}{c}{\paperautospan{1}{MAE$\downarrow$}} & \multicolumn{1}{c}{\paperautospan{1}{Ang$\downarrow$}} & \multicolumn{1}{c}{\paperautospan{1}{Comp$\uparrow$}} & \multicolumn{1}{c}{\paperautospan{1}{CD$\downarrow$}} & \multicolumn{1}{c}{\paperautospan{1}{F@2$\uparrow$}} & \multicolumn{1}{c}{\paperautospan{1}{NC$\uparrow$}} & \multicolumn{1}{c}{\paperautospan{1}{MAE$\downarrow$}} & \multicolumn{1}{c}{\paperautospan{1}{Ang$\downarrow$}} & \multicolumn{1}{c}{\paperautospan{1}{Comp$\uparrow$}} & \multicolumn{1}{c}{\paperautospan{1}{CD$\downarrow$}} & \multicolumn{1}{c}{\paperautospan{1}{F@2$\uparrow$}} & \multicolumn{1}{c}{\paperautospan{1}{NC$\uparrow$}} & \multicolumn{1}{c}{\paperautospan{1}{MAE$\downarrow$}} & \multicolumn{1}{c}{\paperautospan{1}{Ang$\downarrow$}} & \multicolumn{1}{c}{\paperautospan{1}{Comp$\uparrow$}} & \multicolumn{1}{c}{\paperautospan{1}{CD$\downarrow$}} & \multicolumn{1}{c}{\paperautospan{1}{F@2$\uparrow$}} & \multicolumn{1}{c}{\paperautospan{1}{NC$\uparrow$}} \\
    \midrule
    \multicolumn{19}{l}{\textbf{\textit{L0}}} \\
    Coarse & \paperautocell{0.0456} & \paperautocell{18.85} & \paperautocell{0.9984} & \paperautocell{0.0241} & \paperautocell{0.5633} & \paperautocell{0.9073} & \paperautocell{0.0718} & \paperautocell{\textbf{19.14}} & \paperautocell{\textbf{0.9986}} & \paperautocell{\textbf{0.0231}} & \paperautocell{\textbf{0.6349}} & \paperautocell{\textbf{0.9097}} & \paperautocell{0.0587} & \paperautocell{19.00} & \paperautocell{\textbf{0.9985}} & \paperautocell{0.0236} & \paperautocell{0.5991} & \paperautocell{0.9085} \\
    \cellcolor{emphrow}\textbf{Adaptive (ours)} & \paperautocell{\textbf{0.0438}} & \paperautocell{\textbf{18.74}} & \paperautocell{\textbf{0.9986}} & \paperautocell{\textbf{0.0230}} & \paperautocell{\textbf{0.5816}} & \paperautocell{\textbf{0.9110}} & \paperautocell{\textbf{0.0714}} & \paperautocell{19.21} & \paperautocell{0.9984} & \paperautocell{0.0235} & \paperautocell{0.6336} & \paperautocell{0.9090} & \paperautocell{\textbf{0.0576}} & \paperautocell{\textbf{18.98}} & \paperautocell{\textbf{0.9985}} & \paperautocell{\textbf{0.0233}} & \paperautocell{\textbf{0.6076}} & \paperautocell{\textbf{0.9100}} \\
    Fine & \paperautocell{0.0578} & \paperautocell{20.01} & \paperautocell{0.9967} & \paperautocell{0.0313} & \paperautocell{0.4919} & \paperautocell{0.8894} & \paperautocell{0.0892} & \paperautocell{19.89} & \paperautocell{0.9977} & \paperautocell{0.0275} & \paperautocell{0.5849} & \paperautocell{0.8966} & \paperautocell{0.0735} & \paperautocell{19.95} & \paperautocell{0.9972} & \paperautocell{0.0294} & \paperautocell{0.5384} & \paperautocell{0.8930} \\
    \midrule
    \multicolumn{19}{l}{\textbf{\textit{L1}}} \\
    Coarse & \paperautocell{0.0355} & \paperautocell{16.68} & \paperautocell{0.9988} & \paperautocell{\textbf{0.0199}} & \paperautocell{\textbf{0.6476}} & \paperautocell{0.9227} & \paperautocell{0.0631} & \paperautocell{17.26} & \paperautocell{\textbf{0.9991}} & \paperautocell{\textbf{0.0198}} & \paperautocell{\textbf{0.7049}} & \paperautocell{\textbf{0.9232}} & \paperautocell{0.0493} & \paperautocell{16.97} & \paperautocell{\textbf{0.9989}} & \paperautocell{\textbf{0.0199}} & \paperautocell{\textbf{0.6762}} & \paperautocell{0.9229} \\
    \cellcolor{emphrow}\textbf{Adaptive (ours)} & \paperautocell{\textbf{0.0346}} & \paperautocell{\textbf{16.35}} & \paperautocell{\textbf{0.9994}} & \paperautocell{0.0200} & \paperautocell{0.6359} & \paperautocell{\textbf{0.9240}} & \paperautocell{\textbf{0.0569}} & \paperautocell{17.10} & \paperautocell{0.9983} & \paperautocell{0.0199} & \paperautocell{0.6959} & \paperautocell{\textbf{0.9232}} & \paperautocell{\textbf{0.0457}} & \paperautocell{\textbf{16.72}} & \paperautocell{0.9988} & \paperautocell{\textbf{0.0199}} & \paperautocell{0.6659} & \paperautocell{\textbf{0.9236}} \\
    Fine & \paperautocell{0.0407} & \paperautocell{17.23} & \paperautocell{0.9980} & \paperautocell{0.0239} & \paperautocell{0.5893} & \paperautocell{0.9144} & \paperautocell{0.0647} & \paperautocell{\textbf{17.03}} & \paperautocell{0.9981} & \paperautocell{0.0219} & \paperautocell{0.6747} & \paperautocell{0.9185} & \paperautocell{0.0527} & \paperautocell{17.13} & \paperautocell{0.9981} & \paperautocell{0.0229} & \paperautocell{0.6320} & \paperautocell{0.9164} \\
    \midrule
    \multicolumn{19}{l}{\textbf{\textit{L2}}} \\
    Coarse & \paperautocell{0.0335} & \paperautocell{15.71} & \paperautocell{0.9980} & \paperautocell{\textbf{0.0190}} & \paperautocell{\textbf{0.6736}} & \paperautocell{0.9270} & \paperautocell{0.0639} & \paperautocell{16.97} & \paperautocell{\textbf{0.9972}} & \paperautocell{\textbf{0.0200}} & \paperautocell{\textbf{0.7163}} & \paperautocell{0.9237} & \paperautocell{0.0487} & \paperautocell{16.34} & \paperautocell{\textbf{0.9976}} & \paperautocell{\textbf{0.0195}} & \paperautocell{\textbf{0.6950}} & \paperautocell{0.9254} \\
    \cellcolor{emphrow}\textbf{Adaptive (ours)} & \paperautocell{\textbf{0.0324}} & \paperautocell{\textbf{15.31}} & \paperautocell{\textbf{0.9990}} & \paperautocell{\textbf{0.0190}} & \paperautocell{0.6718} & \paperautocell{\textbf{0.9296}} & \paperautocell{\textbf{0.0565}} & \paperautocell{16.33} & \paperautocell{0.9962} & \paperautocell{0.0212} & \paperautocell{0.7155} & \paperautocell{\textbf{0.9247}} & \paperautocell{\textbf{0.0445}} & \paperautocell{\textbf{15.82}} & \paperautocell{\textbf{0.9976}} & \paperautocell{0.0201} & \paperautocell{0.6936} & \paperautocell{\textbf{0.9272}} \\
    Fine & \paperautocell{0.0353} & \paperautocell{15.81} & \paperautocell{0.9923} & \paperautocell{0.0219} & \paperautocell{0.6265} & \paperautocell{0.9220} & \paperautocell{0.0605} & \paperautocell{\textbf{16.20}} & \paperautocell{0.9942} & \paperautocell{0.0209} & \paperautocell{0.7031} & \paperautocell{0.9232} & \paperautocell{0.0479} & \paperautocell{16.00} & \paperautocell{0.9932} & \paperautocell{0.0214} & \paperautocell{0.6648} & \paperautocell{0.9226} \\
    \bottomrule
  \end{tabular}%
  }
\end{table}

\begin{table}[H]
  \centering
  \caption{\textbf{Effect of tile pyramid level selection on Tanks and Temples with 32 input views.} Coarse fixes every cell to the coarsest pyramid level whose image height is at least 512 pixels, and Fine to the full-resolution level; ours matches the level to the cell's voxel spacing. Inputs are the 4K images. F@$\tau$ and CD$/\tau$ use the official per-scene threshold $\tau$, and CD is in meters. The best results are in bold.}
  \label{tab:app-tile-level-tnt}
  \scriptsize
  \setlength{\tabcolsep}{3pt}
  \definecolor{rank1}{HTML}{FFB3B3}
  \definecolor{rank2}{HTML}{FFD9B3}
  \definecolor{rank3}{HTML}{FFFFB3}
  \definecolor{emphrow}{HTML}{EDEDED}
  \ifdefined\paperautocellwidth\else\newlength{\paperautocellwidth}\fi
  \ifdefined\paperautoprobe\else\newlength{\paperautoprobe}\fi
  \settowidth{\paperautocellwidth}{\textbf{00.0000}}
  \settowidth{\paperautoprobe}{Barn}
  \ifdim\dimexpr\paperautoprobe-2\tabcolsep\relax>2\paperautocellwidth  \setlength{\paperautocellwidth}{\dimexpr(\paperautoprobe-2\tabcolsep)/2\relax}\fi
  \settowidth{\paperautoprobe}{Caterpillar}
  \ifdim\dimexpr\paperautoprobe-2\tabcolsep\relax>2\paperautocellwidth  \setlength{\paperautocellwidth}{\dimexpr(\paperautoprobe-2\tabcolsep)/2\relax}\fi
  \settowidth{\paperautoprobe}{Courthouse}
  \ifdim\dimexpr\paperautoprobe-2\tabcolsep\relax>2\paperautocellwidth  \setlength{\paperautocellwidth}{\dimexpr(\paperautoprobe-2\tabcolsep)/2\relax}\fi
  \settowidth{\paperautoprobe}{Ignatius}
  \ifdim\dimexpr\paperautoprobe-2\tabcolsep\relax>2\paperautocellwidth  \setlength{\paperautocellwidth}{\dimexpr(\paperautoprobe-2\tabcolsep)/2\relax}\fi
  \settowidth{\paperautoprobe}{Meetingroom}
  \ifdim\dimexpr\paperautoprobe-2\tabcolsep\relax>2\paperautocellwidth  \setlength{\paperautocellwidth}{\dimexpr(\paperautoprobe-2\tabcolsep)/2\relax}\fi
  \settowidth{\paperautoprobe}{Truck}
  \ifdim\dimexpr\paperautoprobe-2\tabcolsep\relax>2\paperautocellwidth  \setlength{\paperautocellwidth}{\dimexpr(\paperautoprobe-2\tabcolsep)/2\relax}\fi
  \settowidth{\paperautoprobe}{Mean}
  \ifdim\dimexpr\paperautoprobe-2\tabcolsep\relax>2\paperautocellwidth  \setlength{\paperautocellwidth}{\dimexpr(\paperautoprobe-2\tabcolsep)/2\relax}\fi
  \settowidth{\paperautoprobe}{F@$\tau$$\uparrow$}
  \ifdim\dimexpr\paperautoprobe-0\tabcolsep\relax>1\paperautocellwidth  \setlength{\paperautocellwidth}{\dimexpr(\paperautoprobe-0\tabcolsep)/1\relax}\fi
  \settowidth{\paperautoprobe}{CD$\downarrow$}
  \ifdim\dimexpr\paperautoprobe-0\tabcolsep\relax>1\paperautocellwidth  \setlength{\paperautocellwidth}{\dimexpr(\paperautoprobe-0\tabcolsep)/1\relax}\fi
  \settowidth{\paperautoprobe}{CD$/\tau$$\downarrow$}
  \ifdim\dimexpr\paperautoprobe-0\tabcolsep\relax>1\paperautocellwidth  \setlength{\paperautocellwidth}{\dimexpr(\paperautoprobe-0\tabcolsep)/1\relax}\fi
  \providecommand{\paperautospan}[2]{\makebox[\dimexpr#1\paperautocellwidth+#1\tabcolsep+#1\tabcolsep-\tabcolsep-\tabcolsep\relax][c]{#2}}
  \providecommand{\paperautocell}[1]{\paperautospan{1}{#1}}
  \resizebox{\linewidth}{!}{%
  \begin{tabular}{lcccccccccccccc}
    \toprule
     & \multicolumn{2}{c}{\paperautospan{2}{Barn}} & \multicolumn{2}{c}{\paperautospan{2}{Caterpillar}} & \multicolumn{2}{c}{\paperautospan{2}{Courthouse}} & \multicolumn{2}{c}{\paperautospan{2}{Ignatius}} & \multicolumn{2}{c}{\paperautospan{2}{Meetingroom}} & \multicolumn{2}{c}{\paperautospan{2}{Truck}} & \multicolumn{2}{c}{\paperautospan{2}{Mean}} \\
    \cmidrule(lr){2-3}\cmidrule(lr){4-5}\cmidrule(lr){6-7}\cmidrule(lr){8-9}\cmidrule(lr){10-11}\cmidrule(lr){12-13}\cmidrule(lr){14-15}
    Tile level & \multicolumn{1}{c}{\paperautospan{1}{F@$\tau$$\uparrow$}} & \multicolumn{1}{c}{\paperautospan{1}{CD$\downarrow$}} & \multicolumn{1}{c}{\paperautospan{1}{F@$\tau$$\uparrow$}} & \multicolumn{1}{c}{\paperautospan{1}{CD$\downarrow$}} & \multicolumn{1}{c}{\paperautospan{1}{F@$\tau$$\uparrow$}} & \multicolumn{1}{c}{\paperautospan{1}{CD$\downarrow$}} & \multicolumn{1}{c}{\paperautospan{1}{F@$\tau$$\uparrow$}} & \multicolumn{1}{c}{\paperautospan{1}{CD$\downarrow$}} & \multicolumn{1}{c}{\paperautospan{1}{F@$\tau$$\uparrow$}} & \multicolumn{1}{c}{\paperautospan{1}{CD$\downarrow$}} & \multicolumn{1}{c}{\paperautospan{1}{F@$\tau$$\uparrow$}} & \multicolumn{1}{c}{\paperautospan{1}{CD$\downarrow$}} & \multicolumn{1}{c}{\paperautospan{1}{F@$\tau$$\uparrow$}} & \multicolumn{1}{c}{\paperautospan{1}{CD$/\tau$$\downarrow$}} \\
    \midrule
    \multicolumn{15}{l}{\textbf{\textit{L0}}} \\
    Coarse & \paperautocell{0.1554} & \paperautocell{0.0740} & \paperautocell{0.1273} & \paperautocell{0.0730} & \paperautocell{0.0908} & \paperautocell{1.0964} & \paperautocell{0.2996} & \paperautocell{0.0086} & \paperautocell{0.3052} & \paperautocell{0.0365} & \paperautocell{\textbf{0.3289}} & \paperautocell{\textbf{0.0296}} & \paperautocell{0.2178} & \paperautocell{13.0521} \\
    \cellcolor{emphrow}\textbf{Adaptive (ours)} & \paperautocell{0.1463} & \paperautocell{0.0787} & \paperautocell{\textbf{0.1328}} & \paperautocell{0.0805} & \paperautocell{\textbf{0.0965}} & \paperautocell{\textbf{0.3619}} & \paperautocell{\textbf{0.3223}} & \paperautocell{\textbf{0.0073}} & \paperautocell{\textbf{0.3561}} & \paperautocell{\textbf{0.0349}} & \paperautocell{0.2955} & \paperautocell{0.0317} & \paperautocell{\textbf{0.2249}} & \paperautocell{\textbf{8.4522}} \\
    Fine & \paperautocell{\textbf{0.1675}} & \paperautocell{\textbf{0.0619}} & \paperautocell{0.1176} & \paperautocell{\textbf{0.0648}} & \paperautocell{0.0921} & \paperautocell{0.7501} & \paperautocell{0.2672} & \paperautocell{0.0091} & \paperautocell{0.3274} & \paperautocell{0.0390} & \paperautocell{0.2671} & \paperautocell{0.0348} & \paperautocell{0.2065} & \paperautocell{10.5084} \\
    \midrule
    \multicolumn{15}{l}{\textbf{\textit{L1}}} \\
    Coarse & \paperautocell{0.1831} & \paperautocell{0.0849} & \paperautocell{0.1498} & \paperautocell{0.0703} & \paperautocell{0.1021} & \paperautocell{0.7583} & \paperautocell{0.3243} & \paperautocell{0.0077} & \paperautocell{0.3199} & \paperautocell{0.0362} & \paperautocell{\textbf{0.3550}} & \paperautocell{\textbf{0.0286}} & \paperautocell{0.2390} & \paperautocell{10.8031} \\
    \cellcolor{emphrow}\textbf{Adaptive (ours)} & \paperautocell{0.1926} & \paperautocell{0.0641} & \paperautocell{\textbf{0.1752}} & \paperautocell{0.0727} & \paperautocell{\textbf{0.1181}} & \paperautocell{\textbf{0.3355}} & \paperautocell{\textbf{0.4194}} & \paperautocell{\textbf{0.0058}} & \paperautocell{\textbf{0.3721}} & \paperautocell{\textbf{0.0318}} & \paperautocell{0.3523} & \paperautocell{0.0305} & \paperautocell{\textbf{0.2716}} & \paperautocell{\textbf{7.5968}} \\
    Fine & \paperautocell{\textbf{0.2170}} & \paperautocell{\textbf{0.0579}} & \paperautocell{0.1488} & \paperautocell{\textbf{0.0641}} & \paperautocell{0.1164} & \paperautocell{0.6199} & \paperautocell{0.4006} & \paperautocell{0.0061} & \paperautocell{0.3518} & \paperautocell{0.0350} & \paperautocell{0.3367} & \paperautocell{0.0313} & \paperautocell{0.2619} & \paperautocell{9.1992} \\
    \midrule
    \multicolumn{15}{l}{\textbf{\textit{L2}}} \\
    Coarse & \paperautocell{0.1700} & \paperautocell{0.0800} & \paperautocell{0.1442} & \paperautocell{0.0682} & \paperautocell{0.1020} & \paperautocell{0.3401} & \paperautocell{0.3063} & \paperautocell{0.0081} & \paperautocell{0.3026} & \paperautocell{0.0396} & \paperautocell{0.3578} & \paperautocell{\textbf{0.0287}} & \paperautocell{0.2305} & \paperautocell{7.9414} \\
    \cellcolor{emphrow}\textbf{Adaptive (ours)} & \paperautocell{0.1938} & \paperautocell{0.0866} & \paperautocell{\textbf{0.1945}} & \paperautocell{0.0740} & \paperautocell{\textbf{0.1324}} & \paperautocell{\textbf{0.2553}} & \paperautocell{\textbf{0.4904}} & \paperautocell{\textbf{0.0051}} & \paperautocell{\textbf{0.3806}} & \paperautocell{\textbf{0.0308}} & \paperautocell{\textbf{0.3705}} & \paperautocell{0.0304} & \paperautocell{\textbf{0.2937}} & \paperautocell{\textbf{7.4218}} \\
    Fine & \paperautocell{\textbf{0.2002}} & \paperautocell{\textbf{0.0604}} & \paperautocell{0.1814} & \paperautocell{\textbf{0.0607}} & \paperautocell{0.1269} & \paperautocell{0.4057} & \paperautocell{0.4862} & \paperautocell{\textbf{0.0051}} & \paperautocell{0.3662} & \paperautocell{0.0343} & \paperautocell{0.3593} & \paperautocell{0.0304} & \paperautocell{0.2867} & \paperautocell{7.6054} \\
    \bottomrule
  \end{tabular}%
  }
\end{table}

\subsection{Attention window size}
\label{app:window-ablation}

We compare two window sizes for the voxel-queried image attention described in
Appendix~\ref{app:relative-attention}: $w=6$ and $w=2$. Each window size is evaluated with
a separate model trained under the same schedule. Table~\ref{tab:app-window-scannetpp}
shows that both settings perform comparably across all cascade levels with 32 and 256
input views. We therefore adopt $w=2$, which attends to 4 tokens per view instead of 36,
speeding up sampling in the mesh stage by roughly $4\times$.

\begin{table}[H]
  \centering
  \caption{\textbf{Effect of the attention window size $w$ on ScanNet++.} Each voxel attends to the $w\times w$ DINO tokens around its projection in every view. Mean over the 25 smallest, the 25 largest, and all 50 scenes, with the best results in bold.}
  \label{tab:app-window-scannetpp}
  \scriptsize
  \setlength{\tabcolsep}{3pt}
  \definecolor{rank1}{HTML}{FFB3B3}
  \definecolor{rank2}{HTML}{FFD9B3}
  \definecolor{rank3}{HTML}{FFFFB3}
  \definecolor{emphrow}{HTML}{EDEDED}
  \ifdefined\paperautocellwidth\else\newlength{\paperautocellwidth}\fi
  \ifdefined\paperautoprobe\else\newlength{\paperautoprobe}\fi
  \settowidth{\paperautocellwidth}{\textbf{0.0000}}
  \settowidth{\paperautoprobe}{Small (25)}
  \ifdim\dimexpr\paperautoprobe-10\tabcolsep\relax>6\paperautocellwidth  \setlength{\paperautocellwidth}{\dimexpr(\paperautoprobe-10\tabcolsep)/6\relax}\fi
  \settowidth{\paperautoprobe}{Large (25)}
  \ifdim\dimexpr\paperautoprobe-10\tabcolsep\relax>6\paperautocellwidth  \setlength{\paperautocellwidth}{\dimexpr(\paperautoprobe-10\tabcolsep)/6\relax}\fi
  \settowidth{\paperautoprobe}{All (50)}
  \ifdim\dimexpr\paperautoprobe-10\tabcolsep\relax>6\paperautocellwidth  \setlength{\paperautocellwidth}{\dimexpr(\paperautoprobe-10\tabcolsep)/6\relax}\fi
  \settowidth{\paperautoprobe}{2D}
  \ifdim\dimexpr\paperautoprobe-4\tabcolsep\relax>3\paperautocellwidth  \setlength{\paperautocellwidth}{\dimexpr(\paperautoprobe-4\tabcolsep)/3\relax}\fi
  \settowidth{\paperautoprobe}{3D virtual scan}
  \ifdim\dimexpr\paperautoprobe-4\tabcolsep\relax>3\paperautocellwidth  \setlength{\paperautocellwidth}{\dimexpr(\paperautoprobe-4\tabcolsep)/3\relax}\fi
  \settowidth{\paperautoprobe}{MAE$\downarrow$}
  \ifdim\dimexpr\paperautoprobe-0\tabcolsep\relax>1\paperautocellwidth  \setlength{\paperautocellwidth}{\dimexpr(\paperautoprobe-0\tabcolsep)/1\relax}\fi
  \settowidth{\paperautoprobe}{Ang$\downarrow$}
  \ifdim\dimexpr\paperautoprobe-0\tabcolsep\relax>1\paperautocellwidth  \setlength{\paperautocellwidth}{\dimexpr(\paperautoprobe-0\tabcolsep)/1\relax}\fi
  \settowidth{\paperautoprobe}{Comp$\uparrow$}
  \ifdim\dimexpr\paperautoprobe-0\tabcolsep\relax>1\paperautocellwidth  \setlength{\paperautocellwidth}{\dimexpr(\paperautoprobe-0\tabcolsep)/1\relax}\fi
  \settowidth{\paperautoprobe}{CD$\downarrow$}
  \ifdim\dimexpr\paperautoprobe-0\tabcolsep\relax>1\paperautocellwidth  \setlength{\paperautocellwidth}{\dimexpr(\paperautoprobe-0\tabcolsep)/1\relax}\fi
  \settowidth{\paperautoprobe}{F@2$\uparrow$}
  \ifdim\dimexpr\paperautoprobe-0\tabcolsep\relax>1\paperautocellwidth  \setlength{\paperautocellwidth}{\dimexpr(\paperautoprobe-0\tabcolsep)/1\relax}\fi
  \settowidth{\paperautoprobe}{NC$\uparrow$}
  \ifdim\dimexpr\paperautoprobe-0\tabcolsep\relax>1\paperautocellwidth  \setlength{\paperautocellwidth}{\dimexpr(\paperautoprobe-0\tabcolsep)/1\relax}\fi
  \providecommand{\paperautospan}[2]{\makebox[\dimexpr#1\paperautocellwidth+#1\tabcolsep+#1\tabcolsep-\tabcolsep-\tabcolsep\relax][c]{#2}}
  \providecommand{\paperautocell}[1]{\paperautospan{1}{#1}}
  \resizebox{\linewidth}{!}{%
  \begin{tabular}{lcccccccccccccccccc}
    \toprule
     & \multicolumn{6}{c}{\paperautospan{6}{Small (25)}} & \multicolumn{6}{c}{\paperautospan{6}{Large (25)}} & \multicolumn{6}{c}{\paperautospan{6}{All (50)}} \\
    \cmidrule(lr){2-7}\cmidrule(lr){8-13}\cmidrule(lr){14-19}
     & \multicolumn{3}{c}{\paperautospan{3}{2D}} & \multicolumn{3}{c}{\paperautospan{3}{3D virtual scan}} & \multicolumn{3}{c}{\paperautospan{3}{2D}} & \multicolumn{3}{c}{\paperautospan{3}{3D virtual scan}} & \multicolumn{3}{c}{\paperautospan{3}{2D}} & \multicolumn{3}{c}{\paperautospan{3}{3D virtual scan}} \\
    \cmidrule(lr){2-4}\cmidrule(lr){5-7}\cmidrule(lr){8-10}\cmidrule(lr){11-13}\cmidrule(lr){14-16}\cmidrule(lr){17-19}
    Window & \multicolumn{1}{c}{\paperautospan{1}{MAE$\downarrow$}} & \multicolumn{1}{c}{\paperautospan{1}{Ang$\downarrow$}} & \multicolumn{1}{c}{\paperautospan{1}{Comp$\uparrow$}} & \multicolumn{1}{c}{\paperautospan{1}{CD$\downarrow$}} & \multicolumn{1}{c}{\paperautospan{1}{F@2$\uparrow$}} & \multicolumn{1}{c}{\paperautospan{1}{NC$\uparrow$}} & \multicolumn{1}{c}{\paperautospan{1}{MAE$\downarrow$}} & \multicolumn{1}{c}{\paperautospan{1}{Ang$\downarrow$}} & \multicolumn{1}{c}{\paperautospan{1}{Comp$\uparrow$}} & \multicolumn{1}{c}{\paperautospan{1}{CD$\downarrow$}} & \multicolumn{1}{c}{\paperautospan{1}{F@2$\uparrow$}} & \multicolumn{1}{c}{\paperautospan{1}{NC$\uparrow$}} & \multicolumn{1}{c}{\paperautospan{1}{MAE$\downarrow$}} & \multicolumn{1}{c}{\paperautospan{1}{Ang$\downarrow$}} & \multicolumn{1}{c}{\paperautospan{1}{Comp$\uparrow$}} & \multicolumn{1}{c}{\paperautospan{1}{CD$\downarrow$}} & \multicolumn{1}{c}{\paperautospan{1}{F@2$\uparrow$}} & \multicolumn{1}{c}{\paperautospan{1}{NC$\uparrow$}} \\
    \midrule
    \multicolumn{19}{l}{\textbf{\textit{32 Views, L0}}} \\
    $w=6$ & \paperautocell{\textbf{0.0437}} & \paperautocell{\textbf{18.65}} & \paperautocell{0.9985} & \paperautocell{\textbf{0.0230}} & \paperautocell{\textbf{0.5845}} & \paperautocell{\textbf{0.9111}} & \paperautocell{0.0723} & \paperautocell{19.27} & \paperautocell{\textbf{0.9984}} & \paperautocell{0.0238} & \paperautocell{0.6316} & \paperautocell{\textbf{0.9090}} & \paperautocell{0.0580} & \paperautocell{\textbf{18.96}} & \paperautocell{0.9984} & \paperautocell{0.0234} & \paperautocell{\textbf{0.6080}} & \paperautocell{\textbf{0.9100}} \\
    \cellcolor{emphrow}\textbf{$w=2$ (ours)} & \paperautocell{0.0438} & \paperautocell{18.74} & \paperautocell{\textbf{0.9986}} & \paperautocell{\textbf{0.0230}} & \paperautocell{0.5816} & \paperautocell{0.9110} & \paperautocell{\textbf{0.0714}} & \paperautocell{\textbf{19.21}} & \paperautocell{\textbf{0.9984}} & \paperautocell{\textbf{0.0235}} & \paperautocell{\textbf{0.6336}} & \paperautocell{\textbf{0.9090}} & \paperautocell{\textbf{0.0576}} & \paperautocell{18.98} & \paperautocell{\textbf{0.9985}} & \paperautocell{\textbf{0.0233}} & \paperautocell{0.6076} & \paperautocell{\textbf{0.9100}} \\
    \midrule
    \multicolumn{19}{l}{\textbf{\textit{32 Views, L1}}} \\
    $w=6$ & \paperautocell{\textbf{0.0345}} & \paperautocell{16.38} & \paperautocell{\textbf{0.9994}} & \paperautocell{\textbf{0.0200}} & \paperautocell{\textbf{0.6392}} & \paperautocell{\textbf{0.9245}} & \paperautocell{0.0580} & \paperautocell{17.22} & \paperautocell{\textbf{0.9985}} & \paperautocell{0.0201} & \paperautocell{\textbf{0.6968}} & \paperautocell{0.9230} & \paperautocell{0.0463} & \paperautocell{16.80} & \paperautocell{\textbf{0.9990}} & \paperautocell{0.0200} & \paperautocell{\textbf{0.6680}} & \paperautocell{\textbf{0.9237}} \\
    \cellcolor{emphrow}\textbf{$w=2$ (ours)} & \paperautocell{0.0346} & \paperautocell{\textbf{16.35}} & \paperautocell{\textbf{0.9994}} & \paperautocell{\textbf{0.0200}} & \paperautocell{0.6359} & \paperautocell{0.9240} & \paperautocell{\textbf{0.0569}} & \paperautocell{\textbf{17.10}} & \paperautocell{0.9983} & \paperautocell{\textbf{0.0199}} & \paperautocell{0.6959} & \paperautocell{\textbf{0.9232}} & \paperautocell{\textbf{0.0457}} & \paperautocell{\textbf{16.72}} & \paperautocell{0.9988} & \paperautocell{\textbf{0.0199}} & \paperautocell{0.6659} & \paperautocell{0.9236} \\
    \midrule
    \multicolumn{19}{l}{\textbf{\textit{32 Views, L2}}} \\
    $w=6$ & \paperautocell{\textbf{0.0323}} & \paperautocell{15.51} & \paperautocell{0.9988} & \paperautocell{\textbf{0.0189}} & \paperautocell{0.6712} & \paperautocell{0.9293} & \paperautocell{0.0592} & \paperautocell{16.47} & \paperautocell{\textbf{0.9966}} & \paperautocell{0.0229} & \paperautocell{0.7129} & \paperautocell{0.9213} & \paperautocell{0.0458} & \paperautocell{15.99} & \paperautocell{\textbf{0.9977}} & \paperautocell{0.0209} & \paperautocell{0.6921} & \paperautocell{0.9253} \\
    \cellcolor{emphrow}\textbf{$w=2$ (ours)} & \paperautocell{0.0324} & \paperautocell{\textbf{15.31}} & \paperautocell{\textbf{0.9990}} & \paperautocell{0.0190} & \paperautocell{\textbf{0.6718}} & \paperautocell{\textbf{0.9296}} & \paperautocell{\textbf{0.0565}} & \paperautocell{\textbf{16.33}} & \paperautocell{0.9962} & \paperautocell{\textbf{0.0212}} & \paperautocell{\textbf{0.7155}} & \paperautocell{\textbf{0.9247}} & \paperautocell{\textbf{0.0445}} & \paperautocell{\textbf{15.82}} & \paperautocell{0.9976} & \paperautocell{\textbf{0.0201}} & \paperautocell{\textbf{0.6936}} & \paperautocell{\textbf{0.9272}} \\
    \midrule
    \multicolumn{19}{l}{\textbf{\textit{256 Views, L0}}} \\
    $w=6$ & \paperautocell{0.0473} & \paperautocell{19.11} & \paperautocell{0.9987} & \paperautocell{0.0282} & \paperautocell{\textbf{0.4909}} & \paperautocell{0.9078} & \paperautocell{0.0689} & \paperautocell{19.05} & \paperautocell{0.9989} & \paperautocell{0.0234} & \paperautocell{0.6229} & \paperautocell{0.9122} & \paperautocell{0.0581} & \paperautocell{19.08} & \paperautocell{0.9988} & \paperautocell{0.0258} & \paperautocell{0.5569} & \paperautocell{0.9100} \\
    \cellcolor{emphrow}\textbf{$w=2$ (ours)} & \paperautocell{\textbf{0.0469}} & \paperautocell{\textbf{18.98}} & \paperautocell{\textbf{0.9988}} & \paperautocell{\textbf{0.0279}} & \paperautocell{0.4897} & \paperautocell{\textbf{0.9086}} & \paperautocell{\textbf{0.0683}} & \paperautocell{\textbf{18.95}} & \paperautocell{\textbf{0.9990}} & \paperautocell{\textbf{0.0233}} & \paperautocell{\textbf{0.6251}} & \paperautocell{\textbf{0.9125}} & \paperautocell{\textbf{0.0576}} & \paperautocell{\textbf{18.97}} & \paperautocell{\textbf{0.9989}} & \paperautocell{\textbf{0.0256}} & \paperautocell{\textbf{0.5574}} & \paperautocell{\textbf{0.9106}} \\
    \midrule
    \multicolumn{19}{l}{\textbf{\textit{256 Views, L1}}} \\
    $w=6$ & \paperautocell{0.0379} & \paperautocell{\textbf{16.43}} & \paperautocell{0.9993} & \paperautocell{0.0242} & \paperautocell{0.5648} & \paperautocell{0.9210} & \paperautocell{\textbf{0.0570}} & \paperautocell{\textbf{16.58}} & \paperautocell{\textbf{0.9896}} & \paperautocell{\textbf{0.0216}} & \paperautocell{\textbf{0.6902}} & \paperautocell{\textbf{0.9229}} & \paperautocell{0.0474} & \paperautocell{\textbf{16.51}} & \paperautocell{\textbf{0.9945}} & \paperautocell{0.0229} & \paperautocell{0.6275} & \paperautocell{\textbf{0.9219}} \\
    \cellcolor{emphrow}\textbf{$w=2$ (ours)} & \paperautocell{\textbf{0.0375}} & \paperautocell{16.44} & \paperautocell{\textbf{0.9995}} & \paperautocell{\textbf{0.0238}} & \paperautocell{\textbf{0.5726}} & \paperautocell{\textbf{0.9211}} & \paperautocell{0.0571} & \paperautocell{16.68} & \paperautocell{0.9892} & \paperautocell{0.0217} & \paperautocell{0.6874} & \paperautocell{0.9224} & \paperautocell{\textbf{0.0473}} & \paperautocell{16.56} & \paperautocell{0.9943} & \paperautocell{\textbf{0.0227}} & \paperautocell{\textbf{0.6300}} & \paperautocell{0.9217} \\
    \midrule
    \multicolumn{19}{l}{\textbf{\textit{256 Views, L2}}} \\
    $w=6$ & \paperautocell{0.0364} & \paperautocell{15.55} & \paperautocell{\textbf{0.9993}} & \paperautocell{0.0239} & \paperautocell{0.6037} & \paperautocell{0.9251} & \paperautocell{\textbf{0.0546}} & \paperautocell{15.69} & \paperautocell{\textbf{0.9912}} & \paperautocell{\textbf{0.0202}} & \paperautocell{0.7172} & \paperautocell{0.9282} & \paperautocell{\textbf{0.0455}} & \paperautocell{15.62} & \paperautocell{\textbf{0.9952}} & \paperautocell{0.0221} & \paperautocell{0.6604} & \paperautocell{0.9267} \\
    \cellcolor{emphrow}\textbf{$w=2$ (ours)} & \paperautocell{\textbf{0.0361}} & \paperautocell{\textbf{15.49}} & \paperautocell{0.9991} & \paperautocell{\textbf{0.0234}} & \paperautocell{\textbf{0.6167}} & \paperautocell{\textbf{0.9260}} & \paperautocell{0.0557} & \paperautocell{\textbf{15.62}} & \paperautocell{0.9902} & \paperautocell{\textbf{0.0202}} & \paperautocell{\textbf{0.7176}} & \paperautocell{\textbf{0.9283}} & \paperautocell{0.0459} & \paperautocell{\textbf{15.56}} & \paperautocell{0.9947} & \paperautocell{\textbf{0.0218}} & \paperautocell{\textbf{0.6672}} & \paperautocell{\textbf{0.9271}} \\
    \bottomrule
  \end{tabular}%
  }
\end{table}

\subsection{Synthetic training data}
\label{app:training-data-ablation}

\begin{figure}[H]
  \centering
  \setlength{\tabcolsep}{0.5pt}
  \renewcommand{\arraystretch}{1.0}
  \begin{tabular}{ccc}
    \multicolumn{3}{c}{\scriptsize (a) ScanNet++ \texttt{09c1414f1b}} \\[1.5pt]
    \includegraphics[width=\dimexpr(\textwidth-3pt)/3\relax]{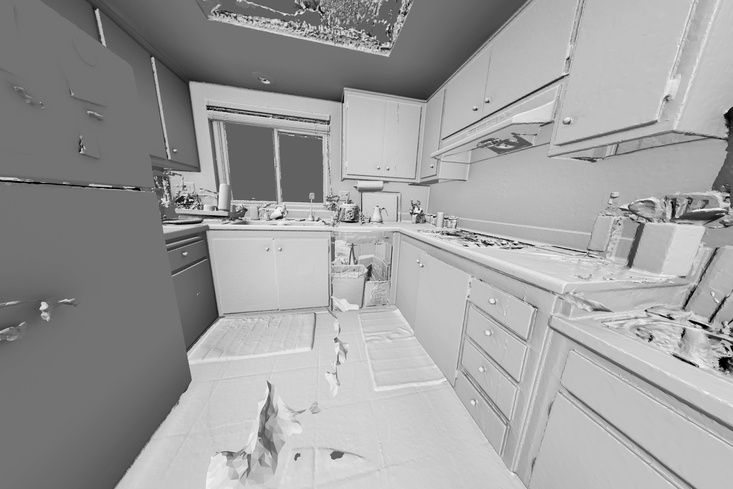} & \includegraphics[width=\dimexpr(\textwidth-3pt)/3\relax]{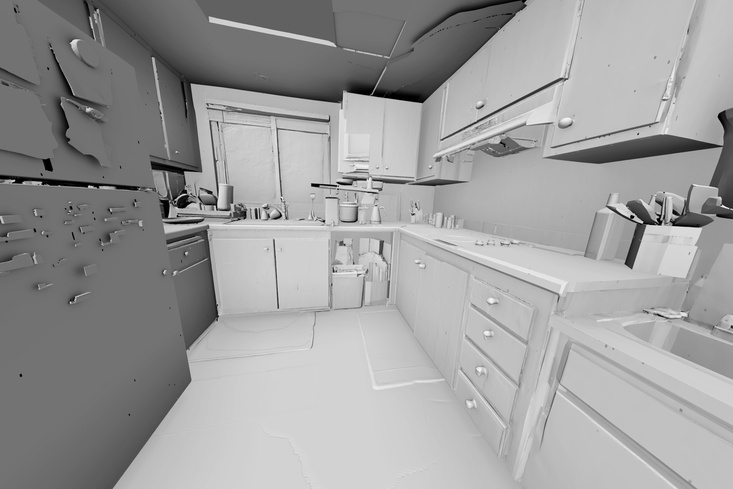} & \includegraphics[width=\dimexpr(\textwidth-3pt)/3\relax]{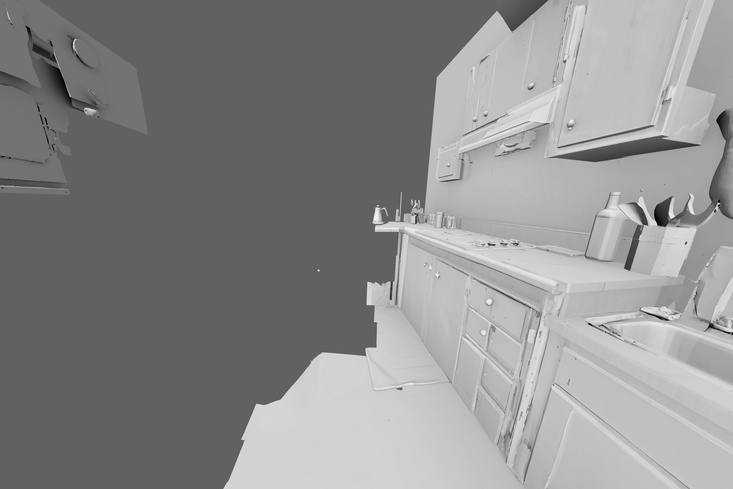} \\[-4.5pt]
    \makebox[\dimexpr(\textwidth-3pt)/3\relax][c]{\scriptsize GT} & \makebox[\dimexpr(\textwidth-3pt)/3\relax][c]{\scriptsize Full} & \makebox[\dimexpr(\textwidth-3pt)/3\relax][c]{\scriptsize w/o house \& grass} \\[8pt]
    \multicolumn{3}{c}{\scriptsize (b) \texttt{caterpillar}} \\[1.5pt]
    \includegraphics[width=\dimexpr(\textwidth-3pt)/3\relax]{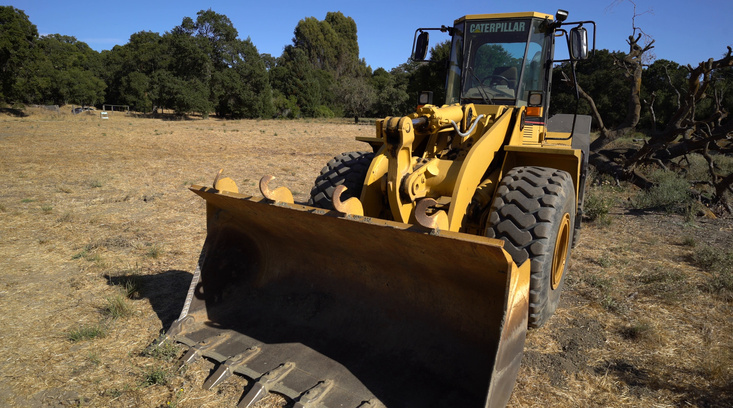} & \includegraphics[width=\dimexpr(\textwidth-3pt)/3\relax]{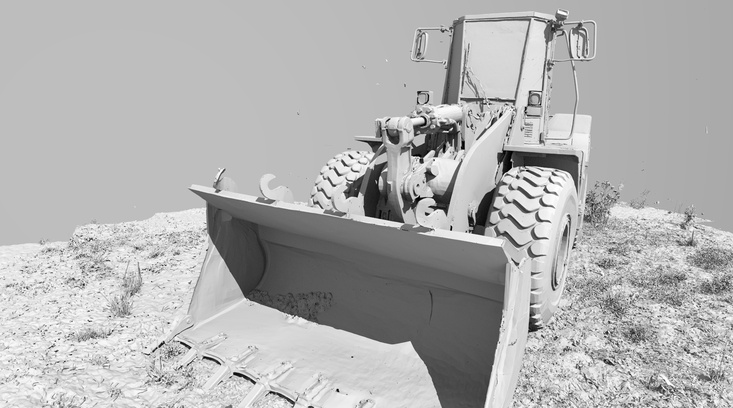} & \includegraphics[width=\dimexpr(\textwidth-3pt)/3\relax]{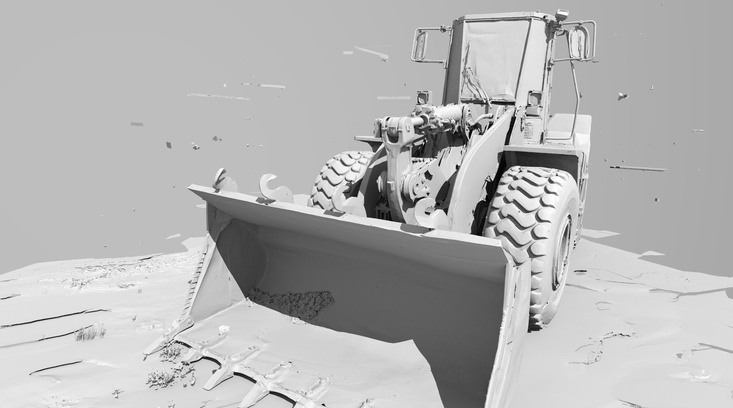} \\[-4.5pt]
    \makebox[\dimexpr(\textwidth-3pt)/3\relax][c]{\scriptsize Image} & \makebox[\dimexpr(\textwidth-3pt)/3\relax][c]{\scriptsize Full} & \makebox[\dimexpr(\textwidth-3pt)/3\relax][c]{\scriptsize w/o house \& grass} \\
  \end{tabular}
  \caption{\textbf{Effect of the synthetic training data.} Meshes at the finest cascade level with 32 input views, from the model trained with our synthetic multi-room houses and grass scenes (Full) and from one trained on SAGE-10K and 3D-FRONT alone (w/o house \& grass). Without the synthetic data, walls of the multi-room ScanNet++ scene disappear, and the grass around the loader becomes a flat surface.}
  \label{fig:app-training-data}
\end{figure}

We compare the model trained on our full data mixture against a variant trained without
synthetic multi-room houses and grass scenes---i.e., using only SAGE-10K and 3D-FRONT
(Appendix~\ref{app:general-training}). This comparison uses $w=6$ instead of the final $w=2$; the two window sizes perform
on par (Appendix~\ref{app:window-ablation}). Figure~\ref{fig:app-training-data} compares the
finest-level reconstructions using 32 input views. Without synthetic multi-room houses,
walls can go missing in multi-room scenes. In the ScanNet++ scene shown at the top of
Figure~\ref{fig:app-training-data}, the model trained without them fails to reconstruct
the left and far walls, whereas the full model reconstructs them. On \texttt{caterpillar}, the
model without grass scenes reduces the grassy ground to a flat, featureless surface,
whereas the full model captures individual grass tufts.

\section{Inference Time}
\label{app:inference-time}

Table~\ref{tab:app-inference-time} reports the inference time for each cascade level on
five large ScanNet++ scenes with 32 input views on a single NVIDIA H200 GPU. Because each level halves the cell size, the number of cells and the total computation can
grow by roughly $8\times$ per level, and by more depending on the cell layout. In practice,
the number of cells grows from 31 to 470 to 2{,}490, and runtime increases by $8.2\times$
from level 0 to level 1 but only by $4.7\times$ from level 1 to level 2, because finer
levels allocate cells only where the parent level detects surfaces. Reconstruction up to level 1 takes an
average of 17 minutes per scene, and up to level 2 takes 89 minutes, which is comparable
to per-scene optimization methods. Flow sampling in the SS and mesh stages accounts for most of the runtime at all levels, whereas view selection, tile preparation,
and parent transfer require less than 3 minutes combined. Peak GPU memory usage is
79~GiB at the finest level.

\begin{table}[H]
  \centering
  \caption{\textbf{Inference time per cascade level on ScanNet++.} Wall-clock seconds on one NVIDIA H200 GPU with 32 input views, averaged over the 5 large scenes. View sel.\ + tiles includes DA3 depth, per-cell view selection, tile cropping and tile feature encoding (DINOv3 and ConvNeXt); parent transfer decodes the parent level to initialize the child SS. Model loading is excluded. Peak is the peak allocated GPU memory.}
  \label{tab:app-inference-time}
  \scriptsize
  \setlength{\tabcolsep}{4pt}
  \begin{tabular}{lrrrrrrrrr}
    \toprule
    & \multicolumn{2}{c}{Count} & \multicolumn{6}{c}{Time (s)} & Memory (GiB) \\
    \cmidrule(lr){2-3}\cmidrule(lr){4-9}\cmidrule(lr){10-10}
    & & & View sel. & Parent &  &  &  & & \\
    Level & \#Cells & \#Tiles & + tiles & transfer & SS & Shape & Decode & Total & Peak \\
    \midrule
    L0 (4\,m) & 31 & 329 & 23 & -- & 38 & 48 & 2.4 & 112 & 36 \\
    L1 (2\,m) & 470 & 473 & 26 & 21 & 466 & 394 & 7.6 & 914 & 57 \\
    L2 (1\,m) & 2,490 & 591 & 27 & 60 & 2,428 & 1,740 & 29 & 4,285 & 79 \\
    \midrule
    Total &  &  & 76 & 82 & 2,932 & 2,181 & 39 & 5,310 (89\,min) & 79 \\
    \bottomrule
  \end{tabular}
\end{table}

\end{document}